\documentclass[times,twocolumn,final,authoryear,svgnames]{elsarticle}

\usepackage{ycviu}
\usepackage{framed,multirow}

\usepackage{latexsym}

\usepackage{times}
\usepackage{epsfig}
\usepackage{graphicx}
\usepackage{amsmath}
\usepackage{amssymb}
\usepackage{booktabs, array}
\usepackage{multirow}
\usepackage{comment}
\usepackage{color, colortbl}
\usepackage{enumitem}
\usepackage{bbding}
\usepackage{pifont}
\usepackage{caption}
\usepackage{subcaption}\usepackage{wasysym}
\newcommand{\tabincell}[2]{\begin{tabular}{@{}#1@{}}#2\end{tabular}}
\usepackage{bbm}

\usepackage{hyperref}
\hypersetup{colorlinks,allcolors=black}

\definecolor{LightYellow}{rgb}{1,1,0.7}

\definecolor{gold}{rgb}{1.0, 0.874, 0}
\definecolor{silver}{rgb}{0.77,0.77,0.77}
\definecolor{brown}{rgb}{0.95, 0.678, 0.4}
\definecolor{gray}{rgb}{0.86, 0.86, 0.86}

\newcommand{\gray}[1]{\colorbox{gray}{\textbf{#1}}}

\newcommand{\netname}{DSR-EI}

\begin{document}

\begin{frontmatter}

\title{Depth Super-Resolution from Explicit and Implicit High-Frequency Features}

\author[1]{Xin \snm{Qiao}} 
\author[1]{Chenyang \snm{Ge} \corref{cor1}}
\cortext[cor1]{Corresponding author: Chenyang Ge
  }
\ead{cyge@mail.xjtu.edu.cn}
\author[2]{Youmin \snm{Zhang}}
\author[1]{Yanhui \snm{Zhou}}
\author[2]{Fabio \snm{Tosi}}
\author[2]{Matteo \snm{Poggi}}
\author[2]{Stefano \snm{Mattoccia}}

\address[1]{Institute of Artificial Intelligence and Robotics, Xi'an Jiaotong University, No.28, West Xianning Road, Xi'an and 710049, China}
\address[2]{Department of Computer Science and Engineering, University of Bologna, Viale Risorgimento, 2, Bologna and 40136, Italy}

\received{1 May 2013}
\finalform{10 May 2013}
\accepted{13 May 2013}
\availableonline{15 May 2013}
\communicated{S. Sarkar}

\begin{abstract}
{Guided depth super-resolution aims at using a low-resolution depth map and an associated high-resolution RGB image to recover a high-resolution depth map. However, restoring precise and sharp edges near depth discontinuities and fine structures is still challenging for state-of-the-art methods. To alleviate this issue, we propose a novel multi-stage depth super-resolution network, which progressively reconstructs HR depth maps from explicit and implicit high-frequency information. We introduce an efficient transformer to obtain explicit high-frequency information. The shape bias and global context of the transformer allow our model to focus on high-frequency details between objects, i.e., depth discontinuities, rather than texture within objects. 
Furthermore, we project the input color images into the frequency domain for additional implicit high-frequency cues extraction. Finally, to incorporate the structural details, we develop a fusion strategy that combines depth features and high-frequency information in the multi-stage-scale framework. Exhaustive experiments on the main benchmarks show that our approach establishes a new state-of-the-art.}\\

\textit{This paper is under consideration at Computer Vision and Image Understanding.}

\end{abstract}

\begin{keyword}
Guided depth super-resolution \sep CNN \sep Transformer \sep Multi-scale \sep High-frequency information
\end{keyword}

\end{frontmatter}

\section{Introduction}

\begin{figure*}	
    \centering	
    \includegraphics[width=0.95\textwidth]{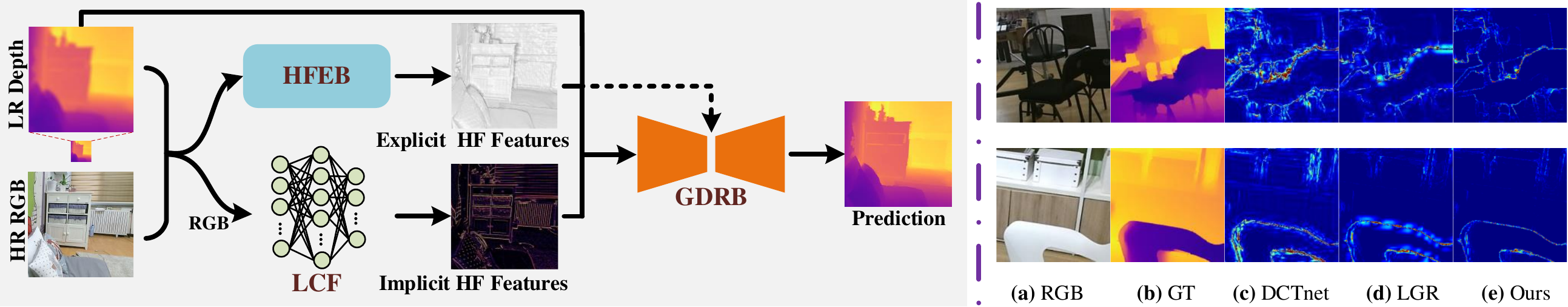} 
    \vspace{-0.3cm}
	\caption{\textbf{Depth Super-Resolution exploiting explicit and implicit high-frequency features.} On the left, an overview of our framework, combining the power of both explicit and implicit high-frequency information extracted from the inputs. On the right, qualitative examples with (a) RGB images, (b) ground truth depth and error maps by existing methods (c -- d) and ours (e).}
	\label{teaser}
\end{figure*}

With the rise of consumer-grade depth cameras, depth maps are employed in various scenarios such as 3D reconstruction \citep{chen20203d, chen2020real}, recognition \citep{cai20103d} and more. Time-of-Flight (ToF) is one of the leading technologies involved in depth sensing, measuring the distance traveled by emitted rays until they reach points in the scenes. However, due to the limitations of physical fabrication, power consumption and costs~\citep{bamji2022review}, the resolution of depth maps usually is often insufficient to fulfill the demand of the downstream applications, such as object detection~\citep{chen2021dpanet} and pose estimation~\citep{ge2019real}. 
In contrast, collecting RGB images at much higher resolution is cheaper. As a result, the guided depth super-resolution task, known as GDSR, has emerged as a crucial solution to this technological limitation, allowing to obtain an accurate high-resolution (HR) depth map from a low-resolution (LR) one, guided by an HR image.

Initially, algorithms addressing this problem were classified into local \citep{kopf2007joint, yang2007spatial, riemens2009multistep, wang2014graph} and global \citep{diebel2005application, park2011high, ferstl2013image, li2016fast}, with the former family being faster, yet suffering in low-textured regions and the latter resulting more robust, at the expense of processing time.
More recently, deep neural networks have become the preferred choice for depth super-resolution \citep{hui2016depth, li2016deep, li2019joint, lutio2019guided, tang2021bridgenet}, although they still struggle to restore sharp and precise edges from LR depth maps reliably, especially when dealing with large upsampling factors. 
This is mainly due to the inadequate guidance provided by High-Frequency (HF) features, implicitly modeled by deep networks, which frequently cause texture copying effects in the upsampled depth maps.
In addition, single-stage multi-scale architectures for this task \citep{ye2020pmbanet, wang2020depth, zuo2020frequency, tang2021joint}, at any given scale, cannot fully leverage fine details encoded at the higher ones, as they are lost due to down-sampling and only partially recovered through skip connections.

In light of the two weaknesses highlighted so far, we aim to improve GDSR by explicitly countering them. For the former, we argue that explicit extraction of HF features, supported by edge detection algorithms such as the Canny operator, can play a crucial role \citep{wang2020depth}. Concerning the latter, multi-stage network design -- which outperforms single-stage counterparts in high-level visual tasks like action segmentation \citep{farha2019ms} and pose estimation \citep{chen2018cascaded}, as well as for low-level vision problems such as image restoration \citep{zamir2021multi, kim2022mssnet} -- can mitigate the information loss issue. However, since features extracted from RGB images need to be considered in addition to depth features, existing multi-stage networks are inadequate for GDSR and should be revised to fuse features from the two domains.

In this paper, we present a \underline{D}epth \underline{S}uper-\underline{R}esolution method leveraging both \underline{E}xplicit and \underline{I}mplicit HF information (\underline{\netname}), which contains two branches: the High-Frequency Extraction Branch (HFEB) and the Guided Depth Restoration Branch (GDRB). The former is designed to model \textbf{explicit} HF features by exploiting dynamic self-calibrated convolutions (DSP) and the power of vision transformers blocks. The latter effectively fuses the guidance from RGB features with depth features to obtain HR depth maps. This is achieved by deploying two novel modules: 1) the Adaptive Feature Fusion Module (AFFM), which counters the HF information loss due to downsampling, and 2) the Low-Cut Filtering (LCF) module, which acts in the frequency domain to improve \textbf{implicit} extraction of HF features. Exhaustive experiments on several standard datasets show the superiority of \netname. In summary, the main contributions of this paper are:

\begin{itemize}[itemsep=2pt,topsep=0pt,parsep=0pt]
\item{The proposed architecture employs a novel efficient transformer for explicit, HF feature extraction. The transformer can accurately capture image details and structures from depth maps.}
\item{In the guided depth restoration branch, we propose a low-cut filtering module that can obtain accurate, {implicit} HF information.}
\item{To counter the information loss issue, we propose an Adaptive Feature Fusion Module located in the middle of the guided depth restoration branch.}
\item{Quantitative and qualitative experimental results demonstrate that our approach establishes a new state-of-the-art in the field of guided depth super-resolution.}
\end{itemize}
Fig. \ref{teaser} provides a high-level view of our framework, followed by examples that anticipate the superior accuracy achieved by \netname{} compared to existing methods \citep{zhao2022discrete,de2022learning}. 
{In case of acceptance, we will make our code publicly available to ease reproducibility.} 

\section{Related Work}
In this section, we first review the literature related to the GDSR task, divided into \textit{conventional} and \textit{learning} methods, as well as to vision transformers.

\begin{figure*}
	\centering
	\includegraphics[width=5.5in]{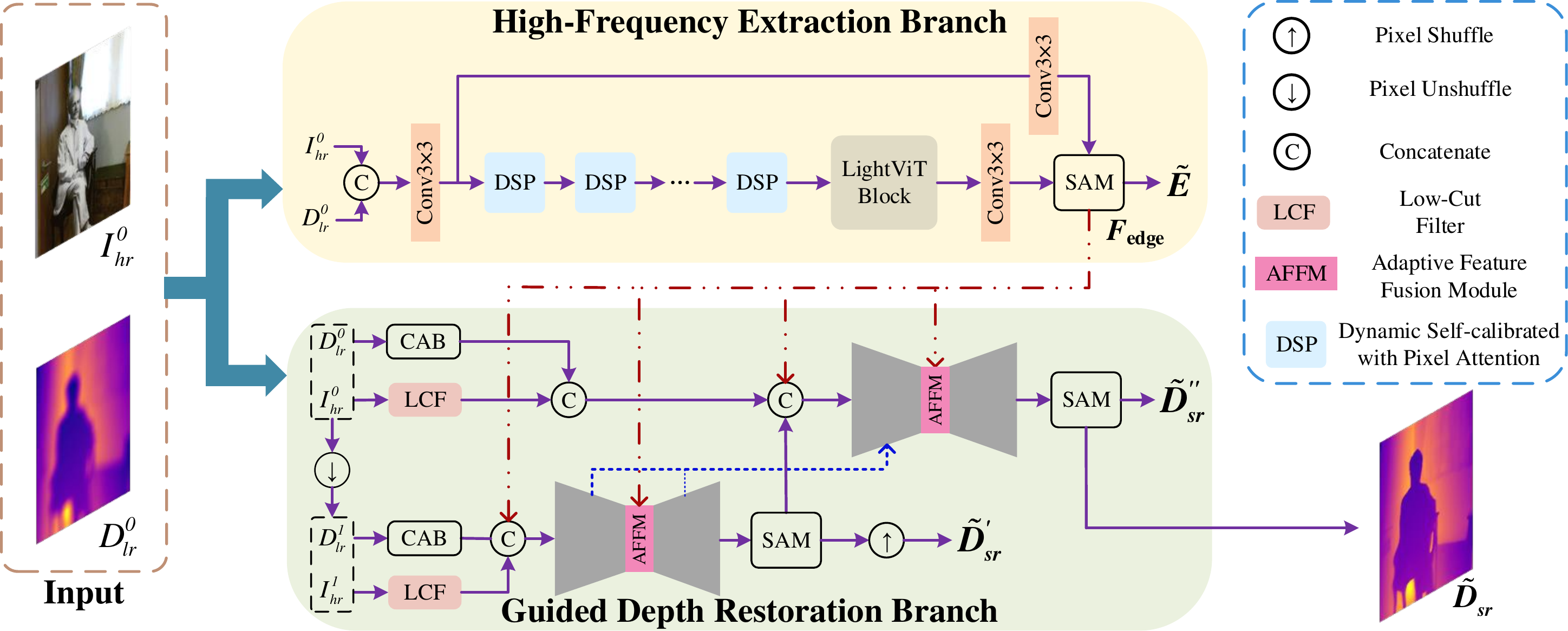}
	\vspace{-0.cm}
	\caption{\textbf{\netname \ architecture.} Rectangles with different colors depict different stages and functions in each stage.}
	\label{framework}
\end{figure*}

\textbf{Conventional Methods.}
Initially, hand-craft models were developed for GDSR, using the edge co-occurrence between the LR depth map and its HR color counterpart as prior. \cite{kopf2007joint} first utilizes a joint bilateral filter, taking guidance cues from color images. The so-called \textit{local} methods followed this pivotal work: \cite{yang2007spatial} enhances the LR depth maps by exploiting registered HR color images, \cite{riemens2009multistep} uses anti-alias image prefiltering built on the multi-stage joint bilateral filter, while graph-based joint bilateral upsampling \citep{wang2014graph} casts GDSR as a regularization problem. 

More accurate solutions, although slower, are represented by \textit{global} methods. The first work in this direction is \cite{diebel2005application}, which employs Markov random fields (MRF) to integrate multi-modal data for LR depth map upsampling. Using the non-local mean filtering method, \cite{park2011high} recovers noisy LR images from a ToF camera to a high-quality image. To be more efficient, \cite{ferstl2013image} exploits Total Generalized Variation (TGV) regularization for GDSR, enabling a high frame rate. \cite{li2016fast} uses fast global smoothing (FGS) to make guided depth interpolation more robust.

\textbf{Learning Methods.}
Earlier methods from this category exploit MRF~\citep{mac2012patch, kiechle2013joint, kwon2015data}. 
However, these techniques rely on manually created dictionaries, whose limited content restricts the capacity of generalizing. More recently, deep learning-based approaches achieved remarkable results and became the preferred choice for GDSR. 
\cite{hui2016depth} designs a multi-scale guided CNN using hierarchical feature extraction to gradually restores blurred edges. To reconstruct sharp edges, the works by \cite{li2016deep,li2019joint} learn salient features from color images using an encoder-decoder structure. In contrast, \cite{lutio2019guided} casts GDSR as a pixel-to-pixel mapping from the HR RGB image to the domain of the LR source image, learned by a multi-layer perceptron. In \cite{ye2020pmbanet}, a multi-branch network with progressive refinement performs adaptive information fusion to restore depth details. \cite{wang2020depth} can quickly upsample depth maps by learning Canny edges, while \cite{zuo2020frequency} proposes a depth-guided affine transformation where the feature refinement is carried out iteratively. 
\cite{tang2021joint} makes use of implicit neural interpolation, \cite{kim2021deformable} develops a deformable kernel network whose outputs are per-pixel kernels, and \cite{zhao2022discrete} proposes a Discrete Cosine Transform Network (DCTNet) to extract multi-modal features effectively. Through graph optimization, \cite{de2022learning} combines the advantages of model-driven and deep learning-based methods. 
Concurrent works exploit recurrent structure attention \citep{yuan2023recurrent} or combine deep learning with anisotropic diffusion \citep{metzger2022guided}.

Despite substantial advancements, these networks are not effective enough at extracting HF guidance from RGB images. Inspired by \cite{liu2021multi}, this paper tackles GDSR leveraging both explicit and implicit HF features guidance. 

\textbf{Vision Transformers.}
\label{Transformer Attention Mechanism}
Transformers, initially designed for natural language processing \citep{vaswani2017attention}, recently gained popularity in computer vision, for tasks such as image recognition \citep{2021An, touvron2021training}, object detection \citep{carion2020end} and semantic segmentation \citep{wang2021pyramid}. 
Vision Transformers (ViTs) learn long-range dependencies across image tokens through self-attention \citep{han2022survey}.
Given the natural advantages of such a mechanism, ViTs targeting low-level vision tasks emerged more recently \citep{zamir2022restormer,lee2022mpvit,pu2022edter}, although requiring much larger amounts of parameters and computing resources. 

\section{\netname{} Framework}

In GDSR, HF information in color images -- complementary to depth maps -- is essential for achieving high performance, which motivates us to seek an efficient method to extract it. In this section, we present our framework that exploits explicit and implicit HF information for depth super-resolution. Then, we introduce the two branches in our network: the High-Frequency Extraction Branch (HFEB) and the Guided Depth Restoration Branch (GDRB). 

\begin{figure}	
	\centering	
	\captionsetup[subfigure]{font=footnotesize,textfont=footnotesize}
	\subfloat{	
		\centering	
		\includegraphics[width=3.2in]{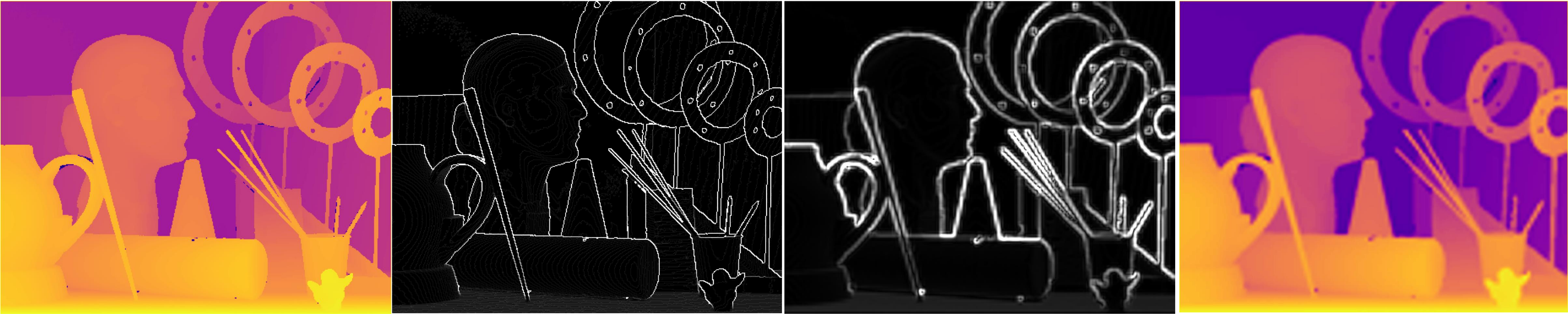}}
	\vspace{-0.cm}
	\caption{\textbf{High-frequency information loss (factor $4\times$).} From left to right, HR depth map and its corresponding gradient map, followed by the gradient map from bicubic upsampled LR depth map and LR depth map itself. HF information is mostly lost in the second gradient map.}
	\label{hf_compare}
\end{figure}

Fig.~\ref{framework} shows an overview of our architecture. Given the LR depth map $D_{lr} \in \mathbb{R}^{h\times w\times 1}$ and the corresponding HR color image $I_{hr} \in \mathbb{R}^{H\times W\times 3}$, we aim at restoring HR depth map $\tilde{D}_{sr}$. Note that $H=s\times h$ and $W=s\times w$, where $s$ denotes the upsampling factor -- e.g., $4\times, 8\times$ or even $16\times$. In our proposed network, the input depth map is firstly upsampled with bicubic interpolation to the same size as $I_{hr}$. At different scales, we denote the corresponding depth maps and color images as $D_{lr}^{i}$ and $I_{hr}^{i}$, respectively, with $s=2^i$. Then, according to the above notation, the input images $D_{lr}^{0}$ and $I_{hr}^{0}$ are fed into the two branches, respectively.  Before being sent to GDRB, both the RGB and depth images are processed by a channel-attention block (CAB) \citep{zhang2018image} and a low-cut filtering (LCF) module, which will be explained in detail in Sec. \ref{GDRB}.

\subsection{High-Frequency Extraction Branch (HFEB)}
We argue HF information is crucial for effective super-solving depth and is often lost by upsampling. The primary goal of HFEB is to produce an accurate gradient map from an LR depth map, with the support of a color image jointly processed with it. 

Indeed, as pointed out in \cite{wang2020depth}, networks for GDSR tend to focus more on depth discontinuities or object boundaries. However, from Fig.~\ref{hf_compare}, we can notice that even with a $4\times$ factor, most high-frequency information vanishes, as shown by the gradient maps extracted from HR and upsampled LR depth maps, leading to severe degradation of the super-solved depth map. Traditional methods tend to transfer texture to depth maps rather than structural details, failing to extract accurate edges. Moreover, methods extracting binary edges \citep{wang2020depth} gather insufficient high-frequency information, yielding sub-optimal results. 

\begin{figure}	
	\centering	
	\captionsetup[subfigure]{font=footnotesize,textfont=footnotesize}
	\subfloat{	
		\centering	
		\includegraphics[width=3.2in]{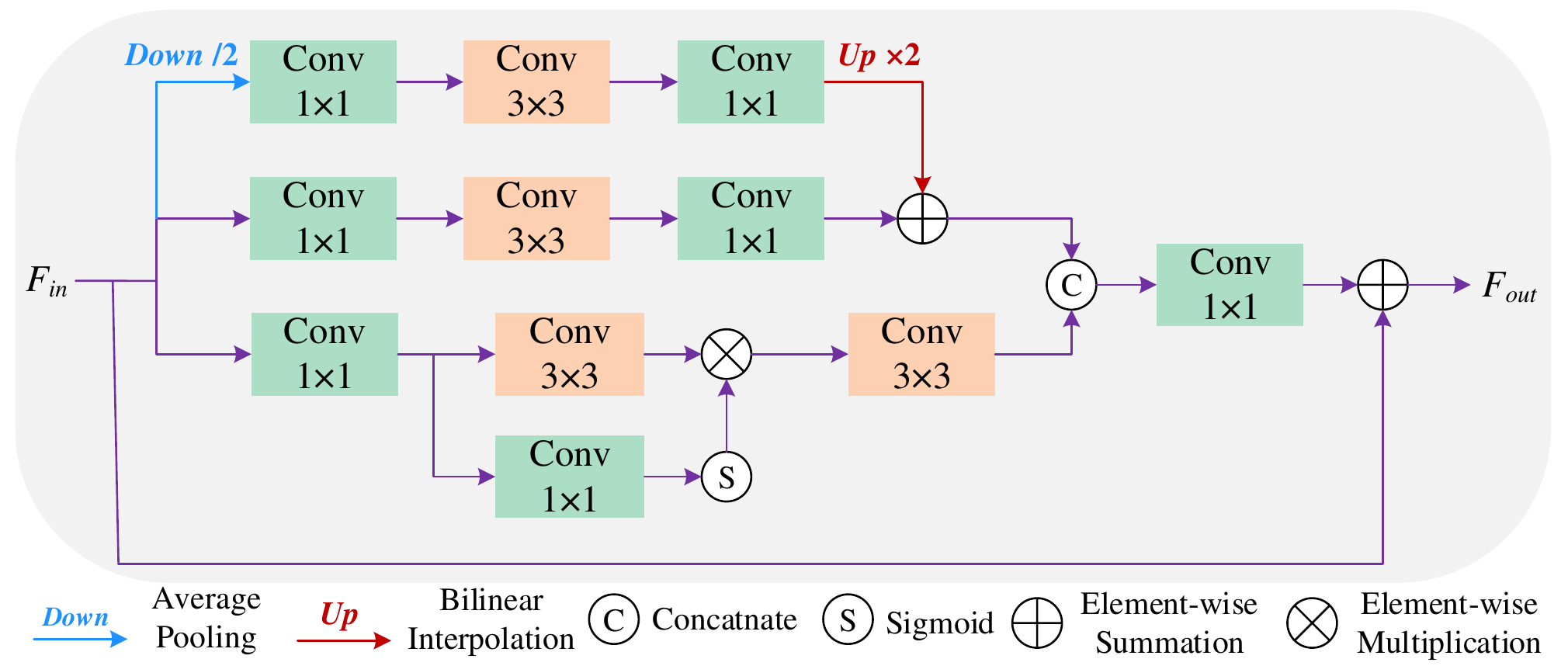}}	
    \vspace{-0.cm}		
	\caption{\textbf{DSP architecture.} Differently from SCPA \citep{zhao2020efficient}, our module processes features at different scales, allowing to extract explicit HF information more effectively.}	
	\label{dsp} 
\end{figure}

The work~\citep{pu2022edter} has shown that transformer-based networks can extract clear and meaningful edges by leveraging both global and local features simultaneously. Considering the sparsity of edge maps, we design an efficient transformer, inspired by dynamic scale policy \citep{wang2019elastic} and self-attention \citep{vaswani2017attention}, to obtain strong HF priors for guiding depth super-resolution. Specifically, our transformer consists of a stack of blocks called dynamic self-calibrated convolution with pixel attention (DSP) and one LightViT block~\citep{huang2022lightvit}. 
To better extract HF features, we design the DSP block, which is inspired by SCPA \citep{zhao2020efficient} and performs self-calibrated convolution with two branches at a single scale. However, unlike SCPA, our DSP block includes an additional branch that enables the processing of features at different scales without incurring extra computational burden, as we will demonstrate empirically in our experiments. Specifically, stacked DSP blocks can be expressed as:
\begin{equation}
    \label{eq_dsp1}
    \Phi_{M}=\mathcal{F}_{DSP}^{M}(\mathcal{F}_{DSP}^{M-1}(\cdot\cdot\cdot\mathcal{F}_{DSP}^{1}(\Phi_0)\cdot\cdot\cdot))
\end{equation}
where $\mathcal{F}_{DSP}^{m}$ denotes the mapping of the $m$-th DSP block, $m\in[1, M]$, $\Phi_0$ and $\Phi_M$ are the input/output features, respectively. As shown in Fig.~\ref{dsp}, each DSP block includes three branches: the upper is the dynamic scale branch, the middle is the flat convolution branch, and the lower is the pixel attention branch. Specifically, we employ three convolutions with $1\times1$ kernel to split the channels, which are further processed by each branch. Note that the dynamic scale branch needs to be downsampled before $1\times1$ convolution. 
{Given the input $\Phi_{m-1}$, we obtain:
\begin{gather}
    \Phi_{m-1}^{1}=Conv_{1\times1}((\Phi_{m-1})\downarrow)  \label{eq_dsp2_1}
    \\
    \Phi_{m-1}^{k}=Conv_{1\times1}(\Phi_{m-1})  \label{eq_dsp2_2}
\end{gather}
where $\Phi_{m-1}^{1}$ is the output from the upper dynamic scale branch, $k=2,3$ denotes the features of the other two branches, $Conv_{1\times1}$ is $1\times1$ convolution, and $\downarrow$ is the downsampling operation.} 
Except for the pixel attention branch, which has features with half the total channels, the other two branches process features with $\frac{1}{4}$ of the channels each. Next, the pixel attention branch obtains features through the pixel attention scheme~\citep{zhao2020efficient}. In contrast, the other two branches extract spatial information with a $3\times3$ flat convolution, followed by a $1\times1$ convolution to restore the number of channels to be the same as the pixel attention branch. Note that the dynamic scale branch needs upsampling after $1\times1$ convolution. Then, the features from the dynamic scale and the flat convolution branches can be fused by summation. After concatenation of the features followed by a $1\times1$ convolution, the DSP finally generates the output features $\Phi_{m}$ in a residual learning fashion. {It can be written as follows:
\begin{gather}
    \Phi_{m}^{1}=Conv_{1\times1}(Conv_{3\times3}(\Phi_{m-1}^{1}))\uparrow     \label{eq_dsp3_1}
    \\
    \Phi_{m}^{2}=Conv_{1\times1}(Conv_{3\times3}(\Phi_{m-1}^{2}))               \label{eq_dsp3_2}
    \\ 
    \Phi_{m}^{3}=Conv_{3\times3}(\Phi_{m-1}^{3}) \odot \sigma(Conv_{1\times1}(\Phi_{m-1}^{3}))
    \label{eq_dsp3_3}
    \\
    \Phi_{m}^{'}=Conv_{3\times3}(\Phi_{m}^{3})
    \label{eq_dsp3_5}
    \\
    \Phi_{m}^{''}=Conv_{3\times3}(\Phi_{m}^{1}\oplus\Phi_{m}^{2})  \label{eq_dsp3_4}
\end{gather}
where $\sigma$ is the sigmoid function, $\odot$ and $\oplus$ are element-wise multiplication and element-wise summation, respectively, and $\uparrow$ denotes the upsampling operation. After concatenation of the features $\Phi_{m}^{'}$ and $\Phi_{m}^{''}$ followed by a $1\times1$ convolution, the DSP finally generates the output features $\Phi_{m}$ in a residual learning manner. This process can be expressed as follows:
\begin{equation}
    \label{eq_dsp1_4}
    \Phi_{m}=Conv_{1\times1}([\Phi_{m}^{'}, \Phi_{m}^{''}]) \oplus \Phi_{m-1}
\end{equation}
where $[\cdot]$ perform concatenation.}

To further enhance the feature representation of the subnetwork, we incorporate LightViT \citep{huang2022lightvit} as the tail module, which utilizes local-global attention broadcast to aggregate information from all tokens, allowing for the efficient integration of global dependencies of local tokens into each image token. Finally, considering that the supervised attention module (SAM) \citep{zamir2021multi} can restore information progressively between stages/branches, we employ it to output the gradient map $E\in \mathbb{R}^{H\times W\times 1}$ and high-frequency features $F_{edge}\in \mathbb{R}^{H\times W\times C}$, used respectively as intermediate output -- allowing for explicit supervision over edges -- and as guidance for GDRB. Under this lightweight design, HFEB can effectively still extract meaningful structural information with different scale receptive fields.

\subsection{Guided Depth Restoration Branch (GDRB)}
\label{GDRB}
As shown in Fig.~\ref{framework}, GDRB is composed of two stages, and each one processes features at three scales, following a coarse-to-fine strategy \citep{gao2019dynamic,sarlin2019coarse}. The two stages are implemented with standard U-net architectures~\citep{ronneberger2015u}. More specifically, a cross-stage feature fusion module~\citep{zamir2021multi} is deployed between the two, which proved to be effective in image restoration and, in our design, allows GDRB to benefit from the intermediate features extracted by HFEB. To prevent aliasing in downsampling, we employ content-aware filtering layers~\citep{zou2022delving} in the encoders. Besides, GDRB deploys some further SAM blocks \citep{zamir2021multi}, allowing valuable features to propagate to the next stage. In addition to depth features, the SAMs of the two stages also output depth maps $\tilde{D}_{sr}^{'}$ and $\tilde{D}_{sr}^{''}$, to which intermediate supervision is provided. Note that input images are downsampled to the lower stage using pixel unshuffling to prevent information loss. Subsequently, the depth map output of this stage is restored at high resolution by employing pixel shuffling.

Based on the above structure, we propose two novel modules: AFFM and LCF. The former fuses gradient features between each encoder/decoder, while the latter supplements additional HF information in an implicit manner. 

\begin{figure}	
	\centering	
	\captionsetup[subfigure]{font=footnotesize,textfont=footnotesize}
	\subfloat{	
		\centering	
		\includegraphics[width=2.9in]{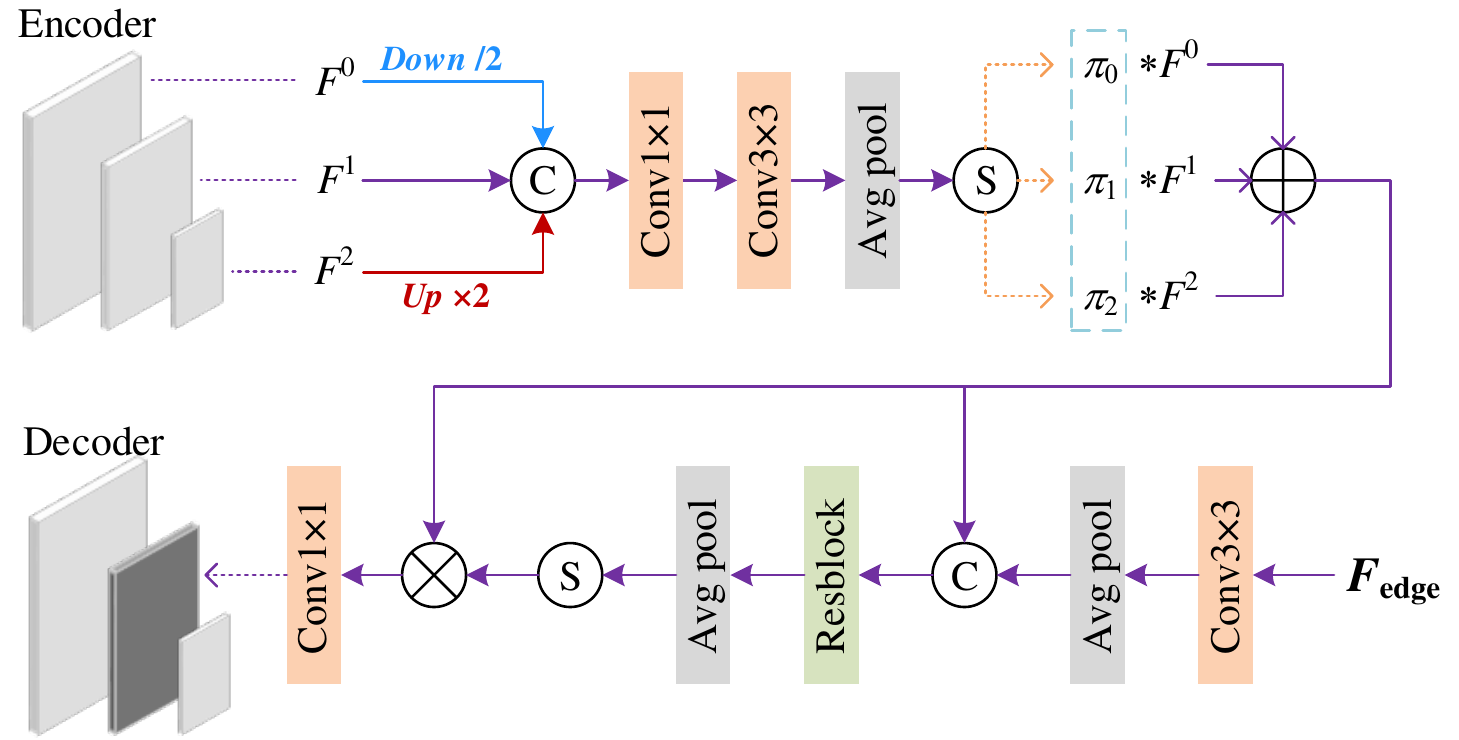}}
	\vspace{-0.cm}
	\caption{\textbf{AFFM architecture, operating at middle scale.} AFFMs for the remaining scales follow the same design.}
	\label{affm} 
\end{figure}

\textbf{Adaptive feature fusion module.} Recent networks such as \cite{ye2020pmbanet,tang2021joint} typically concatenate RGB and depth features directly during feature fusion,  followed by additional operations such as channel attention~\citep{zhang2018image} to capture useful information. In contrast, inspired by \cite{liu2021multi}, we run adaptive feature fusion through AFFM in two steps to strengthen the reconstruction of HF cues, as illustrated in Fig.~\ref{affm}. We differentiate from \cite{liu2021multi} by using dynamic convolution~\citep{chen2020dynamic} to better aggregate depth and HR features. In the first step, we generate dynamic weights $\pi_i, i=0,1,2$, which are then assigned to features from different scales within the current stage. Finally, we perform element-wise summation to obtain the feature maps $F^{'}$. For clarity, the figure shows the module working at the middle scale of the network as an example, with the others sharing the same design. 
{The process is defined as follows:
\begin{gather}
    F_{cat}=Conv_{3\times3}(Conv_{1\times1}([F^0\downarrow,F^1,F^2\uparrow]))       \label{eq_affm1_0}
    \\
    \{\pi_0,\pi_1,\pi_2\}=\sigma(Avgpool(F_{cat})) \label{eq_affm1_1}
    \\
    F^{'}=\pi_0\cdot F^0\downarrow+\pi_1\cdot F^1+\pi_2\cdot F^2\uparrow  \label{eq_affm1_2}
\end{gather}
where $F^{i}, i=0,1,2$ denotes the feature maps from the three scales, and $\downarrow$, $\uparrow$ are respectively downsampling and upsampling operators.}

In the second step, gradient features $F_{edge}$ from HFEB are concatenated with $F^{'}$. Then, per-pixel attention maps $F_{att}$ are generated by a ResBlock \citep{he2016deep} followed by an average pooling operation. These attention maps are then applied directly to the adaptively fused features $F^{'}$ through element-wise multiplication operation. Finally, after $1\times 1$ convolution, the attention-guided features $F_{out}^{i}$ are delivered to the corresponding scale of the current stage. In Fig. \ref{affm}, the output is passed to the middle scale of the decoder. AFFMs working at the other scales send their output to the corresponding scale in the decoder.  
{This step can be formalized as follows:
\begin{gather}
    F^{''}=[Avgpool(Conv_{3\times3}(F_{edge})),F^{'}]  \label{eq_affm2_0}
    \\
    F_{att}=\sigma(Avgpool(ResBlock(F^{''}))) \label{eq_affm2_1}
    \\
    F_{out}^{1}=con_{1\times1}(F^{'}\otimes F_{att})  \label{eq_affm2_2}
\end{gather}
where $\otimes$ is an element-wise multiplication operation and $F_{out}^{1}$ denotes the output features at the middle scale.}

\begin{figure}	
	\centering	
    \includegraphics[width=0.45\textwidth]{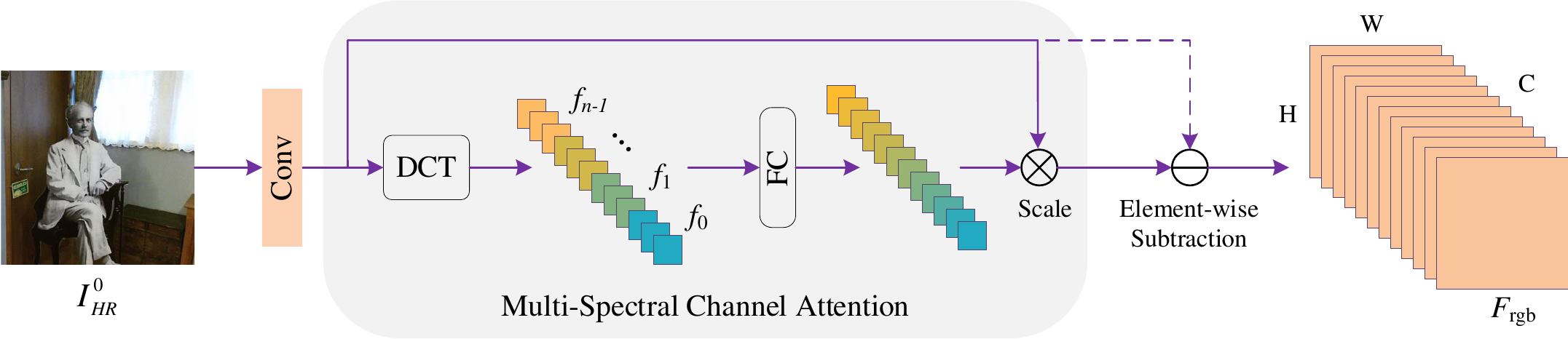}
    \vspace{-0.cm}
	\caption{\textbf{Low-cut filtering module (LCF).} LF features are extracted through DCT and multi-spectral channel attention, and subtracted from the input to retain HF features.}	
	\label{lcf}
\end{figure}

\textbf{Low-cut filtering module.} 
The performance of our method greatly benefits from the explicit gradient information, but some valuable high-frequency information still vanishes. This fact motivates us to consider extracting complementary information in the frequency domain. As a common practice \citep{chang2007reversible,lin2010improving}, we use the low-frequency information of the discrete cosine transform (DCT) to compress images. Based on the design approach proposed in \cite{qin2021fcanet}, we develop a filtering module utilizing feature decomposition in the frequency domain to extract low-frequency components from the input. Specifically, we apply a $1\times1$ convolution followed by a channel split to the input color image $I_{hr}^{0}$. Then, we can obtain assigned frequency components from the output features $[{f}_0, {f}_1, \cdot\cdot\cdot, {f}_{n-1}]$ after DCT. Thus, the multi-spectral channel attention maps are generated by a fully connected layer and sigmoid activation. According to \cite{qin2021fcanet}, the low-frequency information is first assured to pass. Thus, we subtract such a low-frequency component from the input features producing the complementary high-frequency features $F_{rgb}$. Fig.~\ref{lcf} illustrates LCF in detail.  The high-frequency cues extracted from these features enable GDRB to progressively super-resolve LR depth maps into HR ones.

\textbf{Refinement.} To enhance the depth quality further, we optionally feed our final output into NLSPN \citep{park2020non} for refinement. This variant of the method is referred to as \netname$^+$.

\subsection{Training Loss}
Our network is trained in an end-to-end fashion using two loss terms: depth loss $L_d$ and gradient loss $L_g$. The depth loss is defined as:
\begin{equation}
\begin{aligned}
\label{lossd}
    L_d\!=\!\parallel\!(\!\tilde{D}_{sr}\!-\!D_{gt})\!\odot\!\mathbb{I}\!\parallel_1\!&+\lambda_d\cdot\!\parallel\!(\!\tilde{D}_{sr}^{'}\!-\!D_{gt})\!\odot\!\mathbb{I}\!\parallel_1\! \\
        &+\lambda_d\cdot\!\parallel\!(\!\tilde{D}_{sr}^{''}\!-\!D_{gt})\!\odot\!\mathbb{I}\!\parallel_1
\end{aligned}
\end{equation}
where $D_{gt}$ is the ground truth depth, $\tilde{D}_{sr}$, $\tilde{D}_{sr}^{'}$ and $\tilde{D}_{sr}^{''}$ are predicted depth maps from different stages, and $\mathbb{I}$ is pixel validity, as defined in \cite{de2022learning}. We empirically set $\lambda_d=0.2$.
Gradient loss $L_g$ is computed on HEFB output, as:
\begin{equation}
    L_g=\parallel\tilde{E}-E_{gt}\parallel_1
\end{equation}
where $\tilde{E}$ is the predicted gradient map and $E_{gt}$ is the ground truth one, extracted according to \cite{liu2021multi}. Thus, the total loss can be defined as:
\begin{equation}
    L_{total}=L_d+\lambda_g\cdot L_g
\end{equation}
with $\lambda_g$ empirically set to 0.01.
\begin{table*}[t] \scriptsize
	\renewcommand\tabcolsep{8pt} 
	\centering
        \caption{\textbf{Results on Middlebury, NYUv2 and DIML datasets.} The lower the MSE and MAE, the better.}
	\begin{tabular}{@{}lccccccccc@{}}
		\toprule
		  \textbf{Dataset} & \multicolumn{3}{c}{\textbf{Middlebury}} & \multicolumn{3}{c}{\textbf{NYUv2}} & \multicolumn{3}{c}{\textbf{DIML}} \\
            \cmidrule(r){2-4} \cmidrule(r){5-7} \cmidrule(r){8-10}
            \textbf{Methods} & $4\times$ & $8\times$ & $16\times$ & $4\times$ & $8\times$ & $16\times$ &  $4\times$ & $8\times$ & $16\times$    \\
            \midrule
		GF~\citep{he2010guided}          & 33.3 / 1.27 & 40.5 / 1.49 & 67.4 / 2.21 & 114 / 3.91 & 142 / 4.47 & 249 / 6.34 & 25.6 / 1.45 & 34.1 / 1.77 & 66.3 / 2.74   \\
		SD~\citep{ham2017robust}         & 24.9 / 0.46 & 82.5 / 0.86 & 511 / 1.73 & 36.0 / 1.31 & 105 / 2.57 & 533 / 5.07 & 10.5 / 0.40 & 44.9 / 0.83 & 41.1 / 1.91    \\
		P2P~\citep{lutio2019guided}    & 39.8 / 0.79 & 32.7 / 0.82 & 41.5 / 1.24 & 112 / 3.61 & 122 / 3.86 & 219 / 5.40 & 20.7 / 1.15 & 23.0 / 1.26 & 39.3 / 1.78    \\
		MSG~\citep{hui2016depth}         & 4.13 / 0.22 & 10.5 / 0.43 & 34.2 / 1.06 & 6.85 / 0.81 & 24.1 / 1.66 & 84.5 / 3.35 & 1.73 / 0.22 & 4.13 / 0.40 & 13.0 / 0.93  \\
		DKN~\citep{kim2021deformable}    & 4.29 / 0.18 & 11.2 / 0.38 & 47.6 / 1.42 & 11.4 / 1.03 & 29.8 / 1.82 & 115 / 4.01 & 3.47 / 0.33 & 5.47 / 0.45 & 19.3 / 1.20  \\
		FDKN~\citep{kim2021deformable}   & 3.60 / 0.16 & 10.4 / 0.37 & 38.5 / 1.18 & 9.07 / 0.85 & 29.9 / 1.80 & 113 / 3.95 & 2.20 / 0.23 & 5.95 / 0.47 & 20.8 / 1.24  \\
		PMBANet~\citep{ye2020pmbanet}    & 4.72 / 0.25 & 9.48 / 0.38 & 30.6 / 0.89 & 10.8 / 0.93 & 17.2 / 1.38 & 84.9 / 3.26 & 3.05 / 0.31 & 5.87 / 0.47 & 13.8 / 0.87    \\
		  FDSR~\citep{he2021towards}       & 7.72 / 0.35 & 23.2 / 0.69 & 55.4 / 1.51 & 10.1 / 0.94 & 19.5 / 1.38 & 86.4 / 3.35 & 2.75 / 0.29 & 8.40 / 0.66 & 32.9 / 1.66    \\
            JIIF~\citep{tang2021joint}       & 2.70 / 0.11 & 8.01 / 0.27 & 37.5 / 0.98 & 3.28 / 0.52 & 15.2 / 1.29 & 59.9 / 2.81 & 1.19 / 0.16 & 3.65 / 0.32 & 11.7 / 0.81    \\
		DCTNet~\citep{zhao2022discrete}  & 5.00 / 0.24 & 15.1 / 0.57 & 52.3 / 1.50 & 3.63 / 0.68 & 20.9 / 1.79 & 77.0 / 3.61 & 2.09 / 0.31 & 7.08 / 0.65 & 23.4 / 1.75    \\
		LGR~\citep{de2022learning}         & 3.04 / 0.13 & 7.26 / 0.24 & 24.7 / 0.67 & 6.45 / 0.73 & 19.6 / 1.42 & 67.5 / 2.90 & 1.68 / 0.20 & 3.51 / 0.31 & 9.45 / 0.68    \\
		DADA~\citep{metzger2022guided}     & 2.58 / 0.11 & \underline{5.68} / 0.20 & \underline{16.3} / 0.48 & 4.83 / 0.64 & 16.6 / 1.30 & 59.0 / 2.64 & 1.33 / 0.17 & 2.93 / 0.28 & 7.61 / 0.59    \\
		DSR-EI                            & \textbf{2.46} / \underline{0.08} & 6.20 / \textbf{0.18} & \textbf{15.8} / \underline{0.47} & \underline{2.82} / \underline{0.49} & \textbf{11.8} / \underline{1.12} & \underline{47.8} / \underline{2.48} & \underline{0.70} / \underline{0.13} & \underline{2.12} / \textbf{0.22} & \textbf{6.29} / \underline{0.52}    \\
		DSR-EI$^+$                        & \underline{2.56} / \textbf{0.07} & \textbf{5.13} / \textbf{0.18} & 16.6 / \textbf{0.40} & \textbf{2.75} / \textbf{0.47} & \textbf{11.8} / \textbf{1.09} & \textbf{47.14} / \textbf{2.40} & \textbf{0.65} / \textbf{0.12} & \textbf{2.09} / \textbf{0.22} & \underline{6.31} / \textbf{0.50}    \\
            \bottomrule
	\end{tabular}
    \vspace{-0.3cm}
	\label{sota_comparison_mid_nyu_diml}
\end{table*}

\section{Experimental Results}
In this section, we validate the effectiveness of our proposal. We first introduce datasets, metrics and implementation details involved in our evaluation. Then, we compare \netname{} with state-of-the-art methods, conduct an ablation study on our model and, finally, discuss its limitations.

\begin{table}[t] \scriptsize
	\renewcommand\tabcolsep{14pt} 
	\centering
        \caption{\textbf{Results on RGBDD dataset.} We report RMSE, the lower the better.}
	\begin{tabular}{@{}lccc@{}}
	\toprule                
	\textbf{Methods} & $4\times$ & $8\times$ & $16\times$  \\ \midrule
        SDF~\citep{li2016deep}             & 2.00 & 3.23 & 5.16  \\
        SVLRM~\citep{pan2019spatially}     & 3.39 & 5.59 & 8.28  \\
        DJF~\citep{li2016deep}             & 3.41 & 5.57 & 8.15  \\
        DJFR~\citep{li2019joint}           & 3.35 & 5.57 & 7.99  \\
        PAC~\citep{su2019pixel}            & 1.25 & 1.98 & 3.49  \\
        CUNet~\citep{deng2020deep}         & 1.18 & 1.95 & 3.45  \\
        DKN~\citep{kim2021deformable}      & 1.30 & 1.96 & 3.42  \\
        FDKN~\citep{kim2021deformable}     & 1.18 & 1.91 & 3.41  \\
        FDSR~\citep{he2021towards}         & 1.16 & 1.82 & 3.06  \\
        DCTNet~\citep{zhao2022discrete}    & 1.07 & 1.78 & 3.18  \\
        RSAG~\citep{yuan2023recurrent}     & 1.14 & 1.75 & 2.96  \\
        DSR-EI                              & \textbf{0.91} & \textbf{1.37} & \textbf{2.10}  \\
        DSR-EI$^+$                          & \textbf{0.91} & \underline{1.38} & \textbf{2.10}  \\
    \bottomrule
	\end{tabular}
	\vspace{-0.3cm}
	\label{sota_comparison_rgbdd}
\end{table}

\begin{figure*}[t] 
	\centering
	\renewcommand\tabcolsep{1.5pt} 
	\begin{tabular}{cccccccccccc}
	\vspace{-0.1cm}
        \rotatebox[origin=l]{90}{\scriptsize \quad \textbf{4$\times$}} & \includegraphics[height=0.65in]{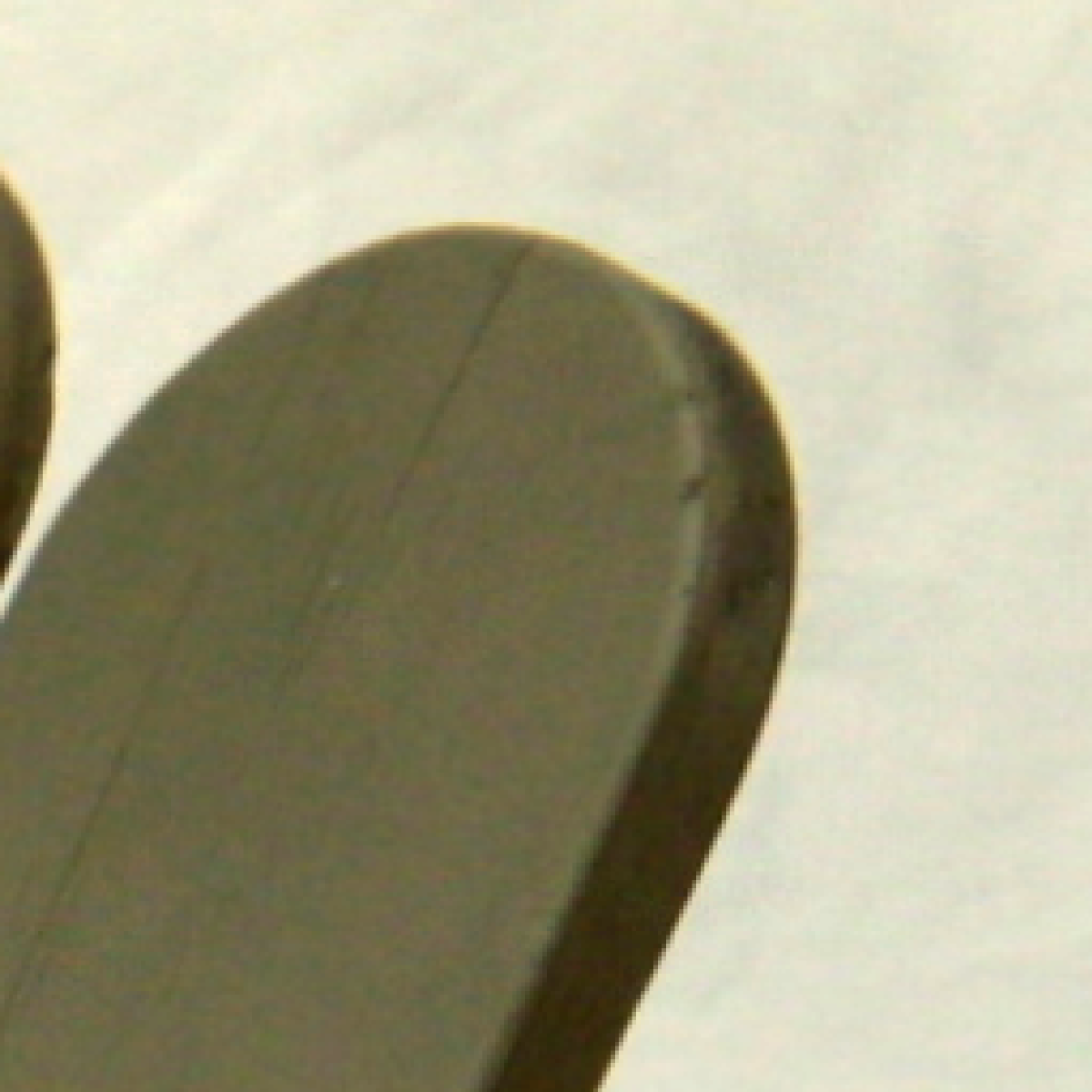}
        \hspace{-1.8mm} & \includegraphics[height=0.65in]{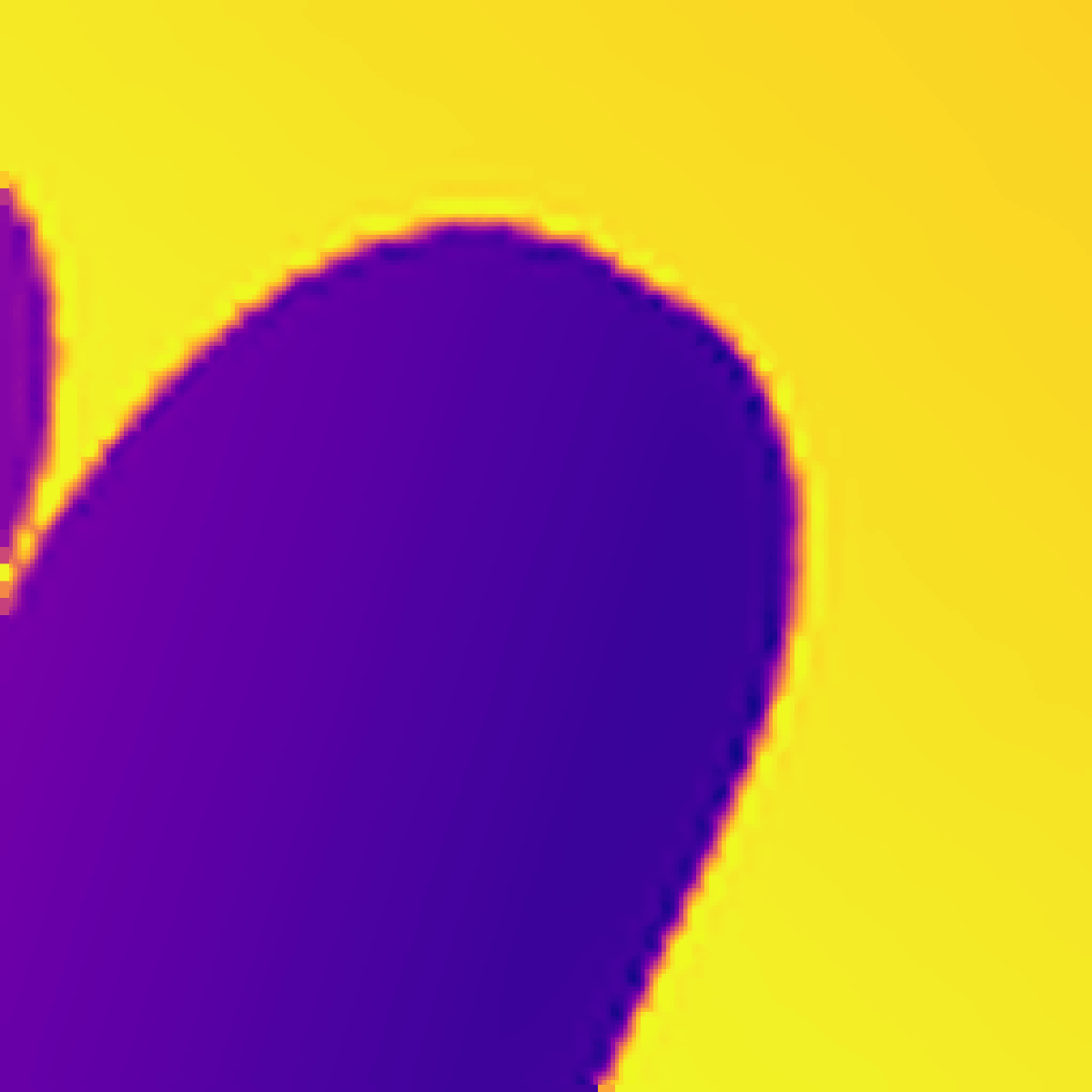}
	\hspace{-1.8mm} &  \includegraphics[height=0.65in]{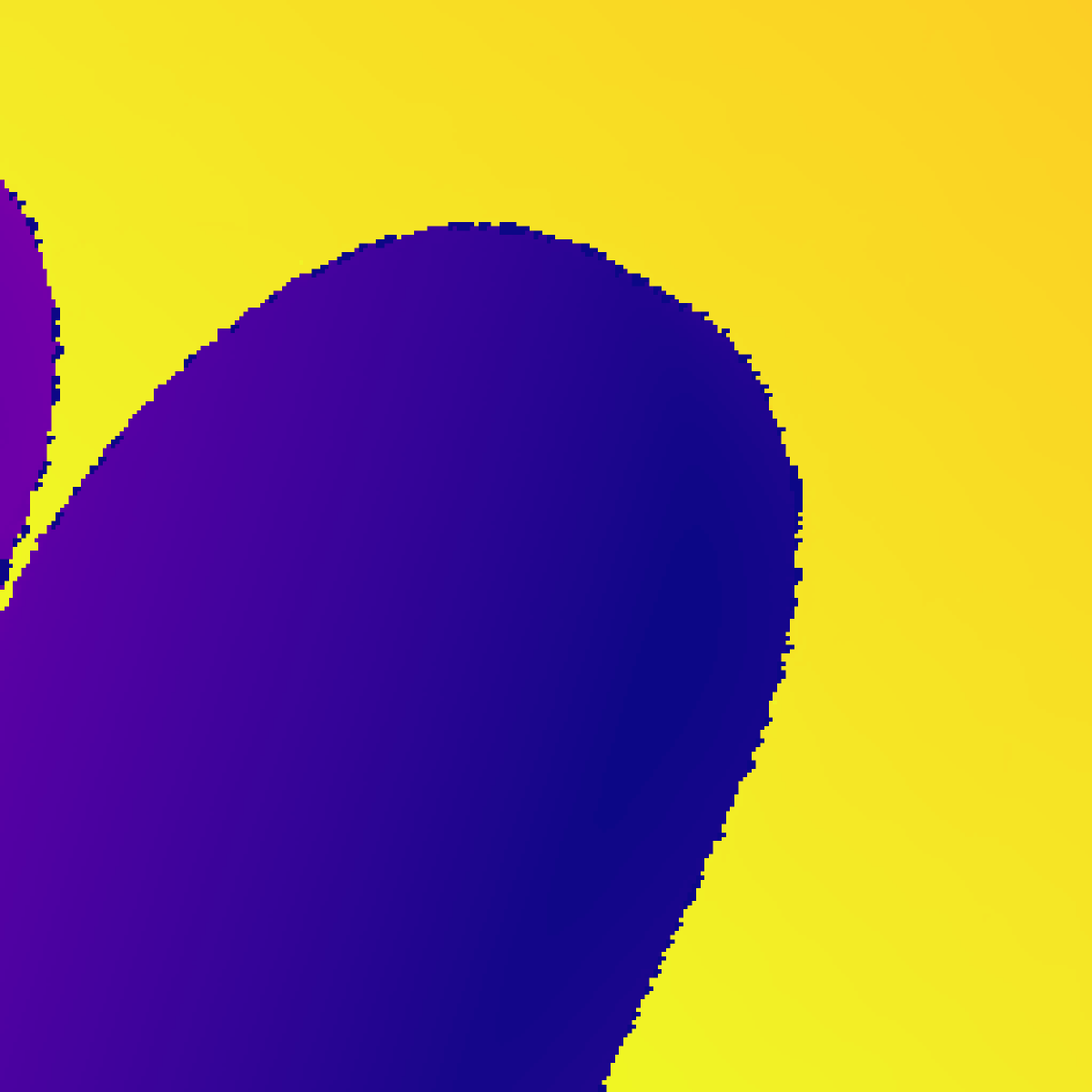}
	\hspace{-1.8mm} & \includegraphics[height=0.65in]{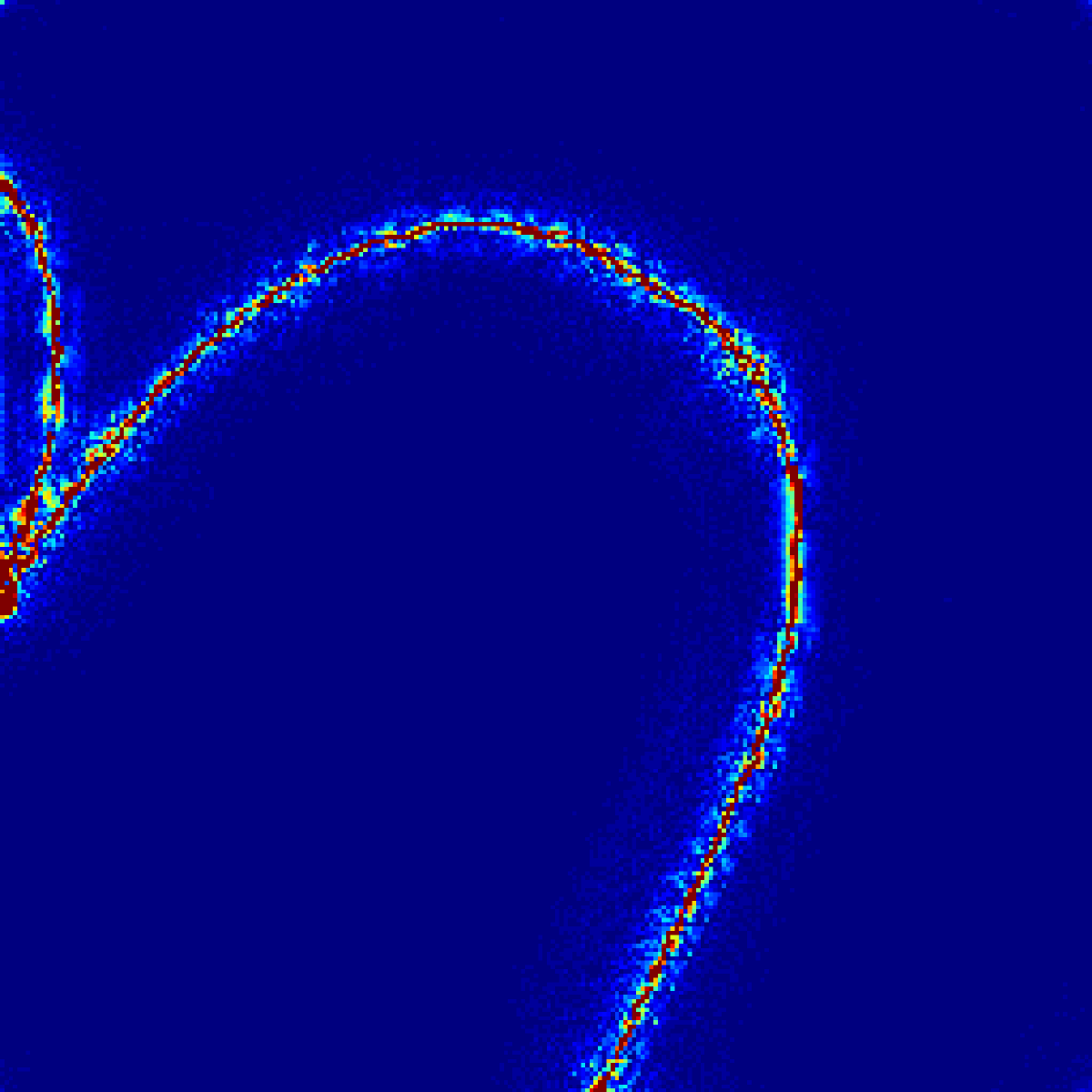}
	\hspace{-1.8mm} & \includegraphics[height=0.65in]{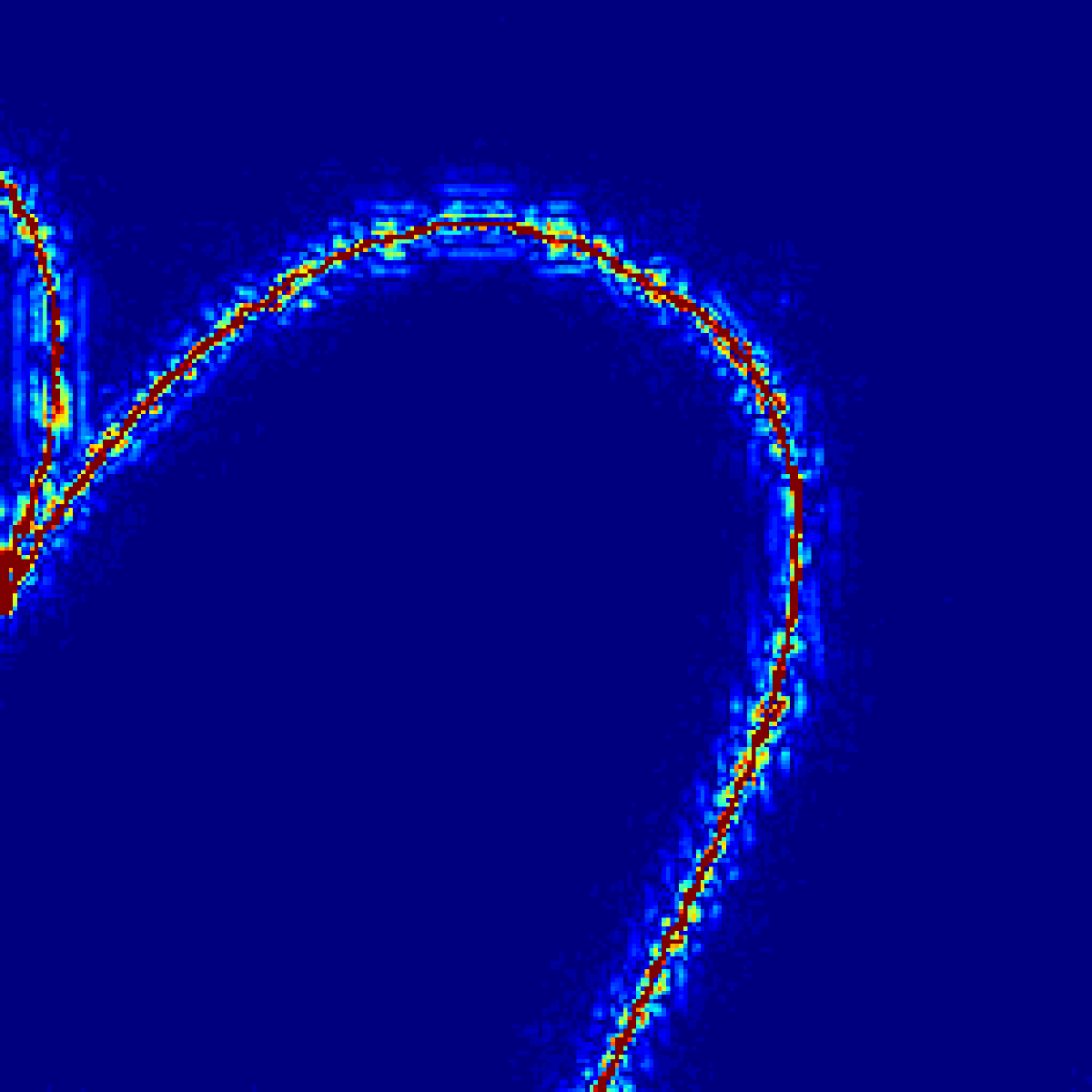}
	\hspace{-1.8mm} & \includegraphics[height=0.65in]{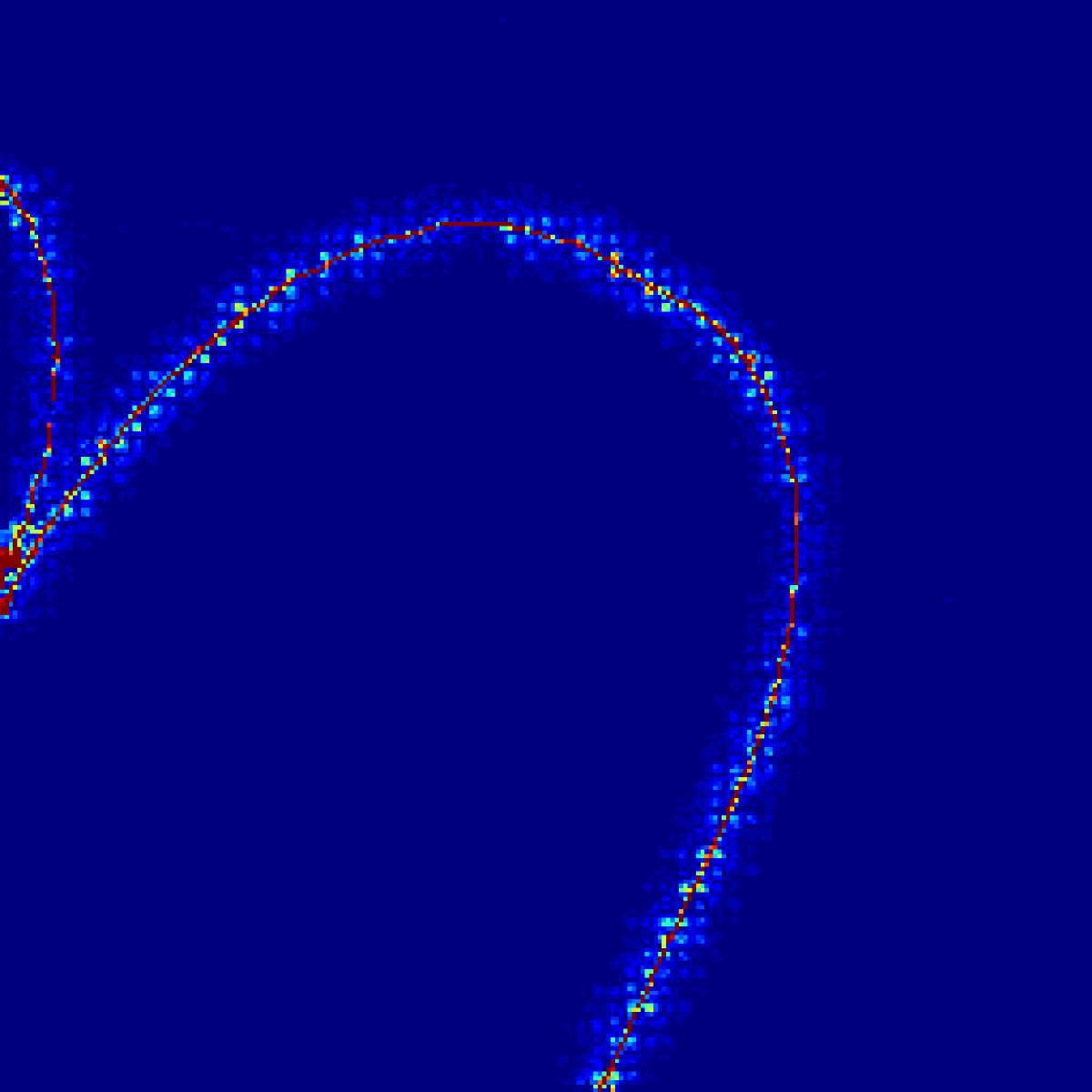}
	\hspace{-1.8mm} & \includegraphics[height=0.65in]{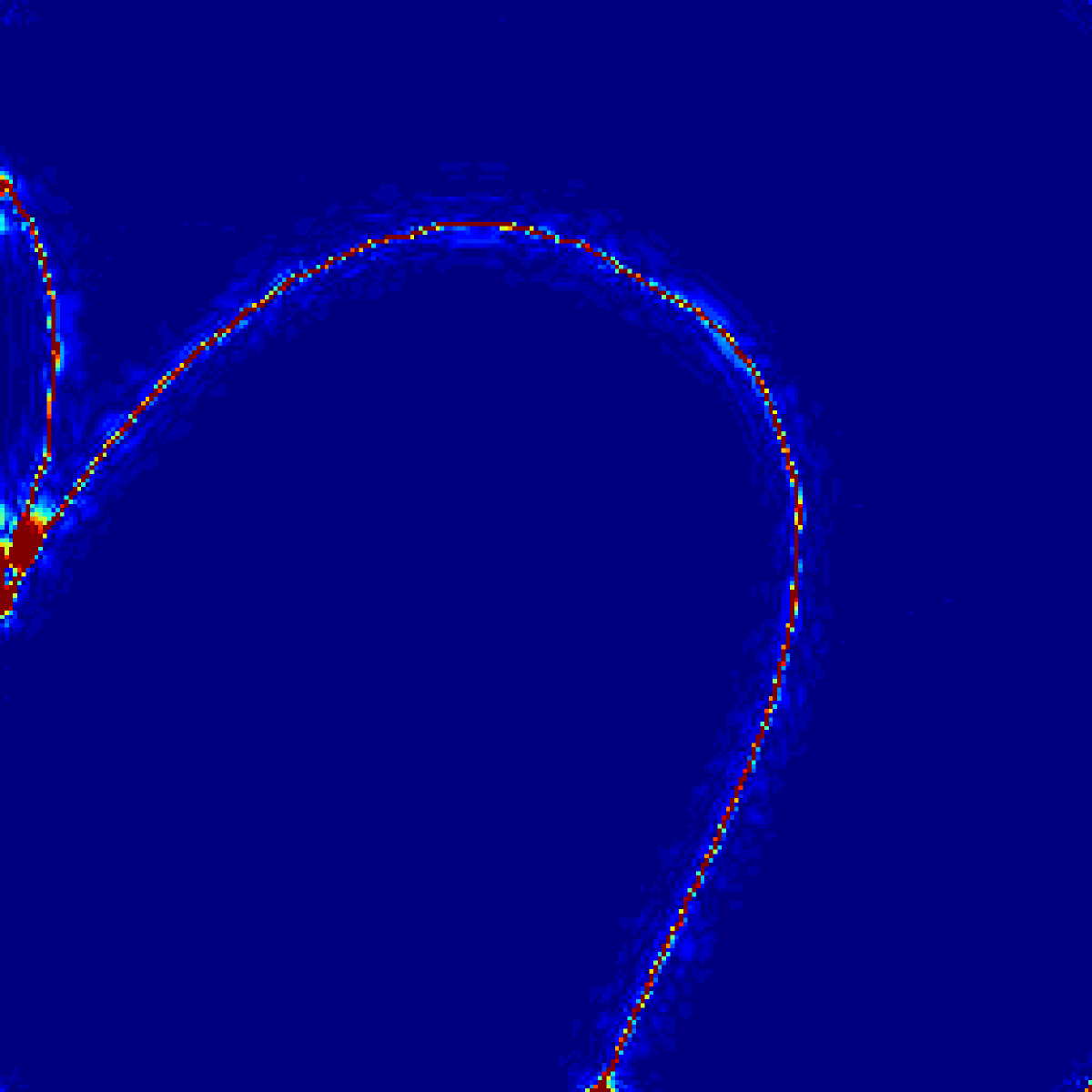}
	\hspace{-1.8mm} & \includegraphics[height=0.65in]{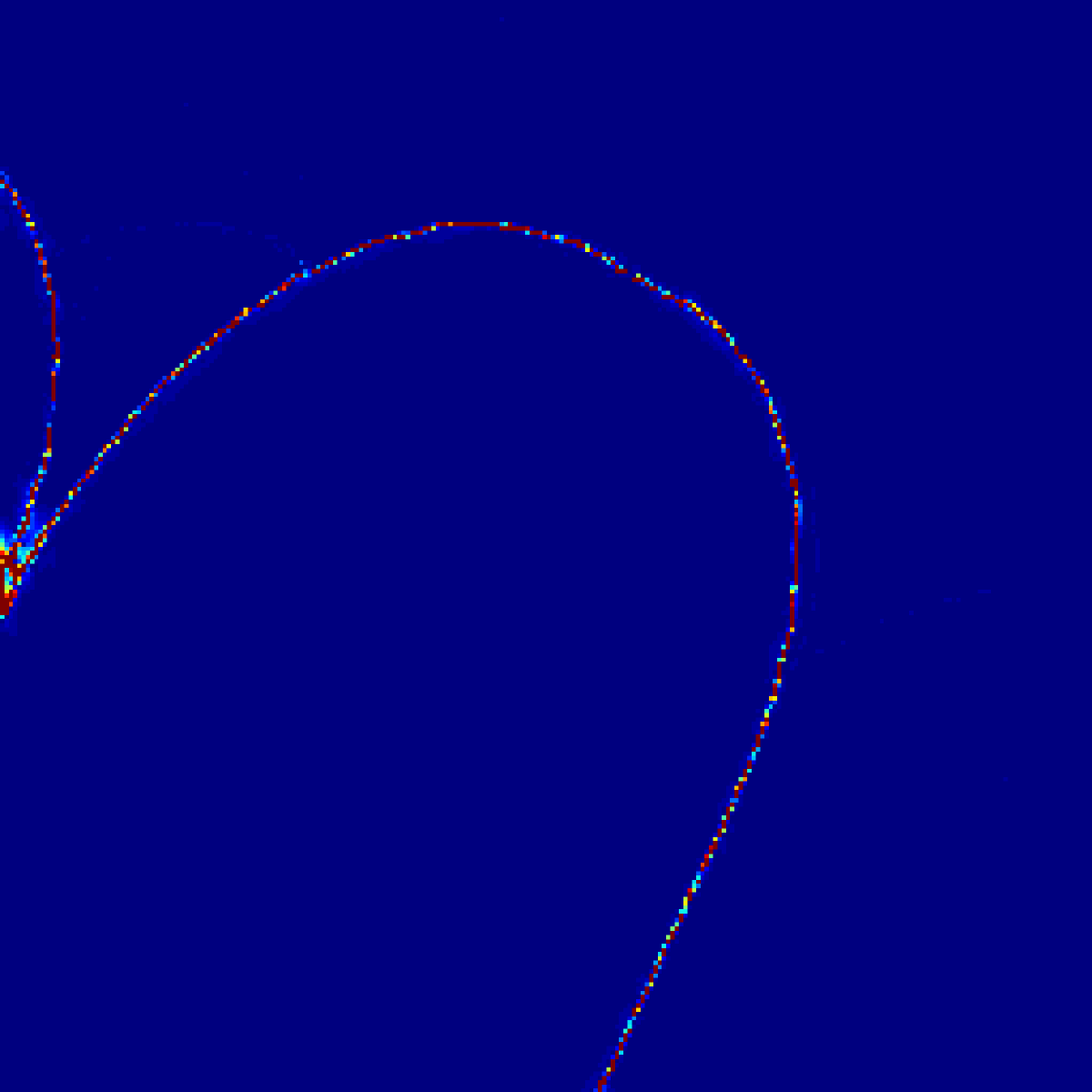}
	\hspace{-1.8mm} & \includegraphics[height=0.65in]{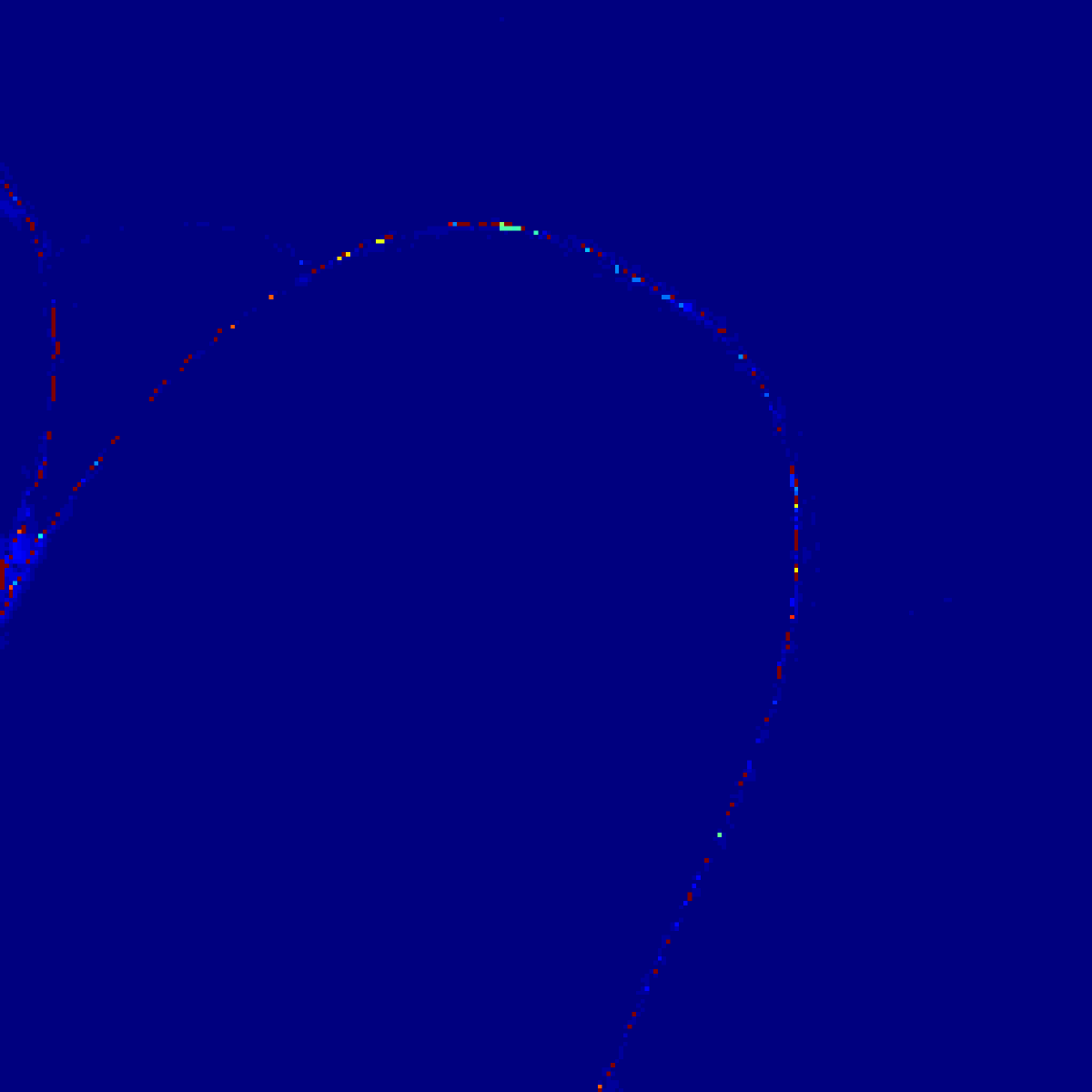}
        
        \hspace{-1.8mm} & \includegraphics[height=0.65in]{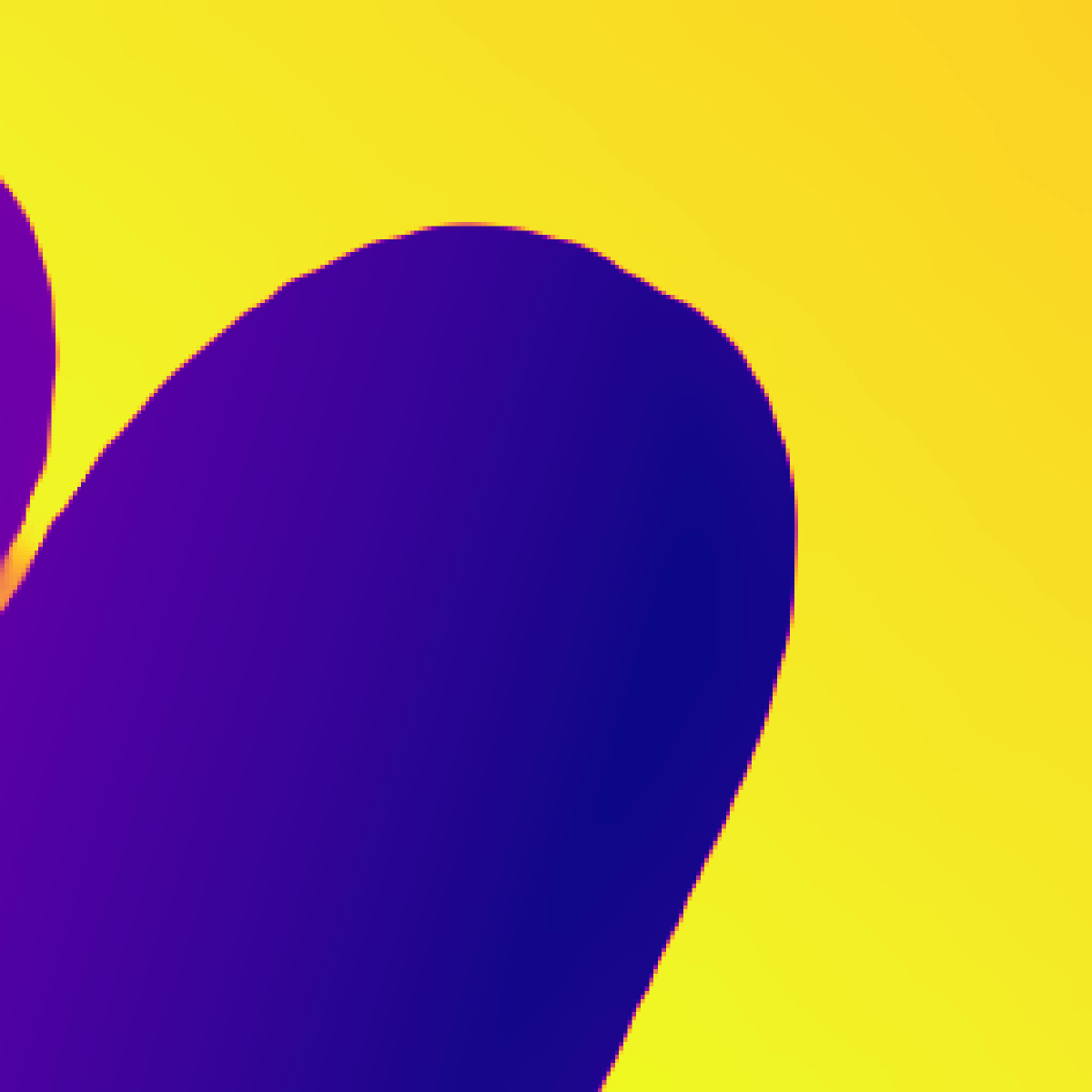}
    \\ \vspace{-0.cm}
    
        \rotatebox[origin=l]{90}{\scriptsize \quad \textbf{Middlebury}} & \includegraphics[height=0.65in]{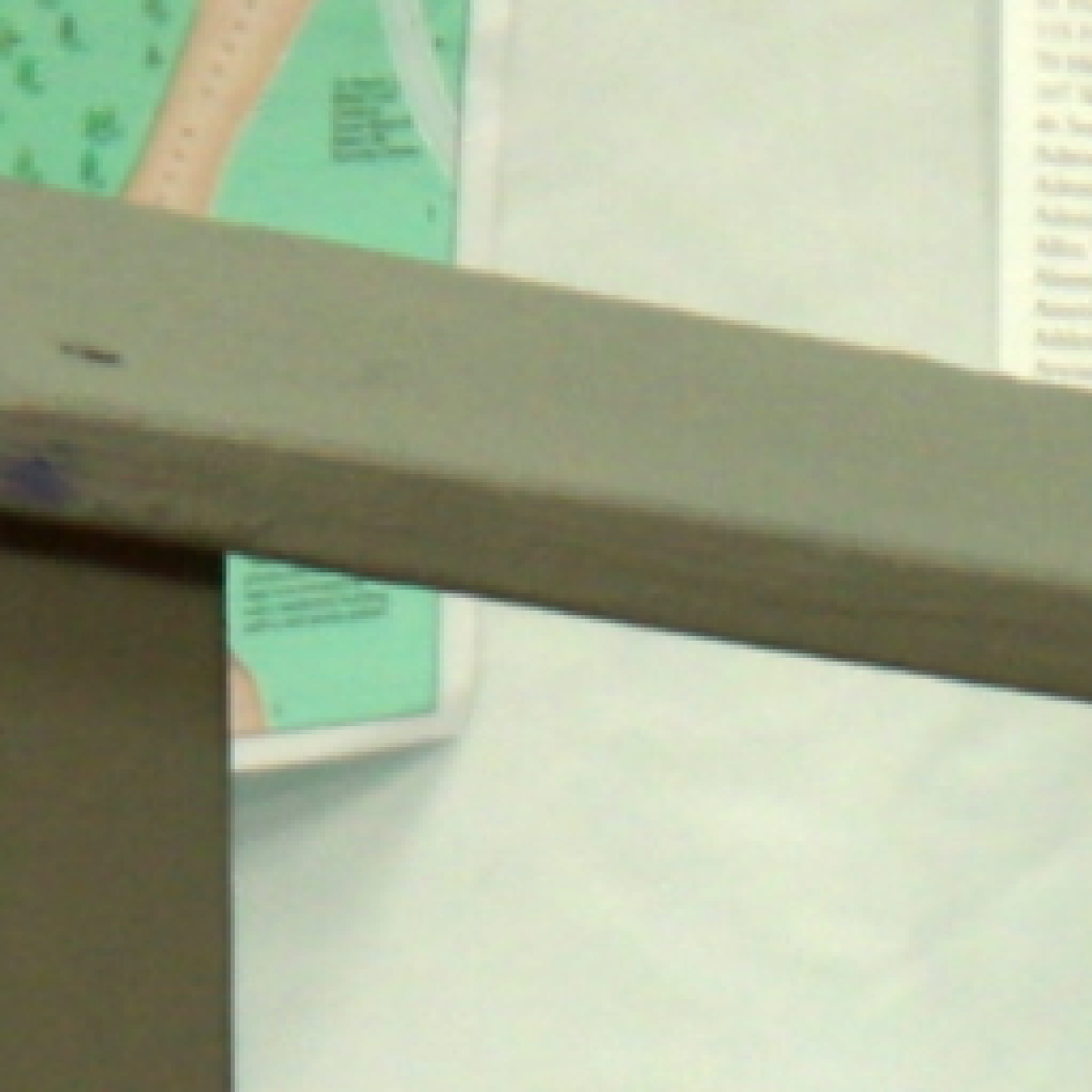}
        \hspace{-1.8mm} & \includegraphics[height=0.65in]{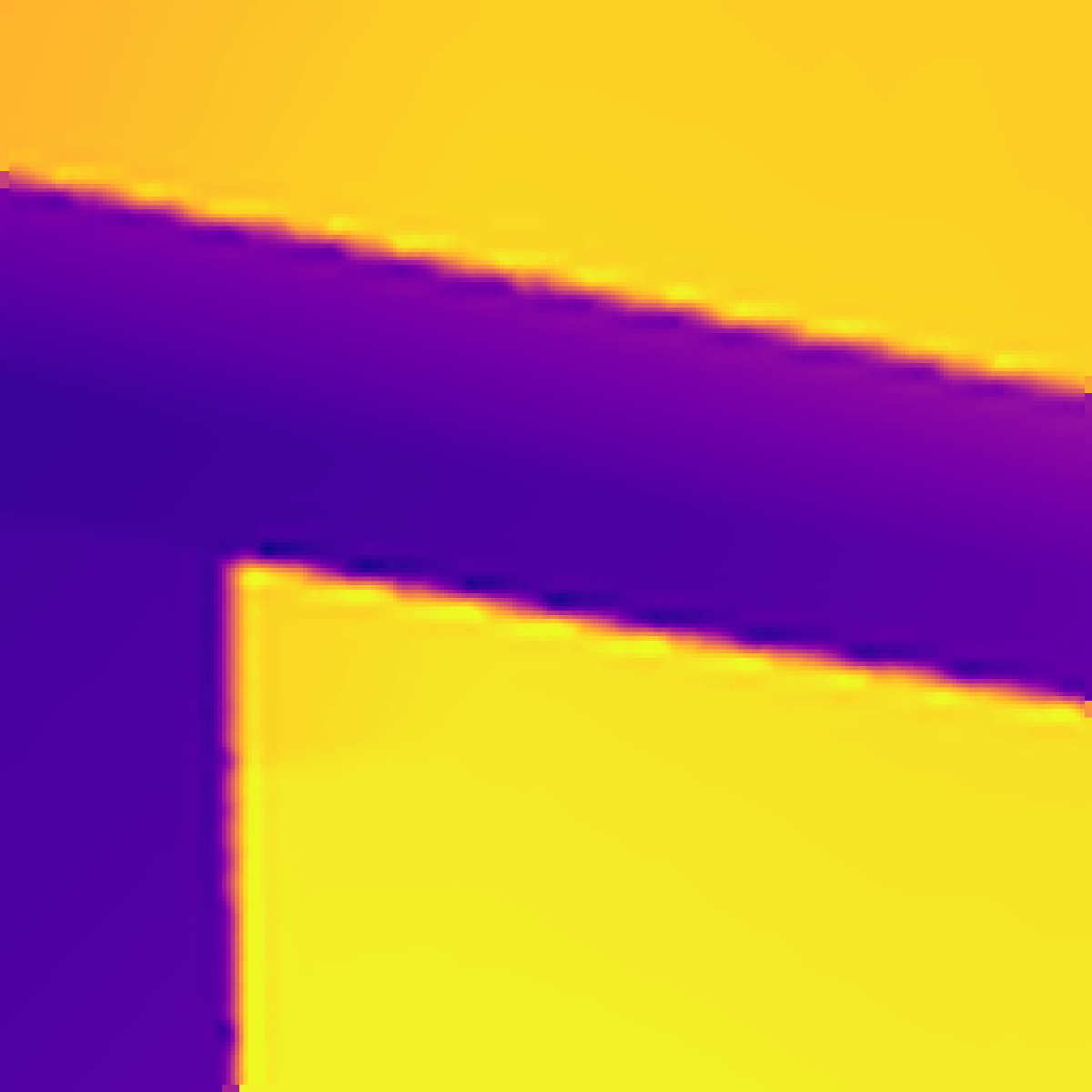}
	\hspace{-1.8mm} &  \includegraphics[height=0.65in]{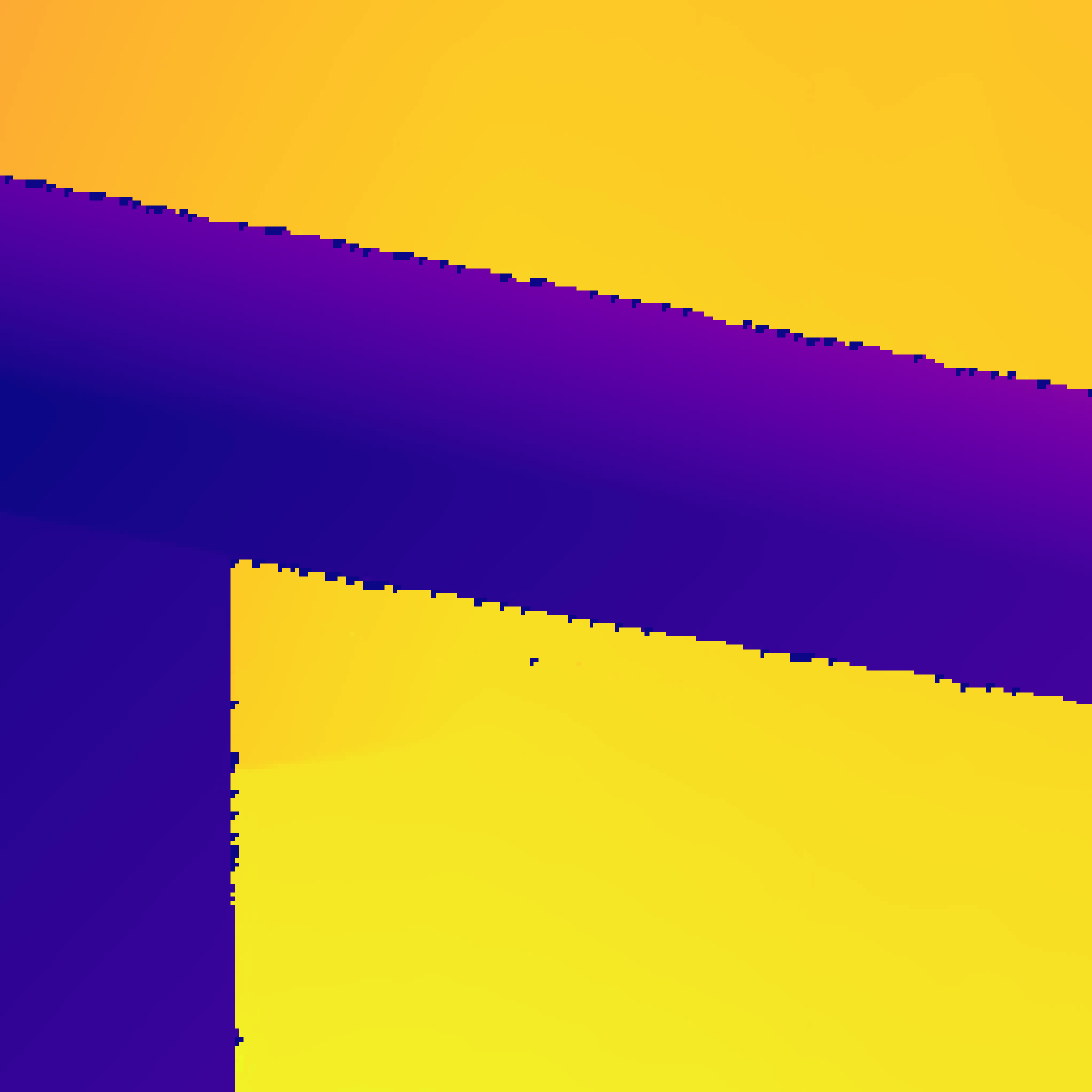}
	\hspace{-1.8mm} & \includegraphics[height=0.65in]{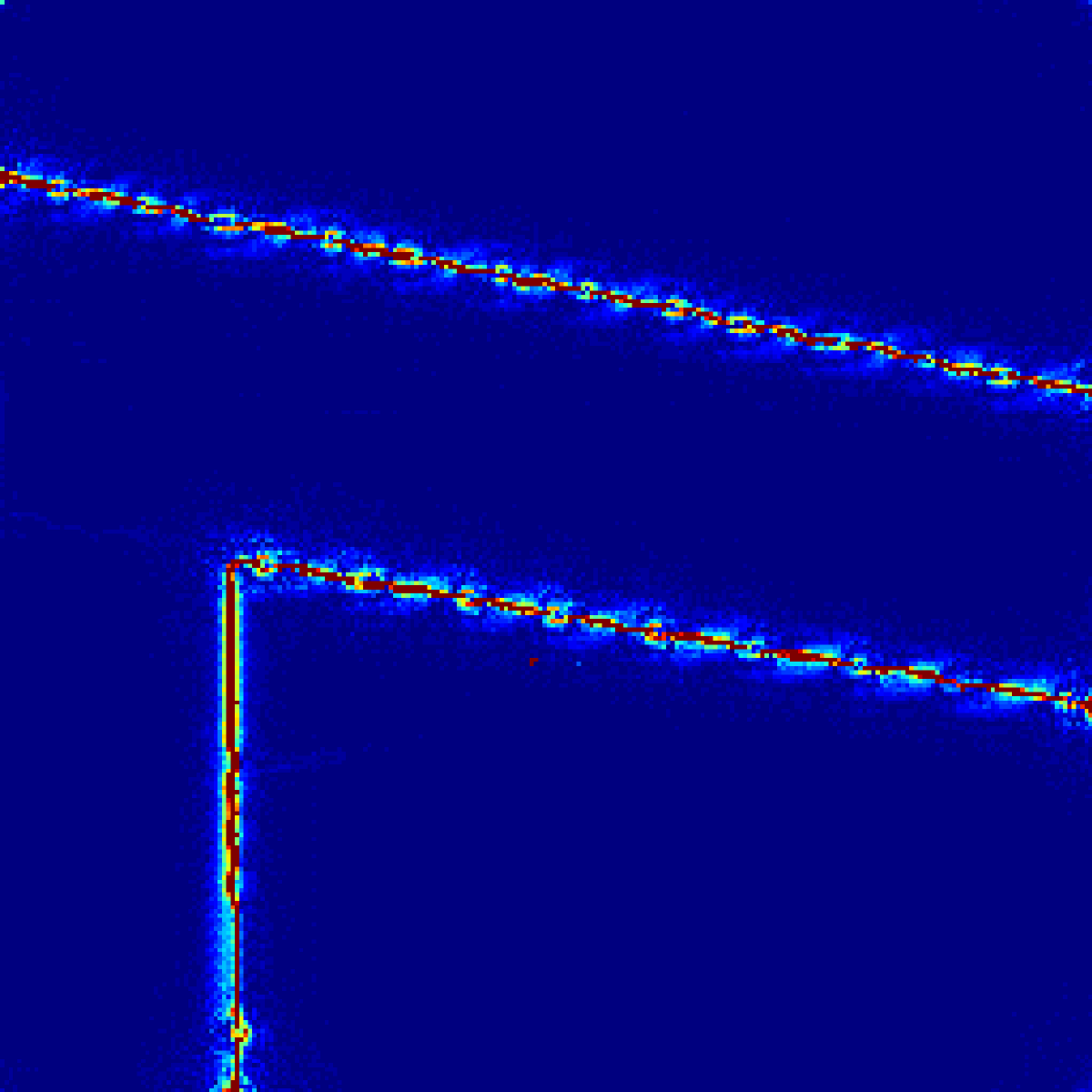}
	\hspace{-1.8mm} & \includegraphics[height=0.65in]{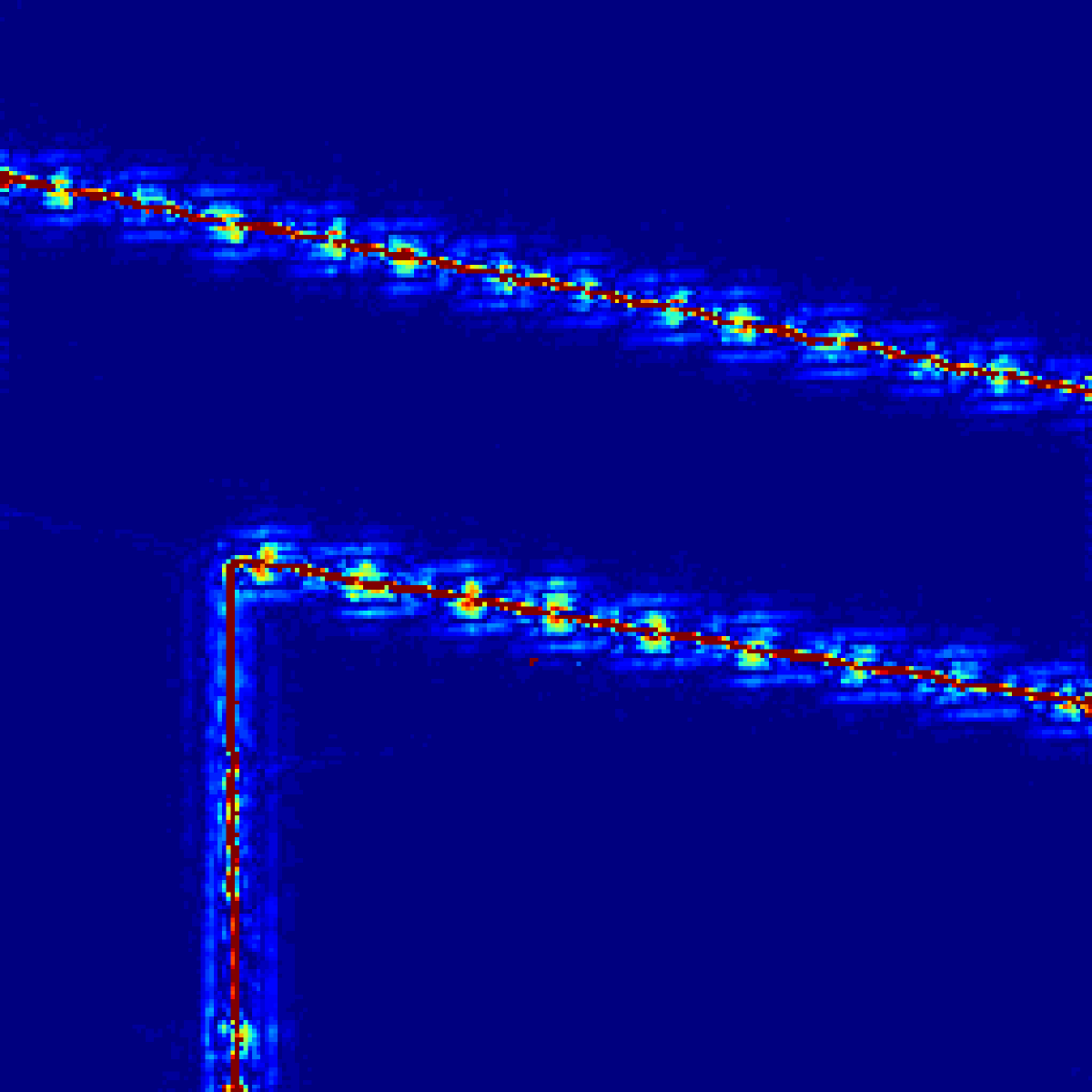}
	\hspace{-1.8mm} & \includegraphics[height=0.65in]{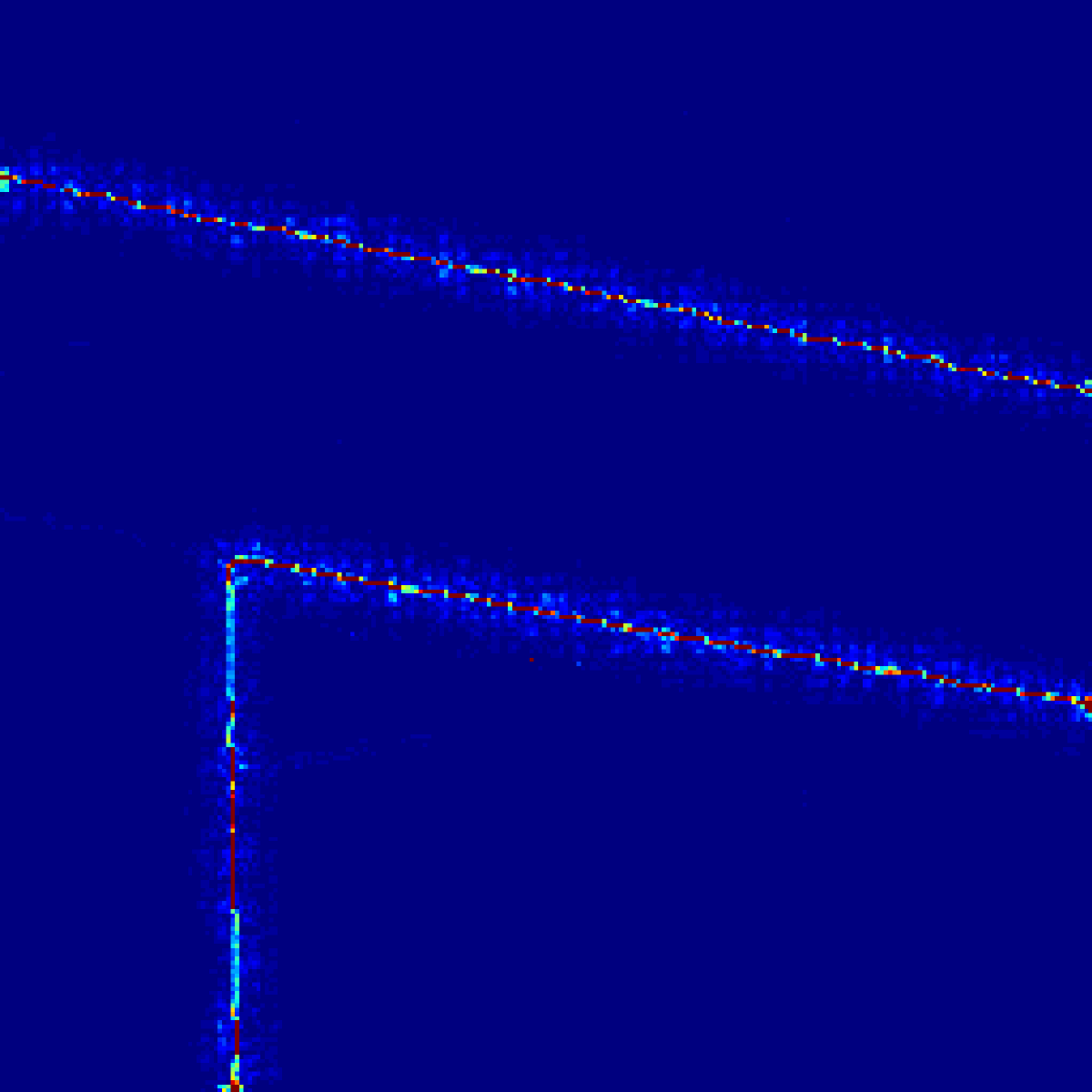}
	\hspace{-1.8mm} & \includegraphics[height=0.65in]{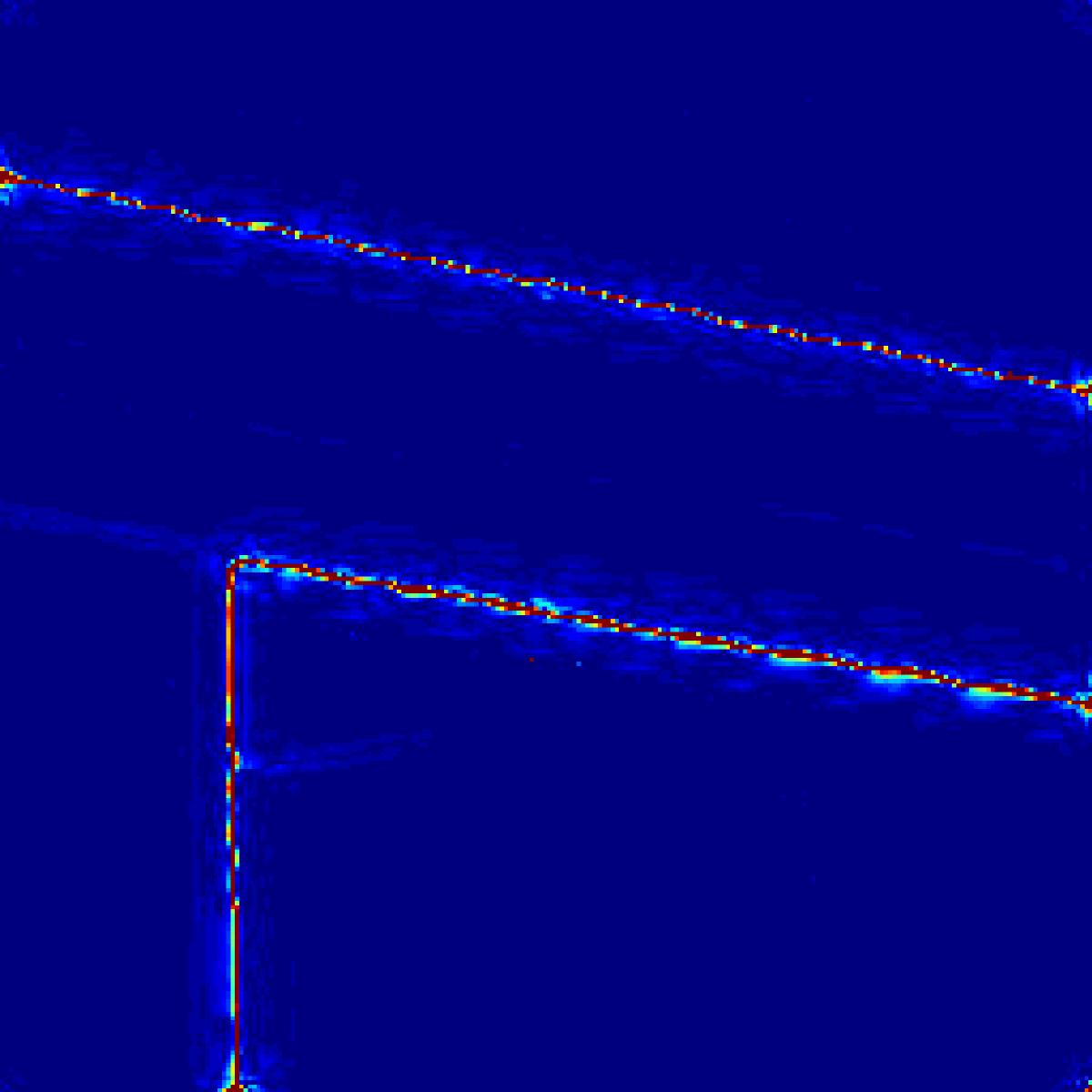}
	\hspace{-1.8mm} & \includegraphics[height=0.65in]{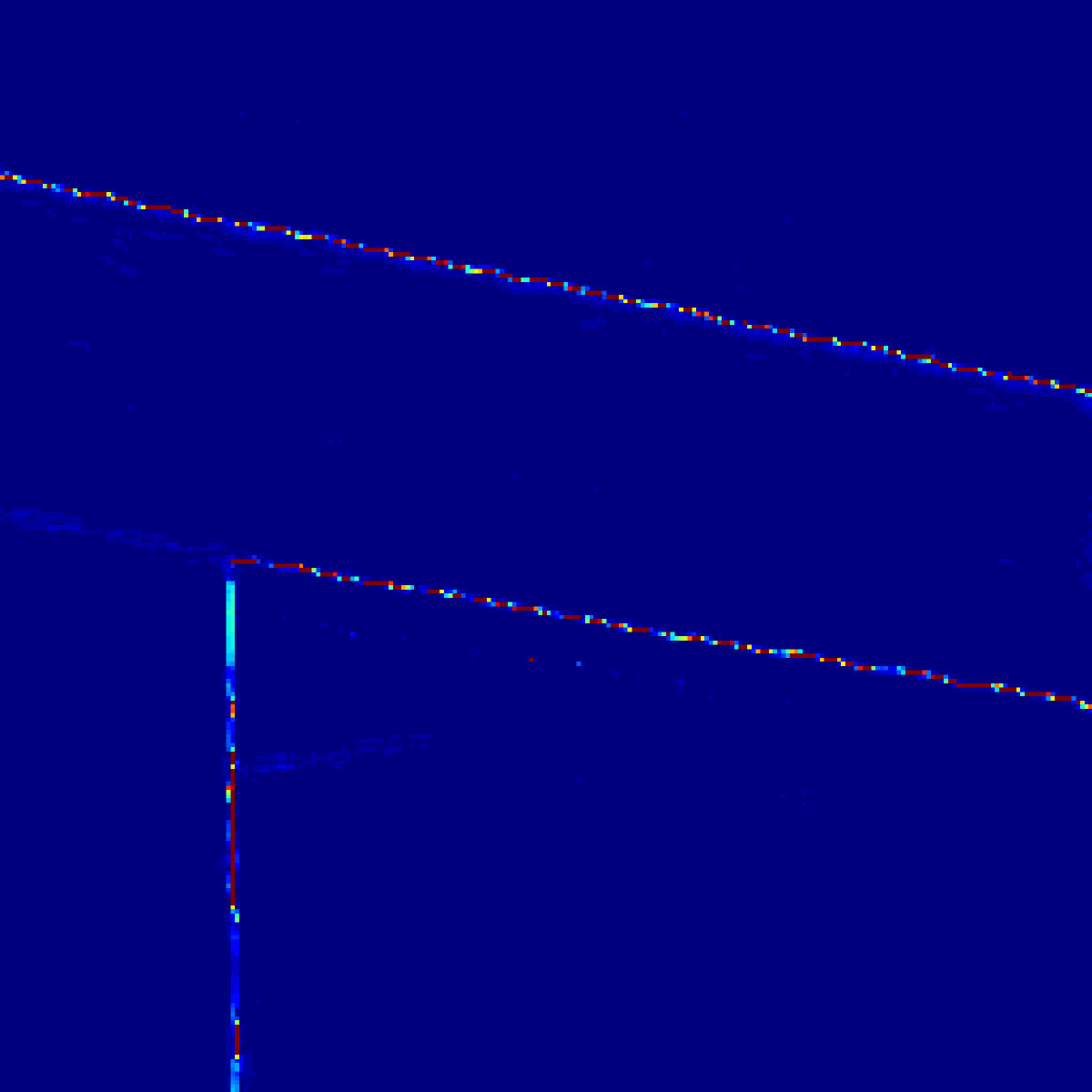}
	\hspace{-1.8mm} & \includegraphics[height=0.65in]{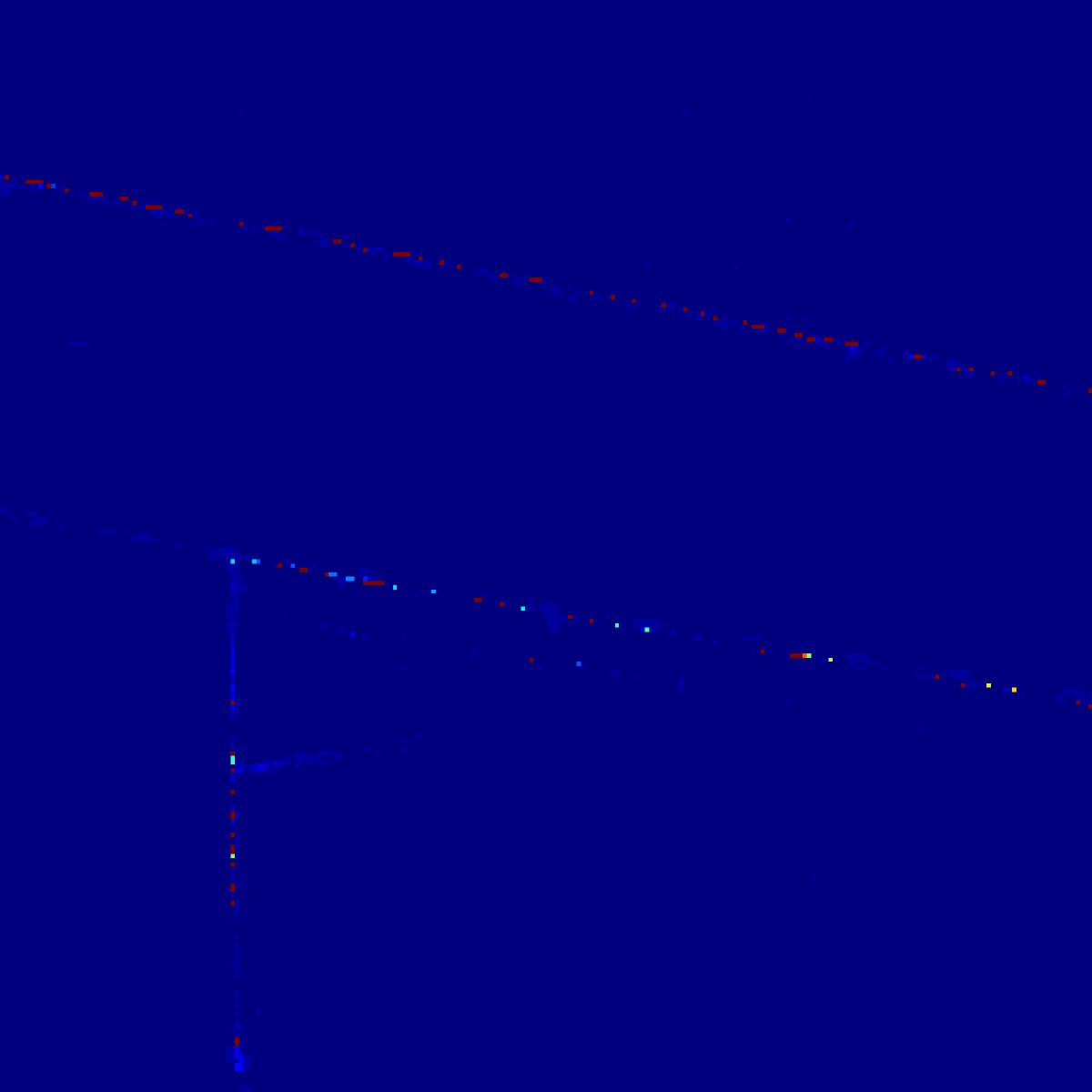}
        
        \hspace{-1.8mm} & \includegraphics[height=0.65in]{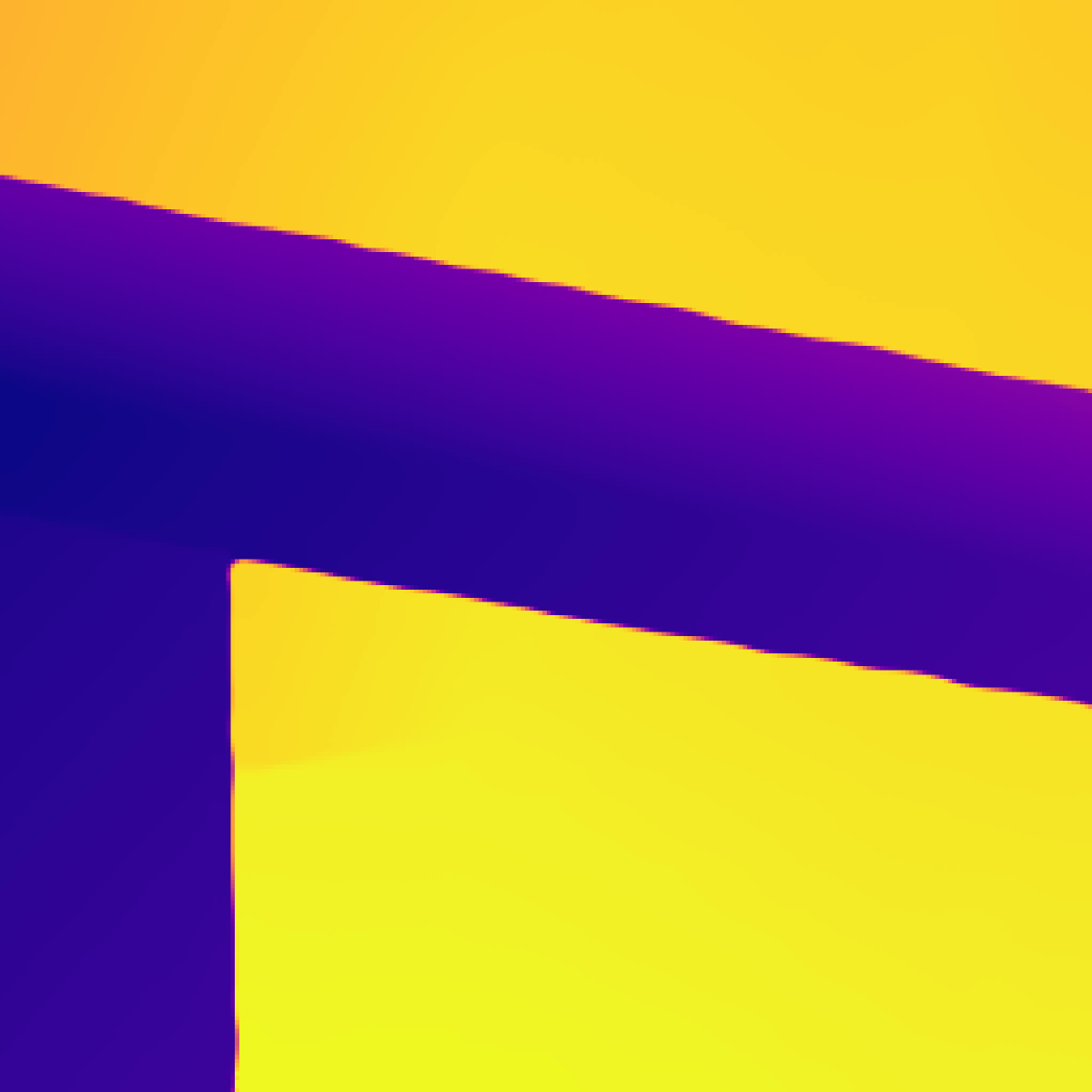}
        \\ \vspace{-0.1cm}
    
        \rotatebox[origin=l]{90}{\scriptsize \quad \textbf{8$\times$}} & \includegraphics[height=0.65in]{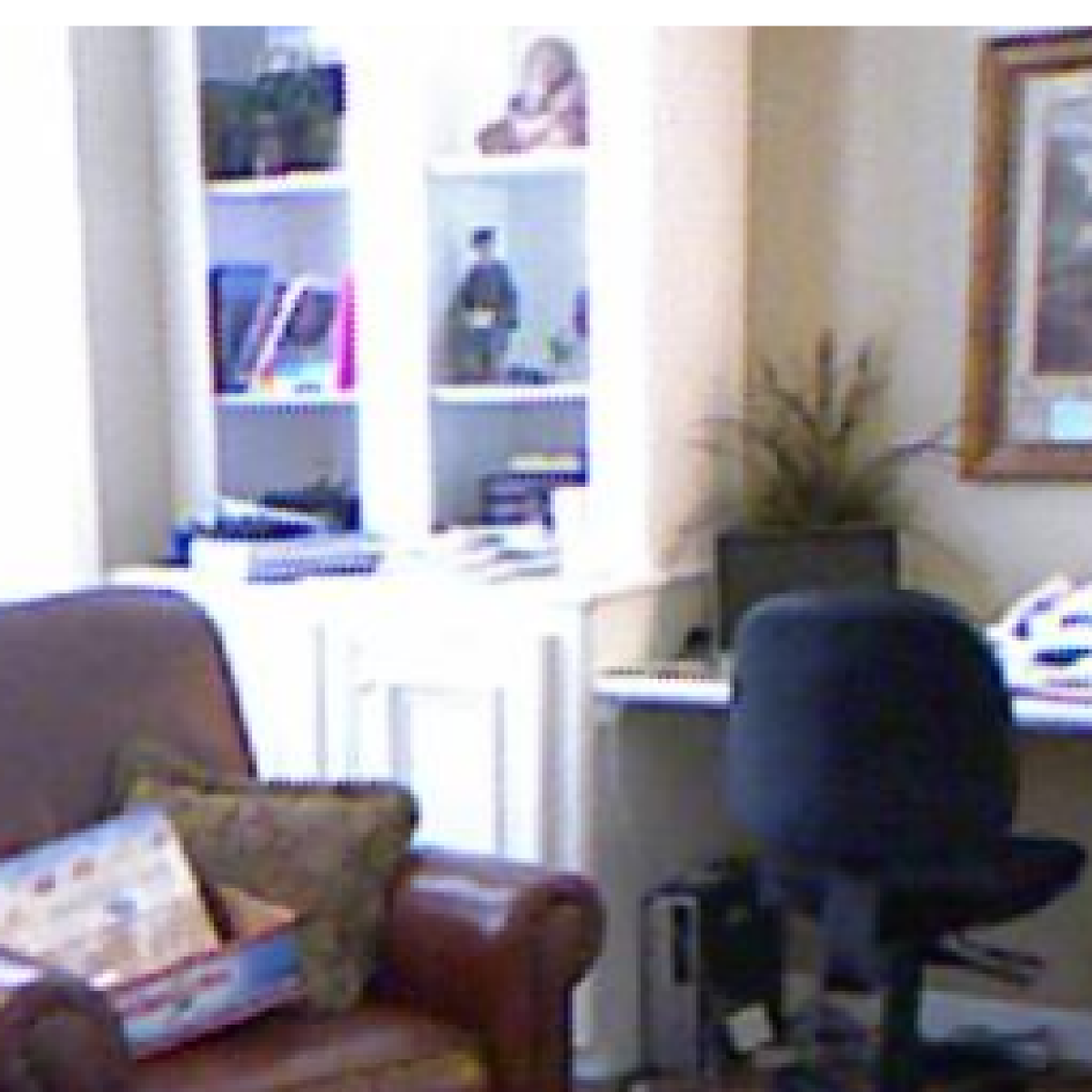}
	\hspace{-1.8mm} & \includegraphics[height=0.65in]{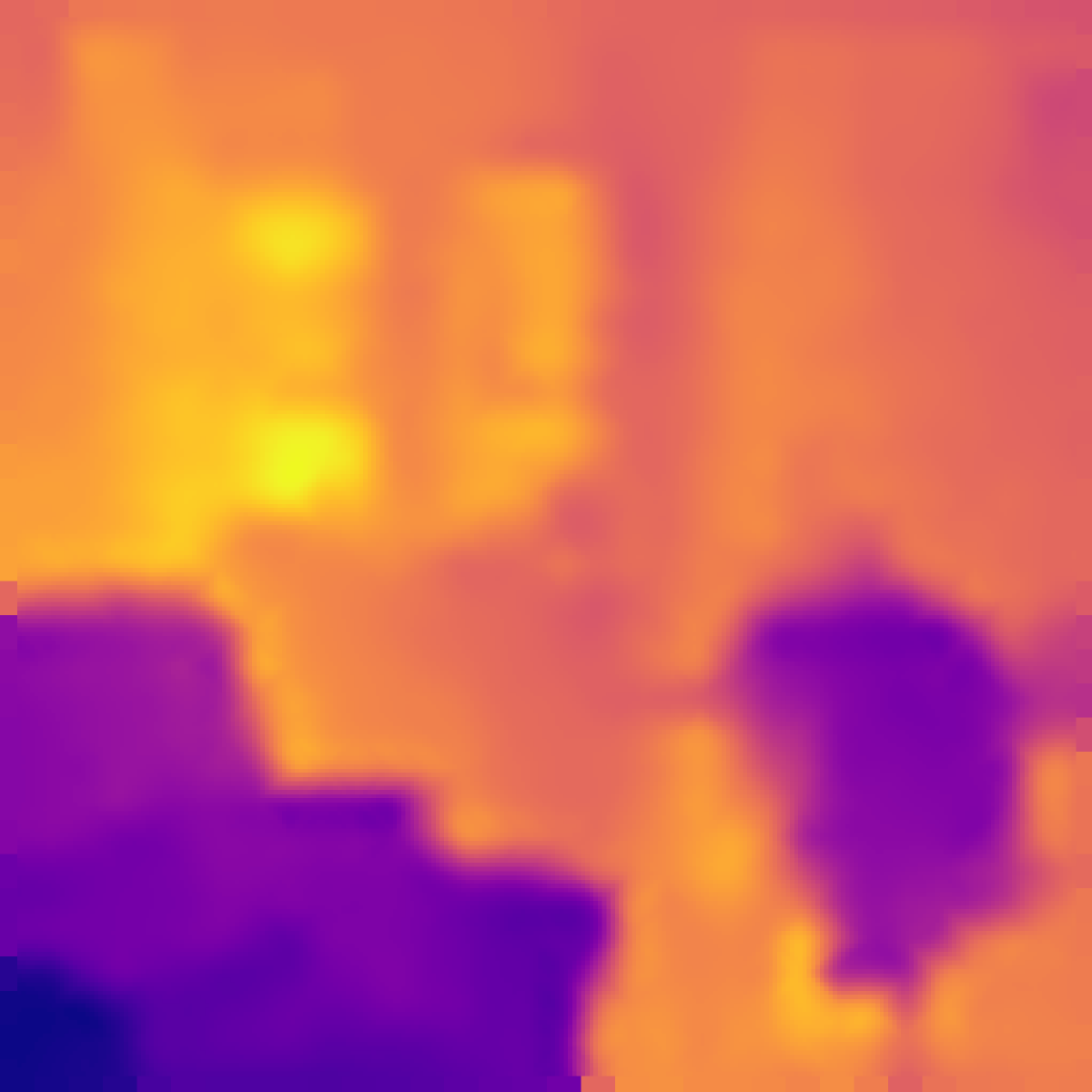}
	\hspace{-1.8mm} & \includegraphics[height=0.65in]{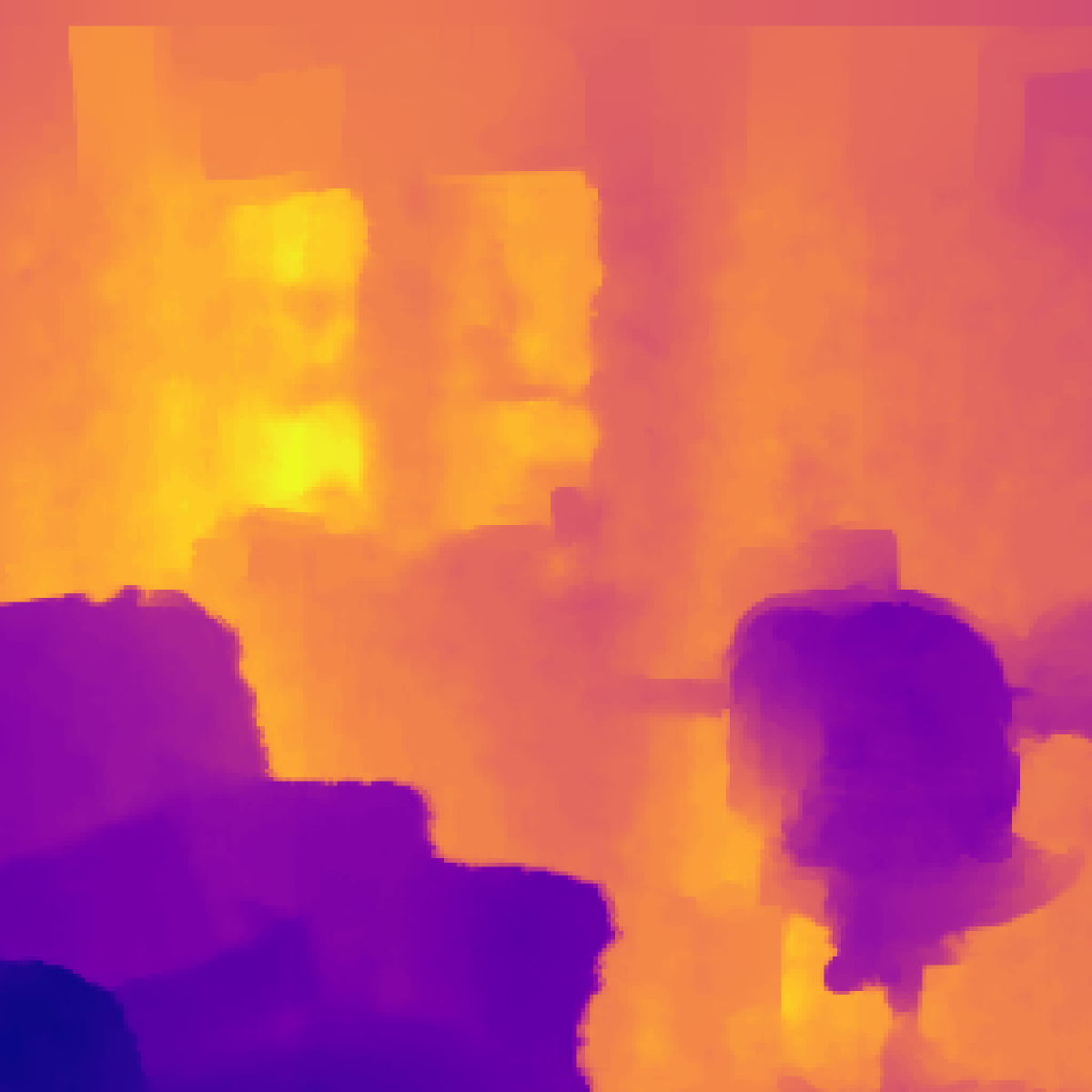}
	\hspace{-1.8mm} & \includegraphics[height=0.65in]{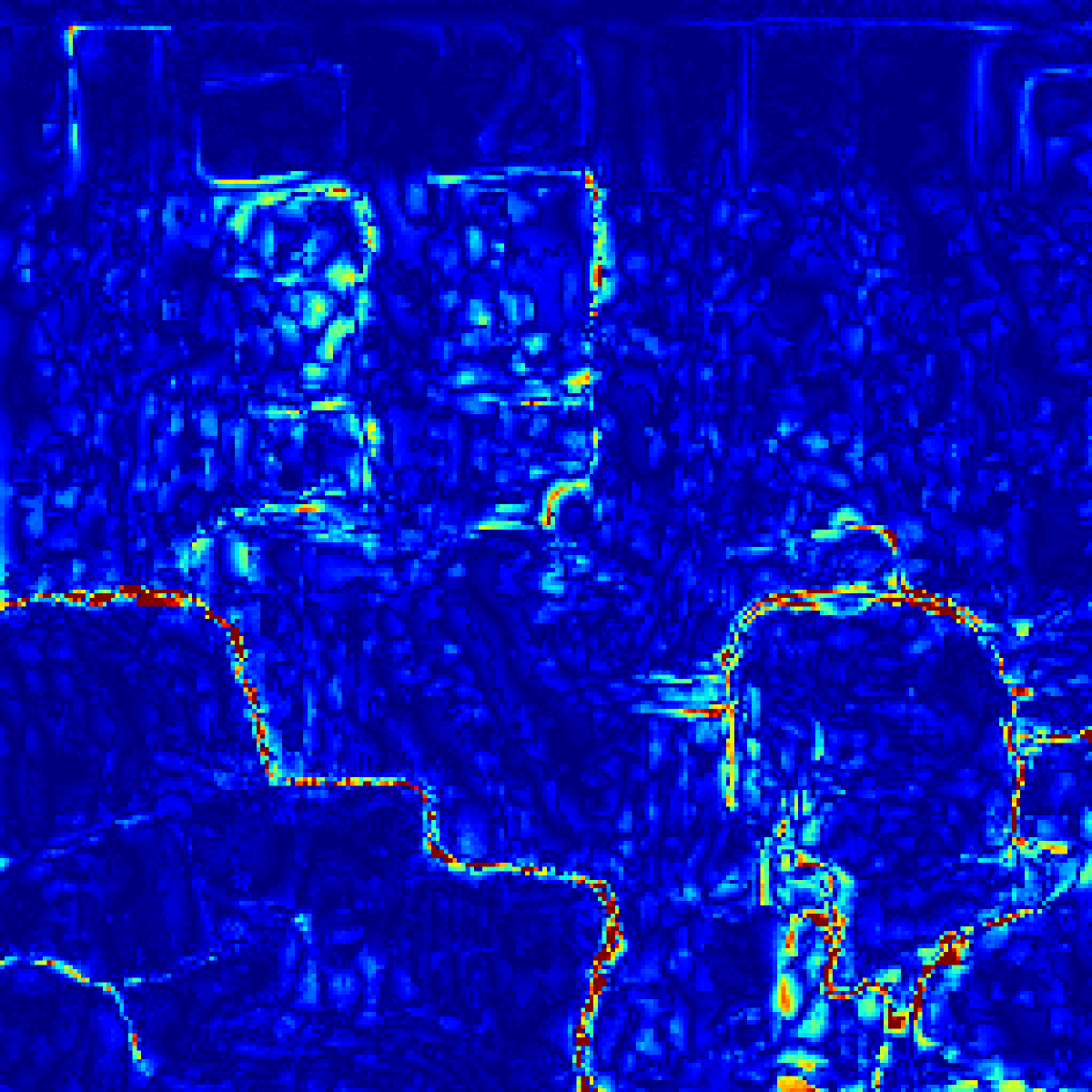}
	\hspace{-1.8mm} & \includegraphics[height=0.65in]{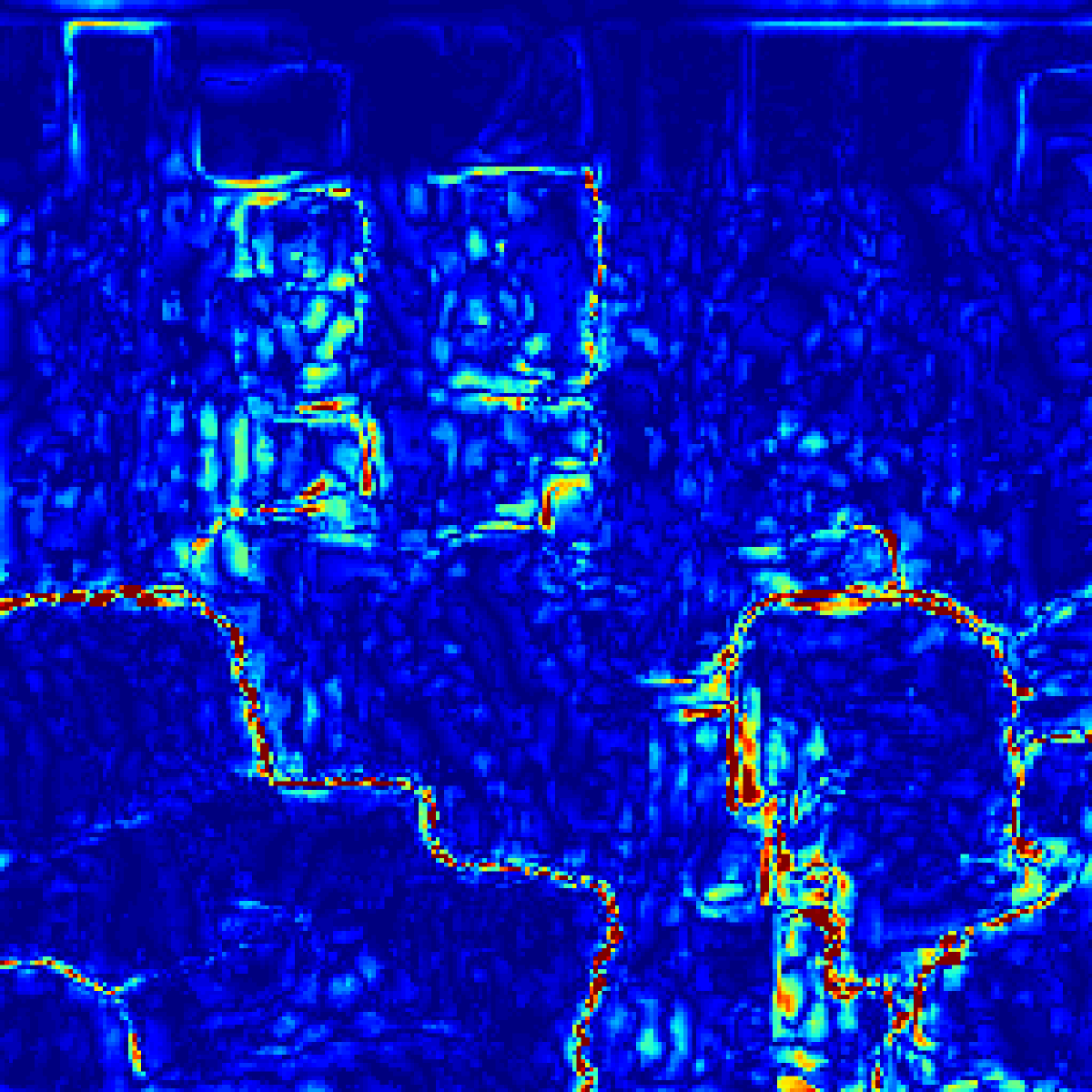}
	\hspace{-1.8mm} & \includegraphics[height=0.65in]{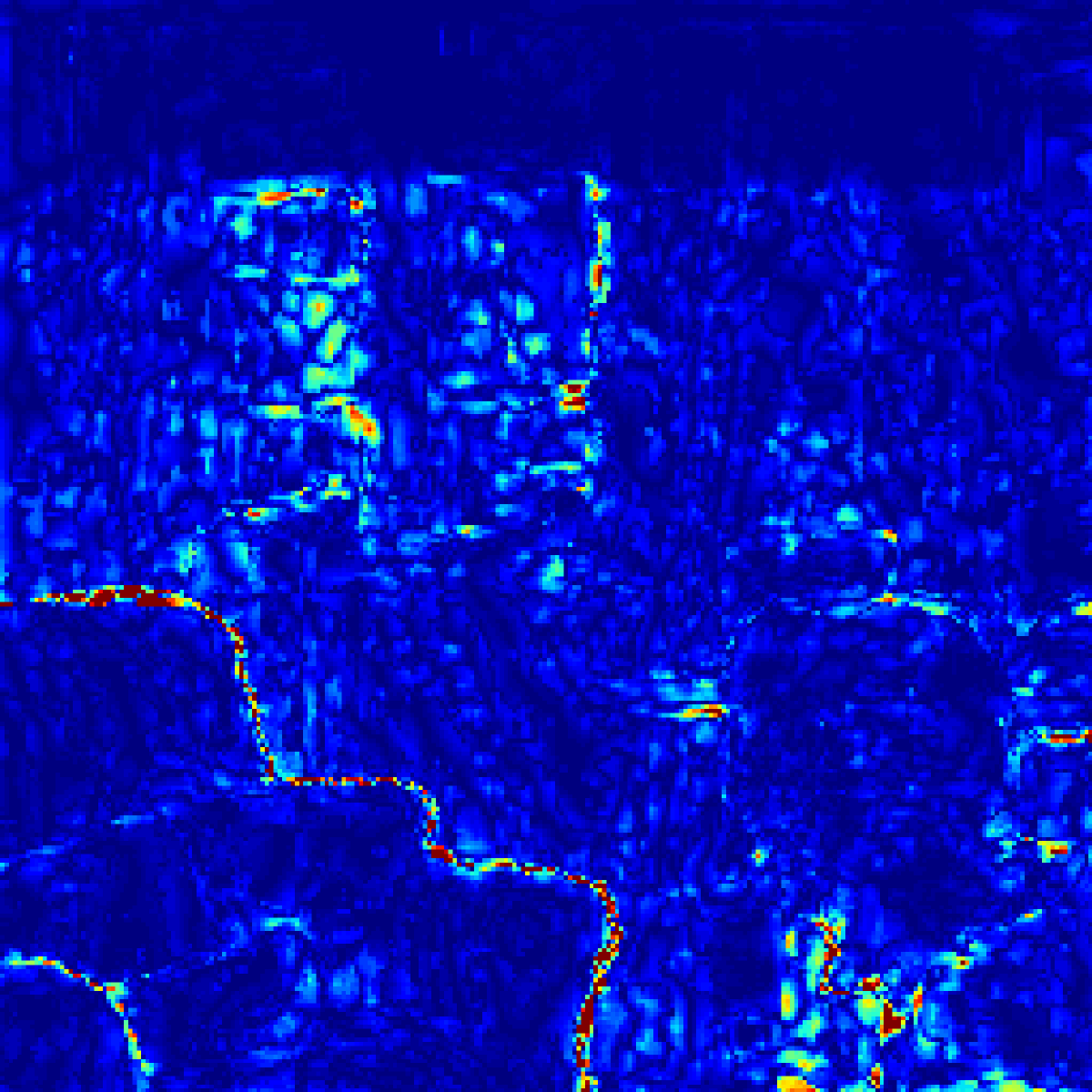}
	\hspace{-1.8mm} & \includegraphics[height=0.65in]{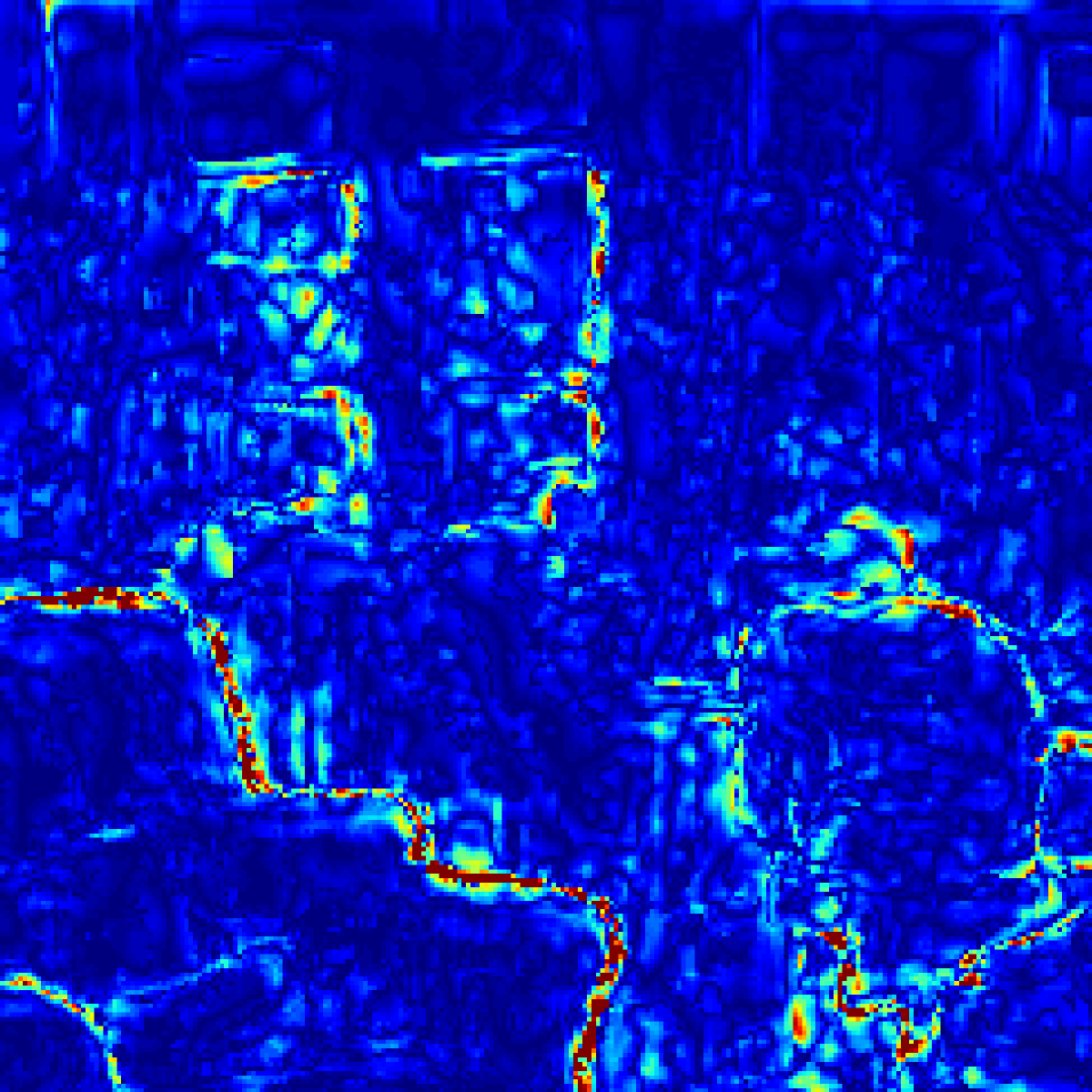}
	\hspace{-1.8mm} & \includegraphics[height=0.65in]{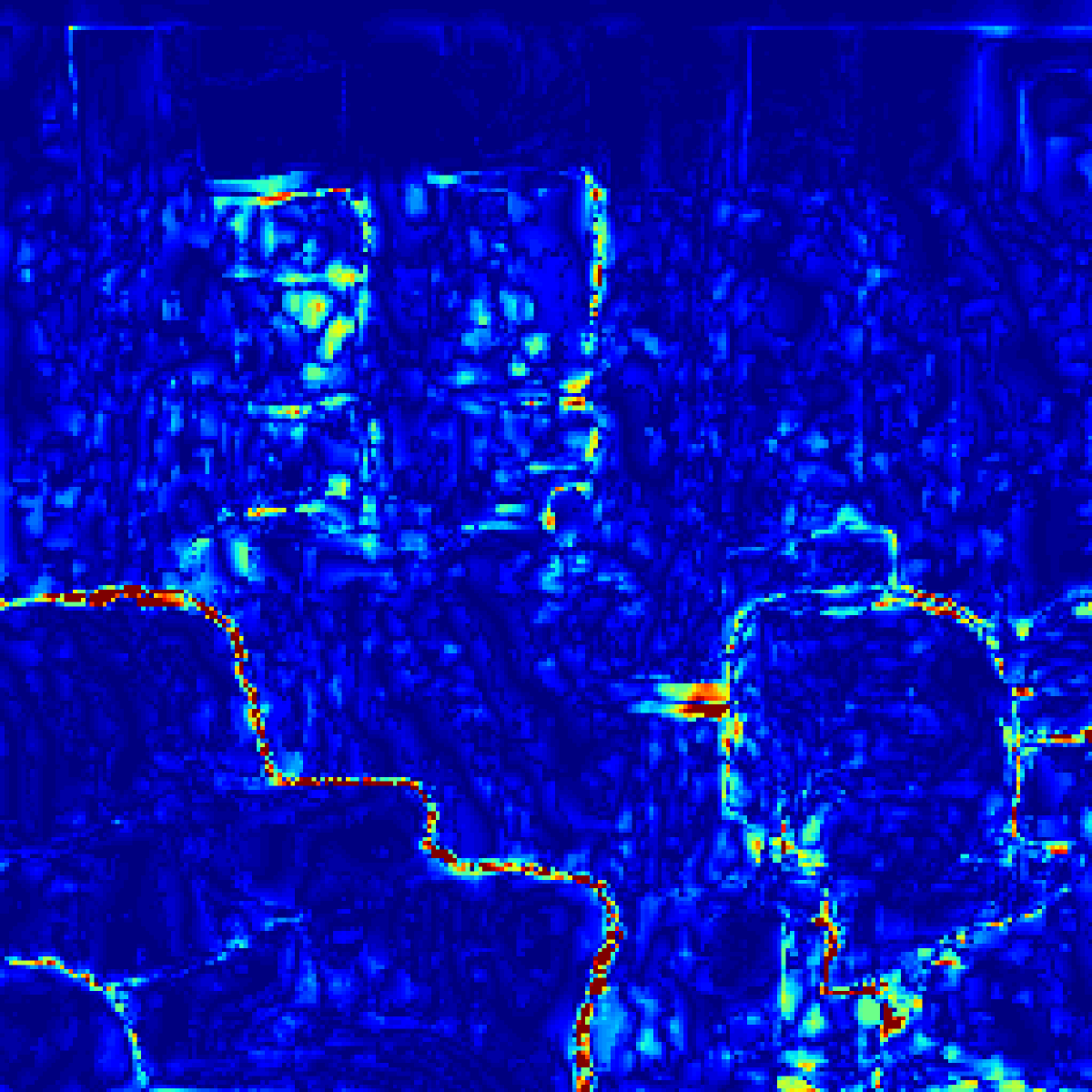}
	\hspace{-1.8mm} & \includegraphics[height=0.65in]{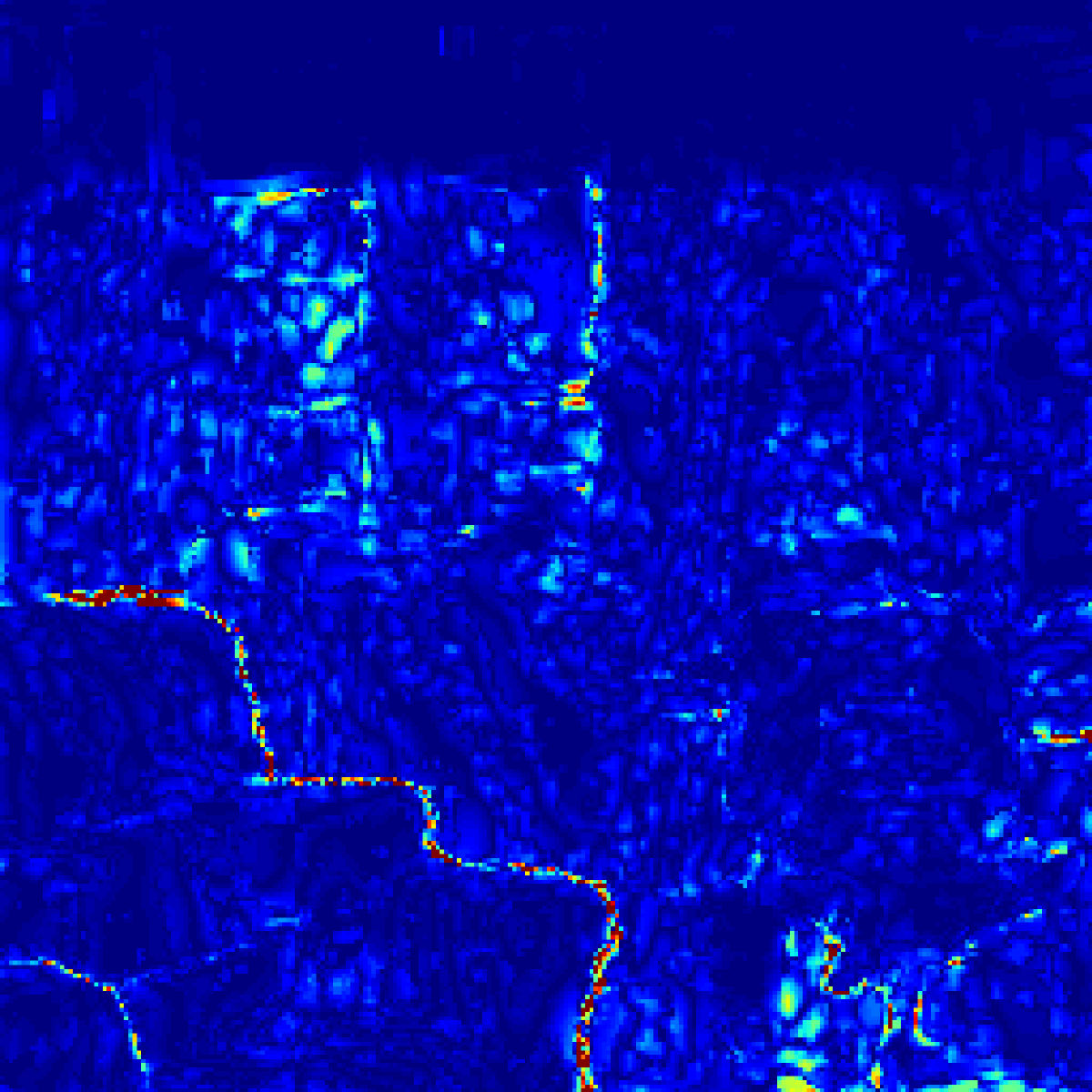}
 
	\hspace{-1.8mm} & \includegraphics[height=0.65in]{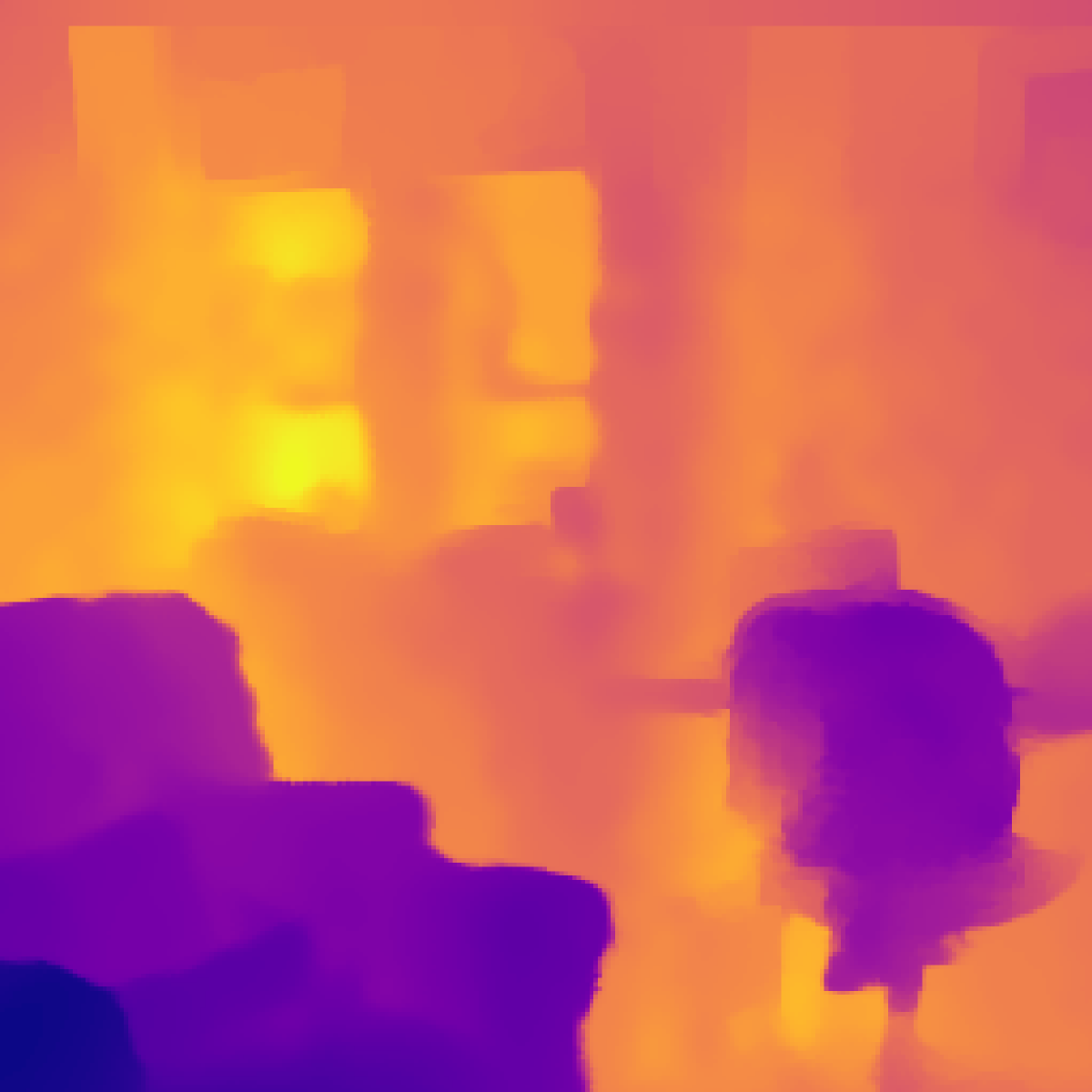}
     \\ \vspace{-0.cm}
    
        \rotatebox[origin=l]{90}{\scriptsize \quad \textbf{NYUv2}} & \includegraphics[height=0.65in]{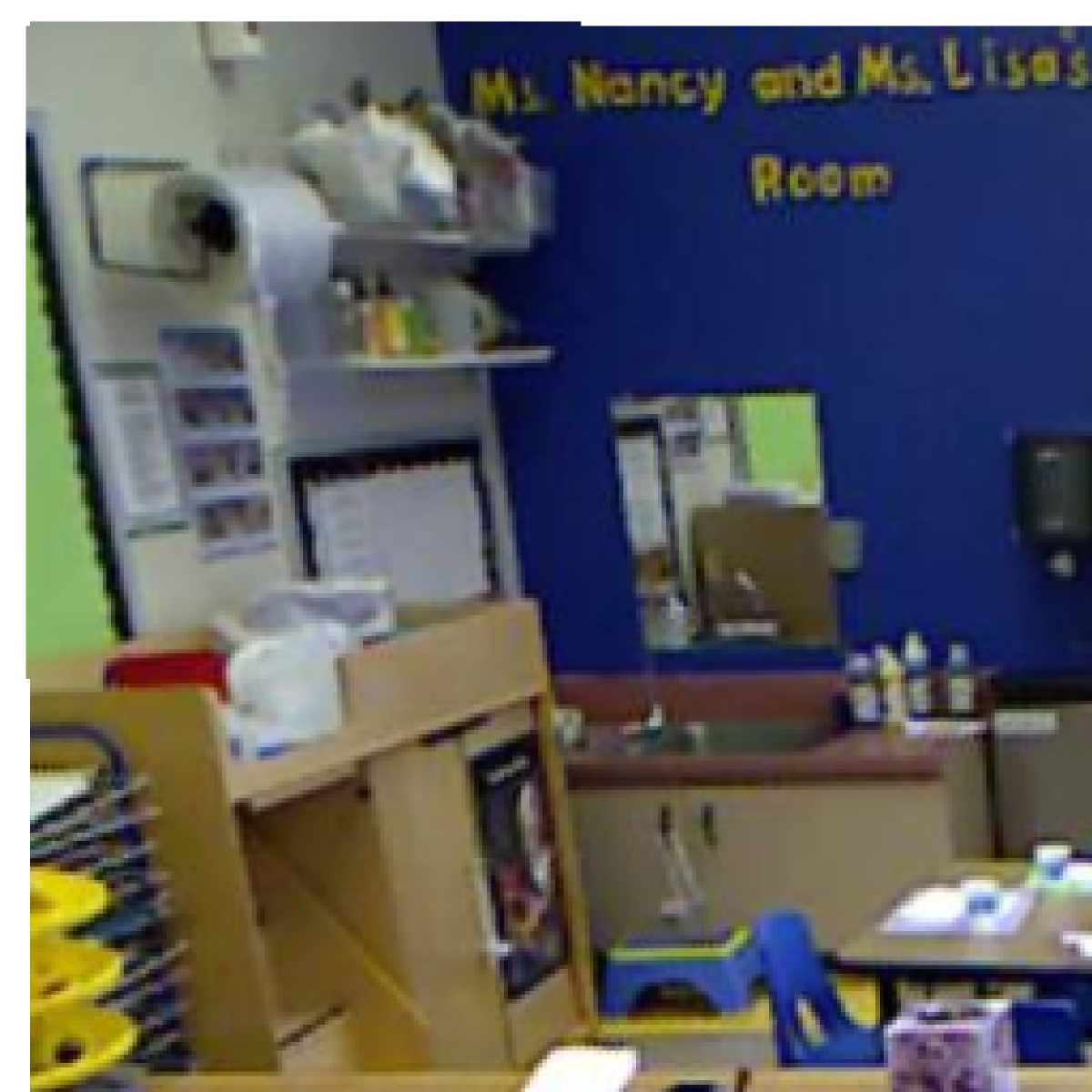}
	\hspace{-1.8mm} & \includegraphics[height=0.65in]{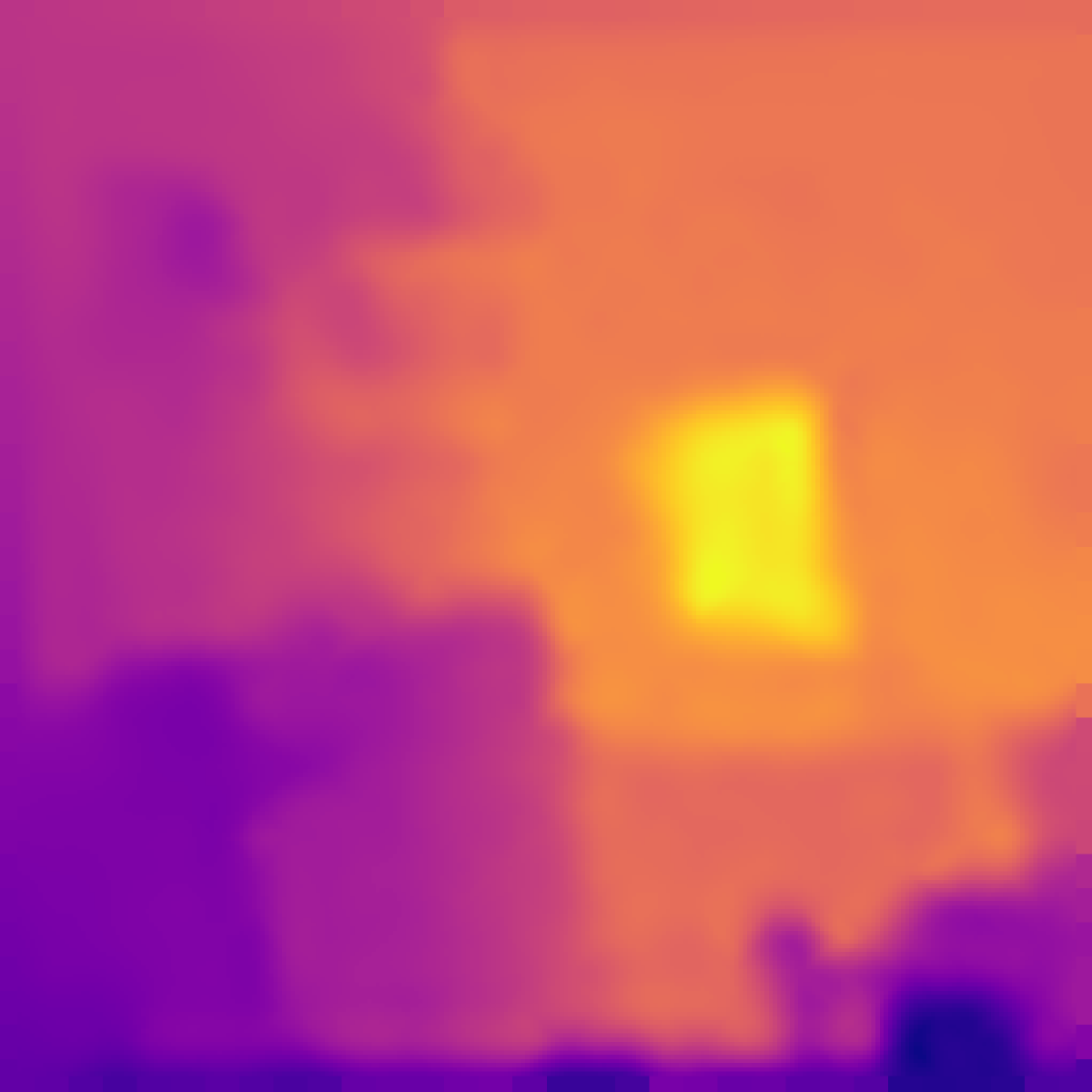}
	\hspace{-1.8mm} & \includegraphics[height=0.65in]{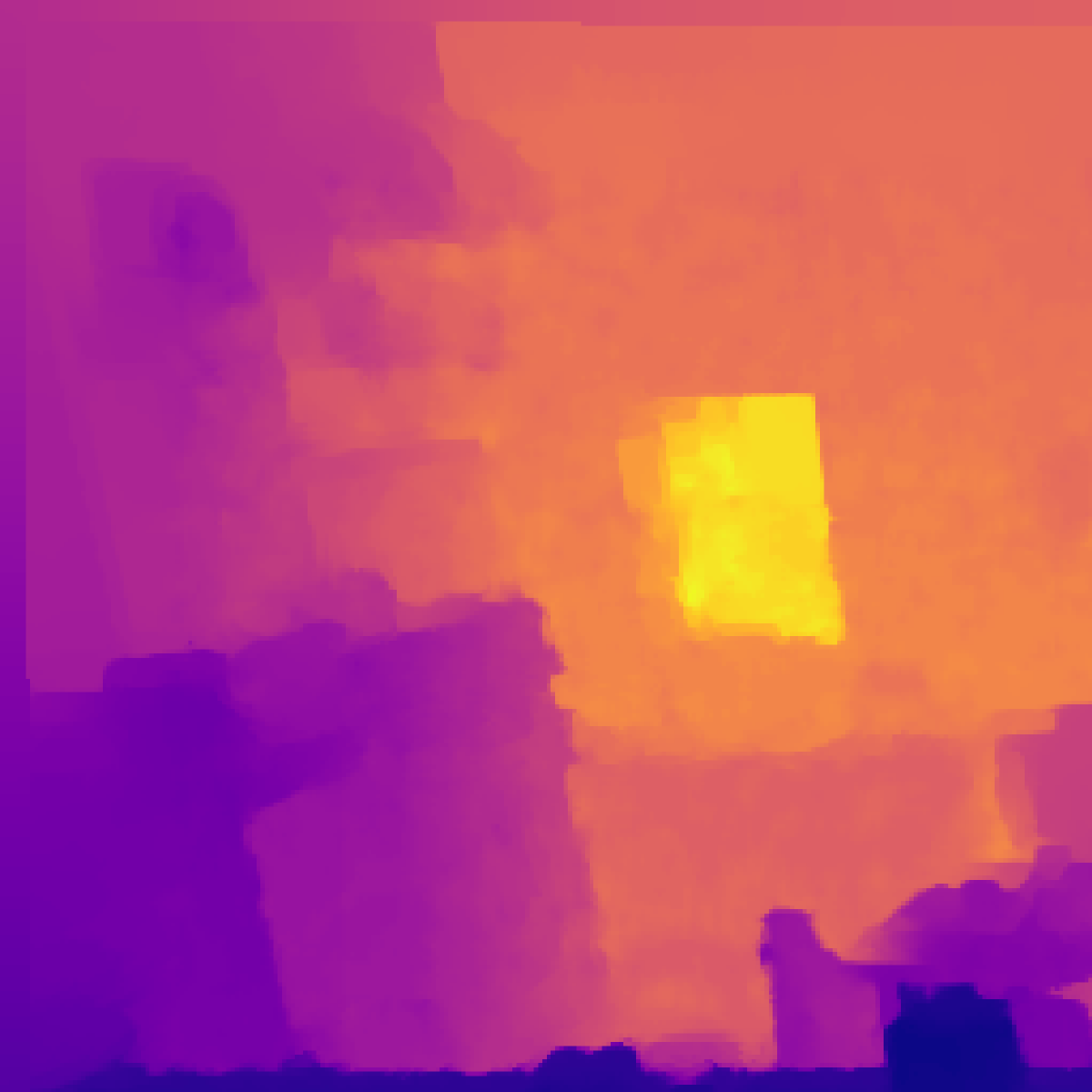}
	\hspace{-1.8mm} & \includegraphics[height=0.65in]{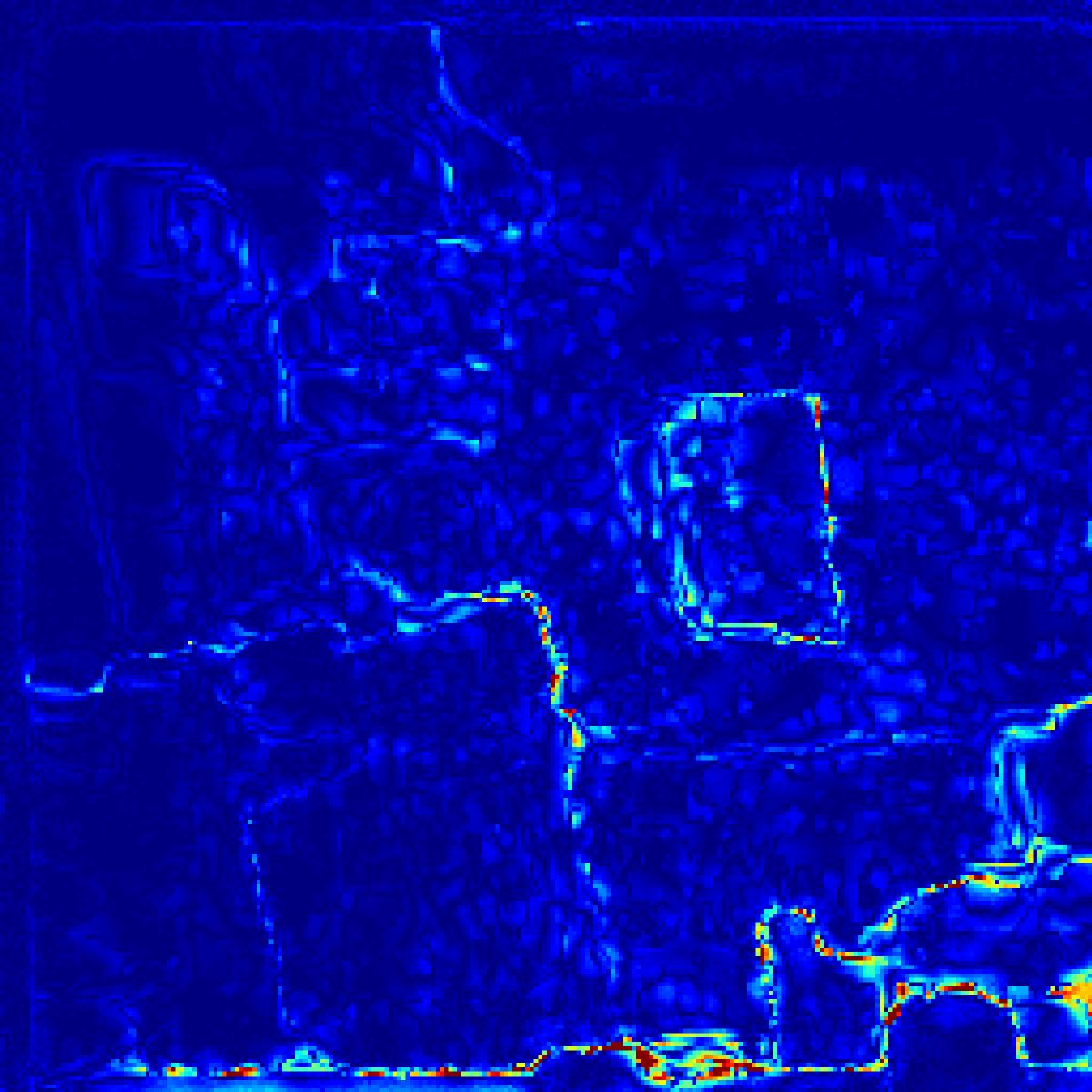}
	\hspace{-1.8mm} & \includegraphics[height=0.65in]{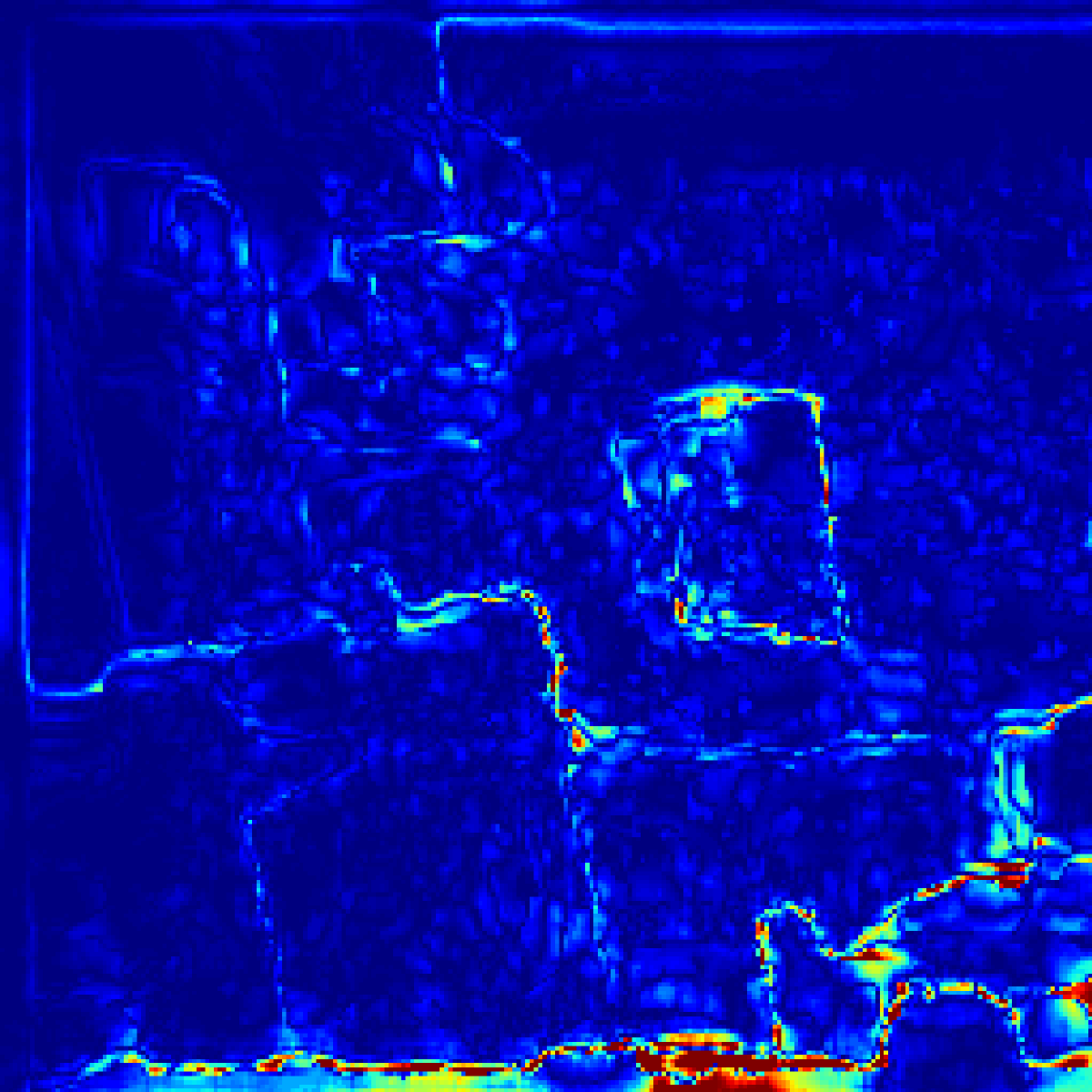}
	\hspace{-1.8mm} & \includegraphics[height=0.65in]{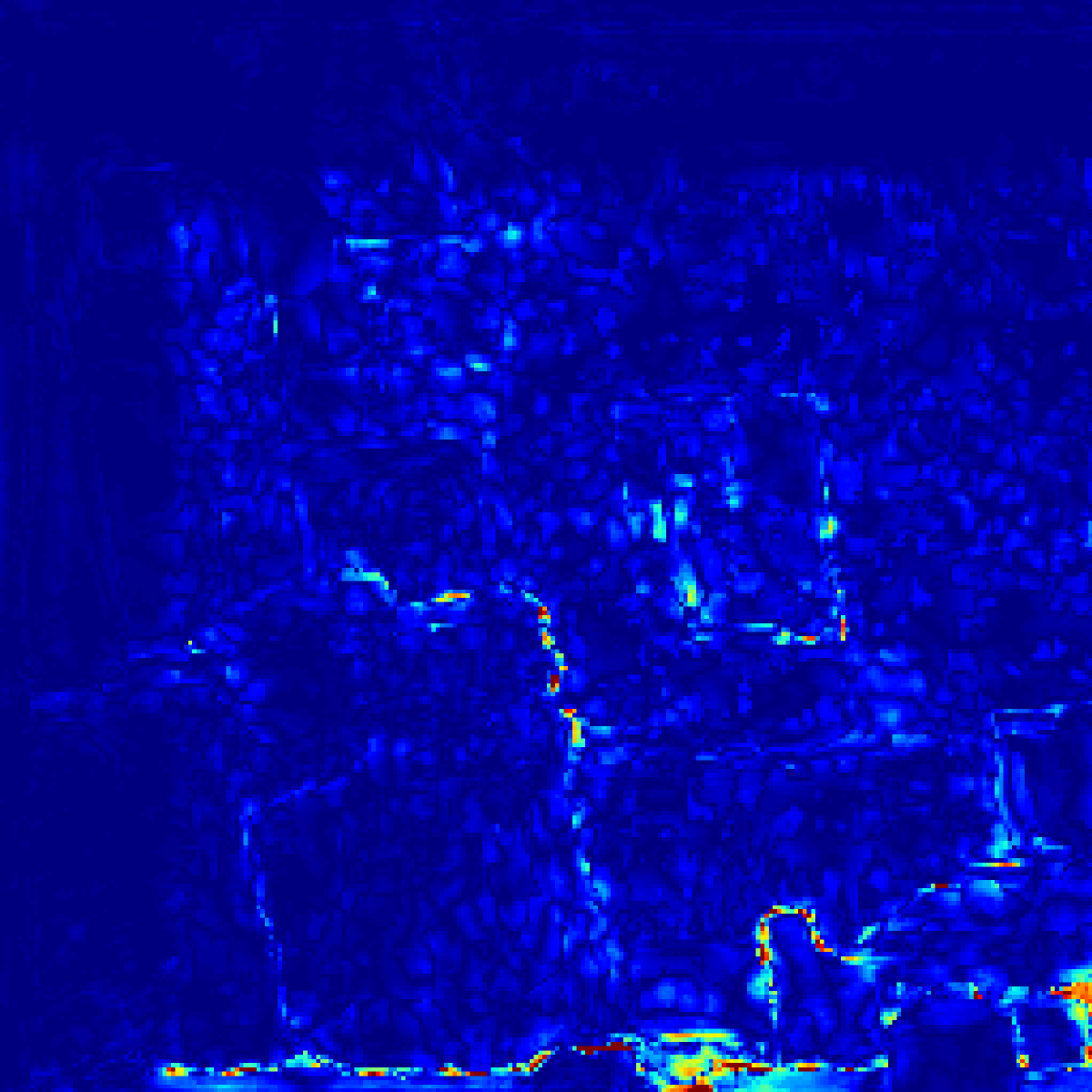}
	\hspace{-1.8mm} & \includegraphics[height=0.65in]{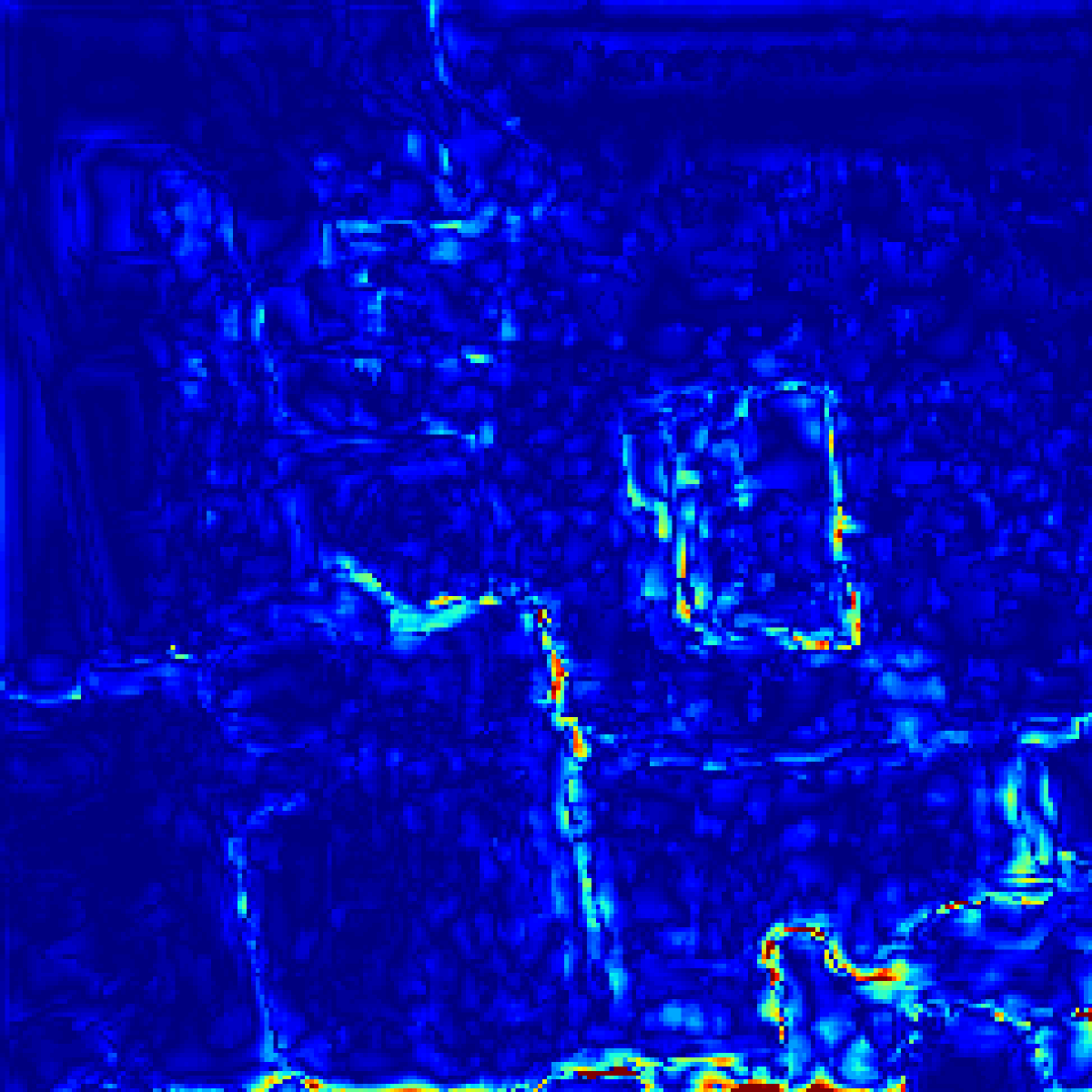}
	\hspace{-1.8mm} & \includegraphics[height=0.65in]{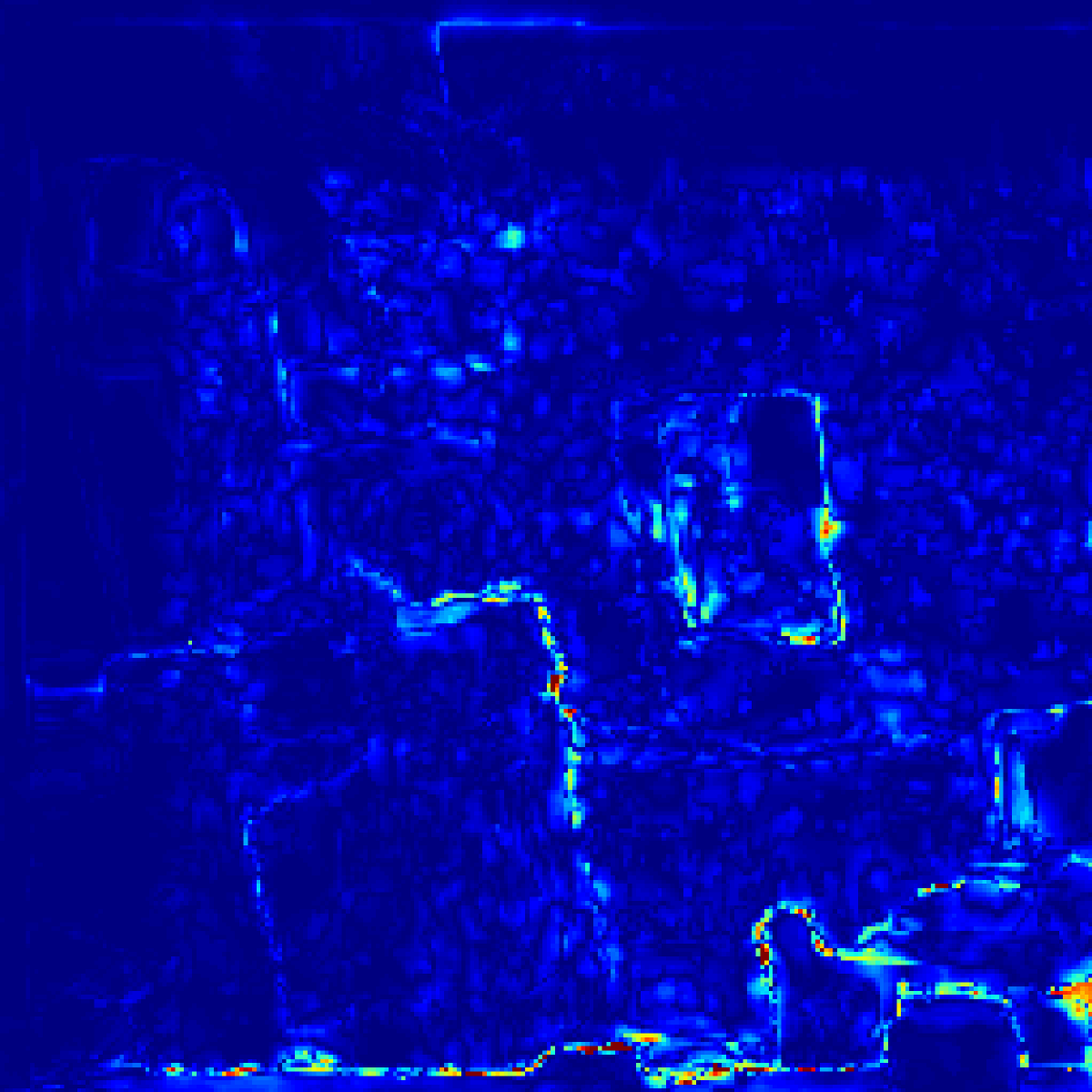}
	\hspace{-1.8mm} & \includegraphics[height=0.65in]{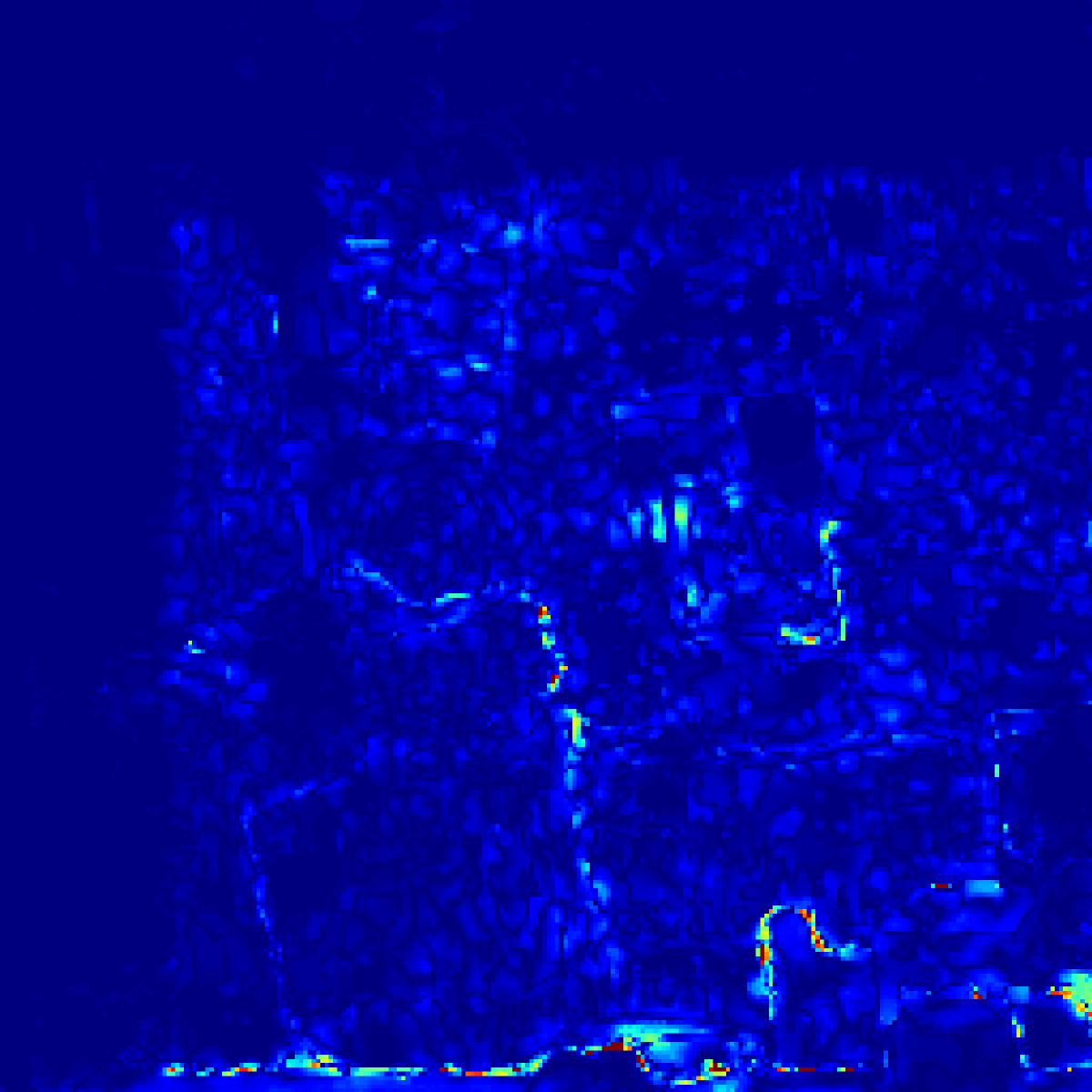}
 
	\hspace{-1.8mm} & \includegraphics[height=0.65in]{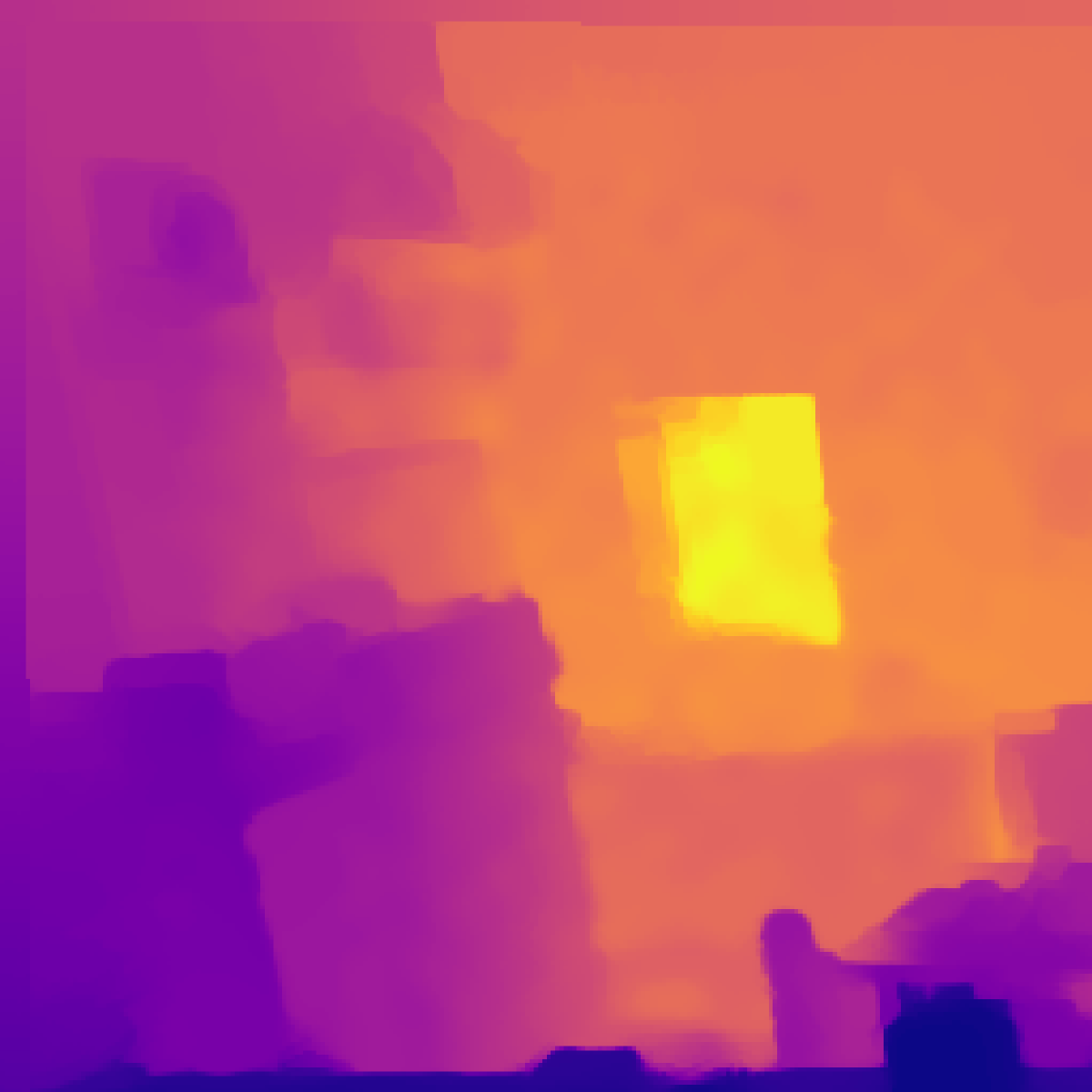}
	\\ \vspace{-0.1cm}
	
        \rotatebox[origin=l]{90}{\scriptsize \quad \textbf{16$\times$}} & \includegraphics[height=0.65in]{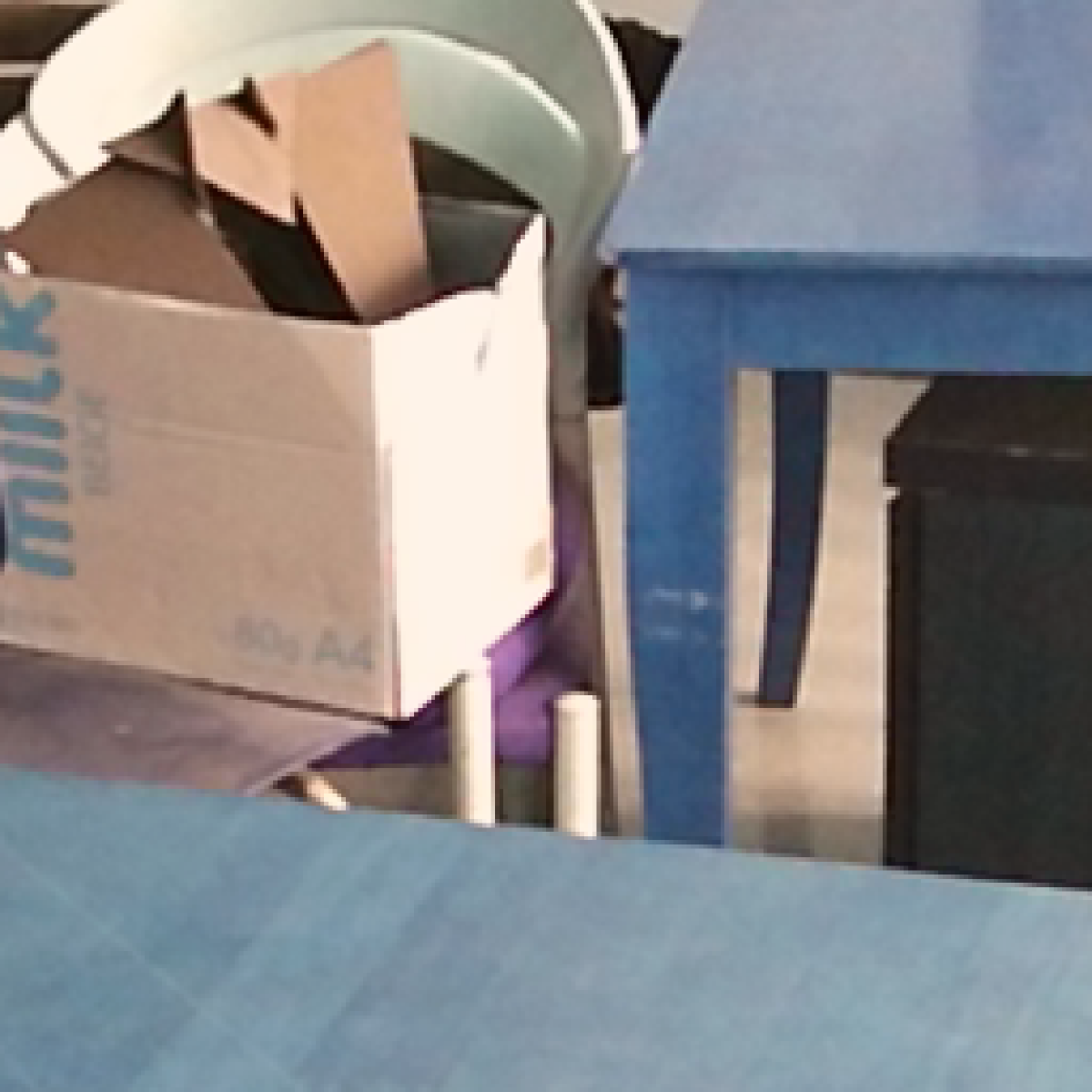}
	\hspace{-1.8mm} & \includegraphics[height=0.65in]{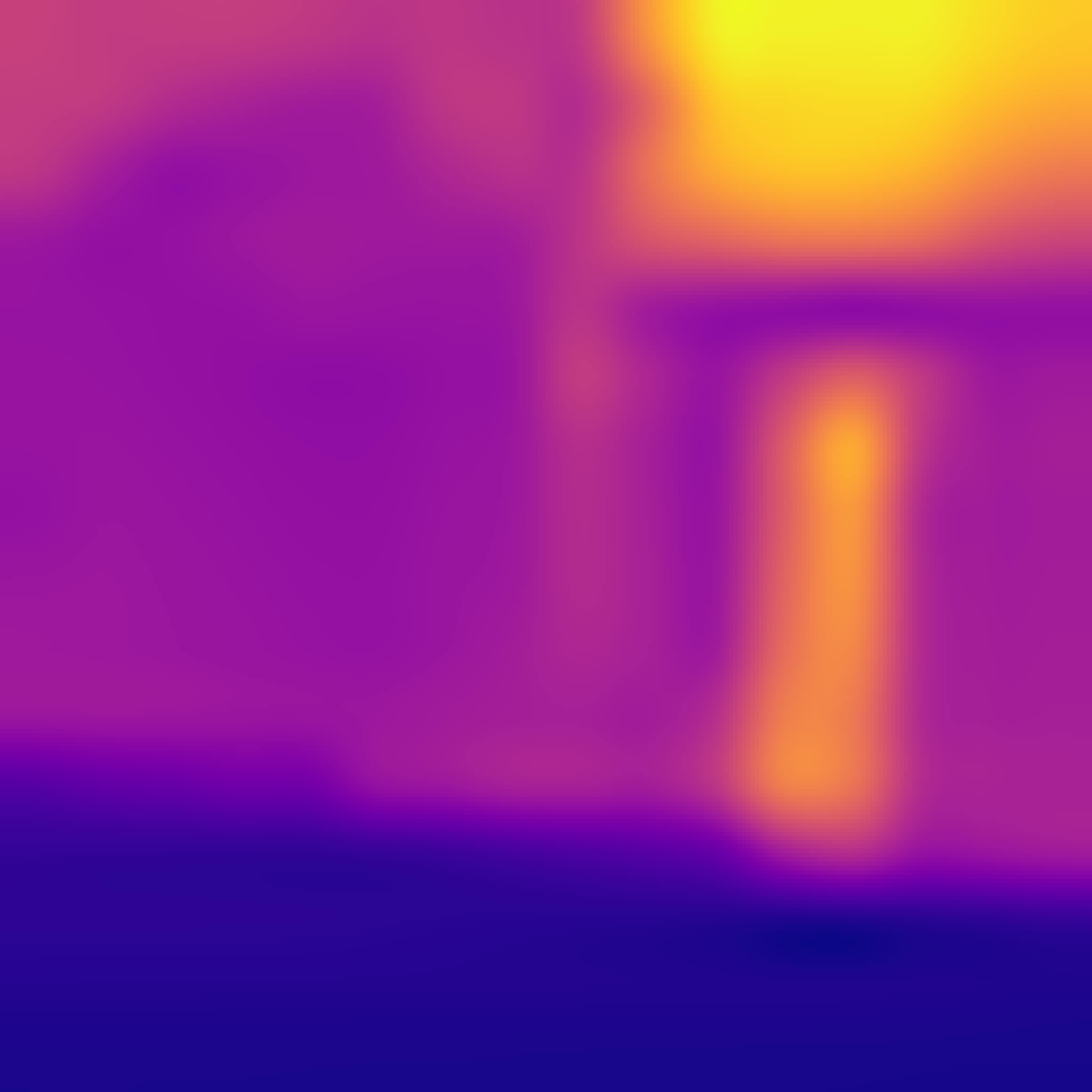}
	\hspace{-1.8mm} & \includegraphics[height=0.65in]{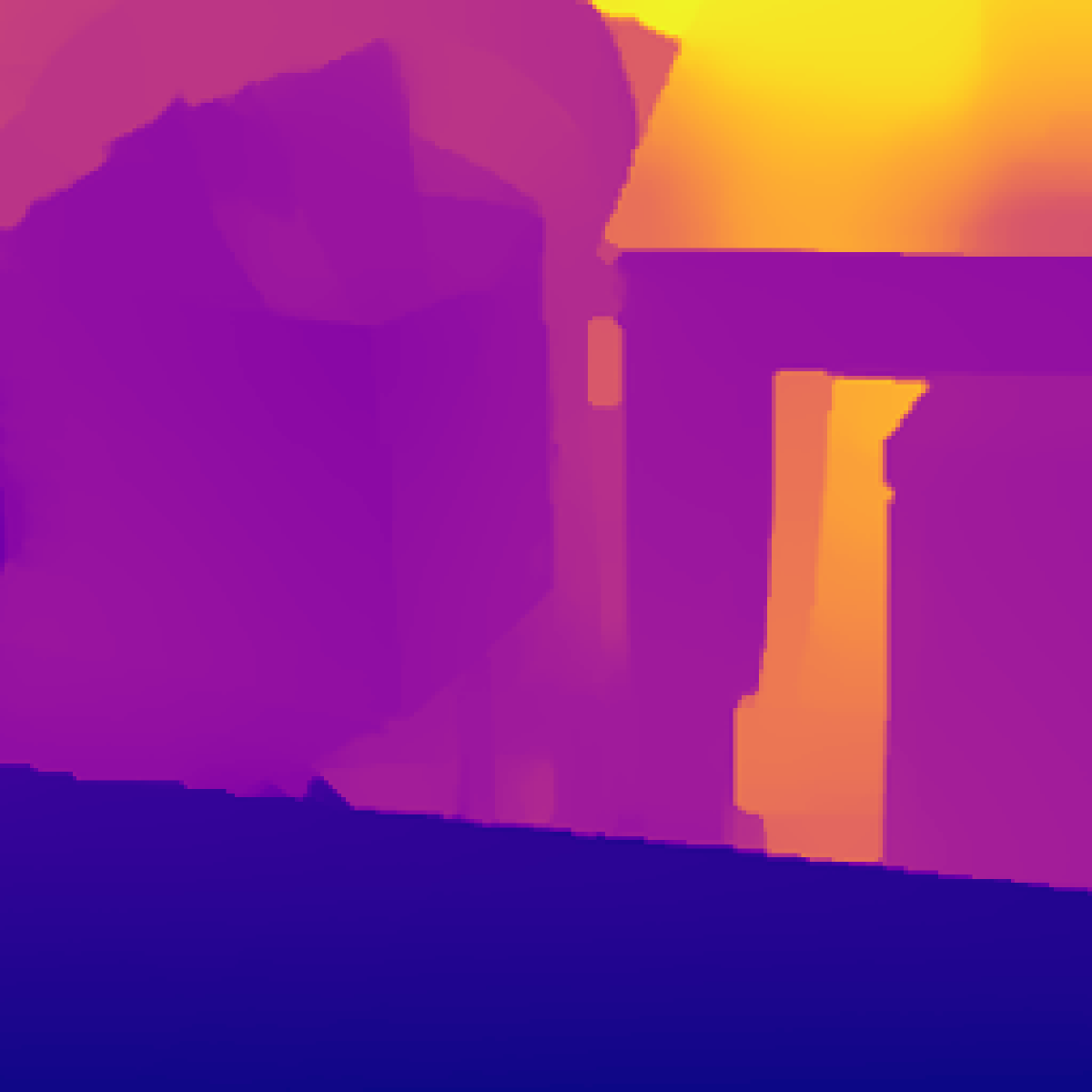}
	\hspace{-1.8mm} & \includegraphics[height=0.65in]{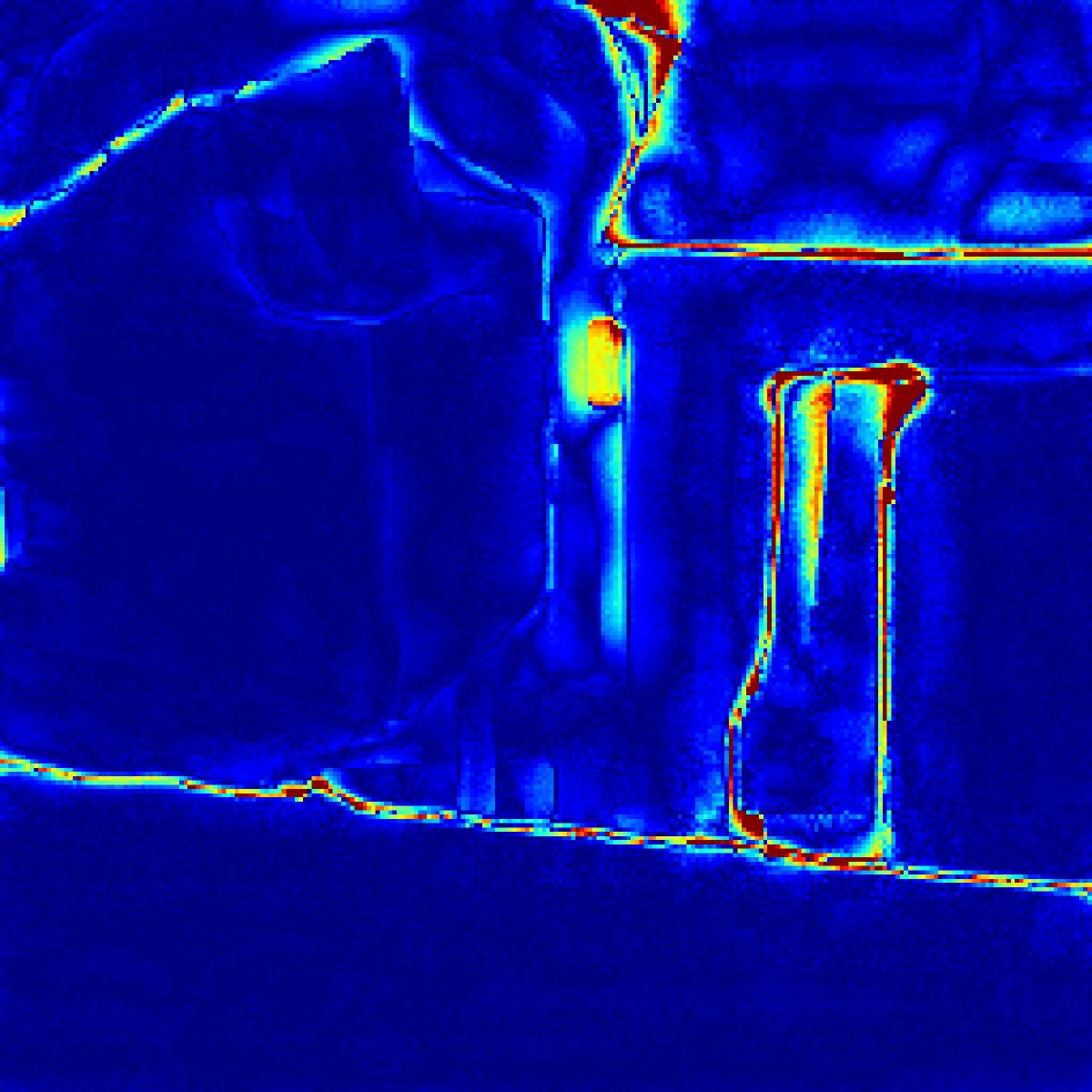}
	\hspace{-1.8mm} & \includegraphics[height=0.65in]{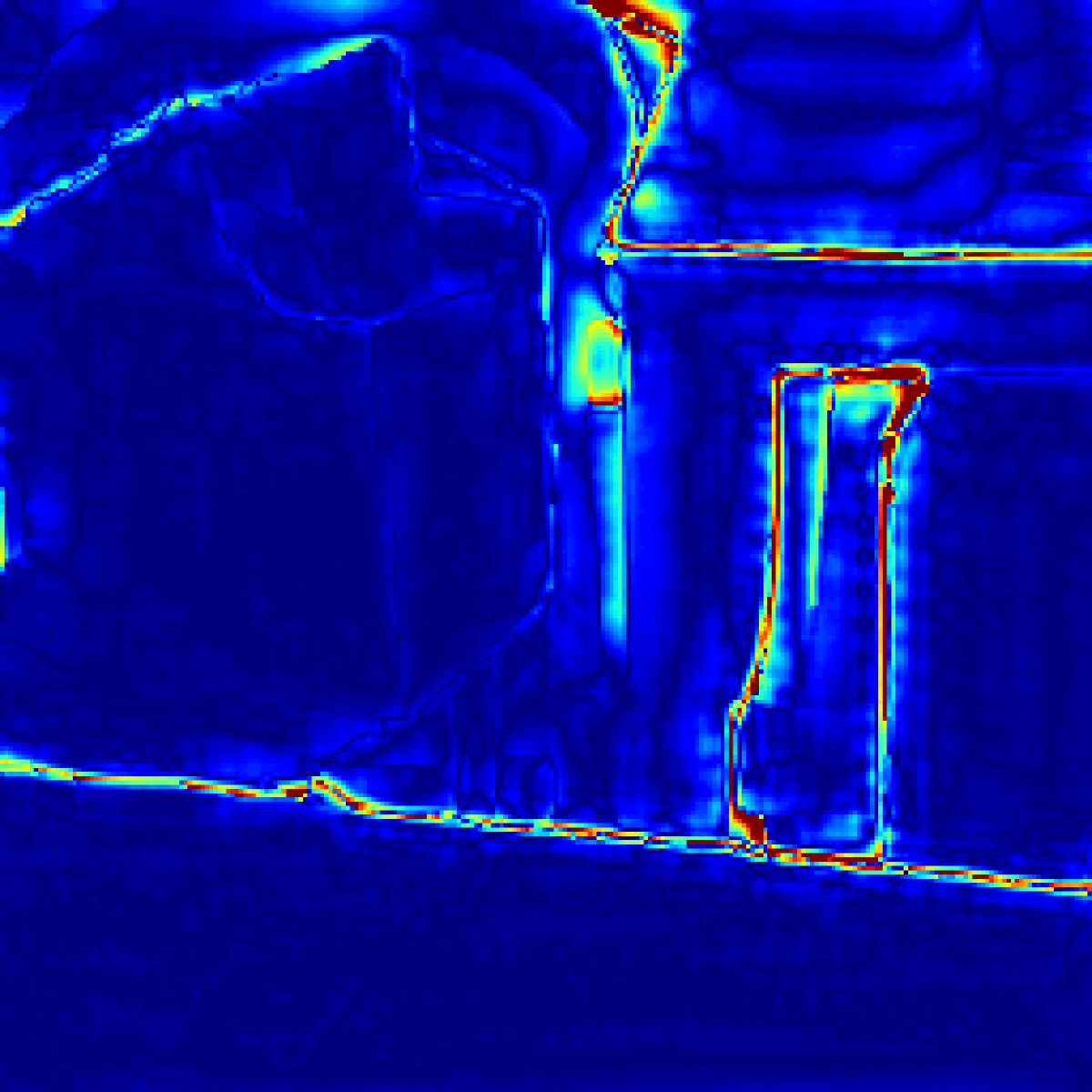}
	\hspace{-1.8mm} & \includegraphics[height=0.65in]{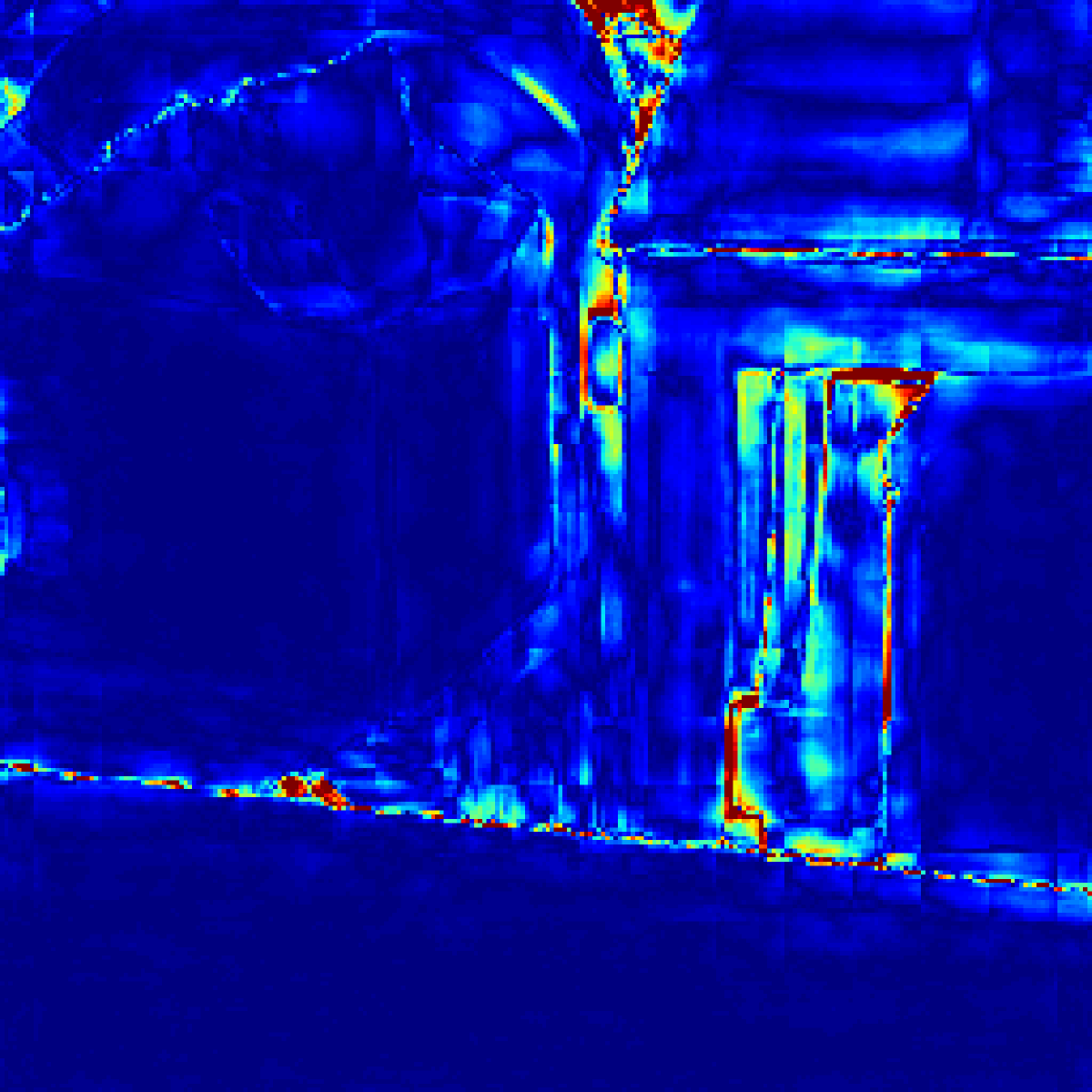}
	\hspace{-1.8mm} & \includegraphics[height=0.65in]{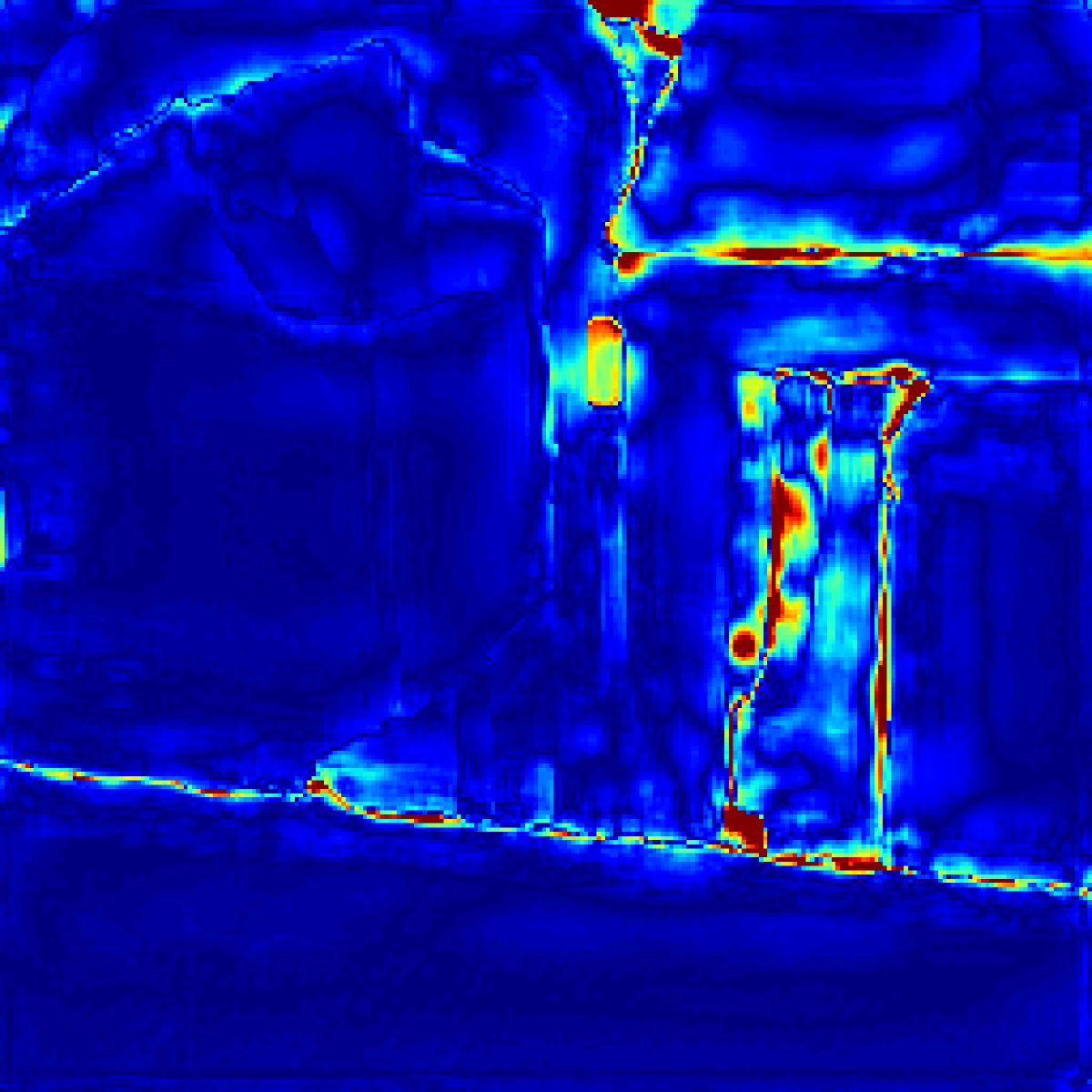}
	\hspace{-1.8mm} & \includegraphics[height=0.65in]{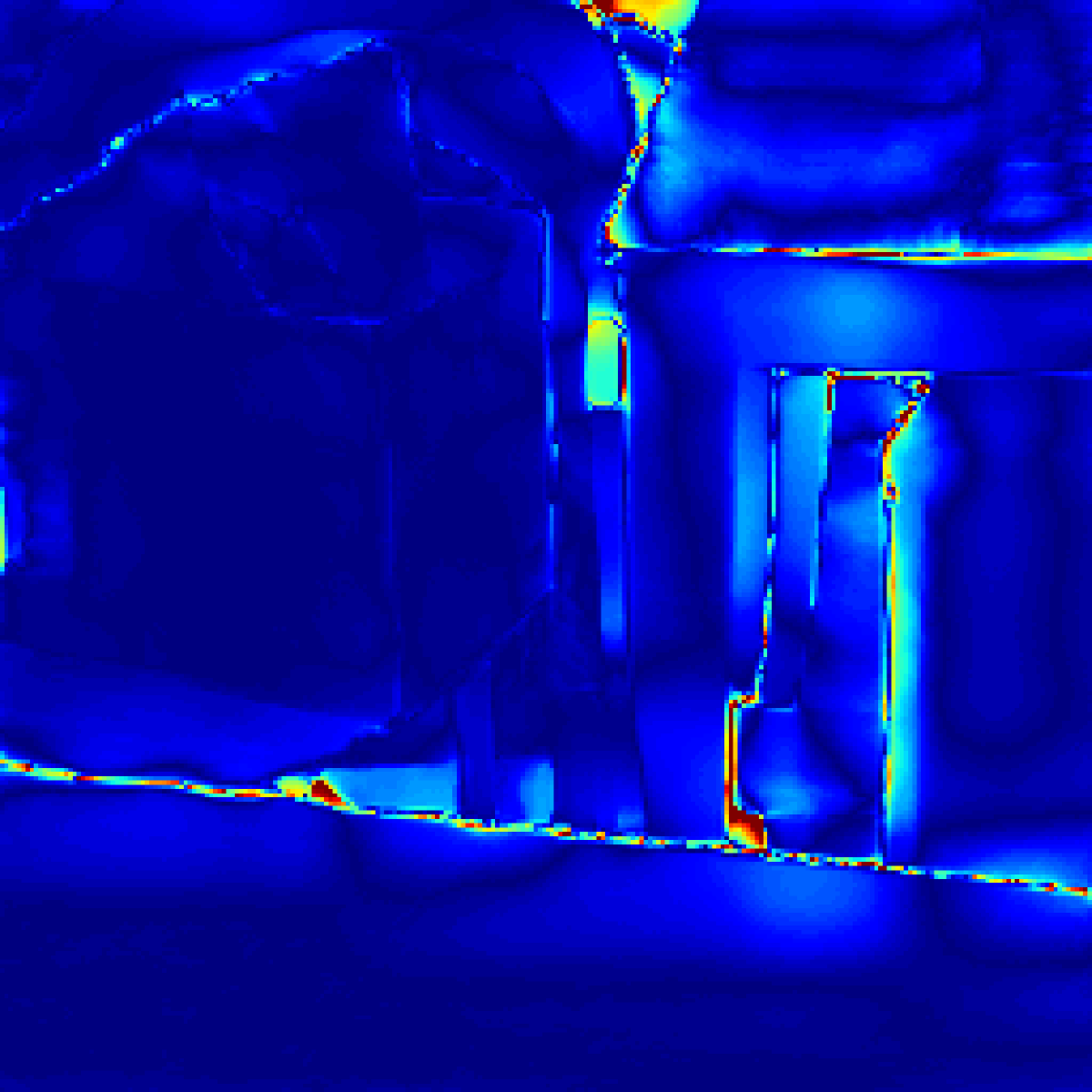}
	\hspace{-1.8mm} & \includegraphics[height=0.65in]{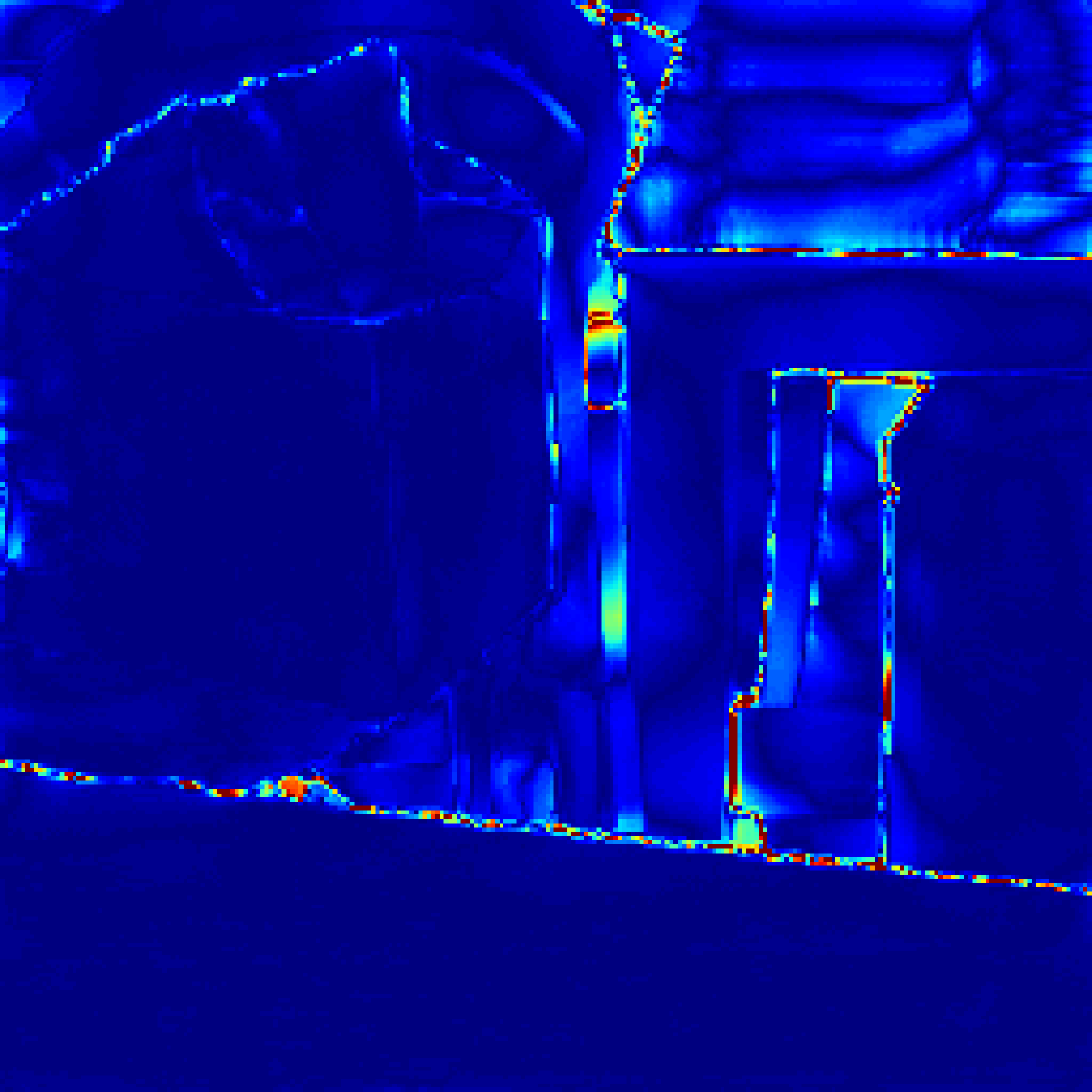}
 
	\hspace{-1.8mm} & \includegraphics[height=0.65in]{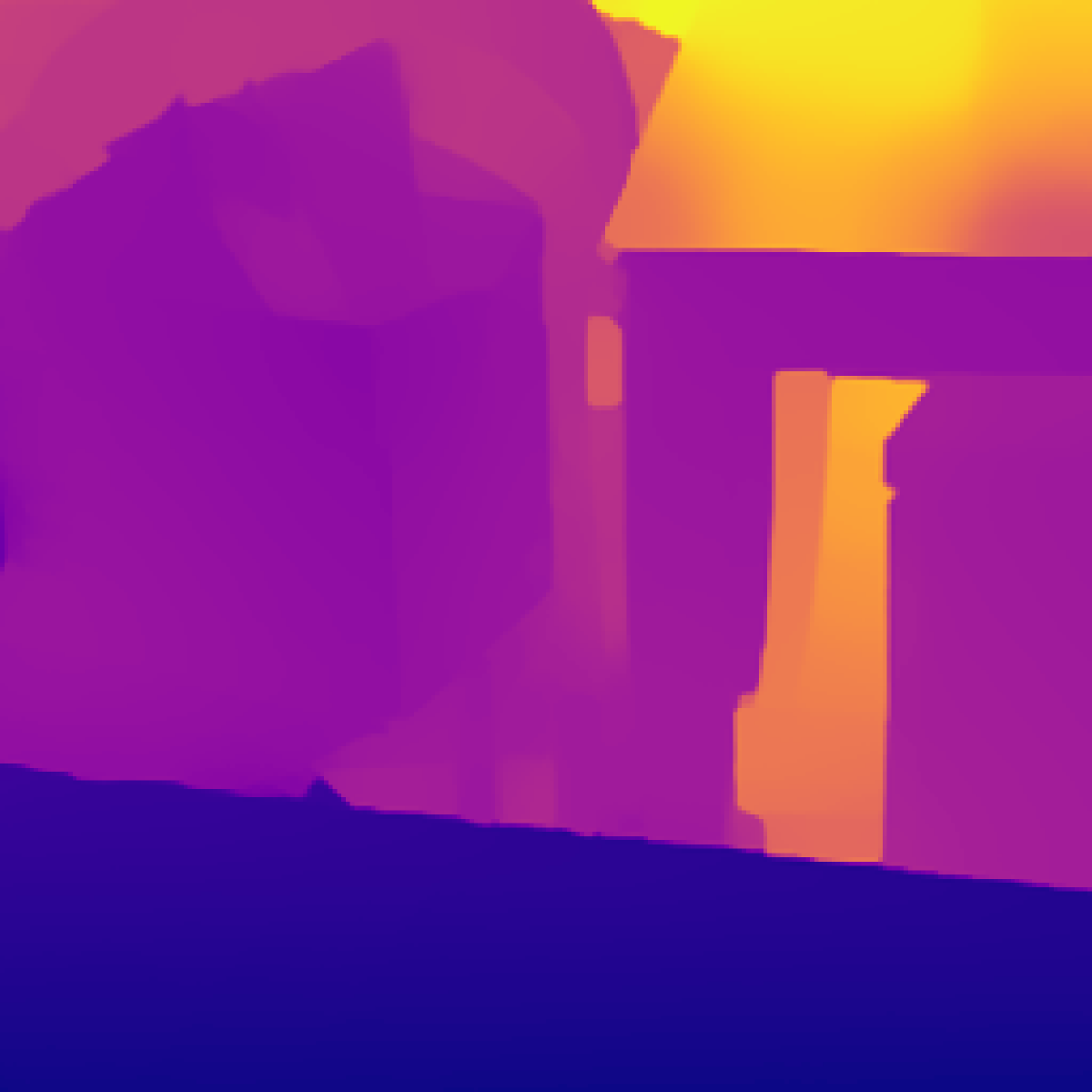}
        \\
        
        \rotatebox[origin=l]{90}{\scriptsize \quad \textbf{DIML}} & \includegraphics[height=0.65in]{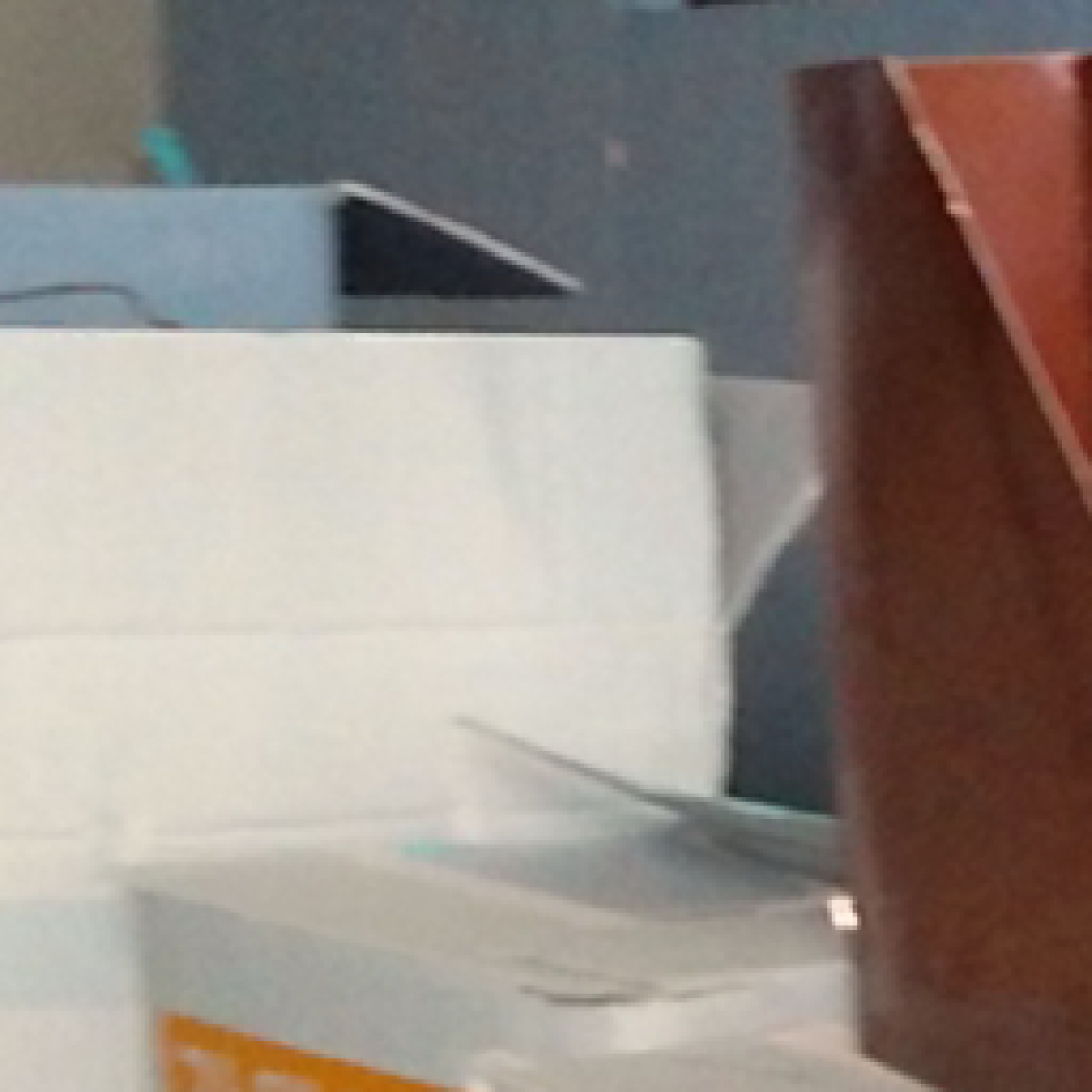}
	\hspace{-1.8mm} & \includegraphics[height=0.65in]{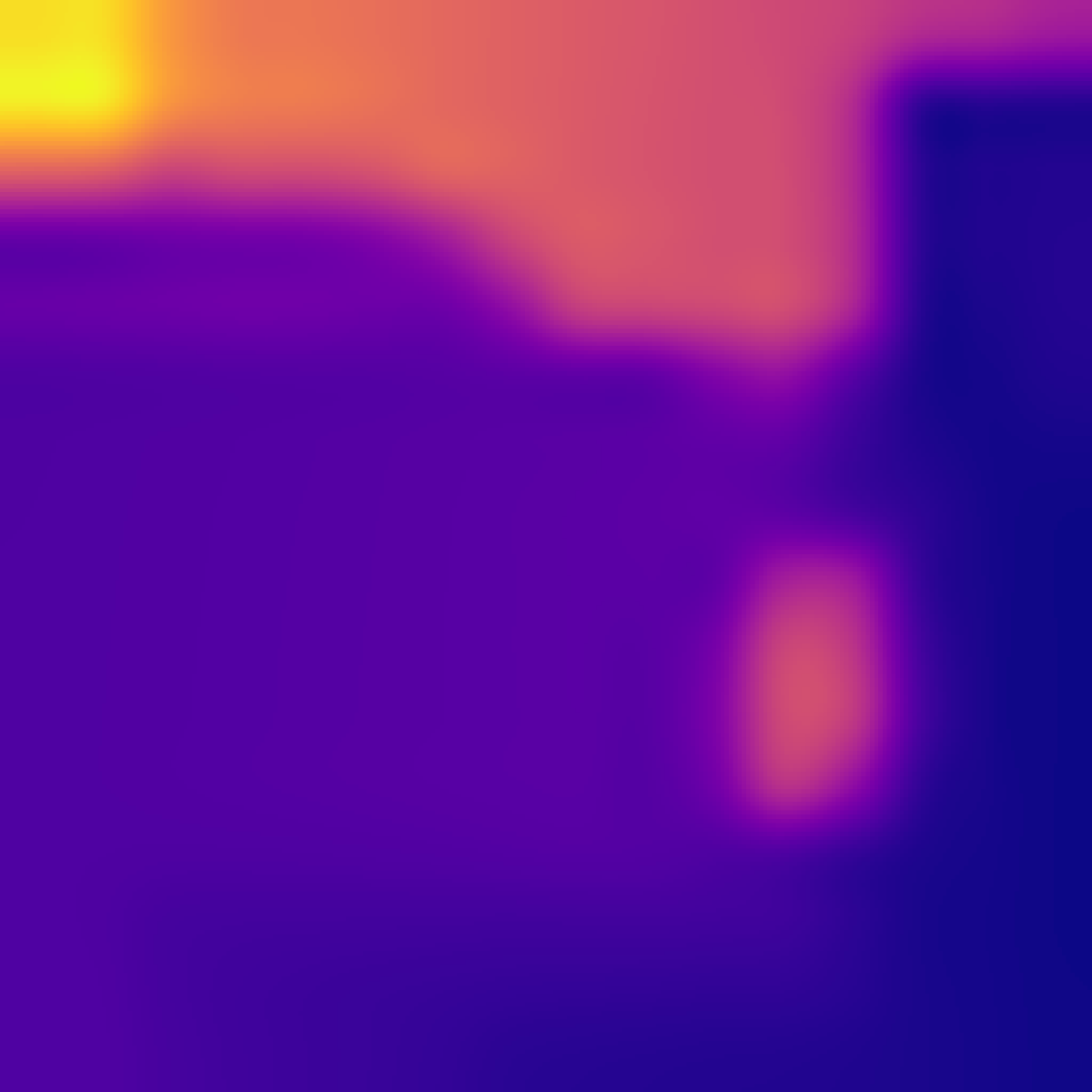}
	\hspace{-1.8mm} & \includegraphics[height=0.65in]{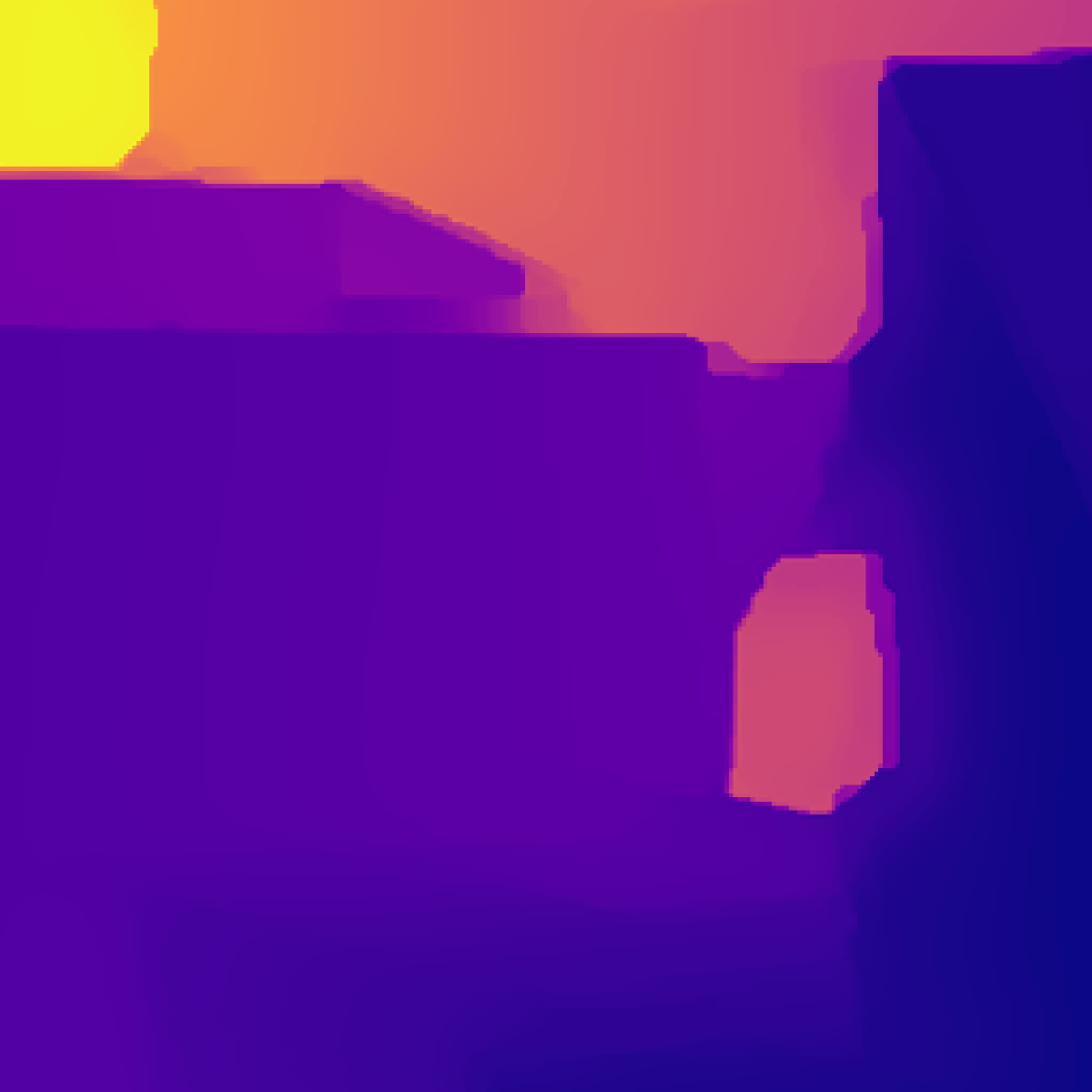}
	\hspace{-1.8mm} & \includegraphics[height=0.65in]{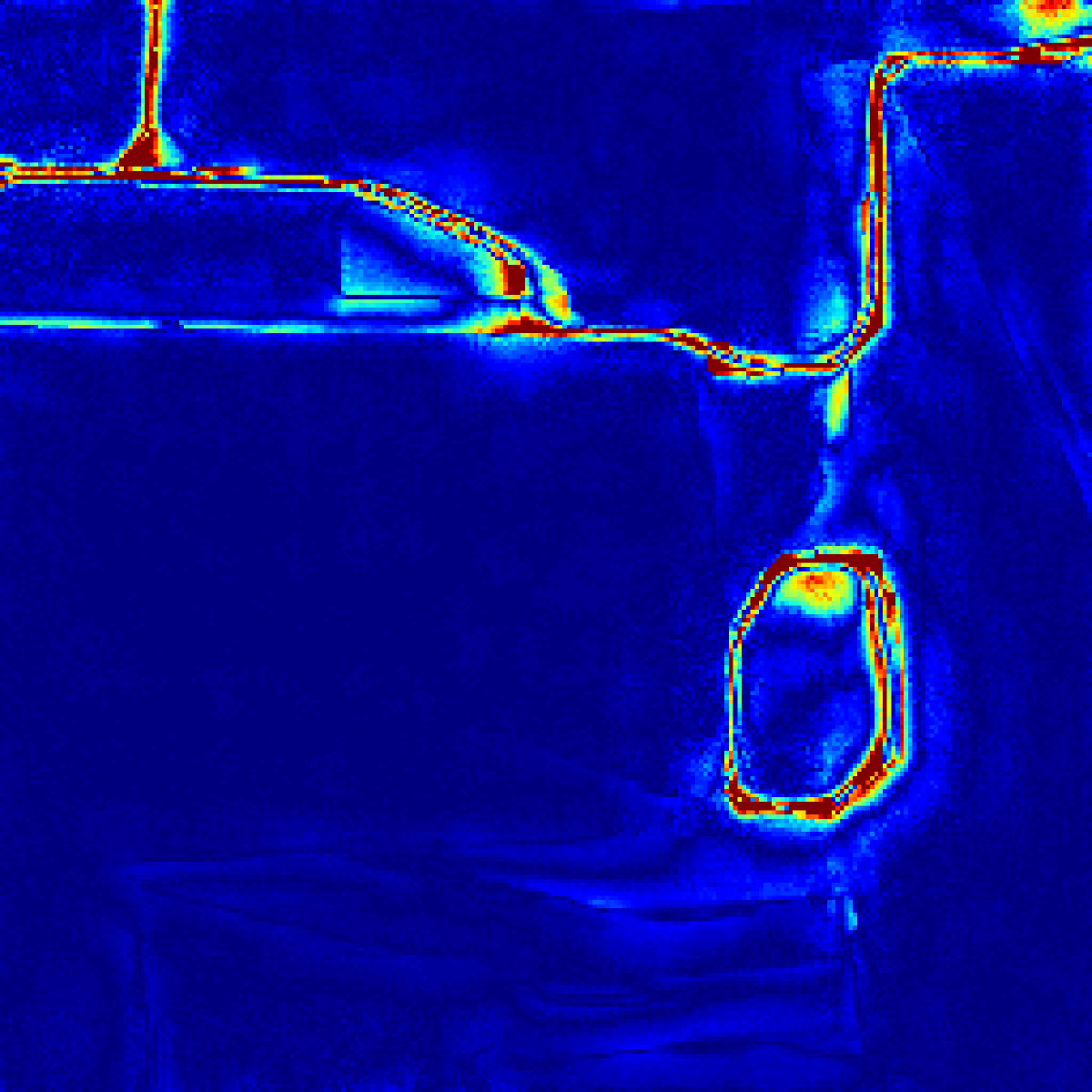}
	\hspace{-1.8mm} & \includegraphics[height=0.65in]{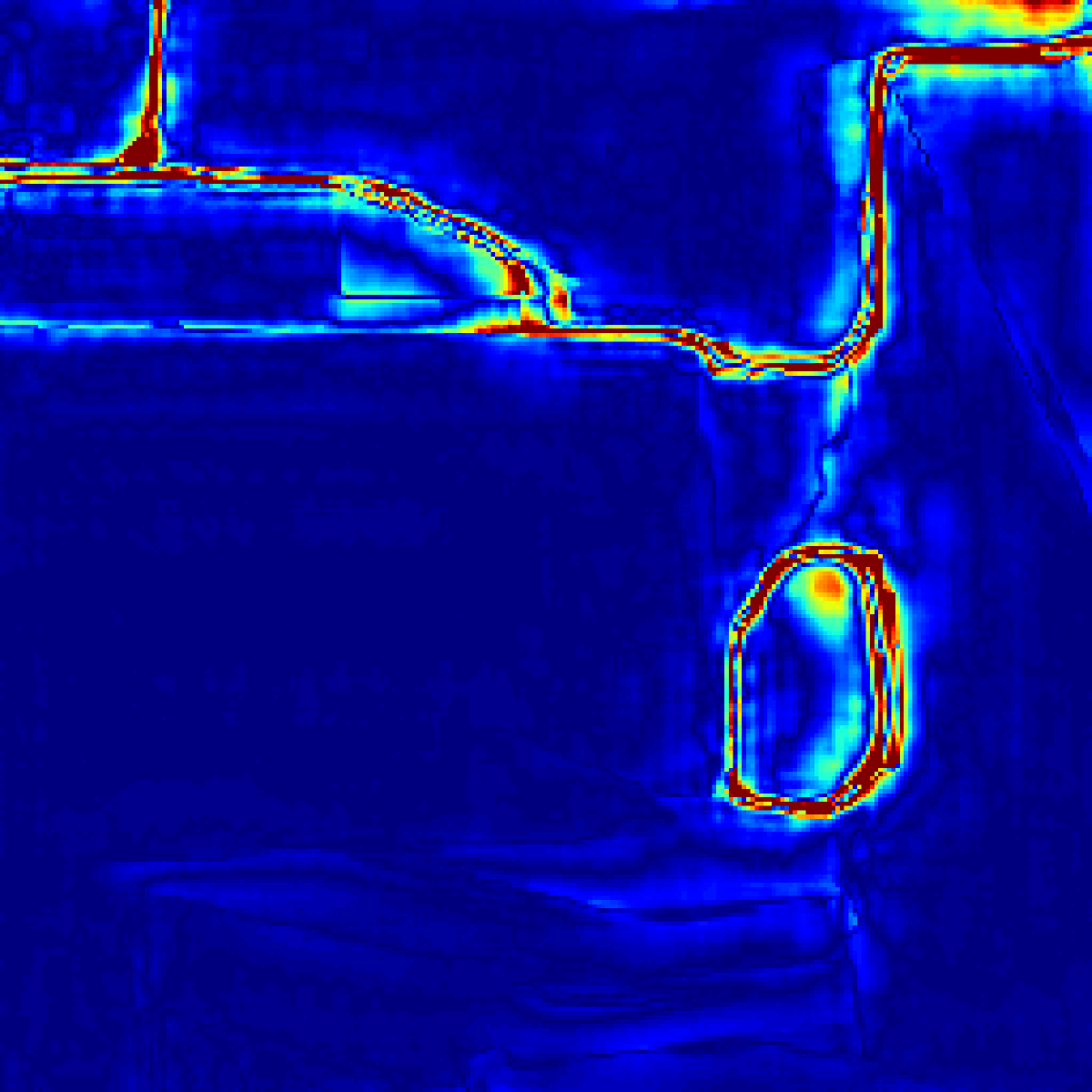}
	\hspace{-1.8mm} & \includegraphics[height=0.65in]{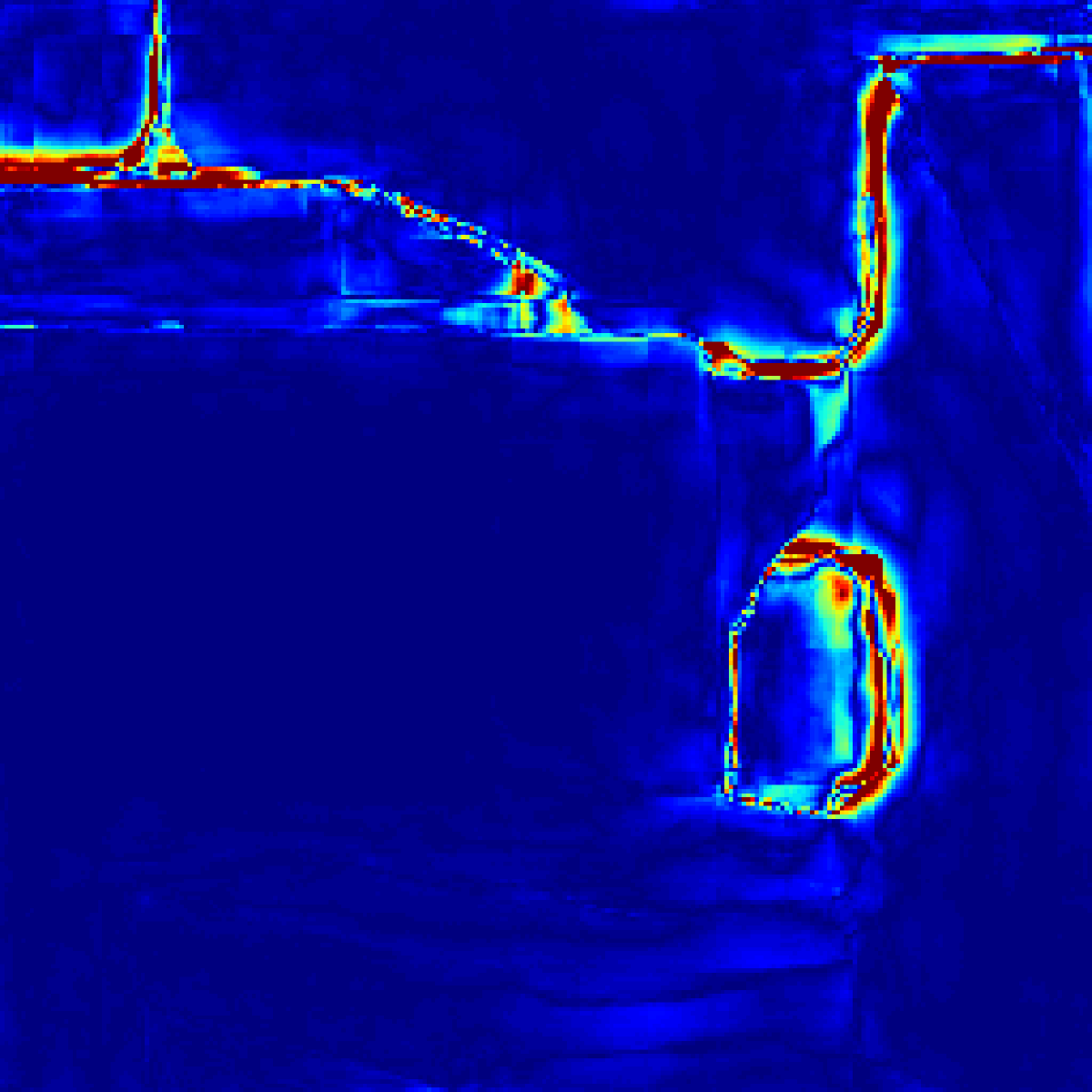}
	\hspace{-1.8mm} & \includegraphics[height=0.65in]{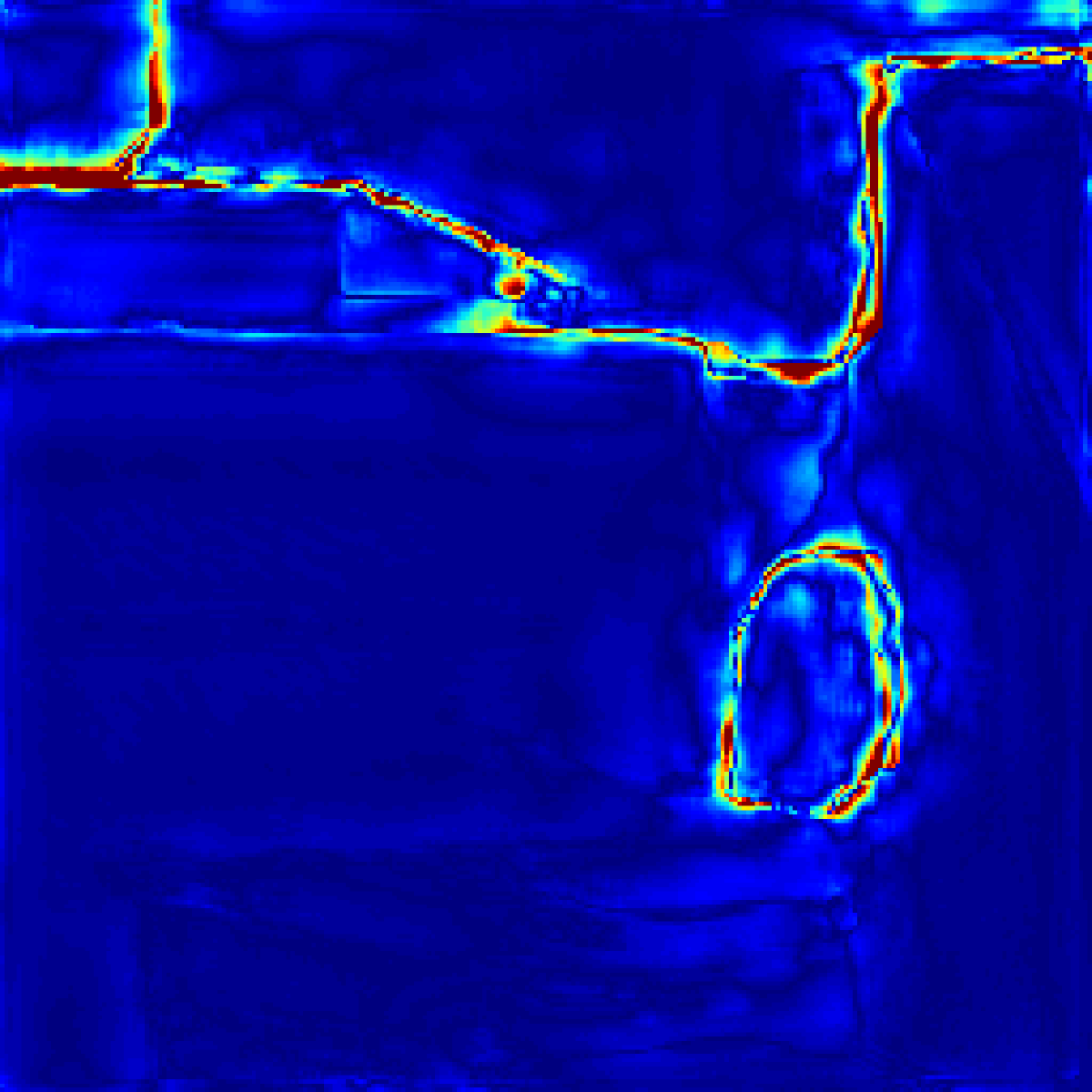}
	\hspace{-1.8mm} & \includegraphics[height=0.65in]{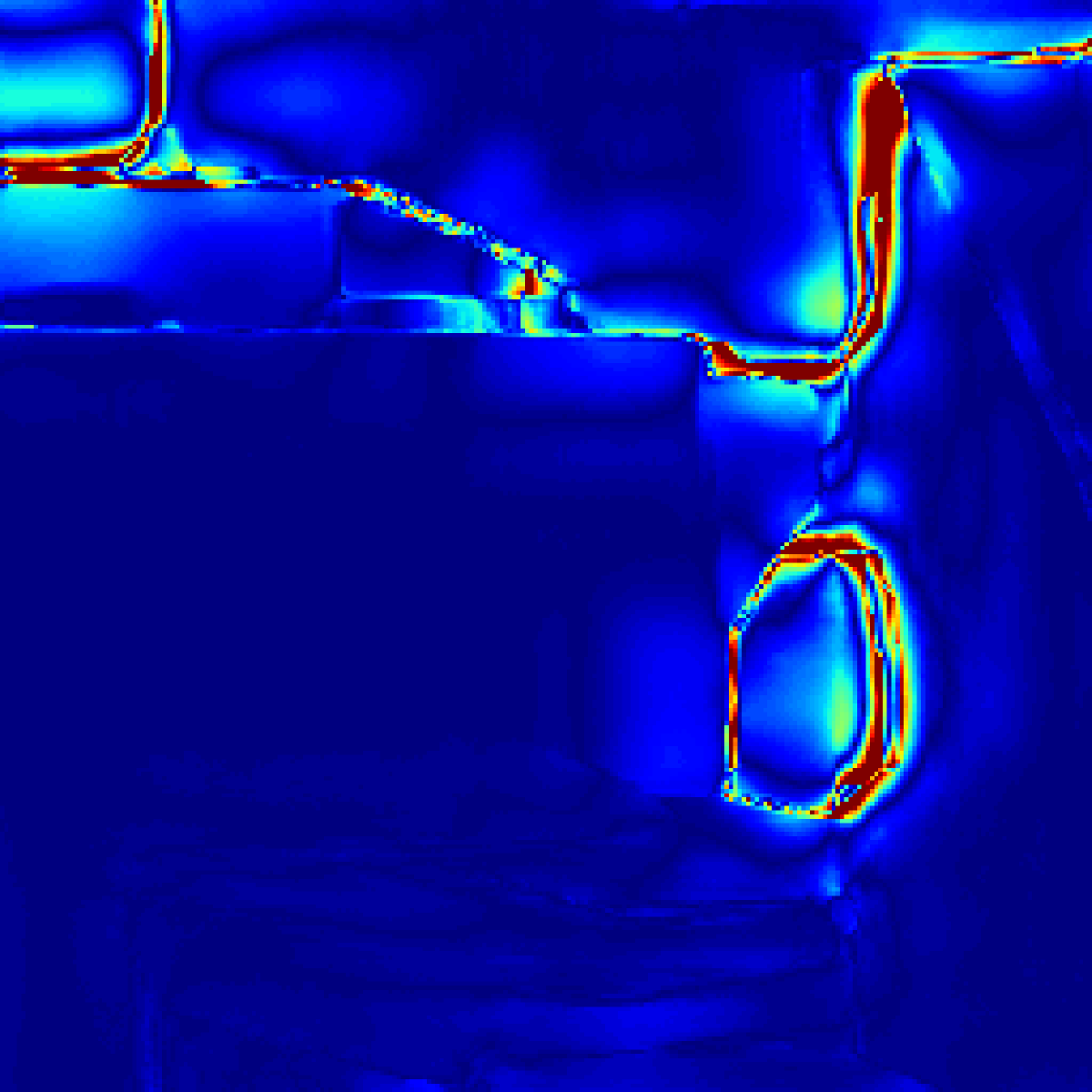}
	\hspace{-1.8mm} & \includegraphics[height=0.65in]{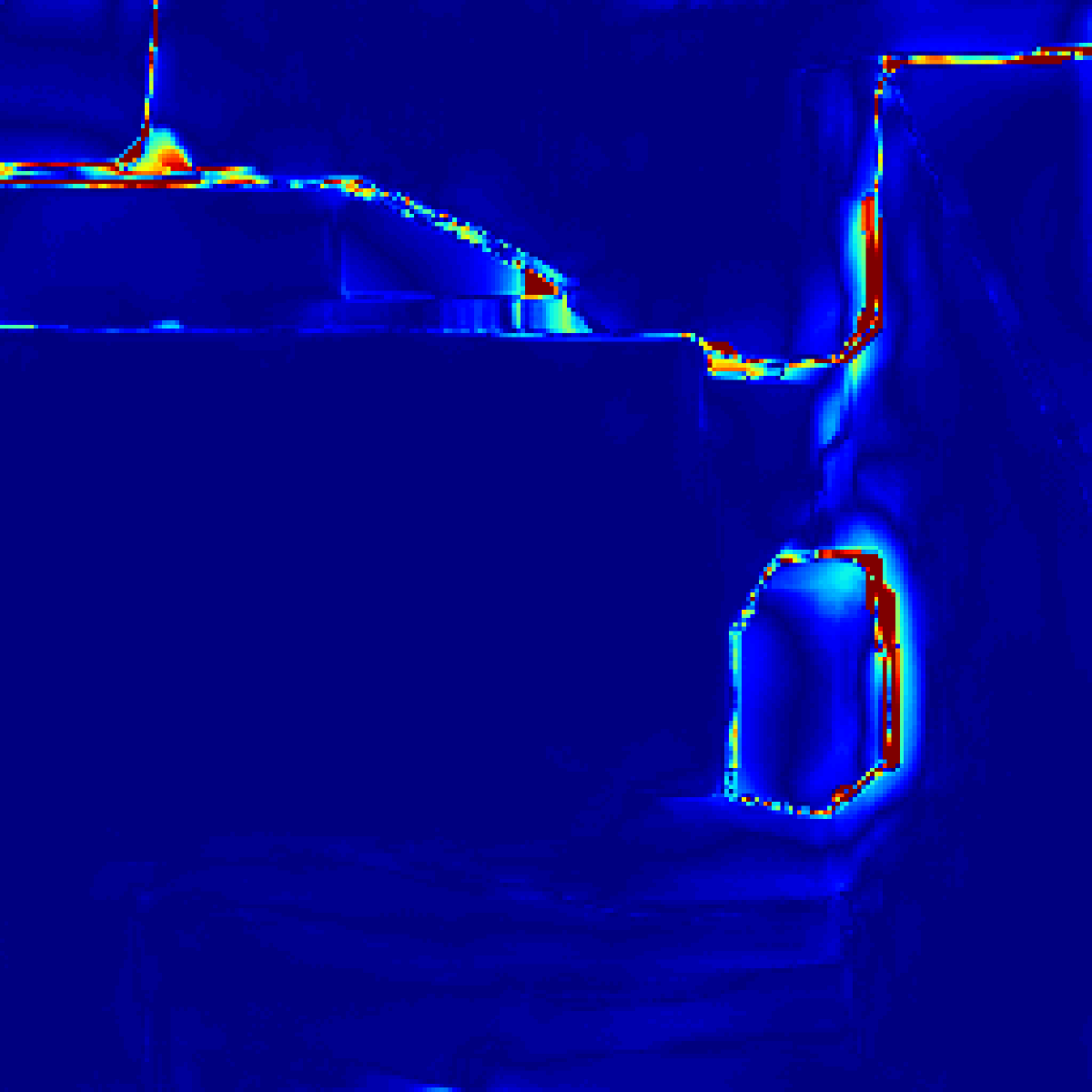}
 
	\hspace{-1.8mm} & \includegraphics[height=0.65in]{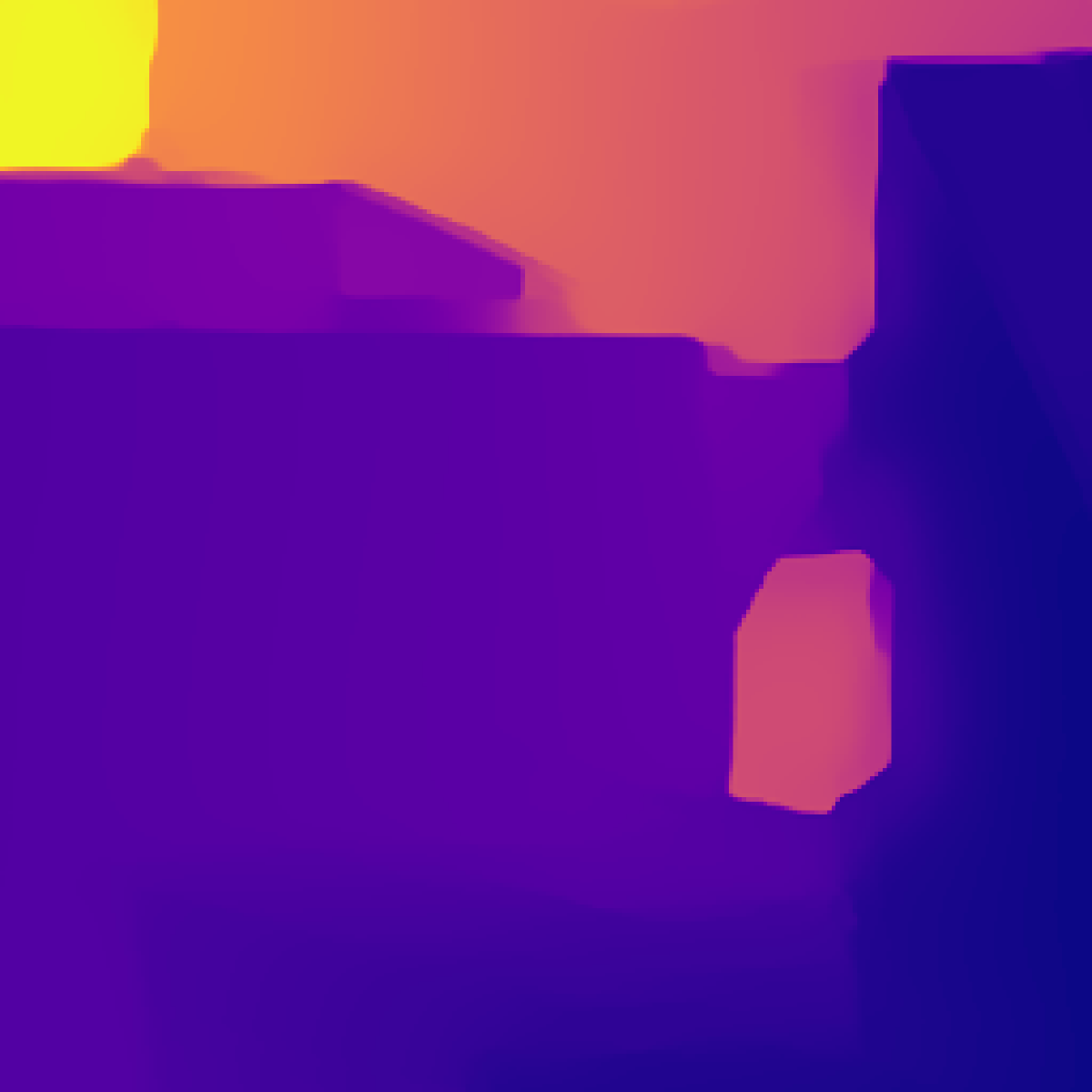}
 \\
	& \scriptsize \textbf{(a)} RGB & \scriptsize \textbf{(b)} Bicubic & \scriptsize \textbf{(c)} GT & \scriptsize \textbf{(d)} PMBA & \scriptsize \textbf{(e)} FDSR & \scriptsize \textbf{(f)} JIIF & \scriptsize \textbf{(g)} DCTNet & \scriptsize \textbf{(h)} LGR & \scriptsize \textbf{(i)} \netname{} & \scriptsize \textbf{(j)} \netname{} (depth)
	\end{tabular}
    \vspace{-0.3cm}
	\caption{\textbf{Qualitative comparison on the Middlebury, NYUv2, and DIML.} From left to right: (a) RGB image, (b) Bicubic upsampled depth map, (c) GT; then, error maps achieved by selected methods: (d) PMBA~\cite{ye2020pmbanet}, (e) FDSR~\cite{he2021towards}, (f) JIIF~\cite{tang2021joint}, (g) DCTNet~\cite{zhao2022discrete}, (h) LGR~\cite{de2022learning}; finally, (i) error maps and (j) predictions by \netname.} 
	\label{fig:sota_comp1}
\end{figure*}

\begin{figure*}[t] 
	\centering
	\renewcommand\tabcolsep{1.5pt} 
	\begin{tabular}{ccccccccc}
	\vspace{-0.1cm}
    
    \rotatebox[origin=l]{90}{\scriptsize \quad \textbf{8$\times$}} & \includegraphics[height=0.62in]{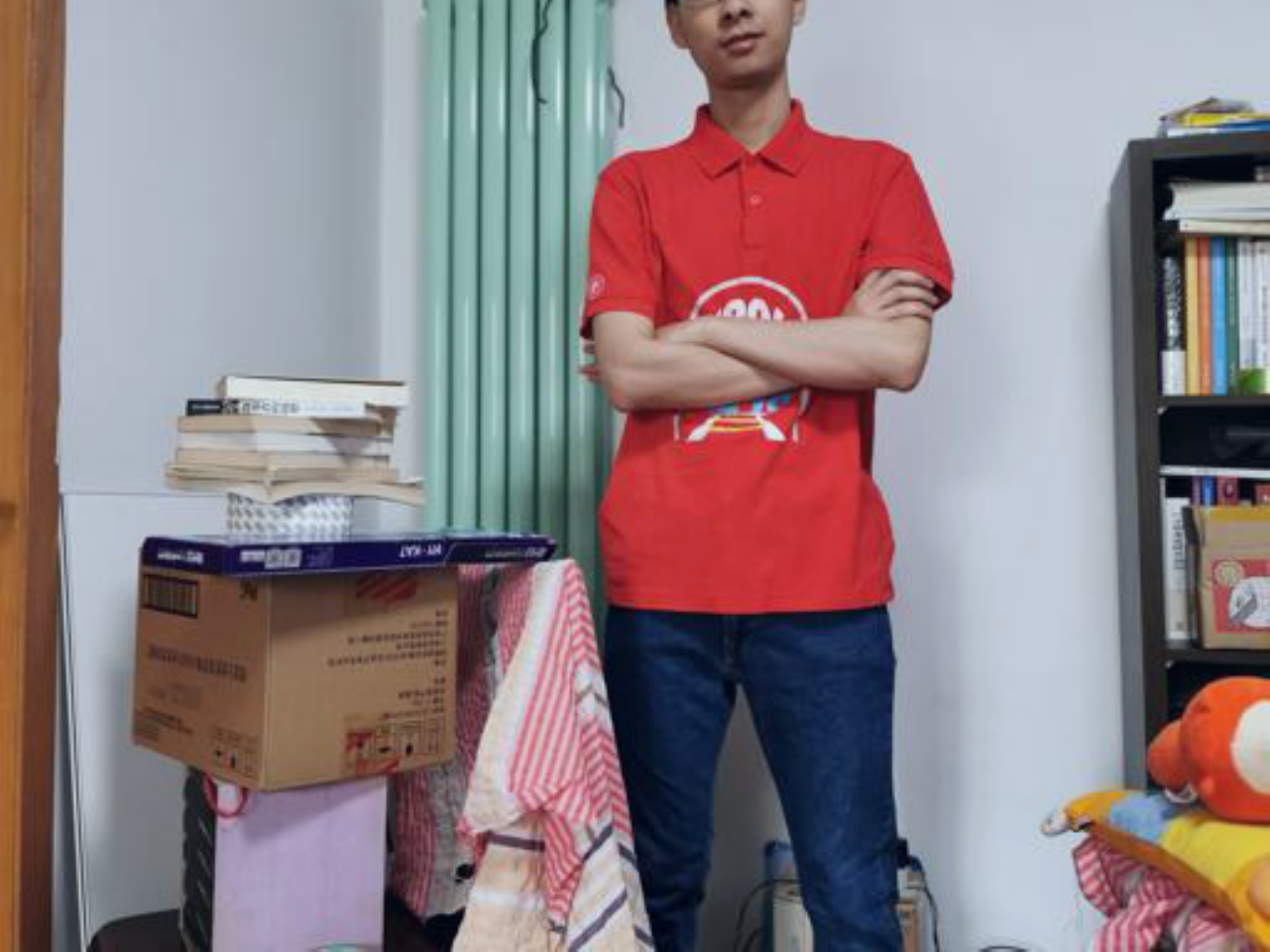}
	\hspace{-1.8mm} & \includegraphics[height=0.62in]{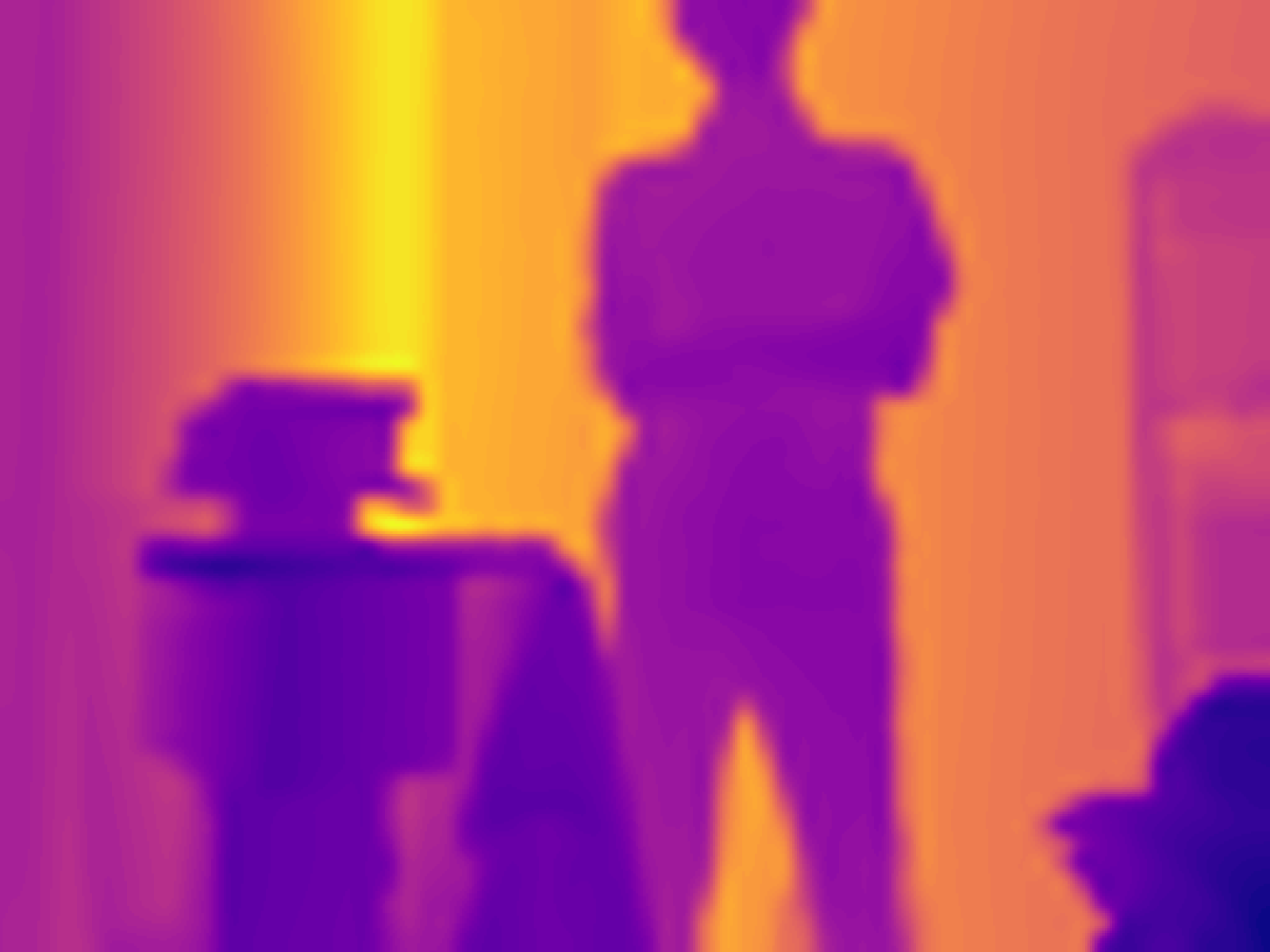}
	\hspace{-1.8mm} & \includegraphics[height=0.62in]{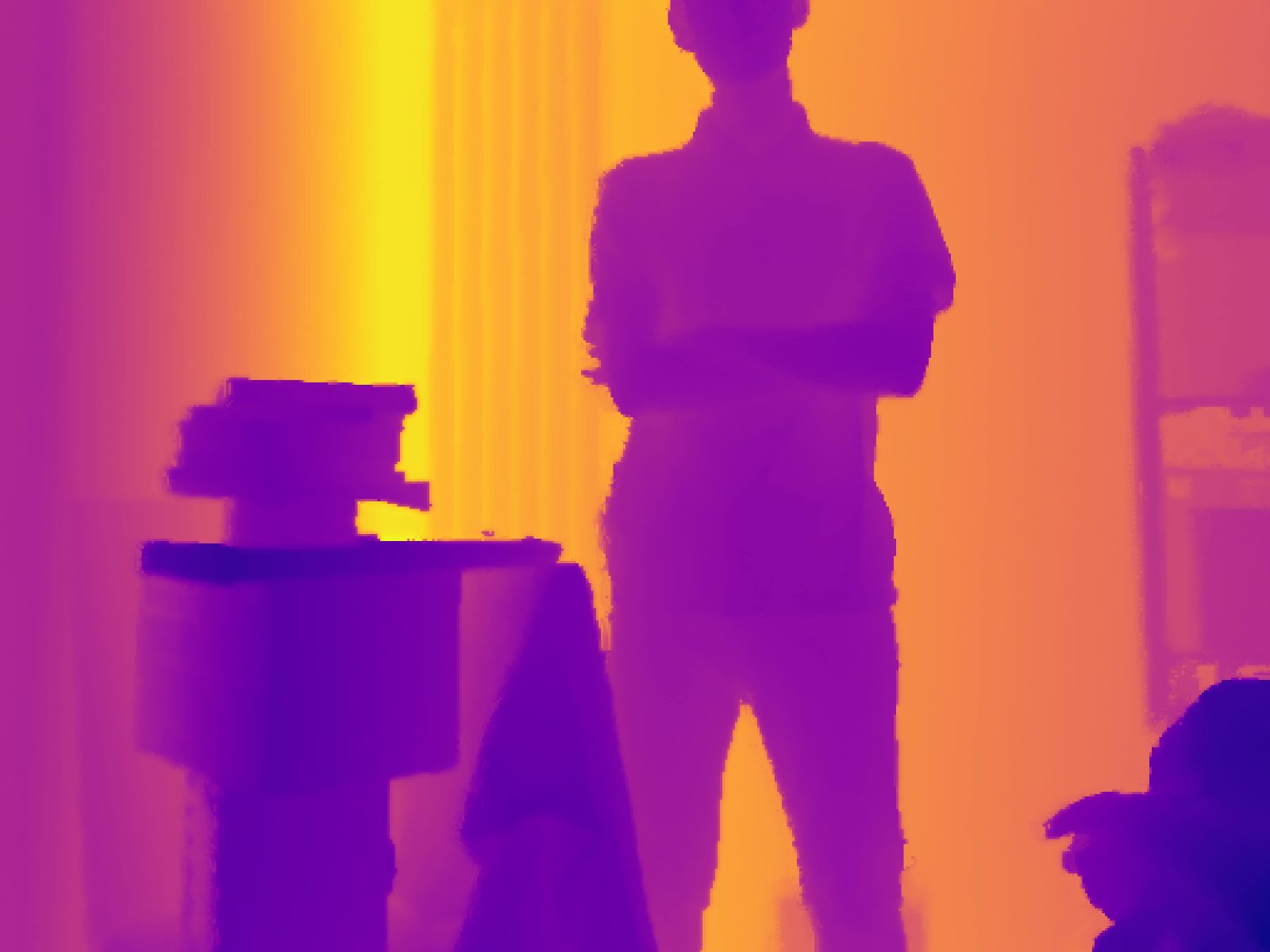}
	\hspace{-1.8mm} & \includegraphics[height=0.62in]{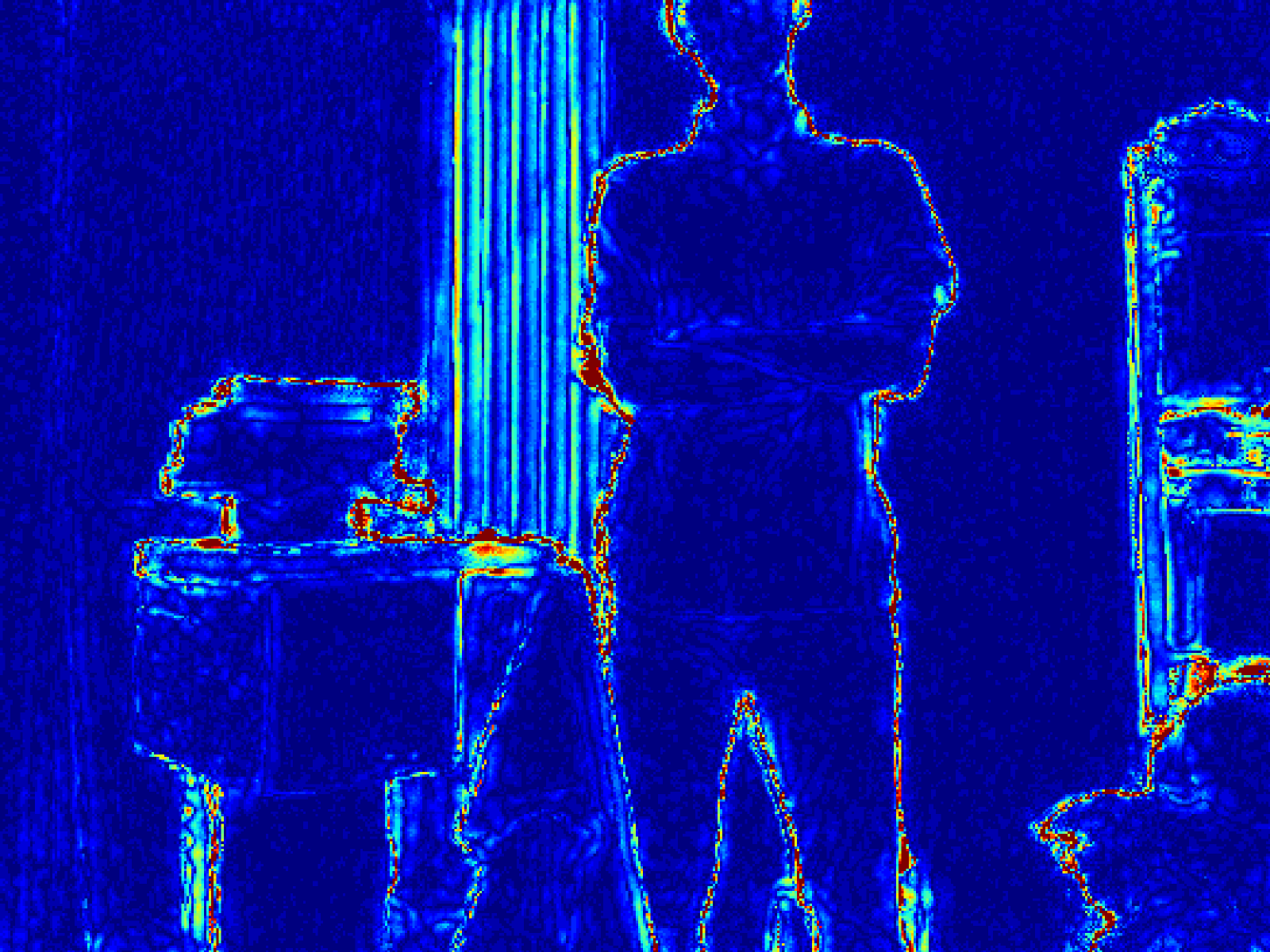}
	\hspace{-1.8mm} & \includegraphics[height=0.62in]{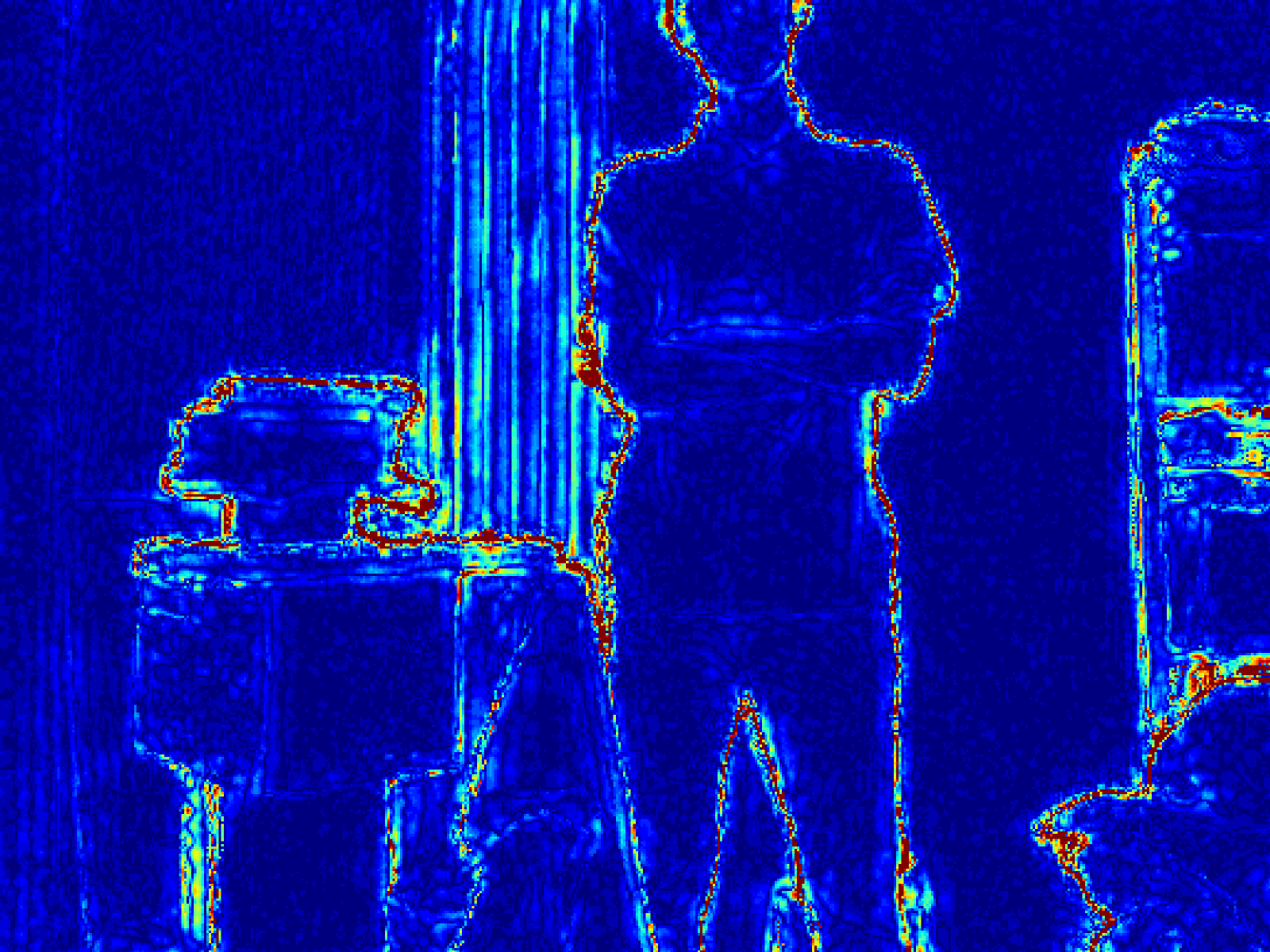}
	\hspace{-1.8mm} & \includegraphics[height=0.62in]{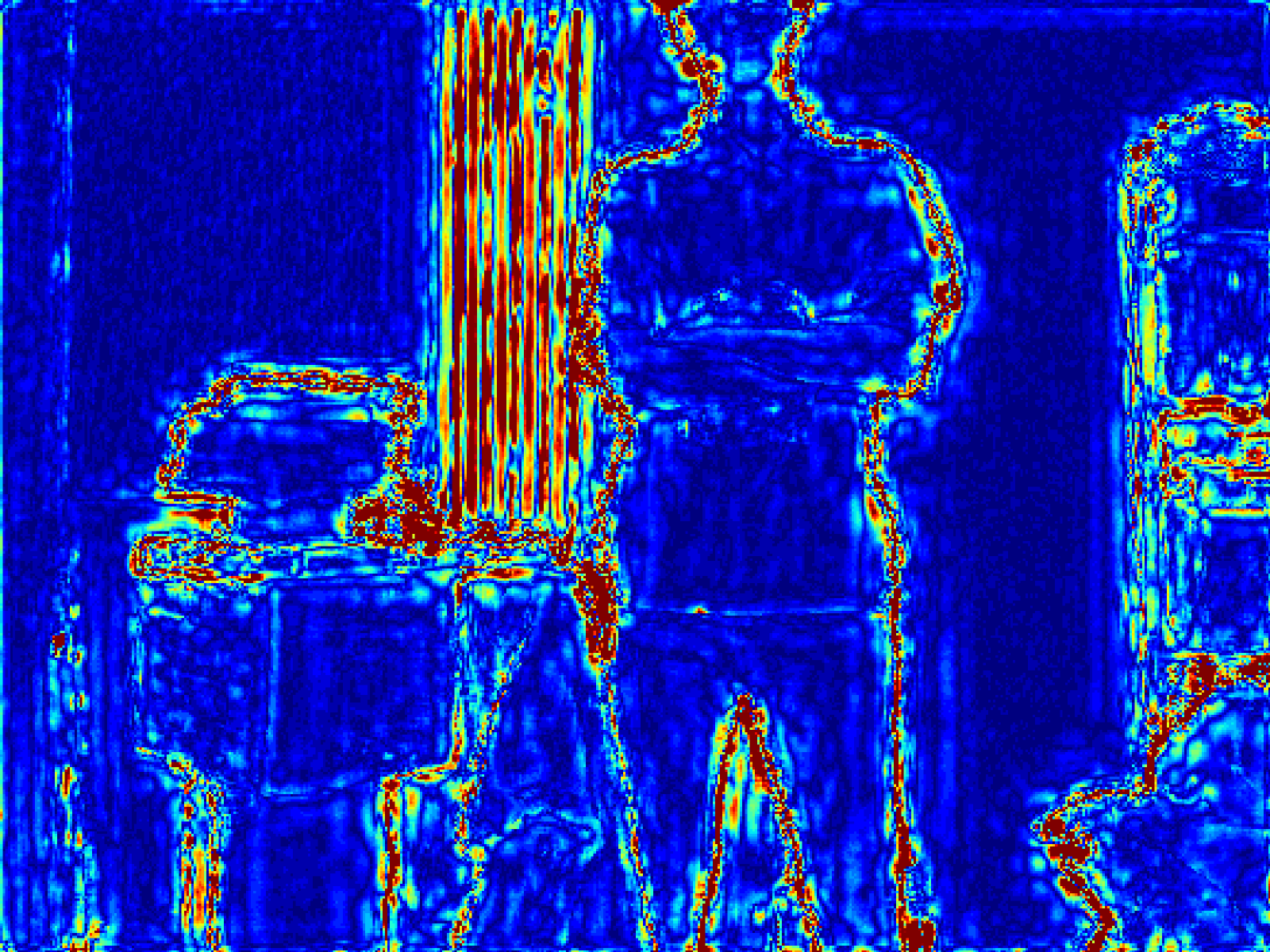}
	\hspace{-1.8mm} & \includegraphics[height=0.62in]{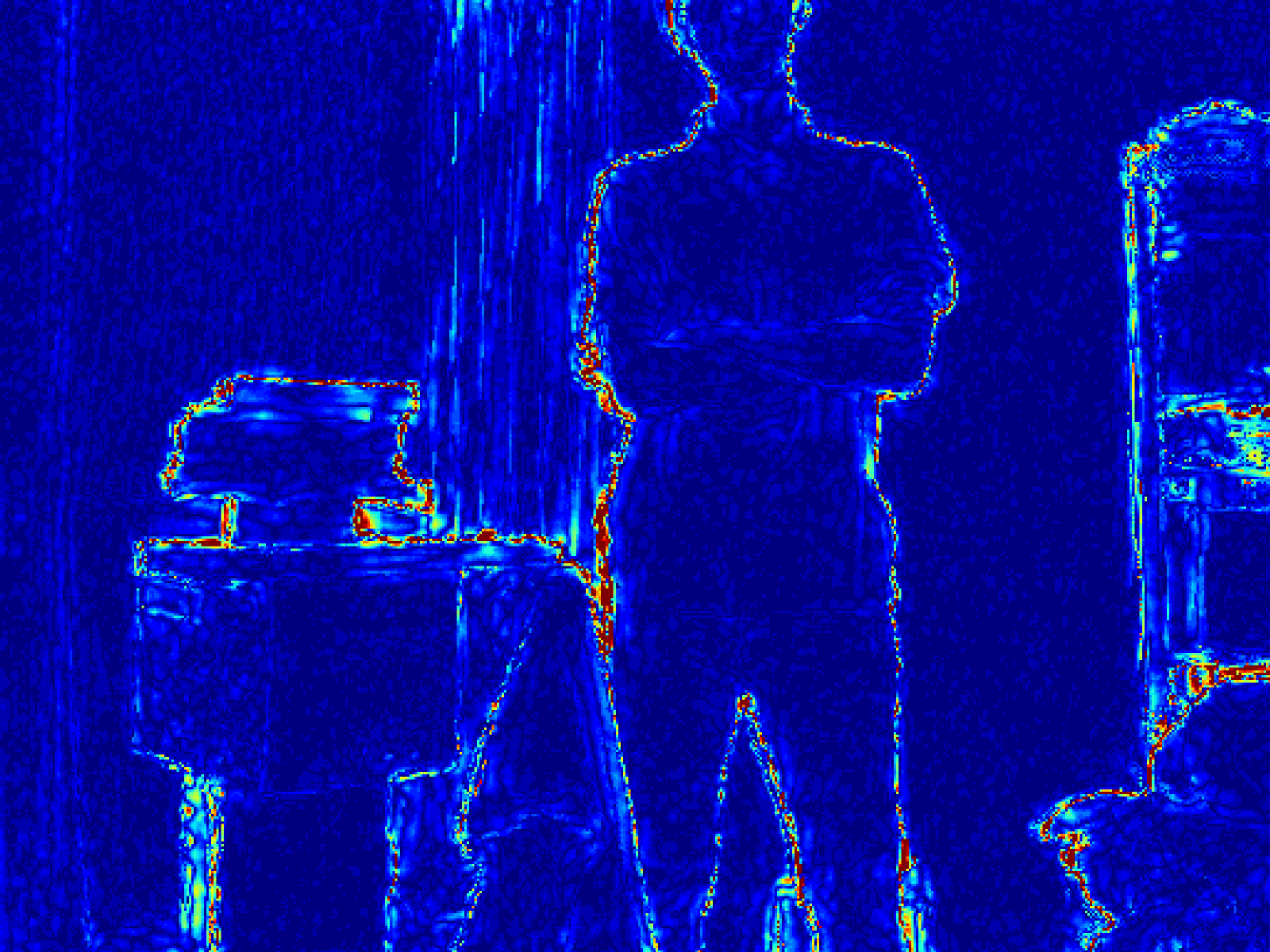}
 
	\hspace{-1.8mm} & \includegraphics[height=0.62in]{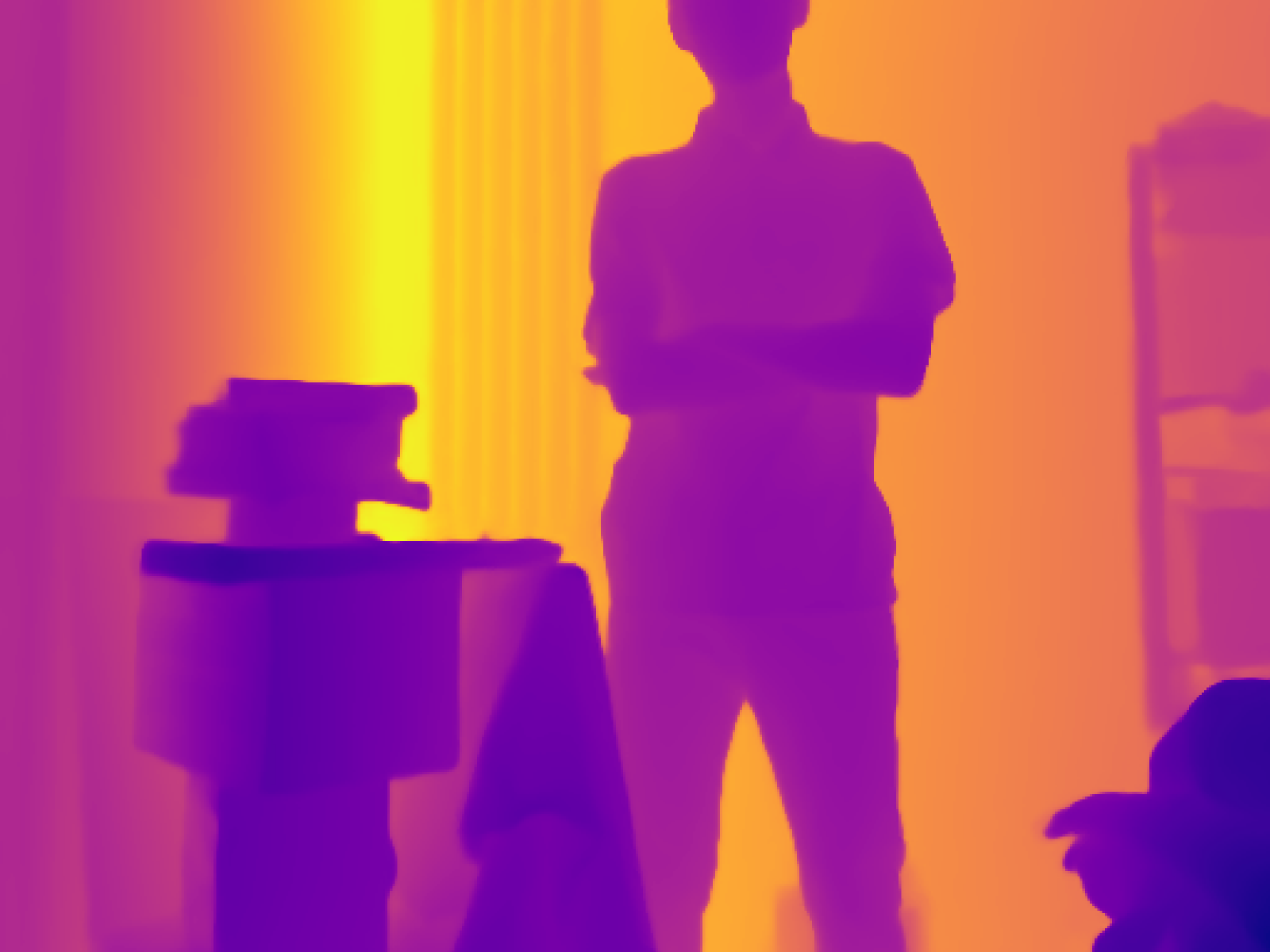}
	\\ 
	
    \rotatebox[origin=l]{90}{\scriptsize \quad \textbf{RGBDD}} & \includegraphics[height=0.62in]{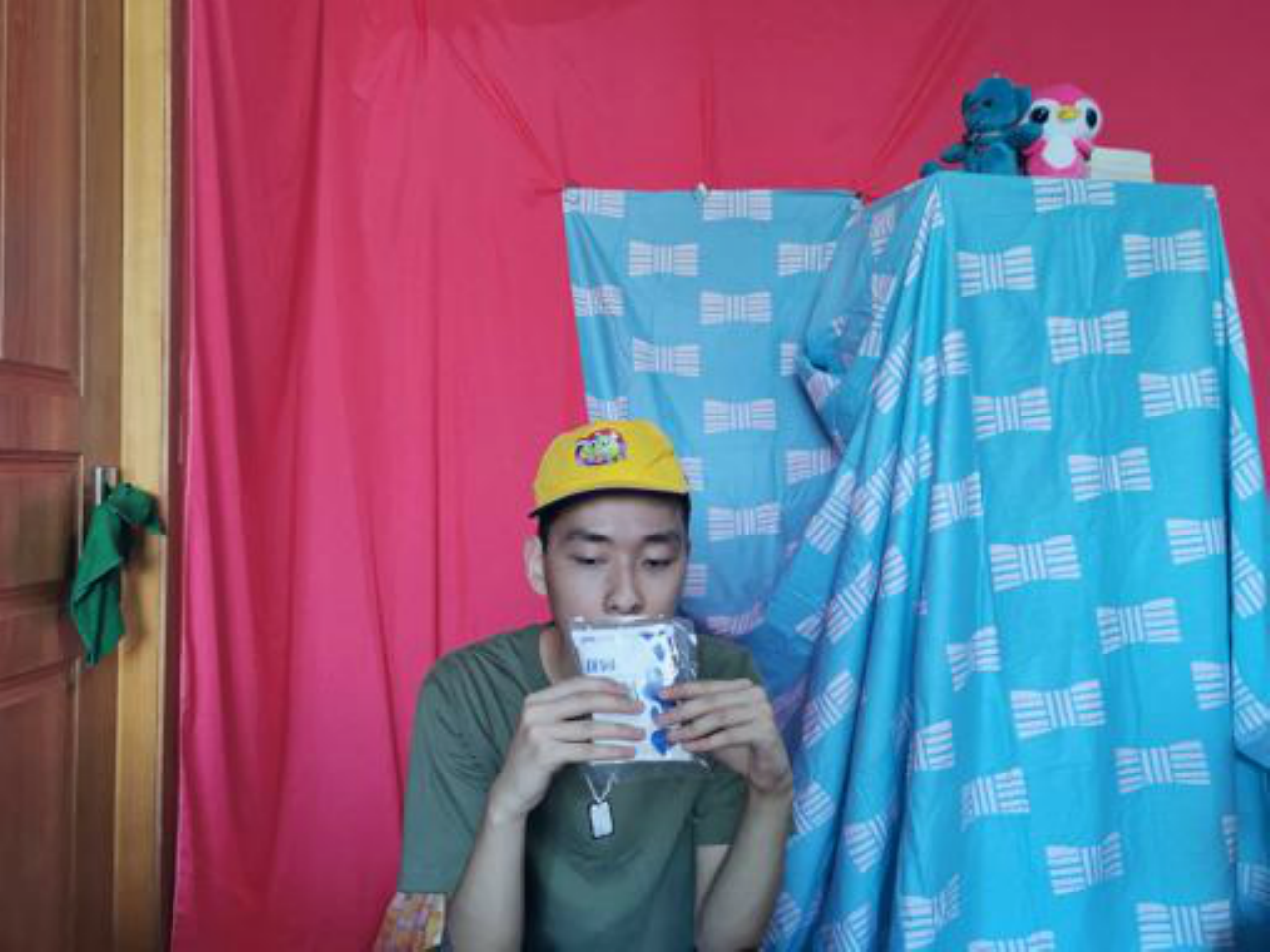}
	\hspace{-1.8mm} & \includegraphics[height=0.62in]{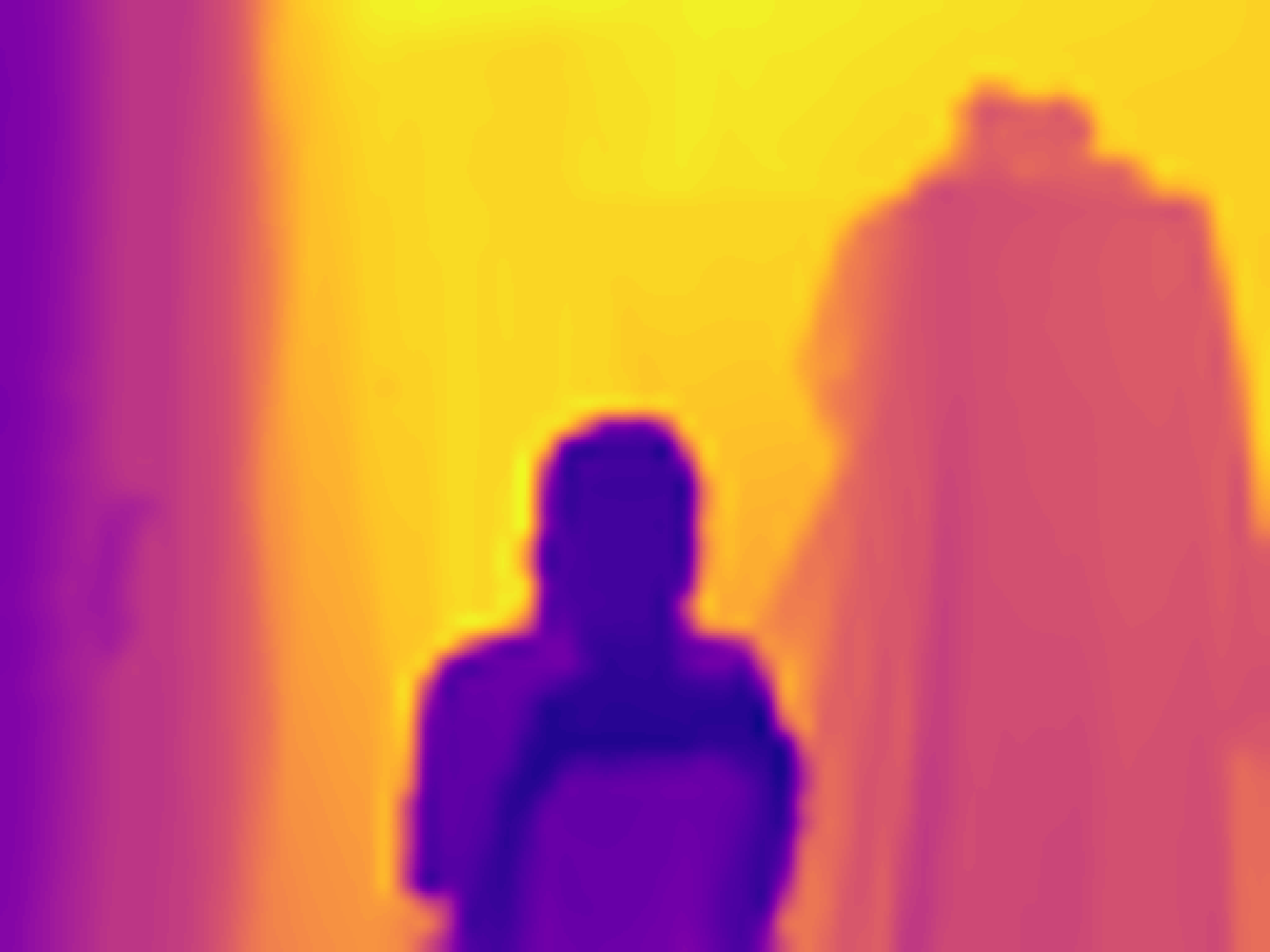}
	\hspace{-1.8mm} & \includegraphics[height=0.62in]{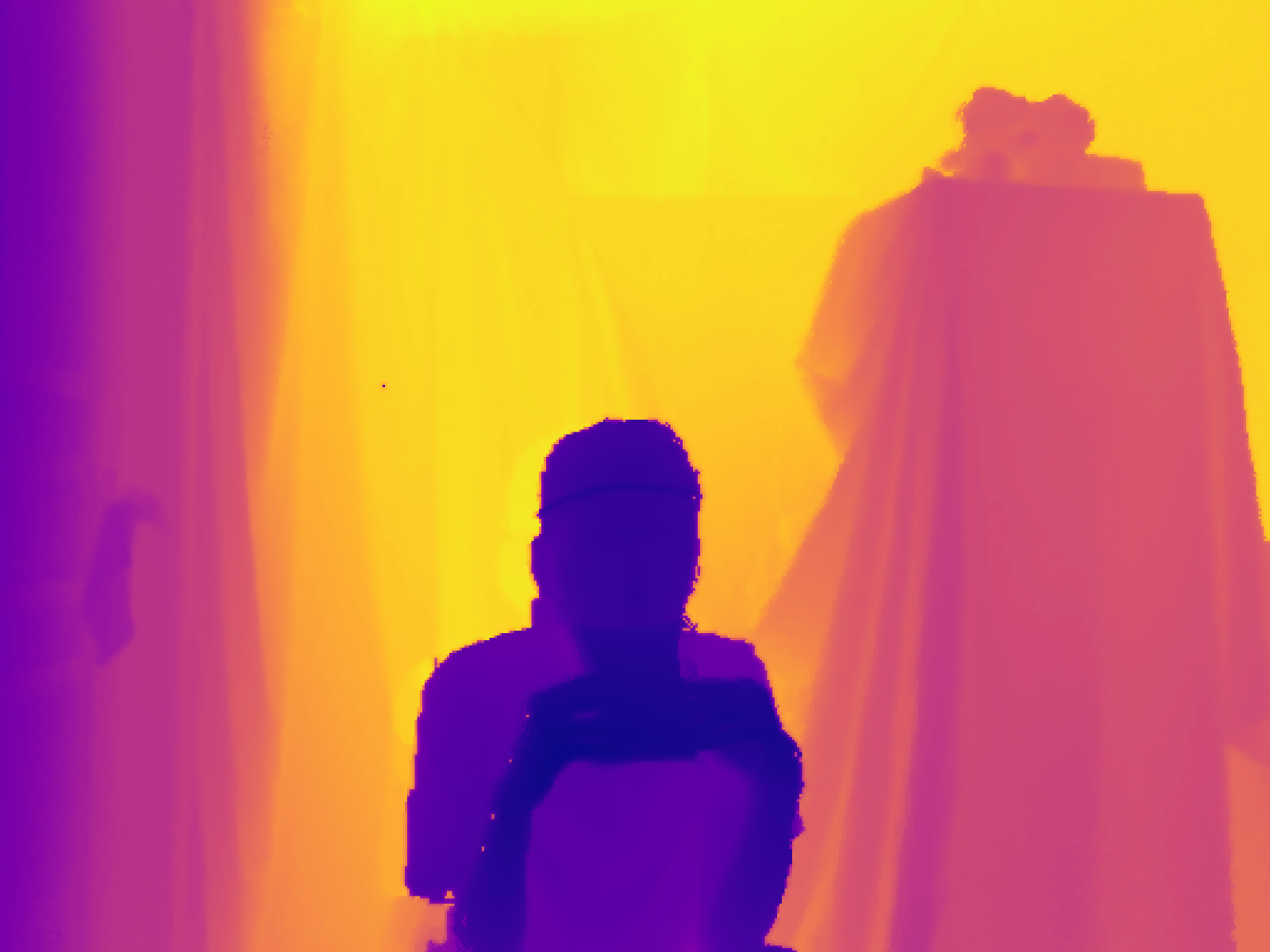}
	\hspace{-1.8mm} & \includegraphics[height=0.62in]{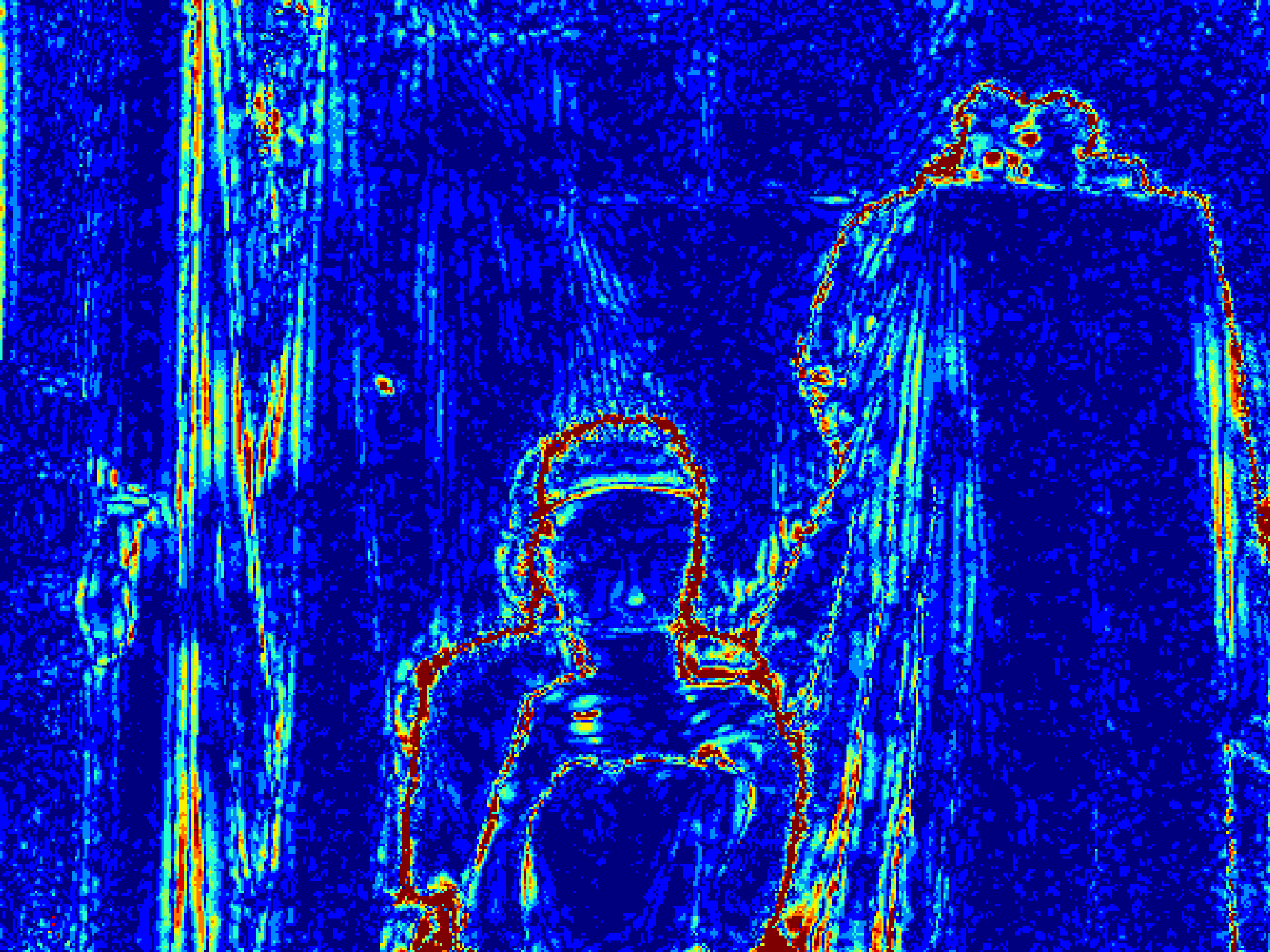}
	\hspace{-1.8mm} & \includegraphics[height=0.62in]{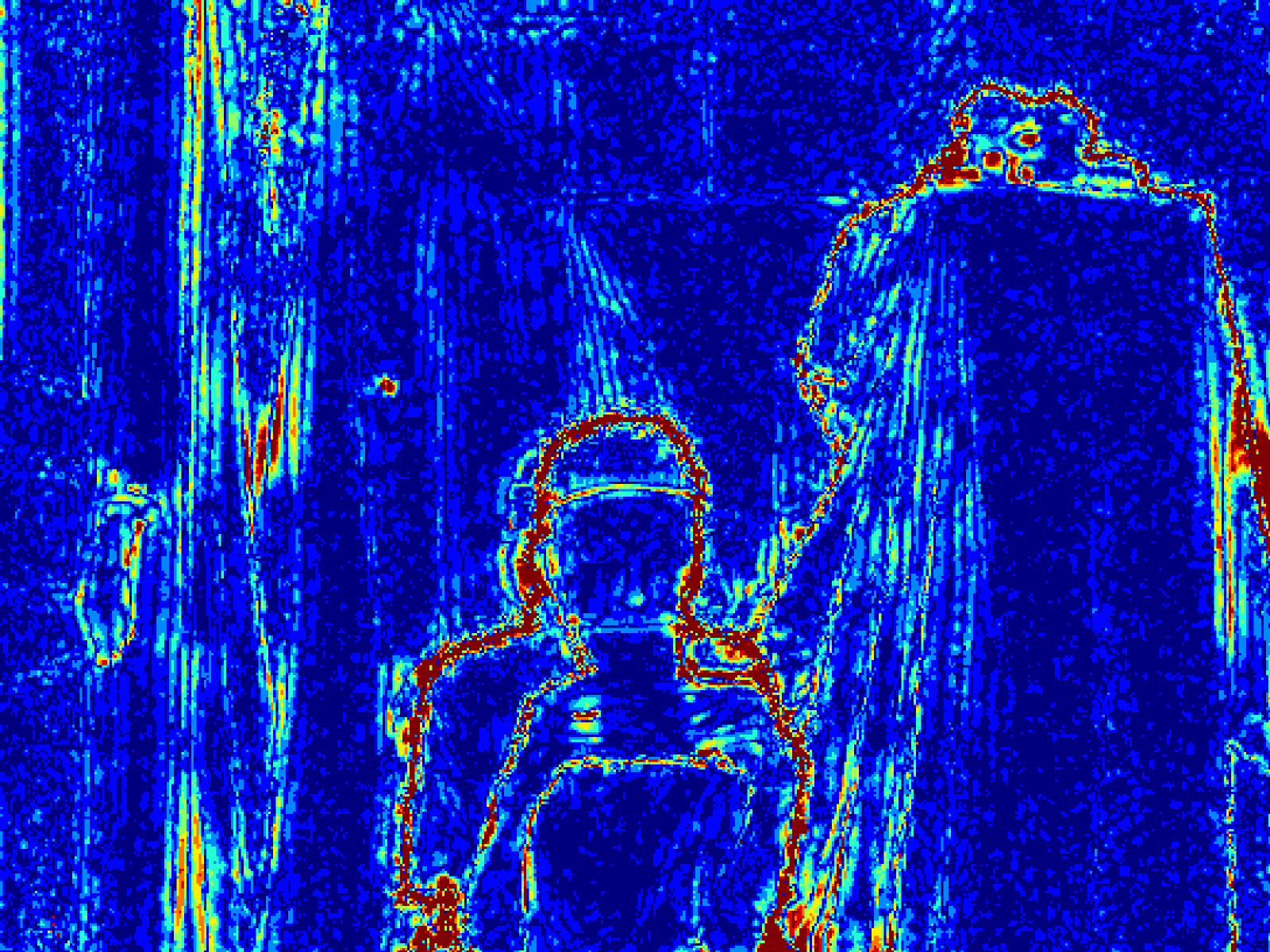}
	\hspace{-1.8mm} & \includegraphics[height=0.62in]{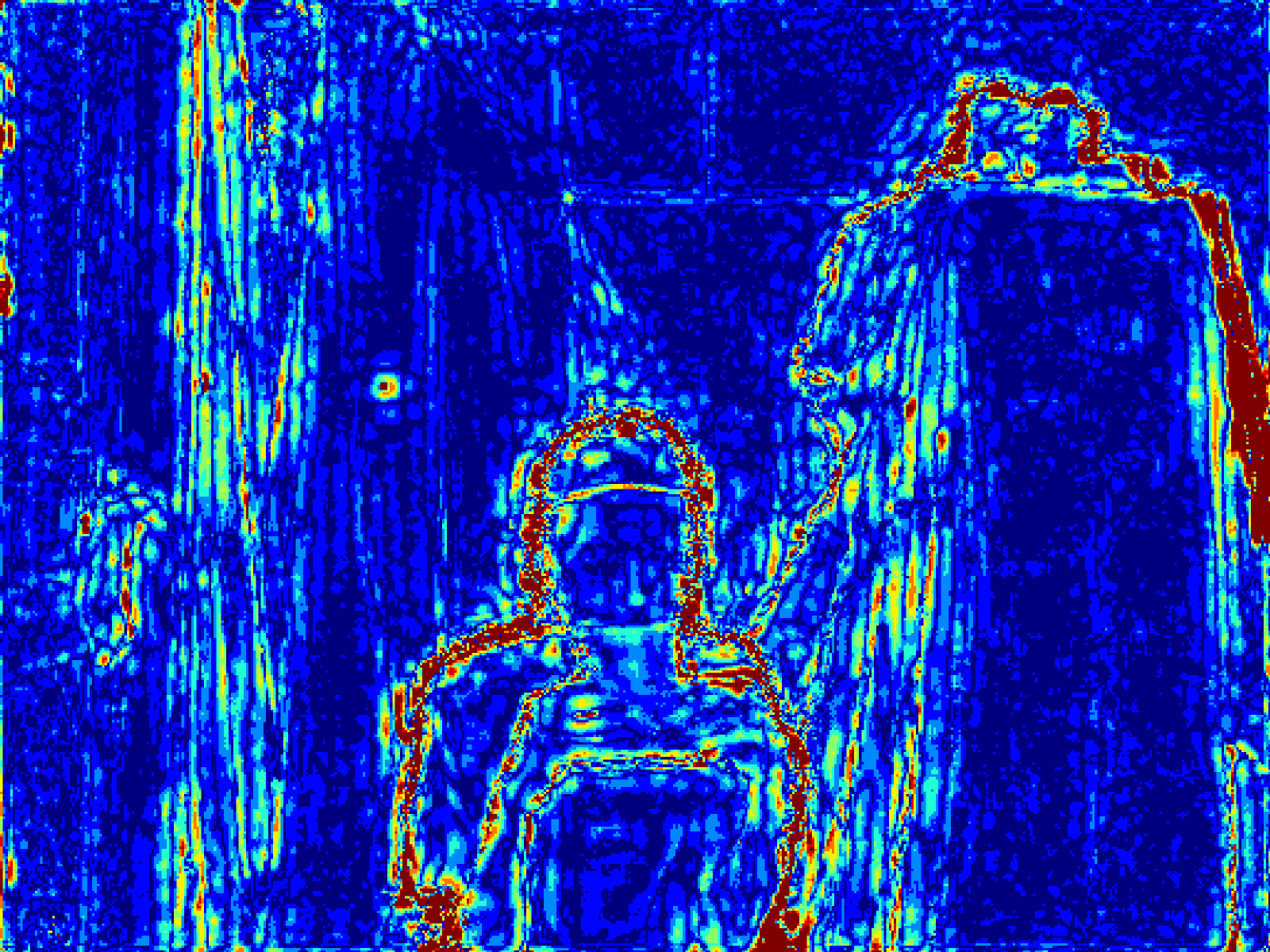}
	\hspace{-1.8mm} & \includegraphics[height=0.62in]{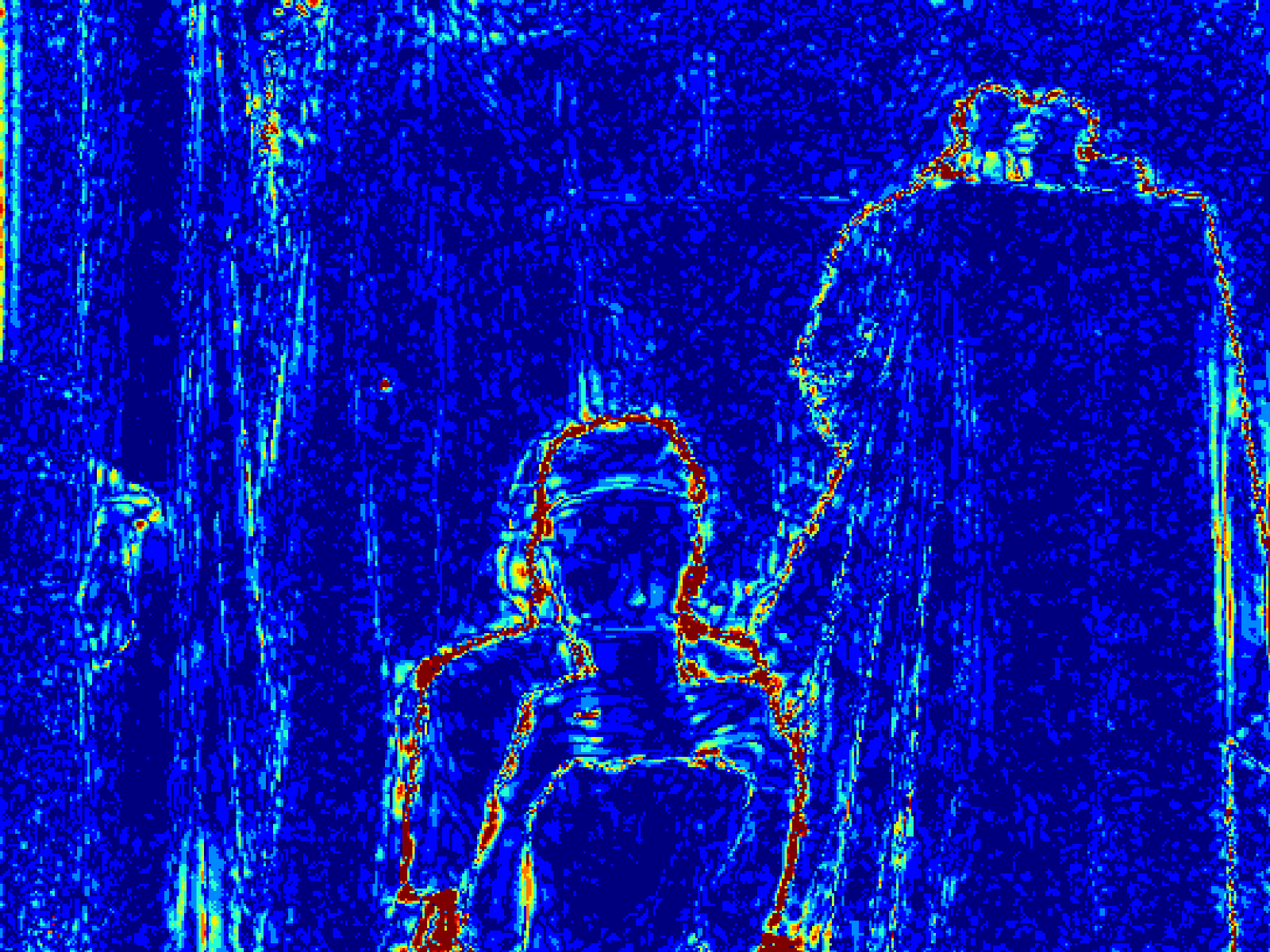}
 
	\hspace{-1.8mm} & \includegraphics[height=0.62in]{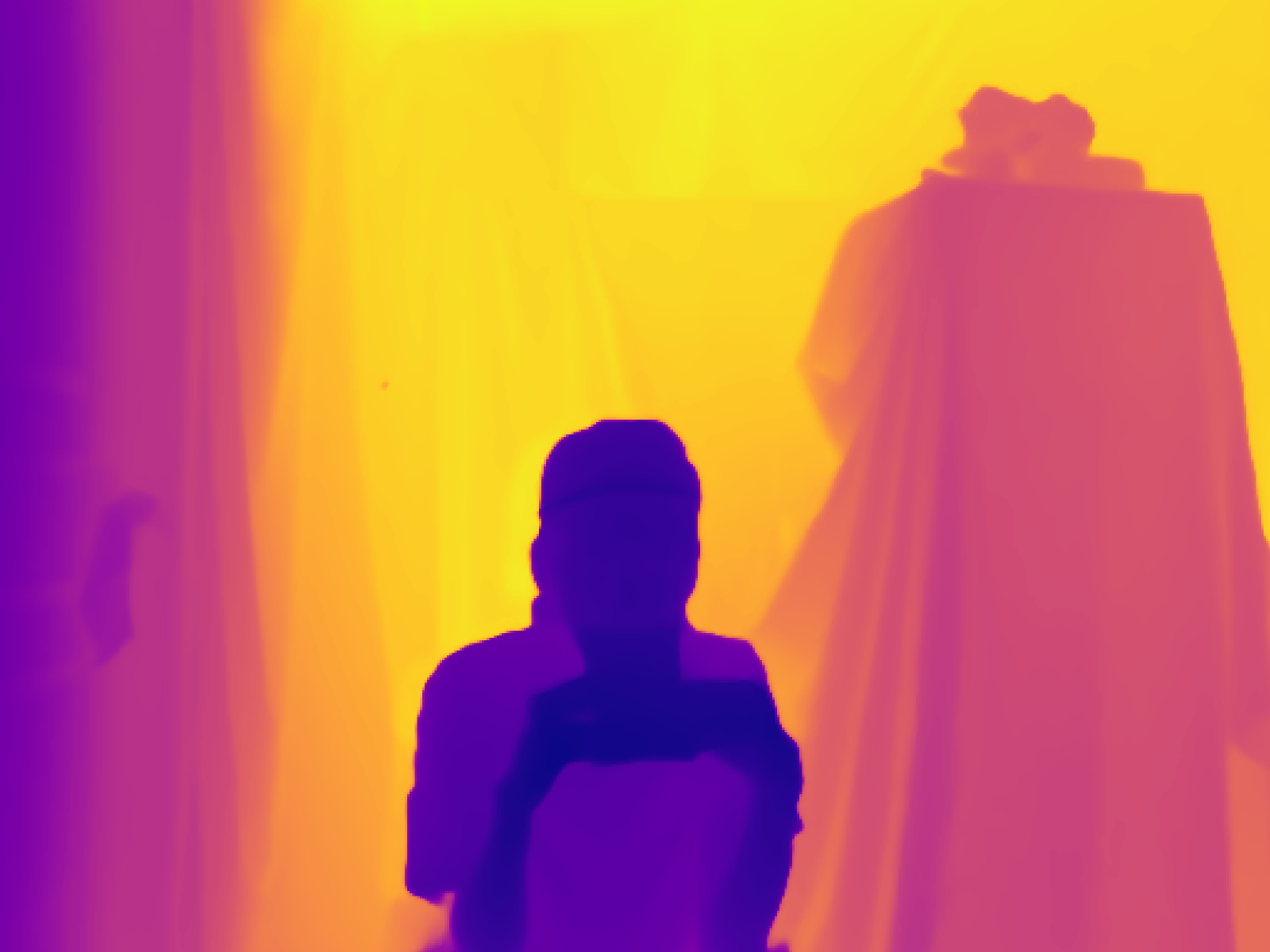}
 \\
	& \scriptsize \textbf{(a)} RGB & \scriptsize \textbf{(b)} Bicubic & \scriptsize \textbf{(c)} GT & \scriptsize \textbf{(d)} FDKN & \scriptsize \textbf{(e)} FDSR & \scriptsize \textbf{(f)} DCTnet & \scriptsize \textbf{(g)} \netname{} & \scriptsize \textbf{(h)} \netname{} (depth)
	\end{tabular}
    \vspace{-0.3cm}
	\caption{\textbf{Qualitative comparison on the RGBDD dataset.} From left to right: (a) RGB image, (b) Bicubic upsampled depth map, (c) GT; then, error maps achieved by selected methods: (d) FDKN~\citep{kim2021deformable}, (e) FDSR~\citep{he2021towards}, (f) DCTNet~\citep{zhao2022discrete}; finally, (g) error maps and (h) predictions by \netname.} 
	\label{fig:rgbdd_comp}
\end{figure*}

\subsection{Datasets and Metrics}
We evaluate \netname{} on four datasets, compared with existing methods when super-solving depth maps by three different upsampling factors: $4\times,\ 8\times$, and $16\times$. 

\textbf{Middlebury}~\citep{scharstein2003high,scharstein2007learning,hirschmuller2007evaluation,scharstein2014high}. We train all learning-based methods using 50 RGB-D images with ground truth from Middlebury 2005, 2006 and 2014 datasets. As in~\cite{de2022learning}, we retain 5 for validation and 5 for testing. 

\textbf{NYUv2}~\citep{silberman2012indoor}. It contains 1449 RGB-D images in total. Following \cite{de2022learning}, we randomly split it into 849 RGB-D images for the training set, 300 for the validation set and 300 for the test set. Compared to \cite{ye2020pmbanet,liu2022pdr}, it comes with a validation set to make the comparison fairer.

\textbf{DIML}~\citep{kim2016structure,kim2017deep,kim2018deep,cho2021deep} consists of 2 million color images and corresponding depth maps from indoor and outdoor scenes. We adopt the same strategy outlined in \cite{de2022learning}, i.e., considering only the indoor data subset, and use 1440 for training, 169 for validation, and 503 for testing.

\textbf{RGBDD}~\citep{he2021towards} is a new real-world dataset for GDSR, which consists of 4811 image pairs. For evaluation, we follow the protocol described in \cite{he2021towards}, using 2215 images (1586 portraits, 380 plants, 249 models) as the training set and 405 images (297 portraits, 68 plants, 40 models) as the test set. 

\textbf{Metrics.} Following \cite{de2022learning}, we compute mean square error (MSE / $cm^2$) and mean absolute error (MAE / $cm$) as metrics on Middlebury, NYUv2 and DIML. For RGBDD, we use root mean square error (RMSE / $cm$) as in \cite{he2021towards}.

\subsection{Implementation Details}
During training, the HR depth maps and the color images are randomly cropped into $256\times 256$ patches. LR depth patches are generated by bicubic interpolation at $64\times 64$, $32\times 32$, $16\times 16$ resolution for $4\times$, $8\times$ and $16\times$ factors, respectively. We randomly extract about 75K, 168K, 223K and 232K patches from Middlebury, NYUv2, DIML and RGBDD for training. Before being fed to the network, depth maps and images are normalized in the [0, 1] range.

We use Pytorch \citep{paszke2019pytorch} to implement and train \netname{}, on a single Nvidia RTX 3090 GPU. The batch size is set to 4, using Adam as the optimizer. The learning rate is initialized to $1\times 10^{-4}$, then performing a 5-epoch warm-up and cosine annealing. We use random rotation, horizontal/vertical flipping as data augmentation. According to the size of the four datasets, we train our network for 1505, 198, 155 and 109 epochs on Middlebury, NYUv2, DIML and RGBDD, respectively. When evaluating results on a specific dataset, we do not perform any pre-training on the others. Following \cite{de2022learning}, testing is performed by processing $256\times256$ patches at a time on Middlebury, NYUv2 and DIML for fairness, while full-resolution images are processed for RGBDD.

\begin{table}[t] \scriptsize
	\renewcommand\tabcolsep{4.3pt} 
	\centering
        \caption{\textbf{Cross-dataset generalization.} All methods are trained on NYUv2 and tested on DIML/Middlebury with factor $8\times$. Middlebury\textit{-HR} is the test set defined in \cite{de2022learning}, Middlebury\textit{-LR} is the one from \cite{tang2021joint}. The lower MSE and MAE, the better. }
	\begin{tabular}{@{}lccc@{}}
		\toprule
		 \textbf{Methods} & DIML & Middlebury\textit{-HR} & Middlebury\textit{-LR}  \\ \midrule
		GF~\citep{he2010guided} & 34.1 \ 1.77 & 40.5 \ 1.49 & 25.6 \ 2.31  \\
		SD~\citep{ham2017robust} & 44.9 \ 0.83 & 82.5 \ 0.86 & 28.8 \ 2.07  \\
		P2P~\citep{lutio2019guided} & 23.0 \ 1.26 & 32.7 \ 0.82 & 15.8 \ 1.73  \\
		MSG~\citep{hui2016depth}  & 5.76 \ 0.51 & 11.0 \ 0.54 & 8.89 \ 1.62  \\
		FDKN~\citep{kim2021deformable} & 6.74 \ 0.53 & 10.0 \ \underline{0.43} & 5.54 \ 0.99  \\
		PMBANet~\citep{ye2020pmbanet} & 7.35 \ 0.59 & \underline{9.62} \ 0.46 & 4.16 \ \underline{0.91}  \\
            FDSR~\citep{he2021towards} & 7.73 \ 0.74 & 18.4 \ 0.73 & 6.92 \ 1.09  \\
            JIIF~\citep{tang2021joint} & \underline{4.10} \ \underline{0.38} & 19.3 \ 0.74 & 4.40 \ 0.92  \\
            DCTNet~\citep{zhao2022discrete} & 5.64 \ 0.77 & 17.5 \ 0.77 & 6.96 \ 1.15  \\ 
            LGR~\citep{de2022learning} & 4.95 \ 0.40 & \textbf{8.25} \ \textbf{0.35} & 5.94 \ 1.11  \\
            \netname$^+$  & \textbf{3.72} \ \textbf{0.36} & 14.6 \ 0.54 & \textbf{3.44} \ \textbf{0.87}  \\         
        \bottomrule
	\end{tabular}
	\vspace{-0.cm}
	\label{cross-data_comparison}
\end{table}

\subsection{Comparison with State-of-the-Art}
We compare \netname{} to GF \citep{he2010guided}, SD \citep{ham2017robust}, P2P \citep{lutio2019guided}, MSG \citep{hui2016depth}, DKN and its fast implementation FDKN \citep{kim2021deformable}, PMBANet \citep{ye2020pmbanet}, FDSR \citep{he2021towards}, JIIF \citep{tang2021joint}, DCTNet \citep{zhao2022discrete}, LGR \citep{de2022learning}, and finally to DADA~\citep{metzger2022guided} on Middlebury, NYUv2 and DIML datasets. We could not compare with PDRNet \citep{liu2022pdr} under the same setting because the source code is unavailable at the time of writing. For the other methods, we use the results from \citep{de2022learning} or the officially published codes, and results from \citep{yuan2023recurrent,metzger2022guided} for concurrent works. On the RGBDD dataset, the proposed network is compared to SDF~\citep{li2016deep}, SVLRM \citep{pan2019spatially}, DJF~\citep{li2016deep}, DJFR~\citep{li2019joint}, PAC~\citep{su2019pixel}, CUNet~\citep{deng2020deep}, FDKN~\citep{kim2021deformable}, DKN~\citep{kim2021deformable}, FDSR~\citep{he2021towards}, DCTNet~\citep{zhao2022discrete} and RASG~\citep{yuan2023recurrent}. To be fair with DCTNet~\citep{zhao2022discrete}, we downsample depth maps as the LR input. 
{When reporting results, we highlight \textbf{absolute} and \underline{second} best methods for each metric on each dataset.}

\begin{figure*}[ht] 
	\centering
	\renewcommand\tabcolsep{1.5pt} 
	\begin{tabular}{cccccccccccc}
	\vspace{-0.cm}
        \rotatebox[origin=l]{90}{\scriptsize \quad \textbf{DIML}} & \includegraphics[height=0.65in]{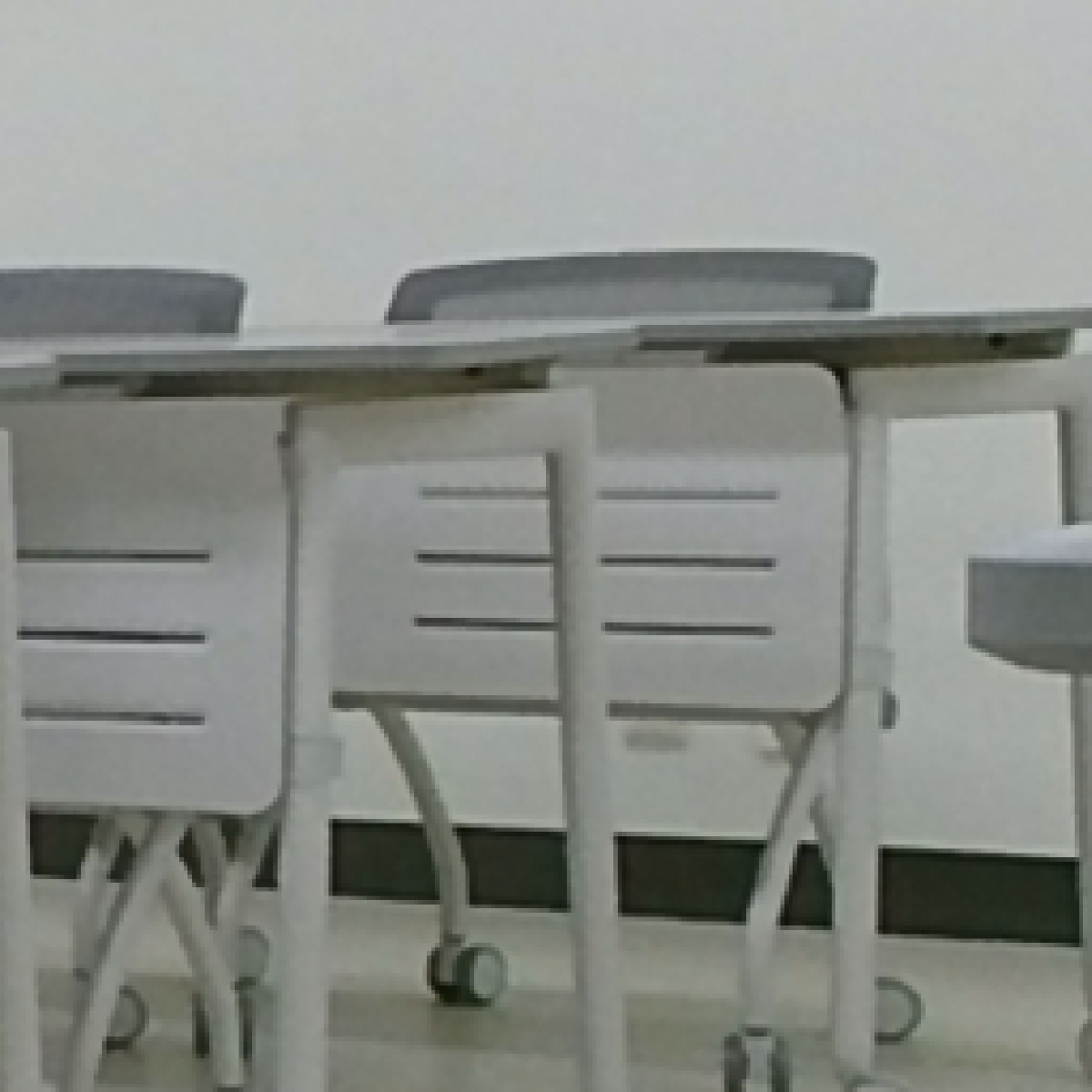}
        \hspace{-1.8mm} & \includegraphics[height=0.65in]{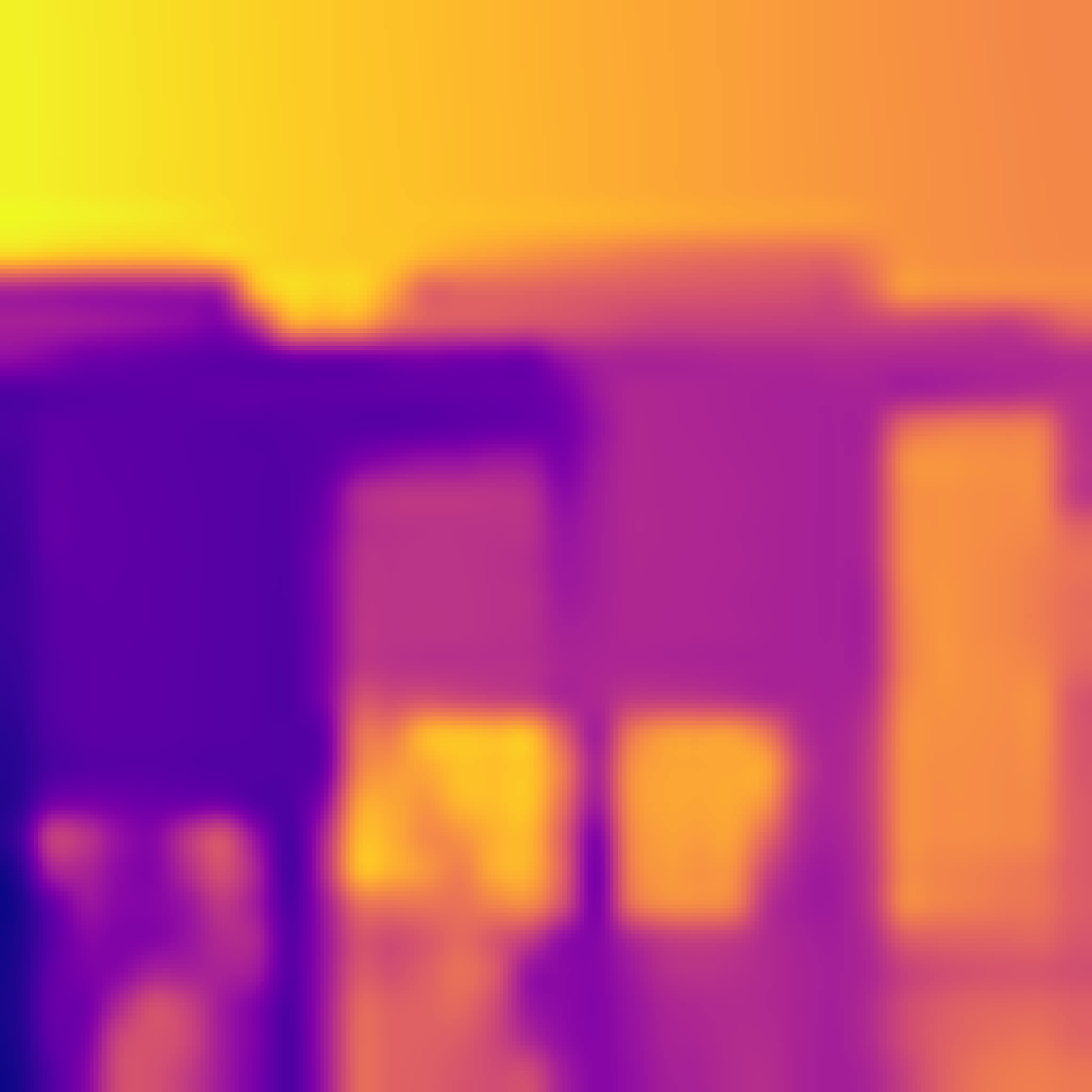}
	\hspace{-1.8mm} &  \includegraphics[height=0.65in]{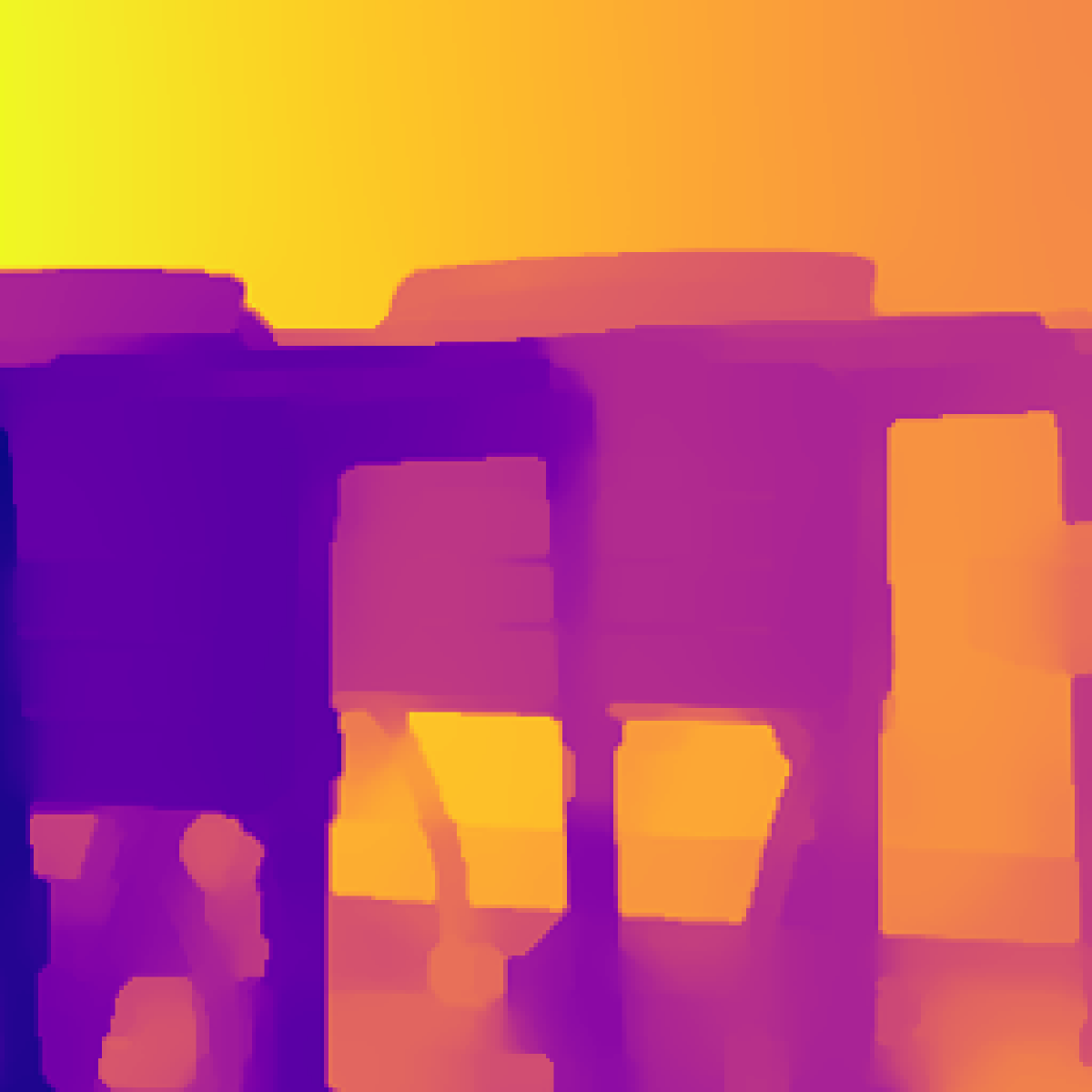}
	\hspace{-1.8mm} & \includegraphics[height=0.65in]{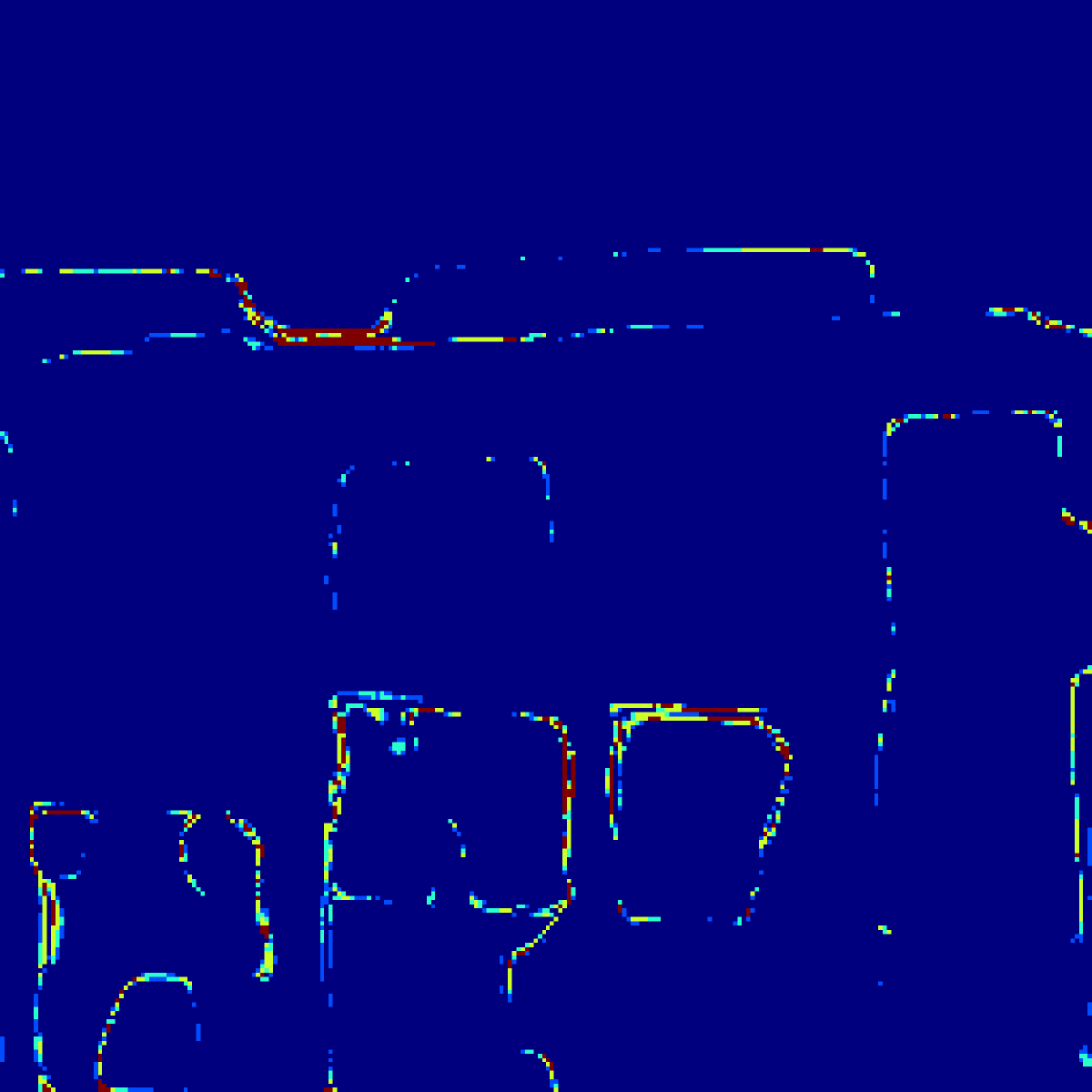}
	\hspace{-1.8mm} & \includegraphics[height=0.65in]{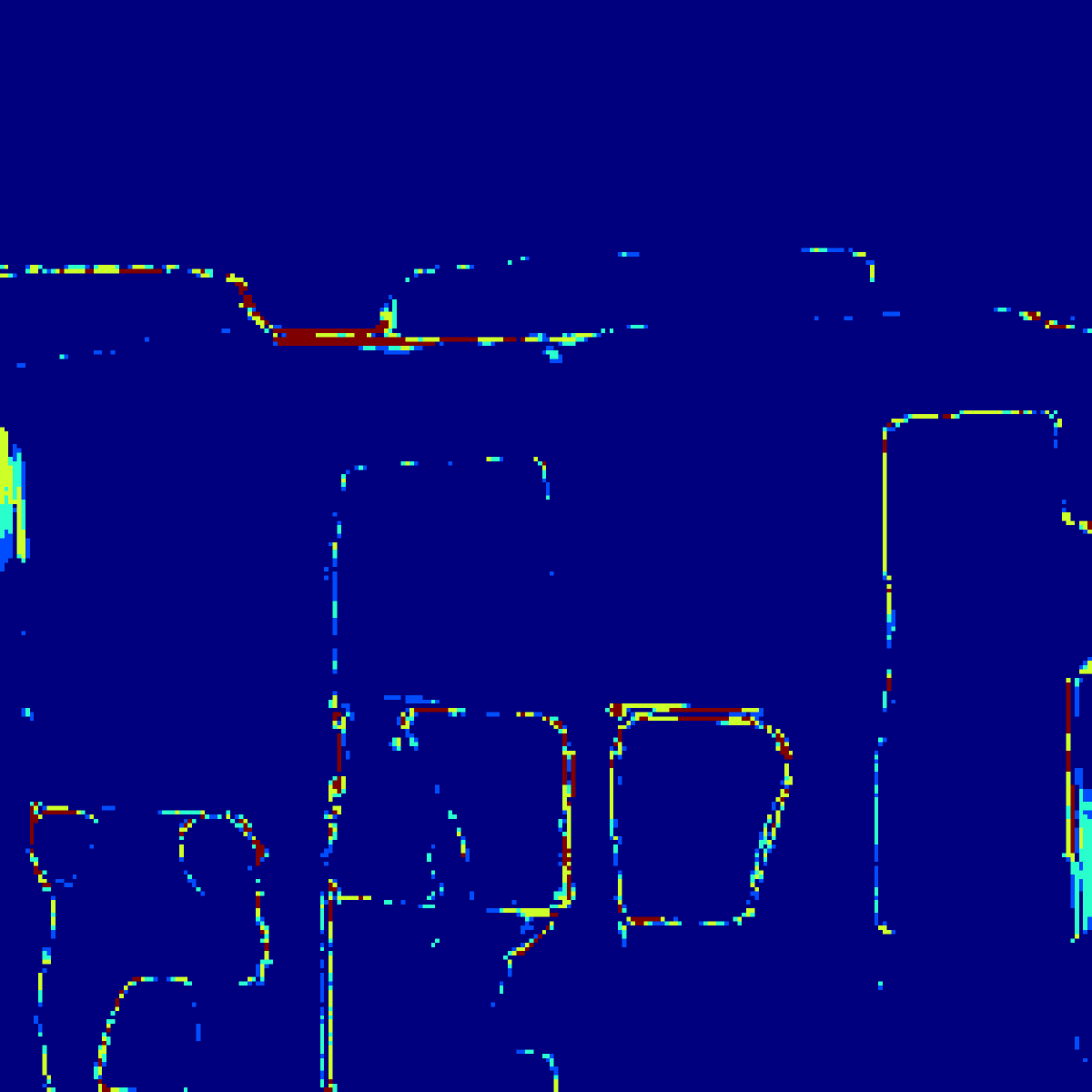}
	\hspace{-1.8mm} & \includegraphics[height=0.65in]{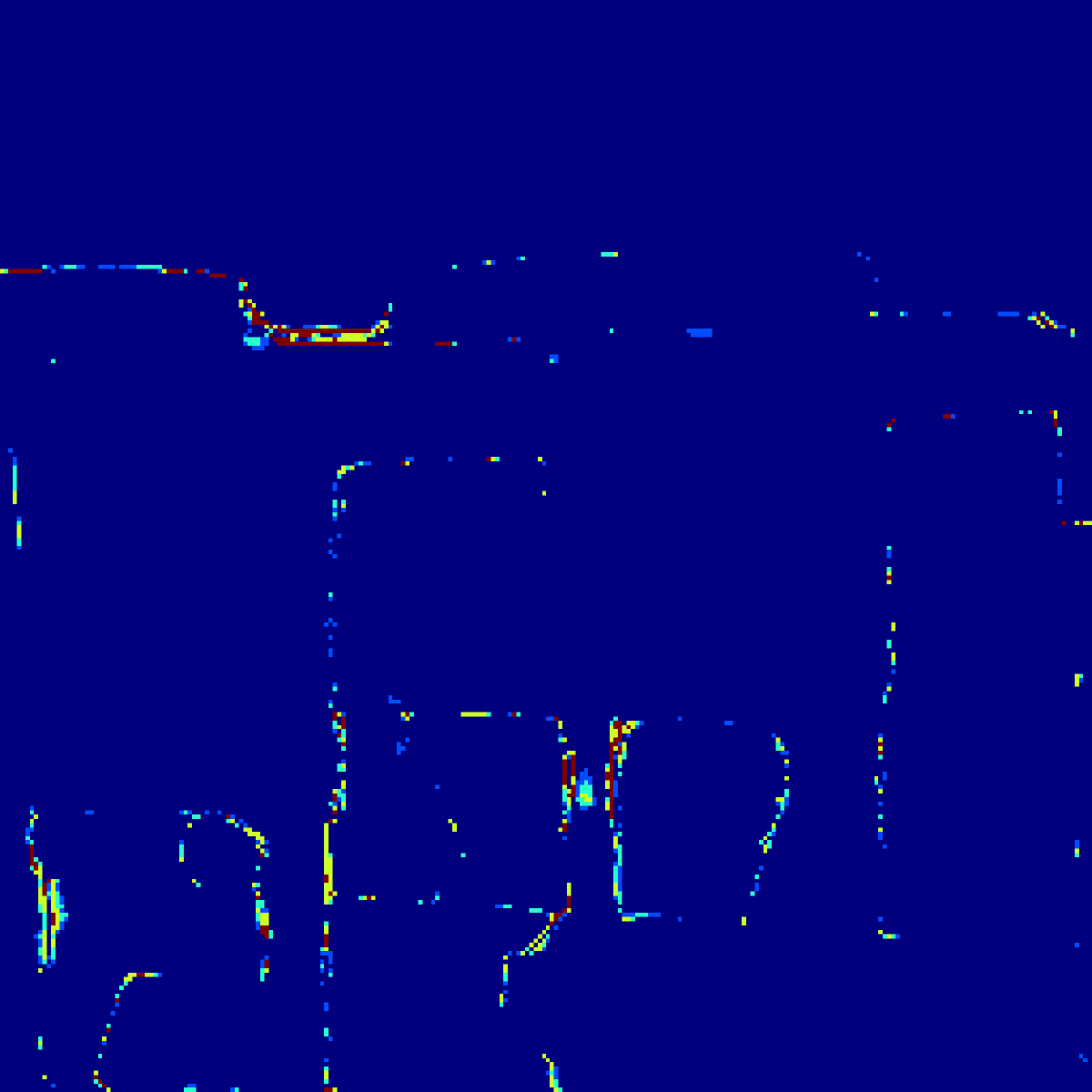}
	\hspace{-1.8mm} & \includegraphics[height=0.65in]{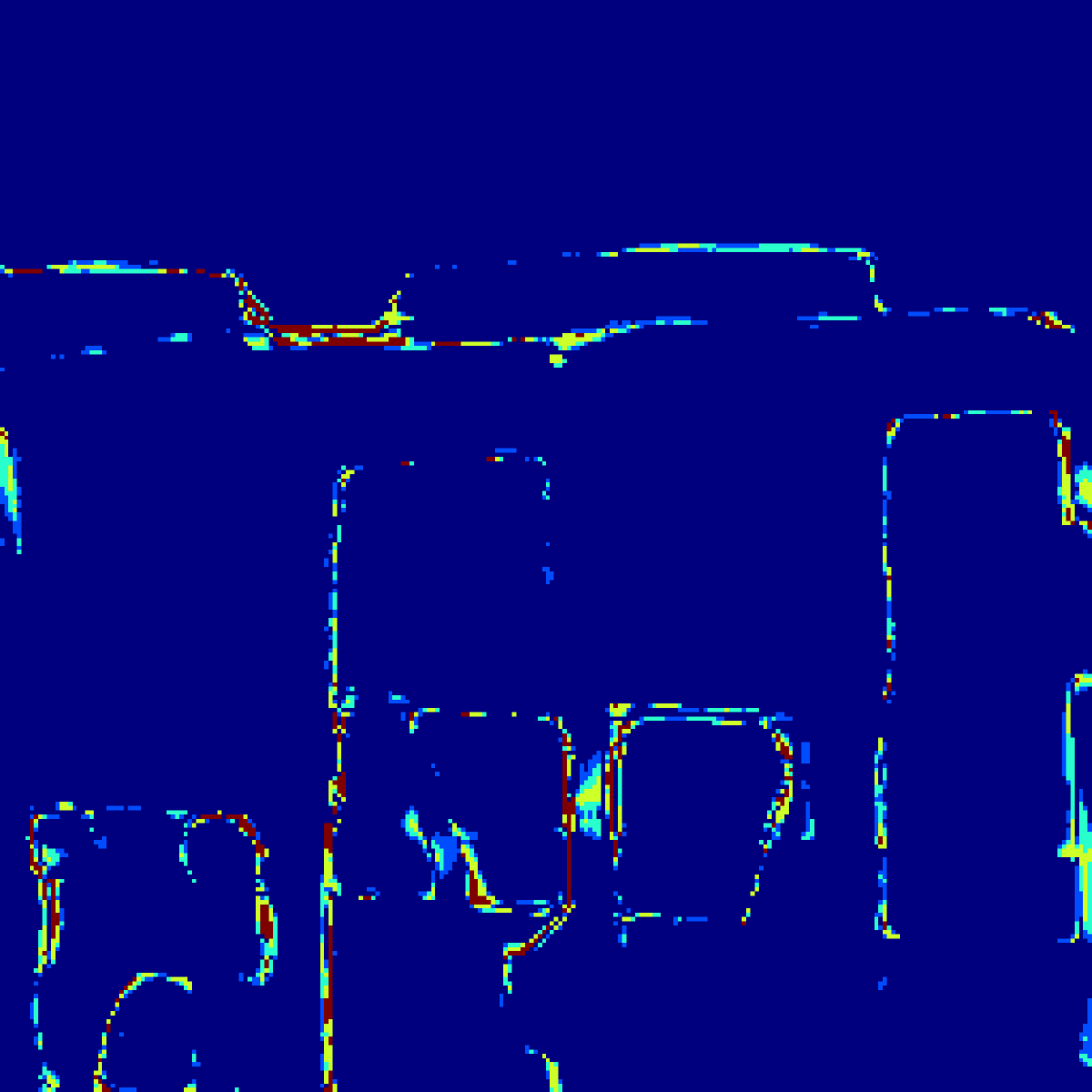}
	\hspace{-1.8mm} & \includegraphics[height=0.65in]{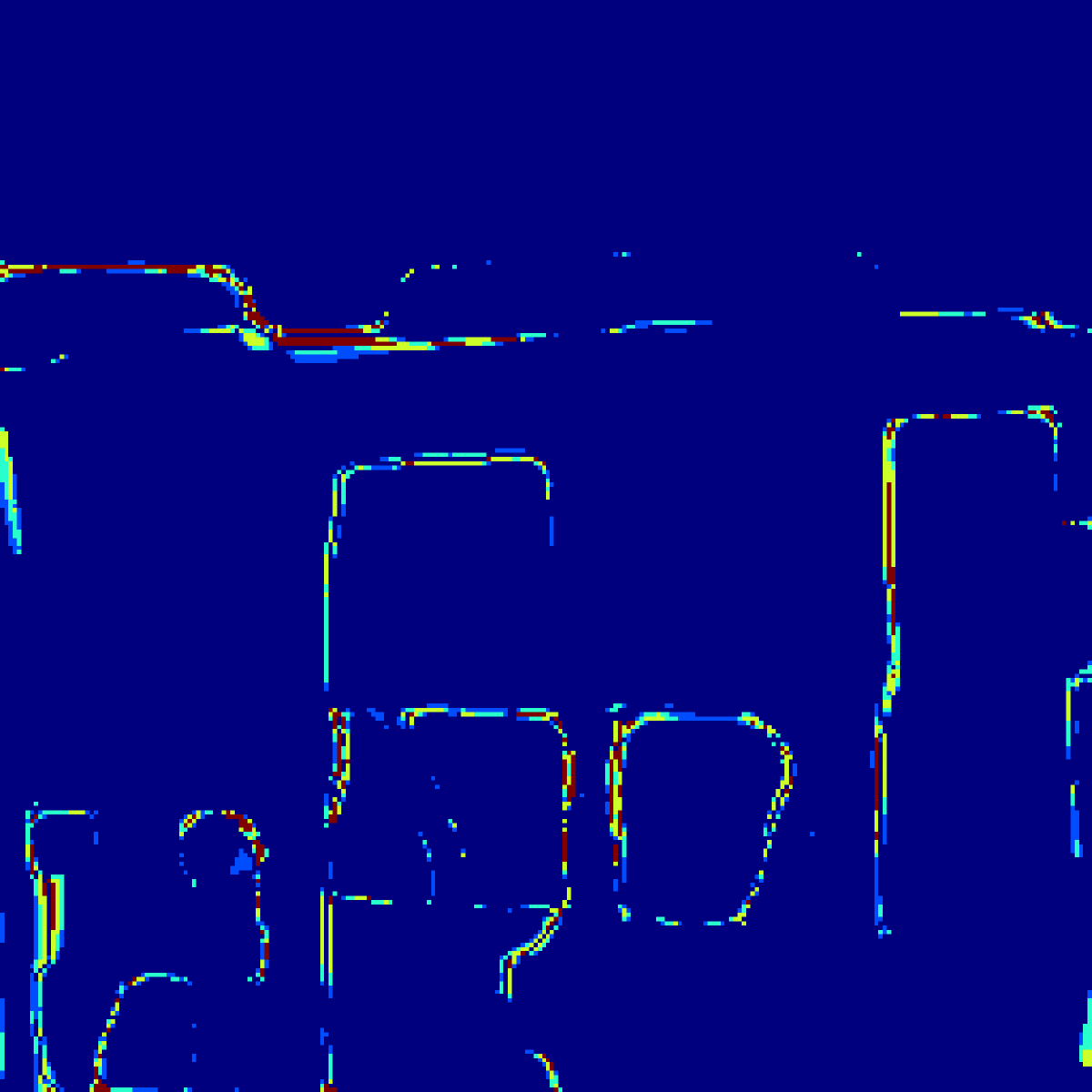}
	\hspace{-1.8mm} & \includegraphics[height=0.65in]{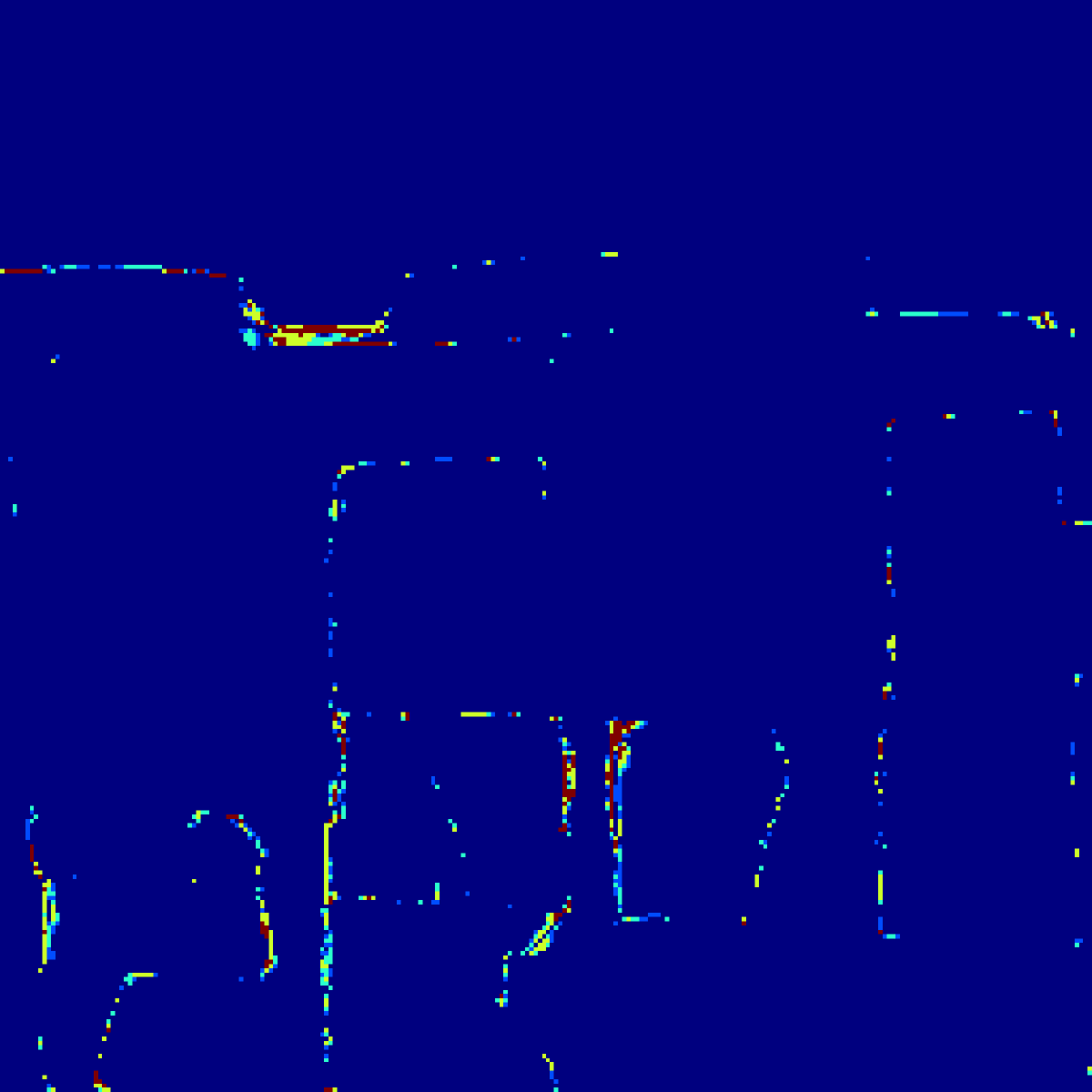}
        
        \hspace{-1.8mm} & \includegraphics[height=0.65in]{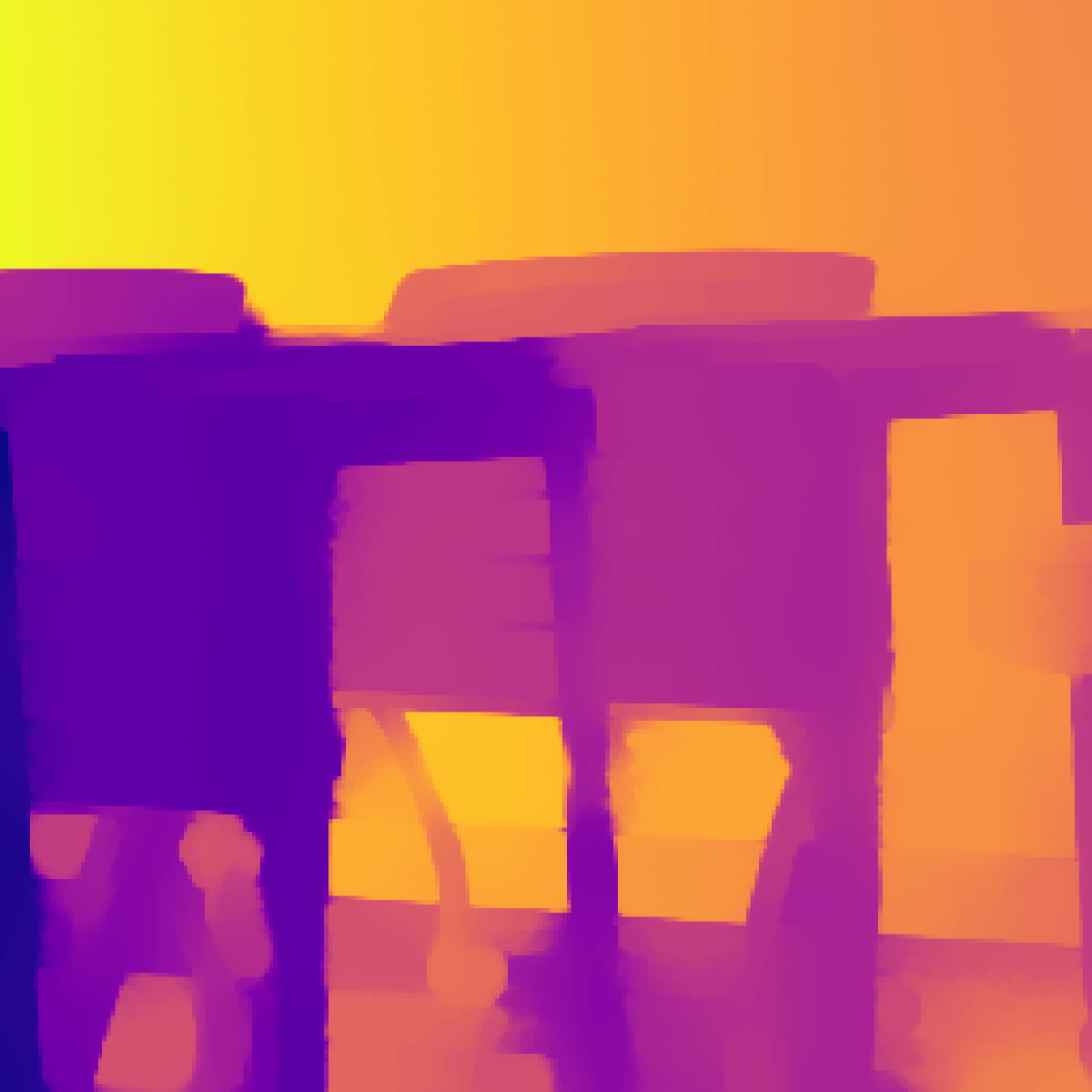}
    \\ \vspace{-0.cm}
    
        \rotatebox[origin=l]{90}{\scriptsize  \textbf{\textit{Middlebury-HR}}} & \includegraphics[height=0.65in]{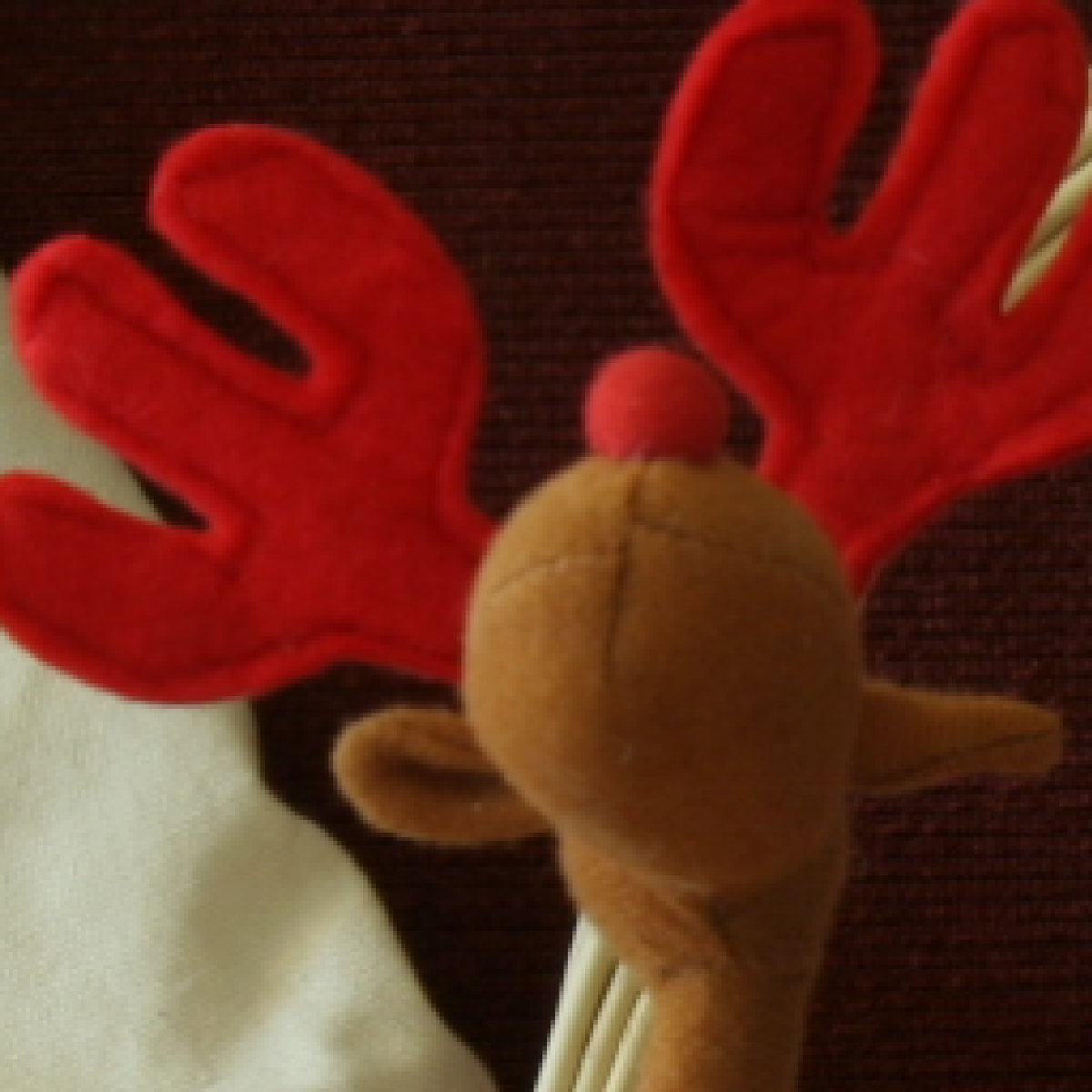}
        \hspace{-1.8mm} & \includegraphics[height=0.65in]{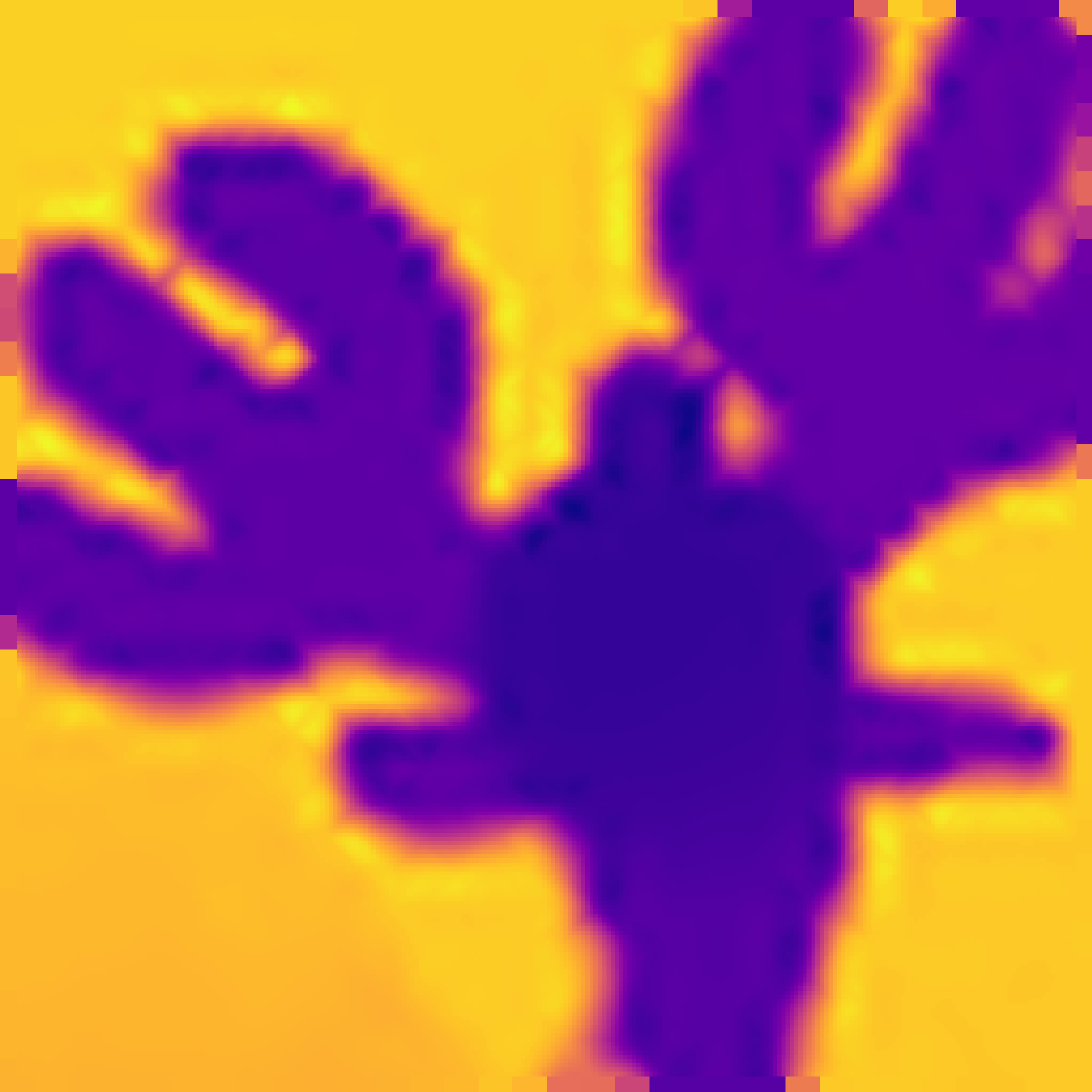}
	\hspace{-1.8mm} &  \includegraphics[height=0.65in]{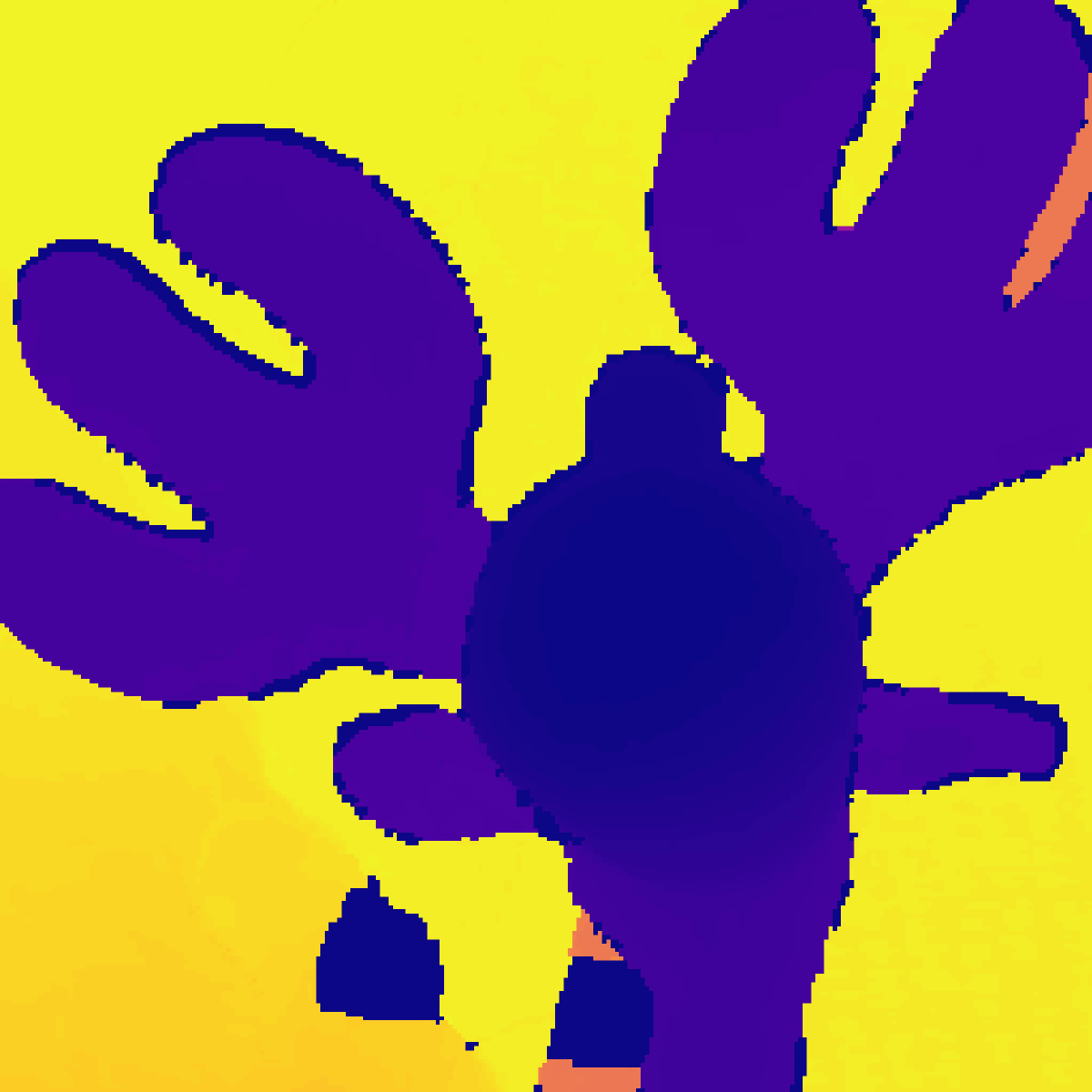}
	\hspace{-1.8mm} & \includegraphics[height=0.65in]{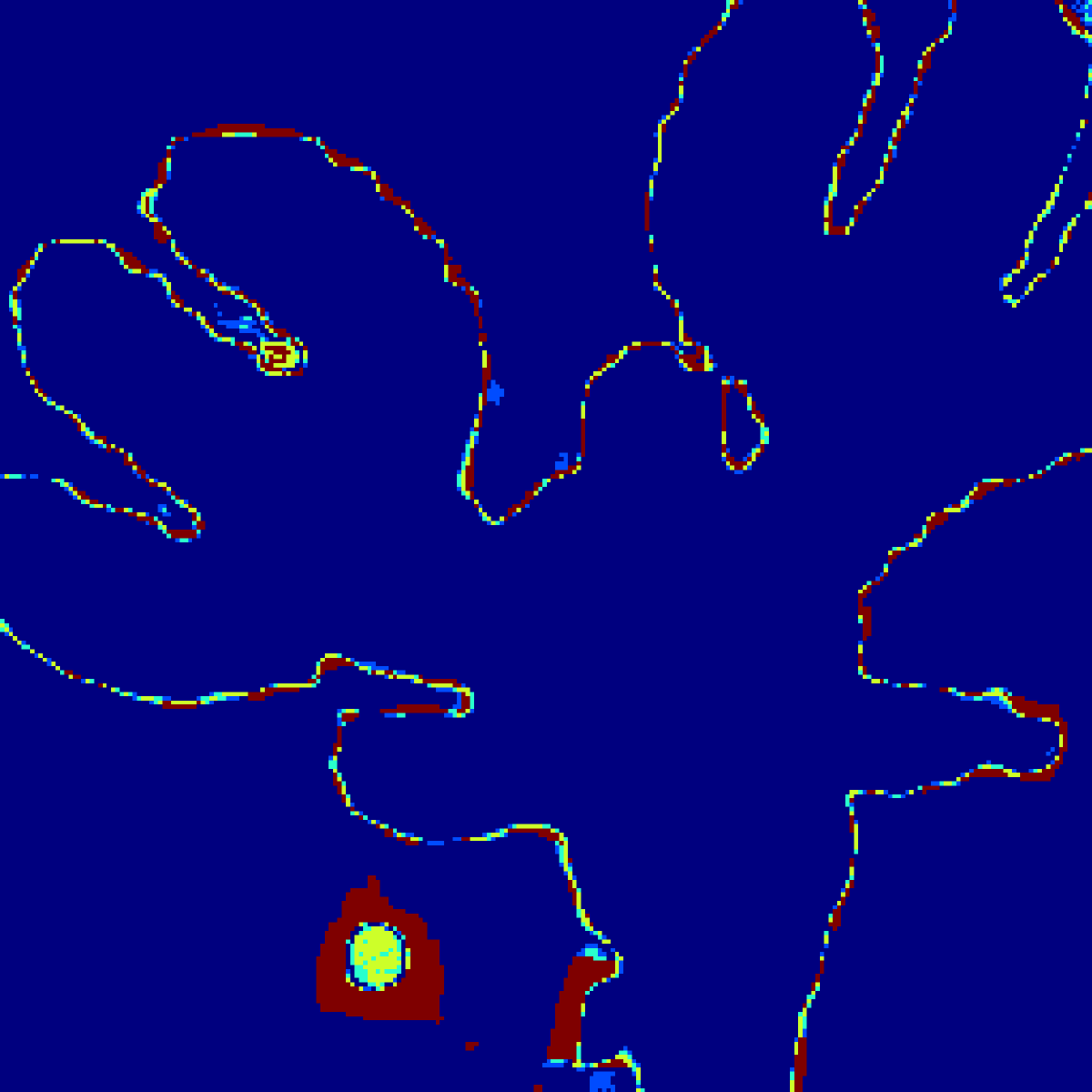}
	\hspace{-1.8mm} & \includegraphics[height=0.65in]{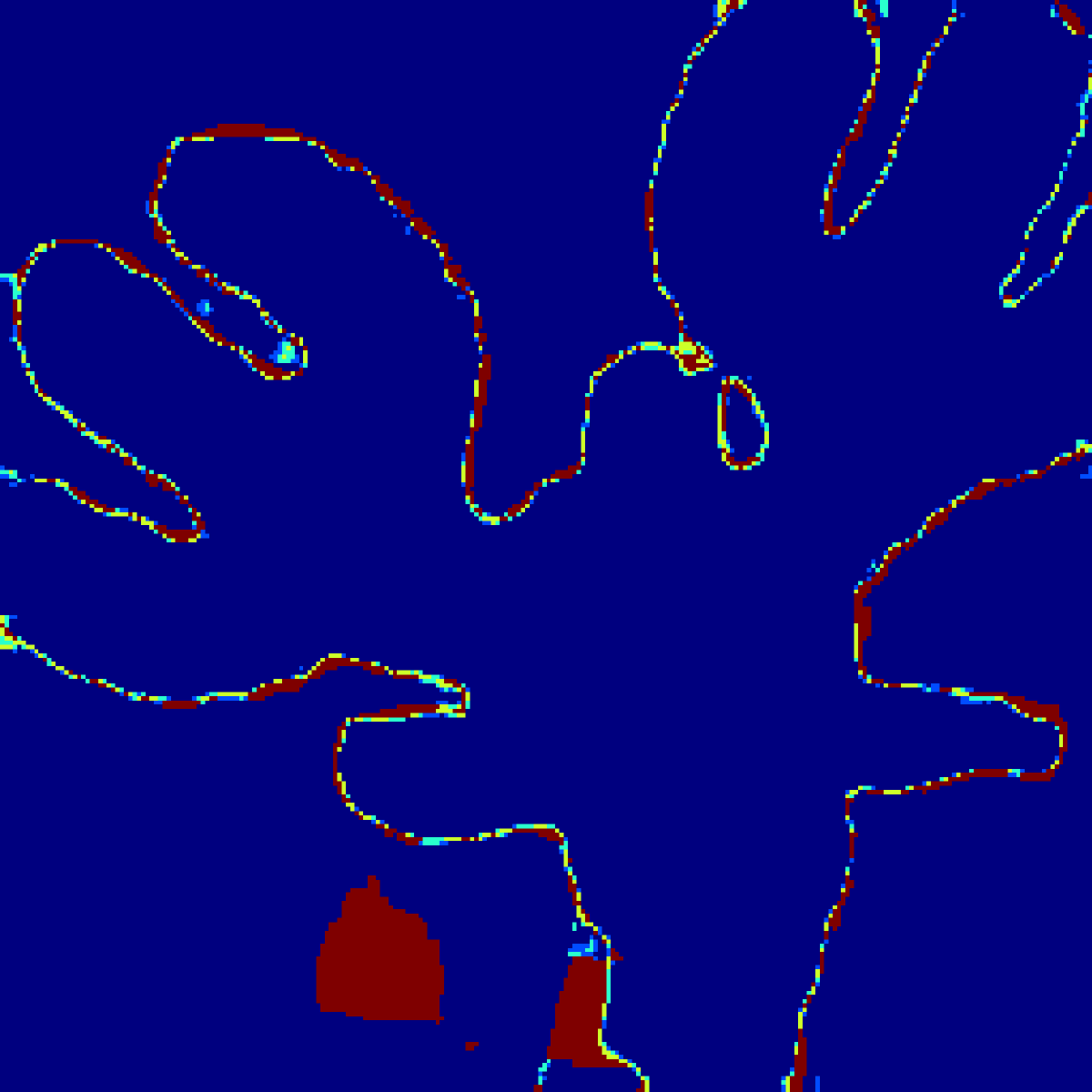}
	\hspace{-1.8mm} & \includegraphics[height=0.65in]{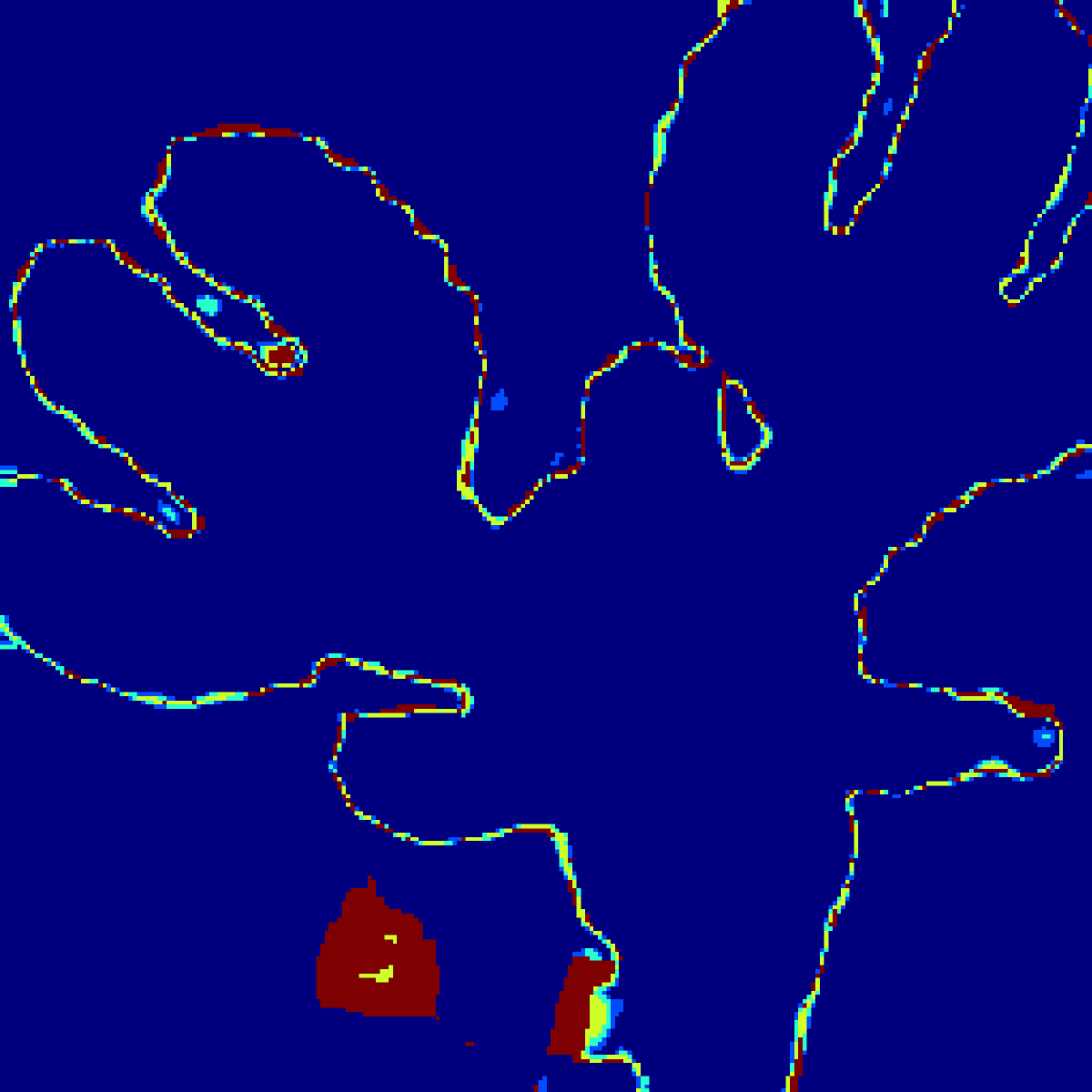}
	\hspace{-1.8mm} & \includegraphics[height=0.65in]{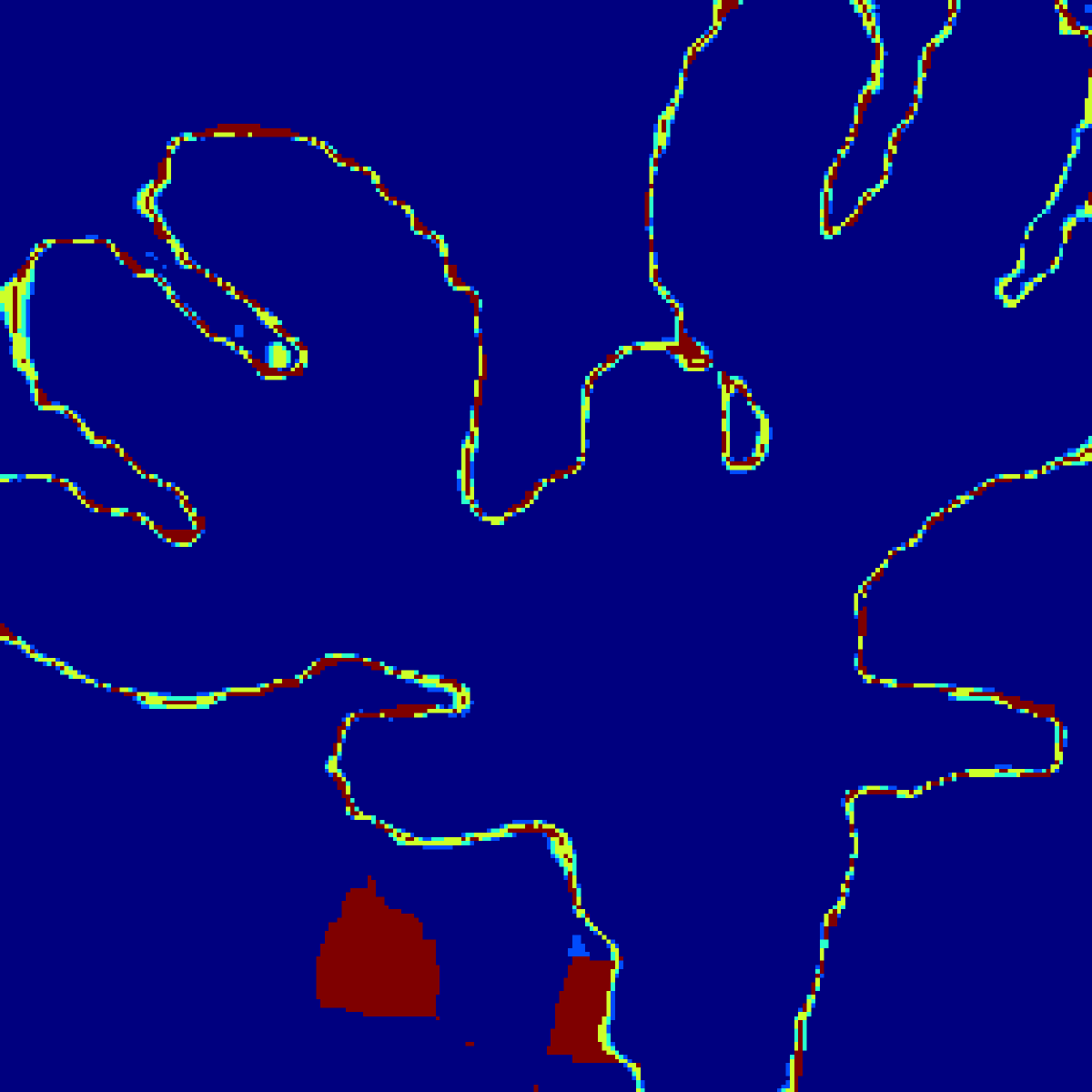}
	\hspace{-1.8mm} & \includegraphics[height=0.65in]{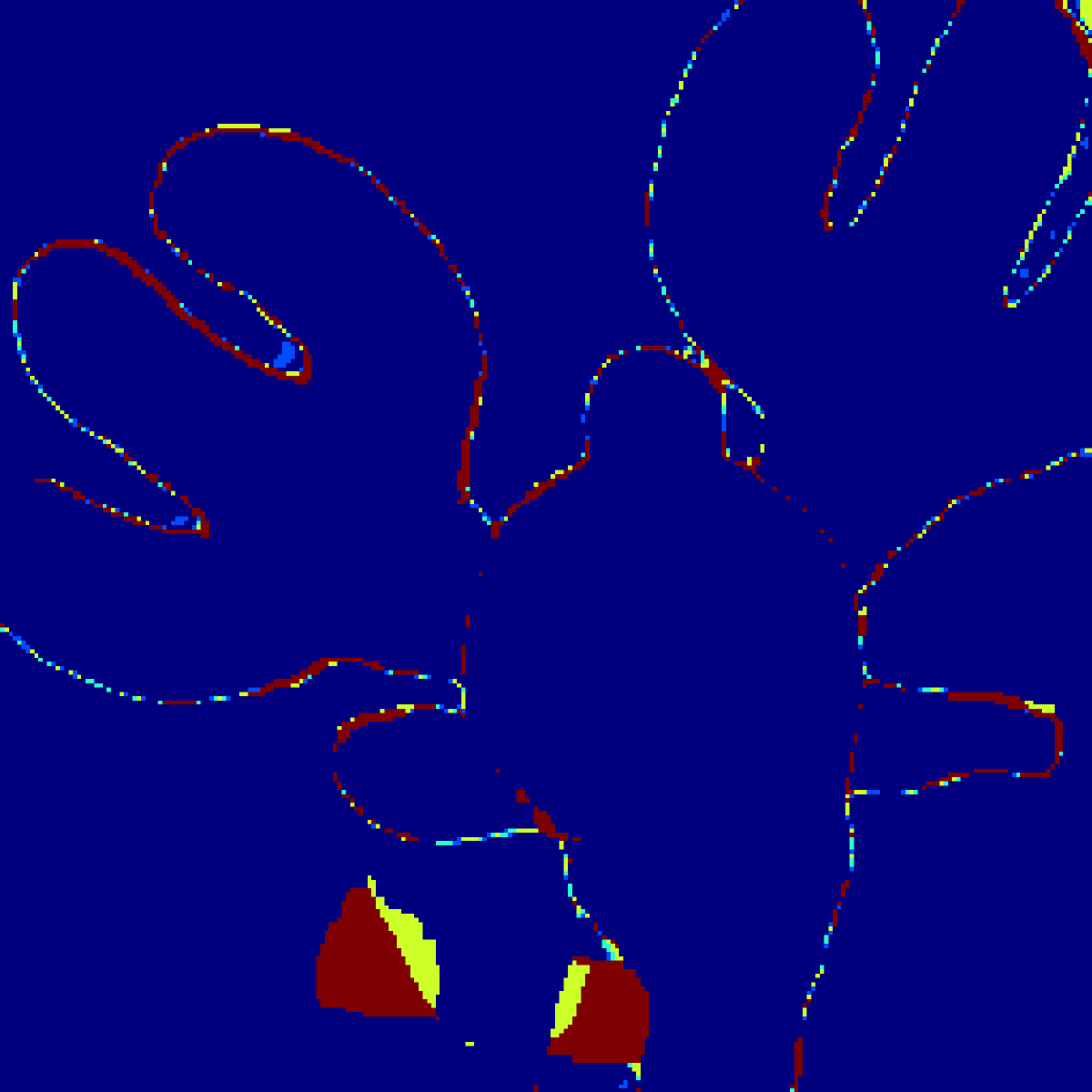}
	\hspace{-1.8mm} & \includegraphics[height=0.65in]{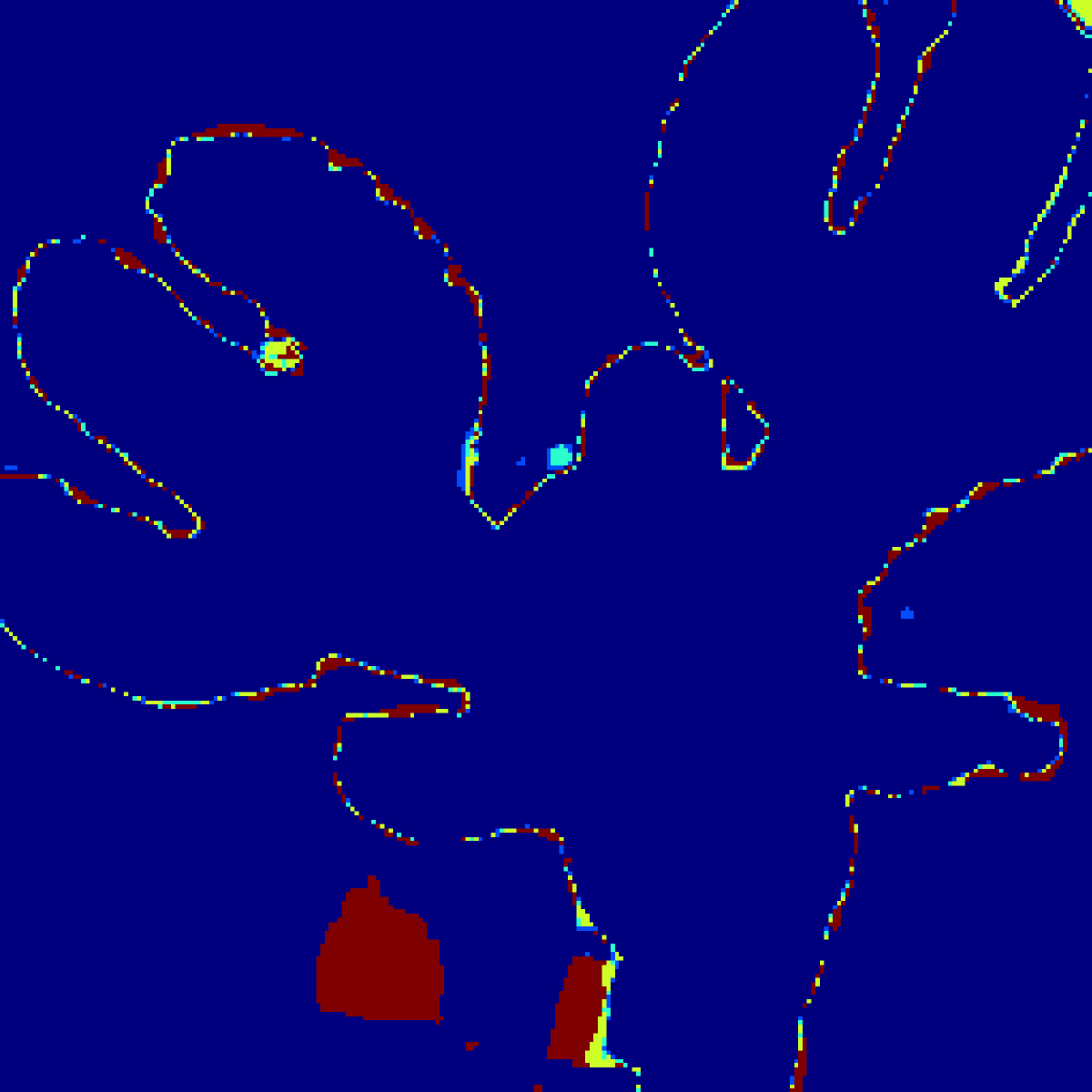}
        
        \hspace{-1.8mm} & \includegraphics[height=0.65in]{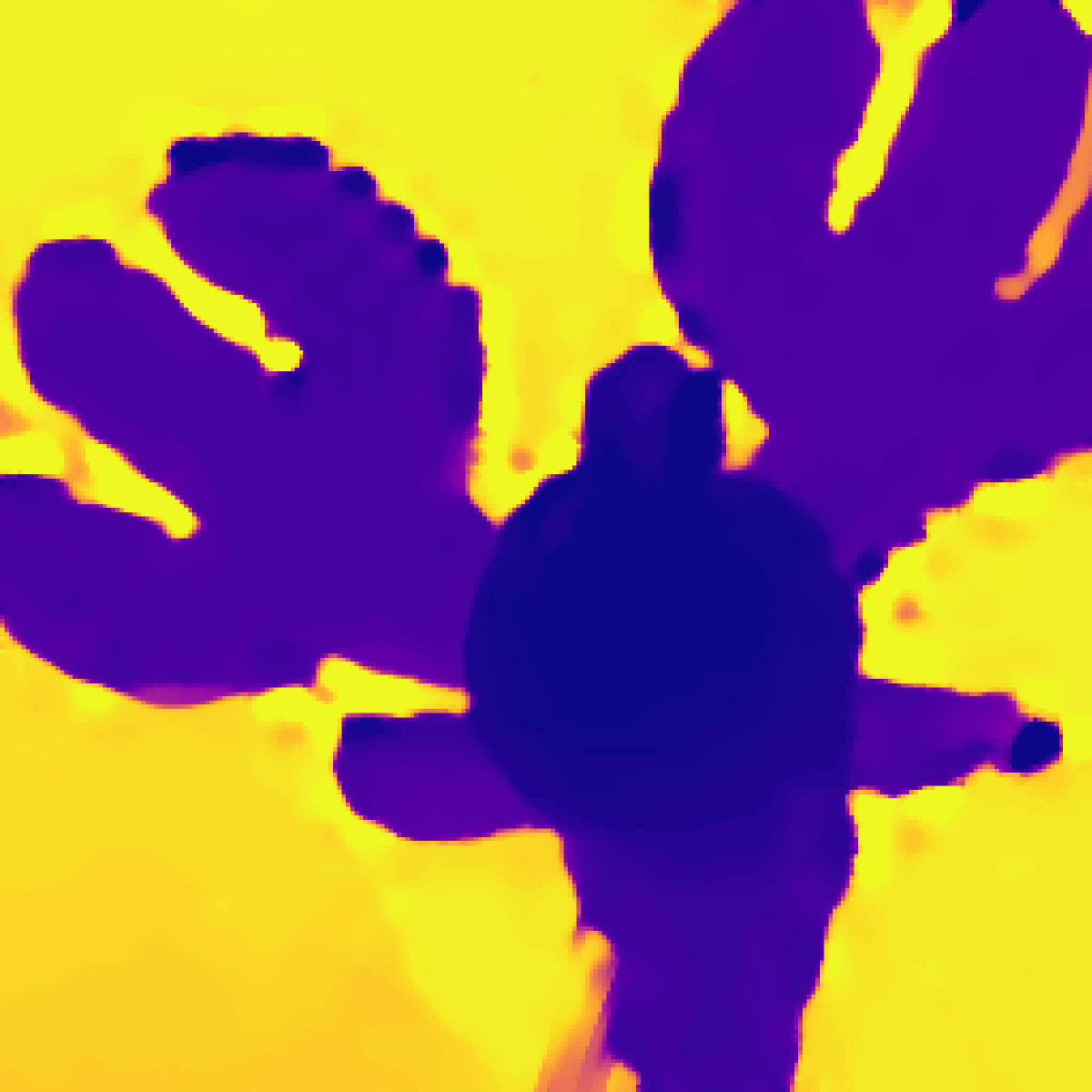}
        \\ \vspace{-0.cm}
    
        \rotatebox[origin=l]{90}{\scriptsize \textbf{\textit{Middlebury-LR}}} & \includegraphics[height=0.65in]{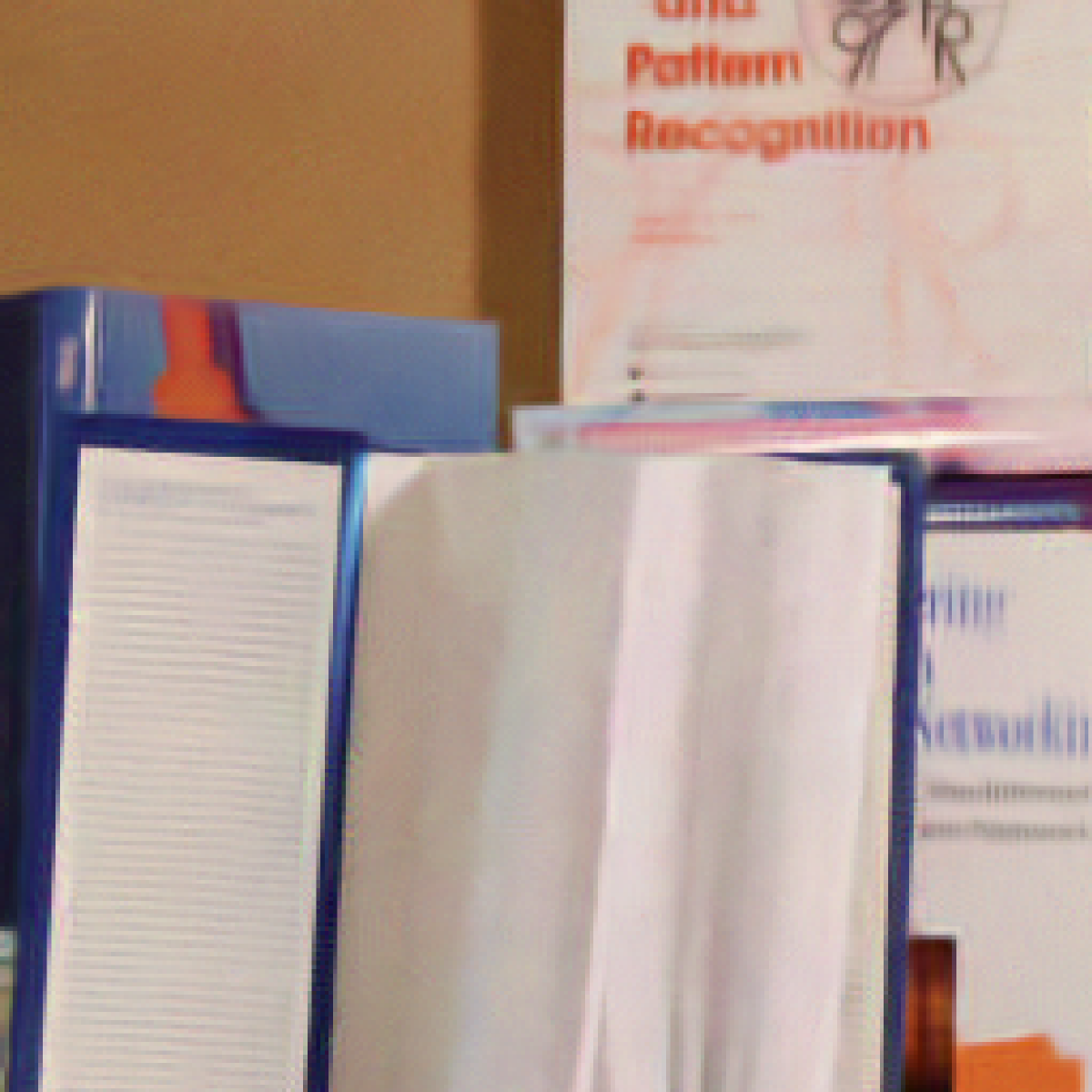}
	\hspace{-1.8mm} & \includegraphics[height=0.65in]{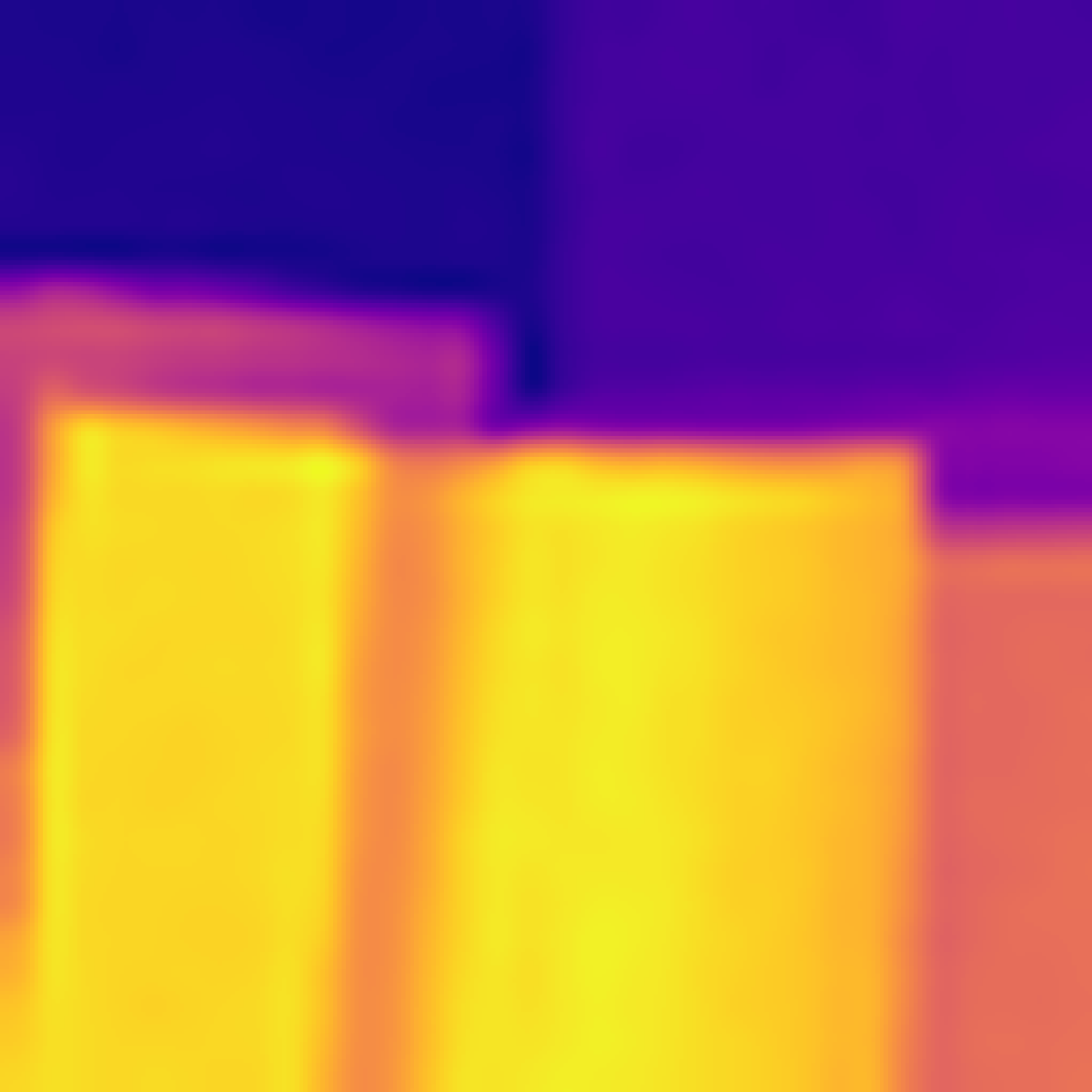}
	\hspace{-1.8mm} & \includegraphics[height=0.65in]{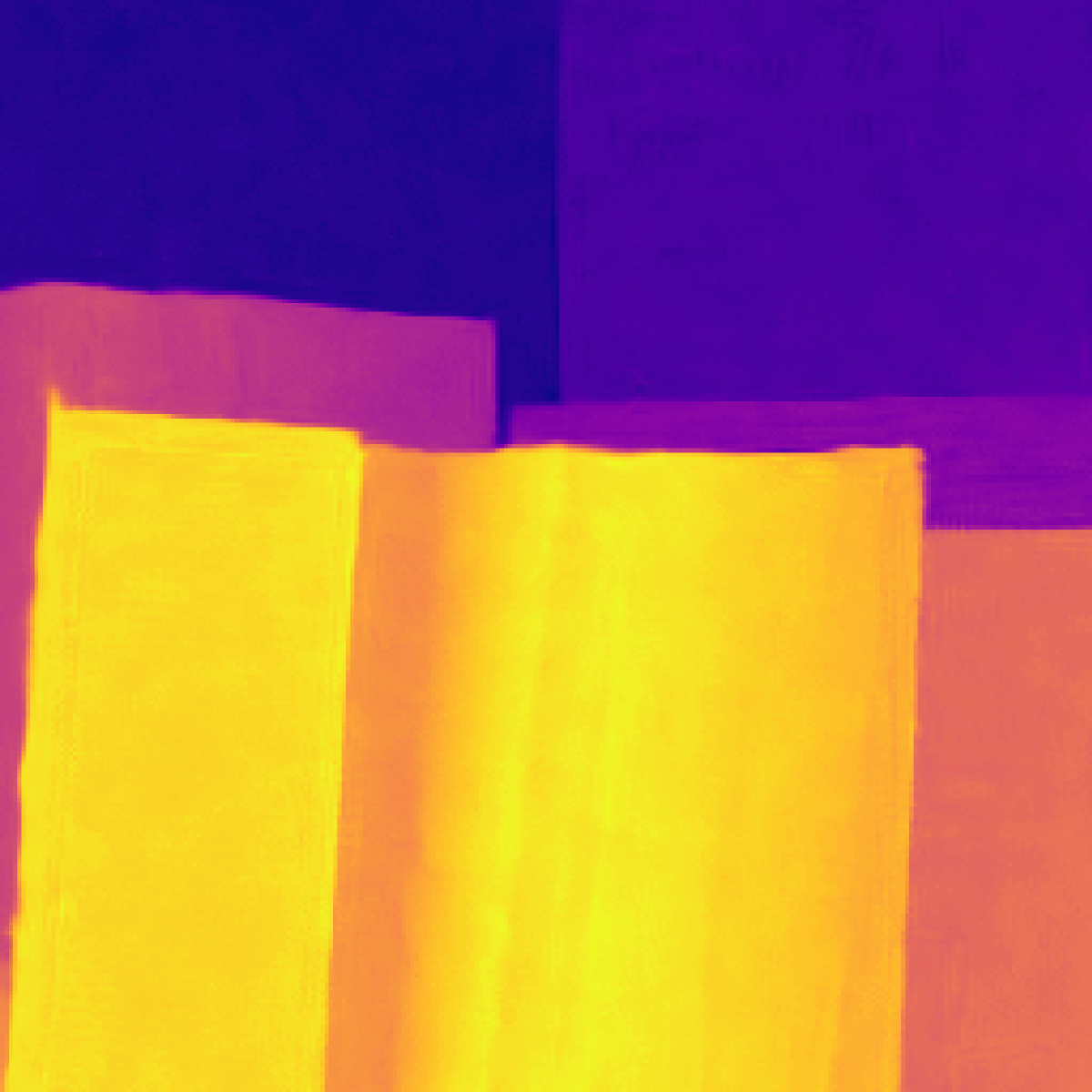}
	\hspace{-1.8mm} & \includegraphics[height=0.65in]{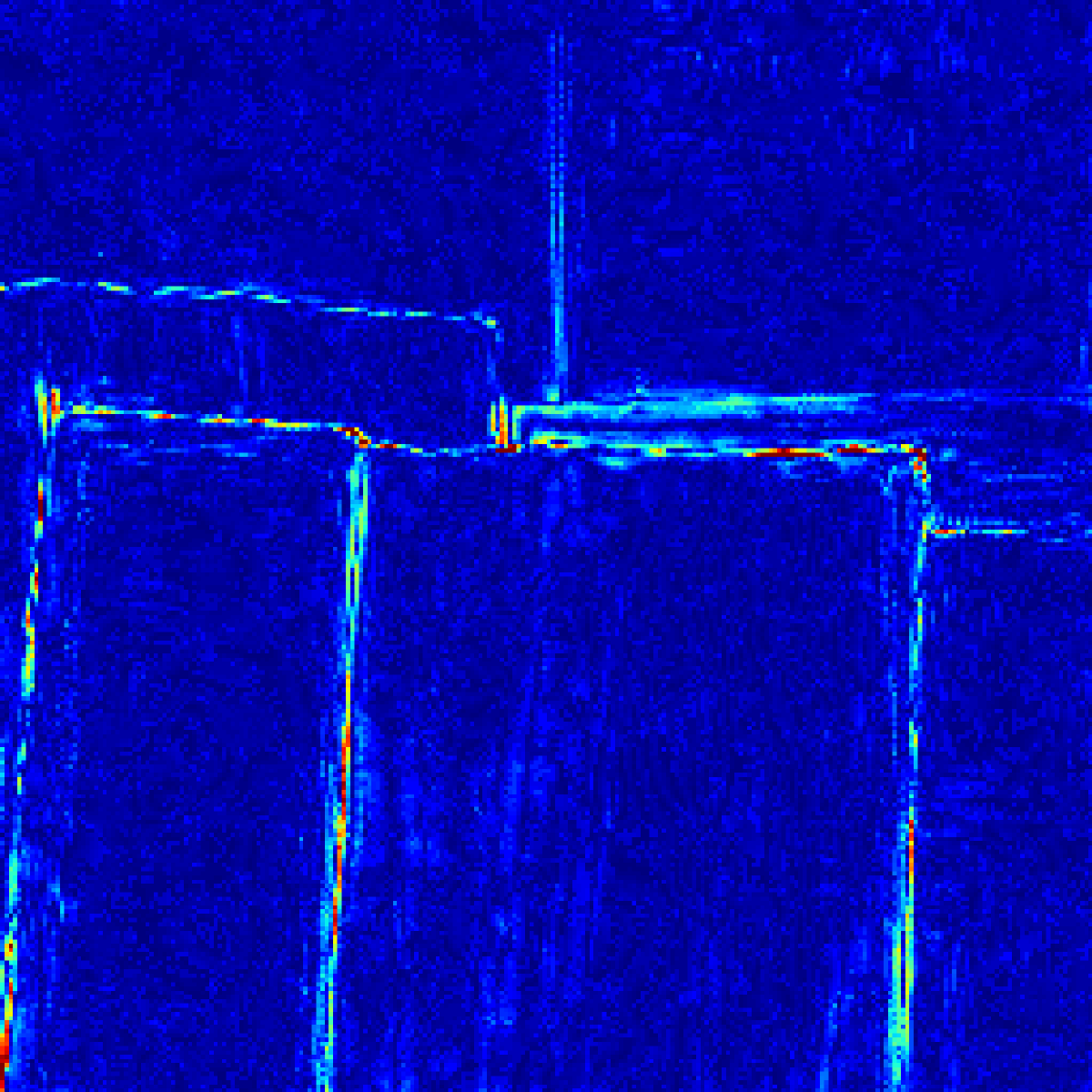}
	\hspace{-1.8mm} & \includegraphics[height=0.65in]{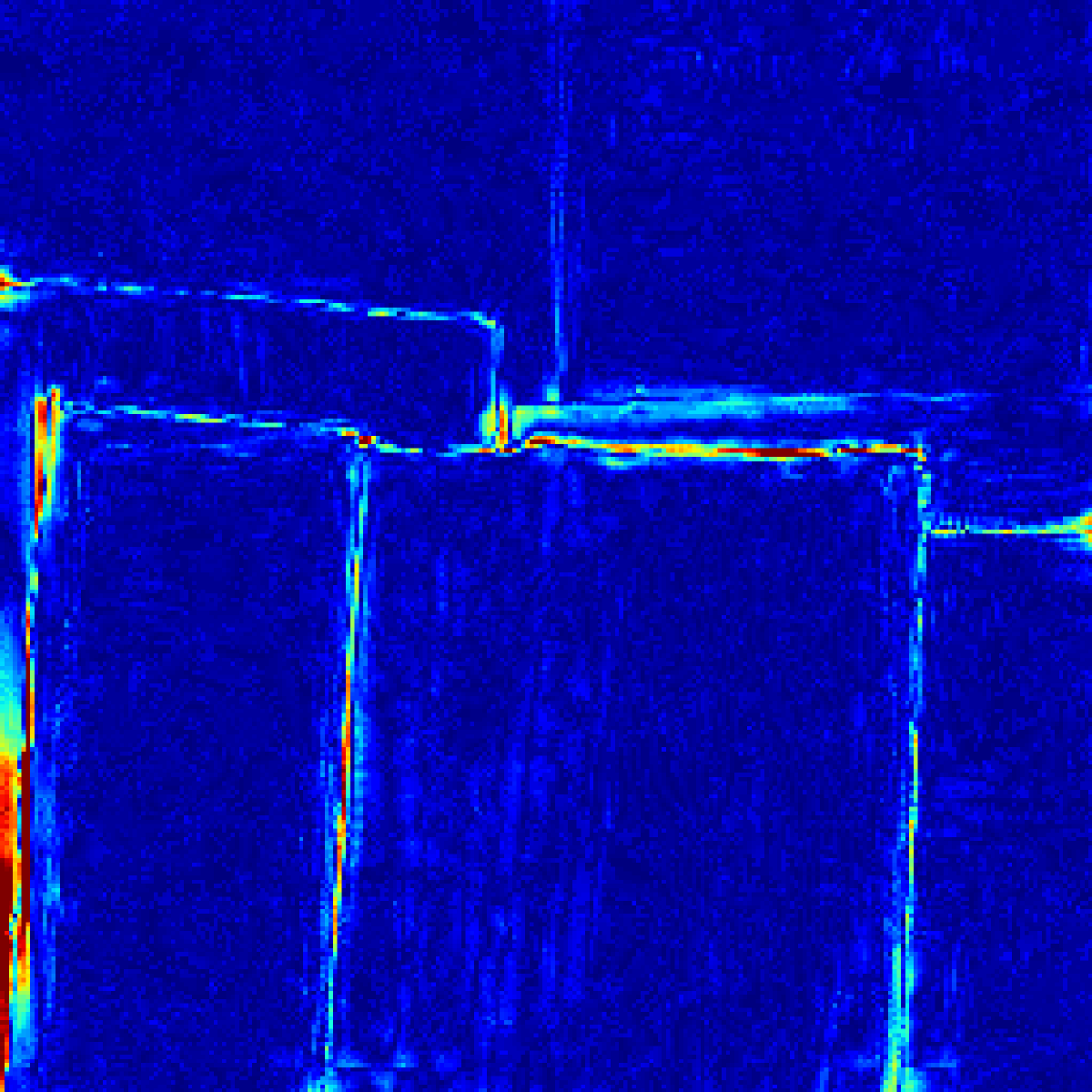}
	\hspace{-1.8mm} & \includegraphics[height=0.65in]{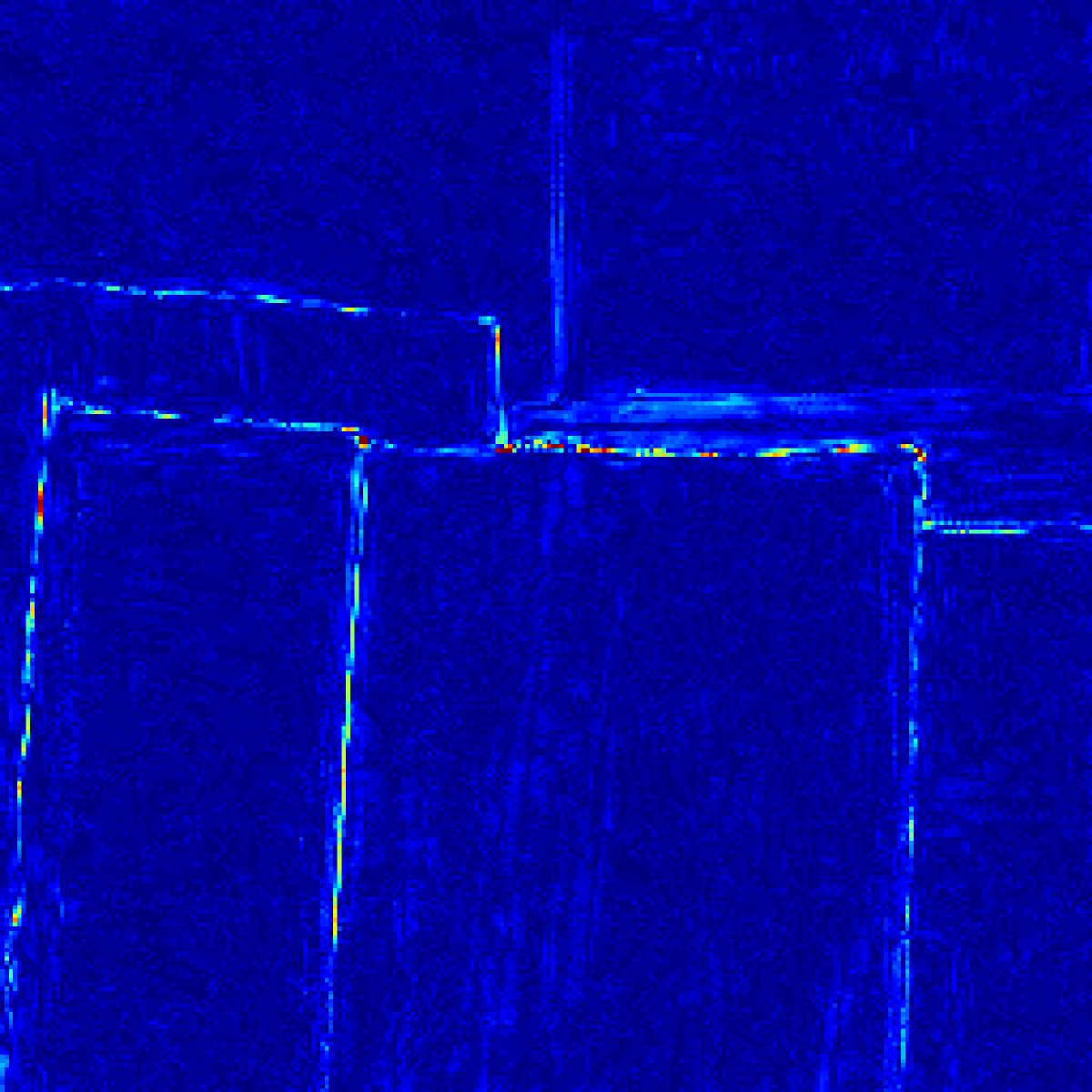}
	\hspace{-1.8mm} & \includegraphics[height=0.65in]{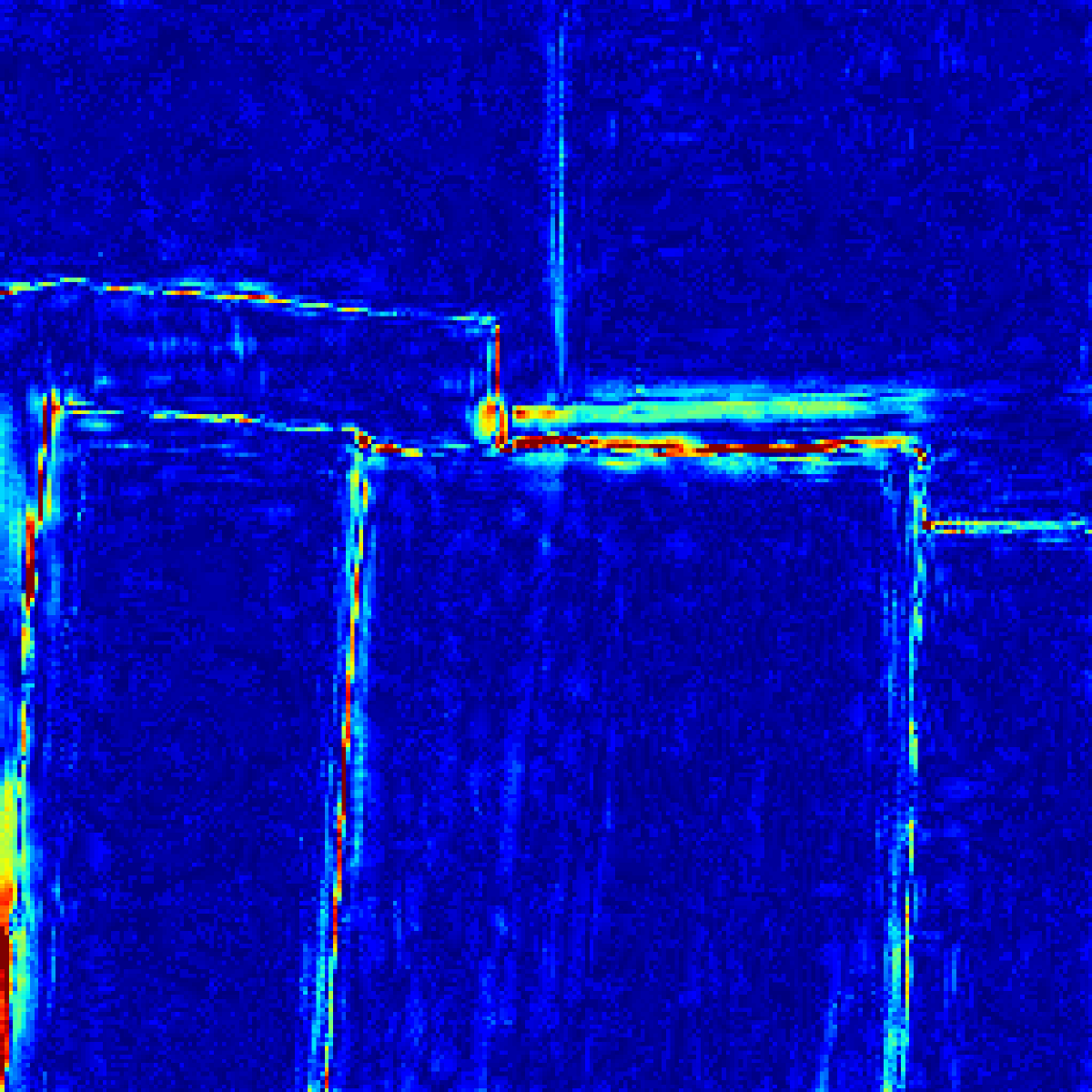}
	\hspace{-1.8mm} & \includegraphics[height=0.65in]{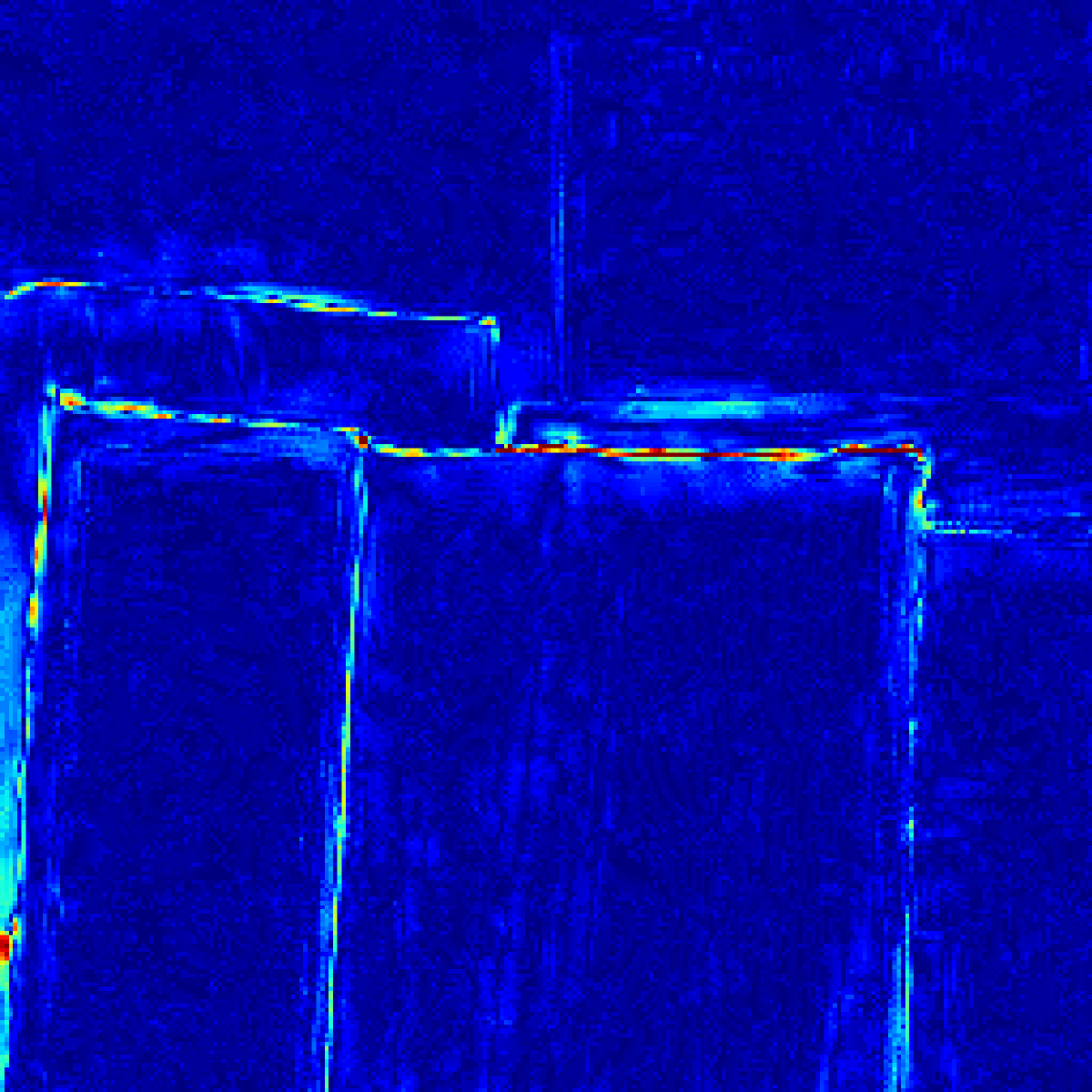}
	\hspace{-1.8mm} & \includegraphics[height=0.65in]{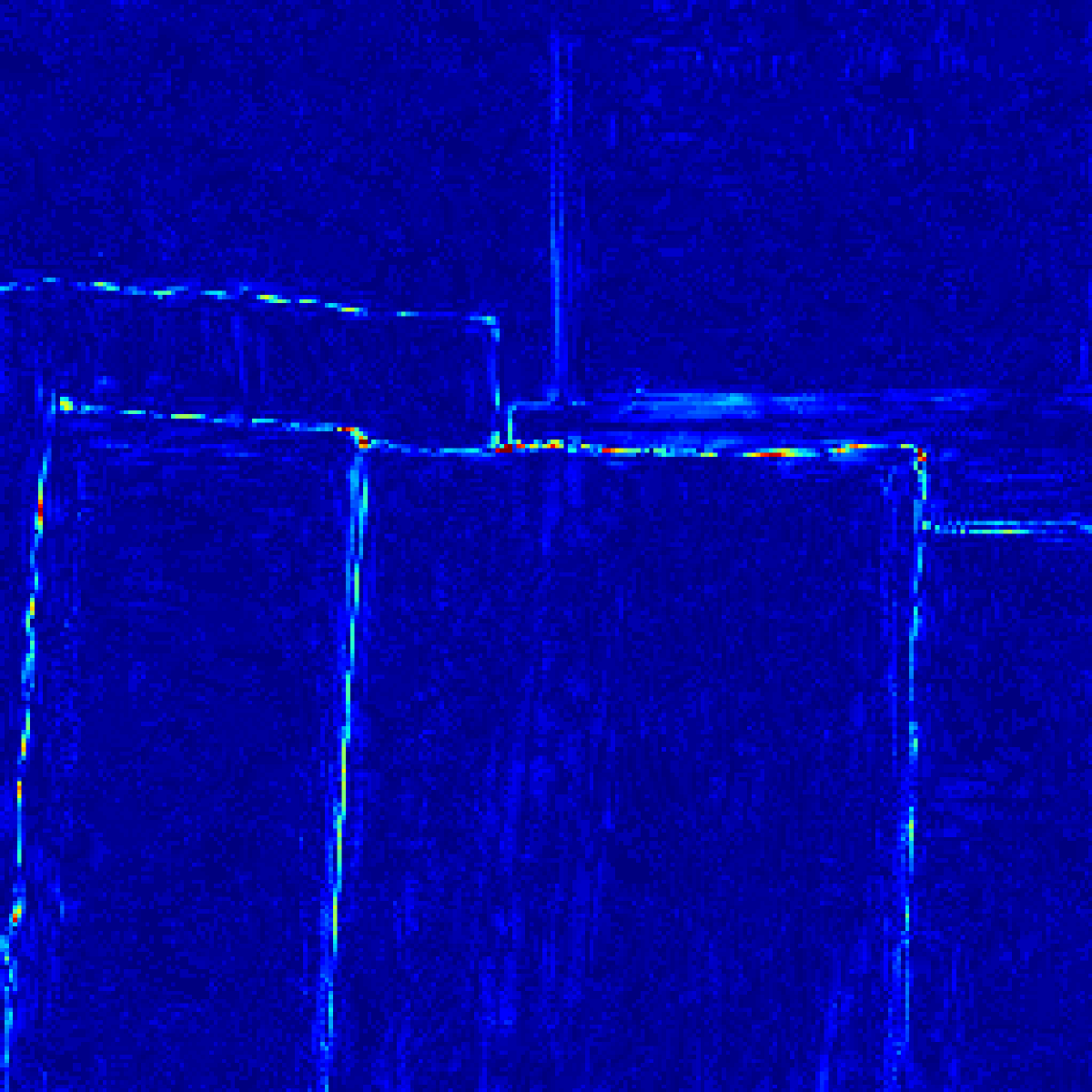}
 
	\hspace{-1.8mm} & \includegraphics[height=0.65in]{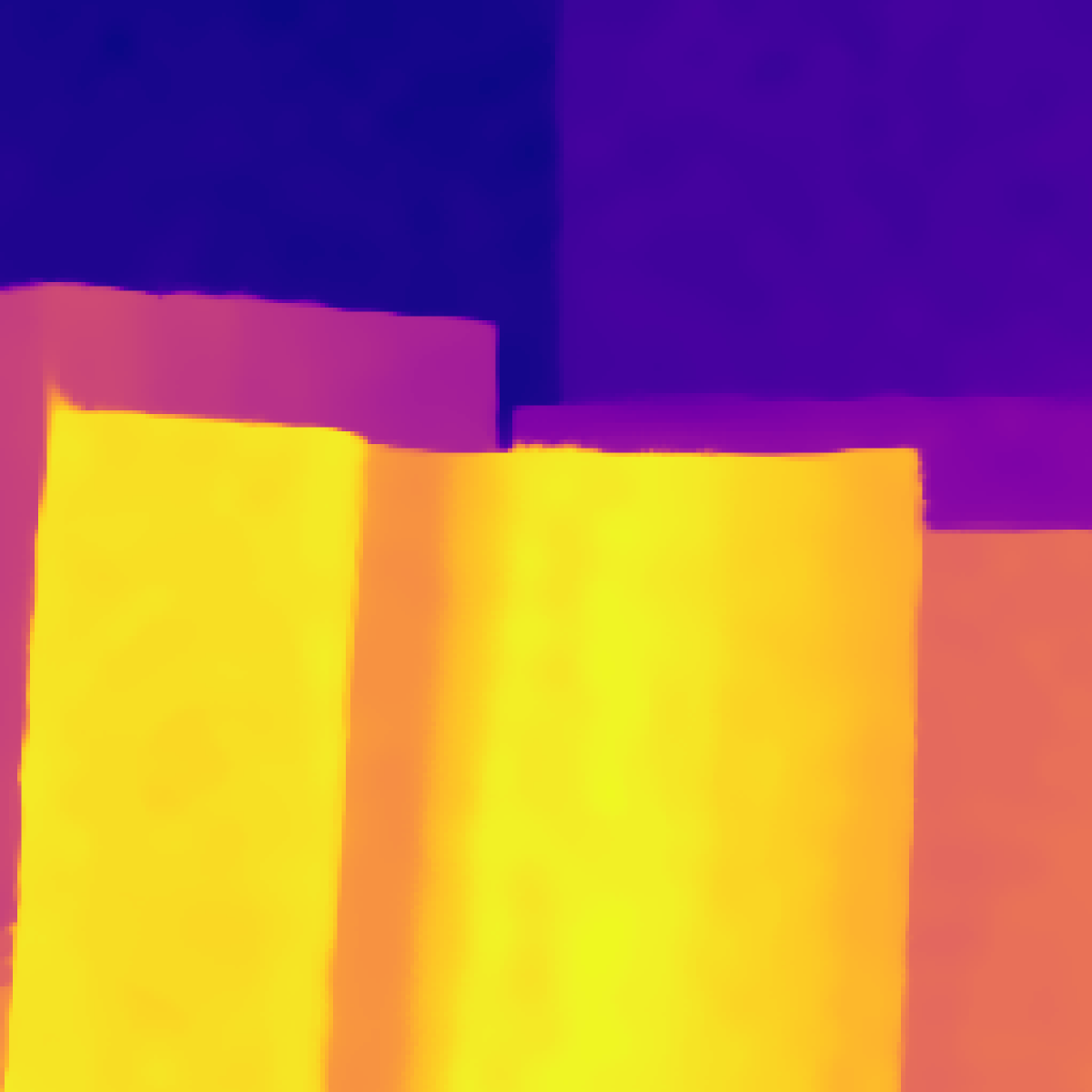}
 \\
	& \scriptsize \textbf{(a)} RGB & \scriptsize \textbf{(b)} Bicubic & \scriptsize \textbf{(c)} GT & \scriptsize \textbf{(d)} PMBA & \scriptsize \textbf{(e)} FDSR & \scriptsize \textbf{(f)} JIIF & \scriptsize \textbf{(g)} DCTNet & \scriptsize \textbf{(h)} LGR & \scriptsize \textbf{(i)} \netname{} & \scriptsize \textbf{(j)} \netname{} (depth)
	\end{tabular}
    \vspace{-0.3cm}
	\caption{\textbf{Visual comparison on cross-dataset generalization (scaling factor $8\times$).} The top, middle and last row show the error maps on the DIML dataset, the \textit{Middlebury-HR} dataset and the \textit{Middlebury-LR} dataset, respectively. From left to right: (a) RGB image, (b) Bicubic upsampled depth map, (c) GT; then, error maps achieved by selected methods: (d) PMBA~\cite{ye2020pmbanet}, (e) FDSR~\cite{he2021towards}, (f) JIIF~\cite{tang2021joint}, (g) DCTNet~\cite{zhao2022discrete}, (h) LGR~\cite{de2022learning}; finally, (i) error maps and (j) predictions by \netname.} 
	\label{fig:cross_dataset}
\end{figure*}

\textbf{Quantitative Comparison.} Tabs. \ref{sota_comparison_mid_nyu_diml} and \ref{sota_comparison_rgbdd} report the accuracy of super-solved depth maps at factors $4\times$, $8\times$ and $16\times$ on the four datasets. As expected, learning-based methods show a significant improvement over traditional methods \citep{he2010guided,ham2017robust,lutio2019guided}. \netname{} vastly outperforms any existing network, with larger gaps in accuracy with the increasing of the upsampling factor. This can be attributed to the limitations affecting existing methods, i.e., 1) the guidance of either explicit or implicit RGB features alone being insufficient; 2) multi-modal information fusion on a single scale being not flexible enough to deal with complex scenes. Both limitations are fully addressed by \netname, which consistently outperforms concurrent works \citep{metzger2022guided,yuan2023recurrent}. 

The margin is consistent both on perfect (Middlebury) and noisy datasets (NYUv2, DIML, RGBDD), with the latter being a more challenging, realistic benchmark. Although \netname$^+$ is definitely the absolute best, its margin over \netname{} is negligible, with tiny gains yielded by NLSPN with respect to our main modules. Indeed, \netname{} alone consistently outperforms any other approach already.

\textbf{Qualitative Comparison.}
Fig.~\ref{fig:sota_comp1} provides qualitative comparisons of the GDSR results across multiple datasets, i.e., NYUv2, Middlebury, and DIML, which cover various types of scenarios and noise levels. We can notice that our model can extract boundaries and details from the RGB image more accurately. Specifically, on the depth discontinuities in the two topmost rows, \netname{}$^+$ introduces fewer artifacts around the edges of objects where specular reflections occur, which means that our network is more robust in removing texture-copy effects from RGB images compared with other methods. On the two samples selected from NYUv2, our network produces fewer errors in recovering fine structures and details. For example, in the fourth row of this figure, there are many tiny objects whose shape and structure are degraded due to downsampling. Other methods may produce artifacts and inaccurate depth boundaries, while our method has a clear advantage in recovering fine-grained depth details. Fig.~\ref{fig:rgbdd_comp} also reports two examples on the RGBDD dataset. In this case, we notice fewer errors in the background, e.g., on the curtain. 

\begin{figure}[t]	
	\centering	
	\subfloat[RGB]{	
		\centering	
		\label{cross_dataset} 
		\includegraphics[height=0.8in]{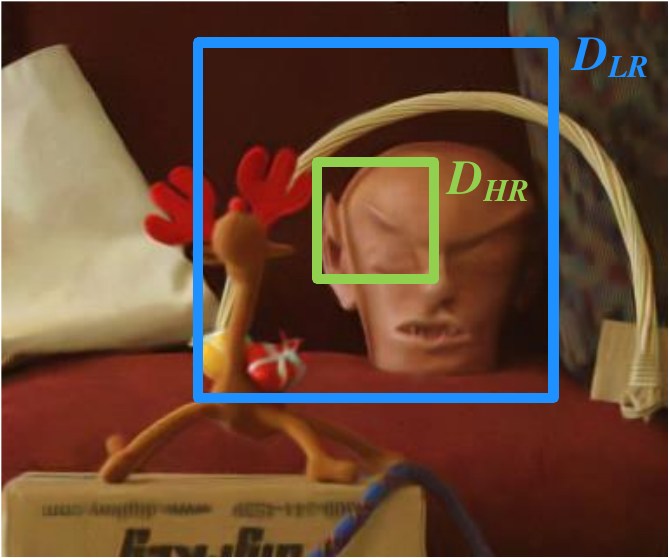}}	
	\hspace{-2mm}
	\subfloat[$D_{hr}$]{	
		\centering	
		\label{HR}
		\includegraphics[width=0.8in]{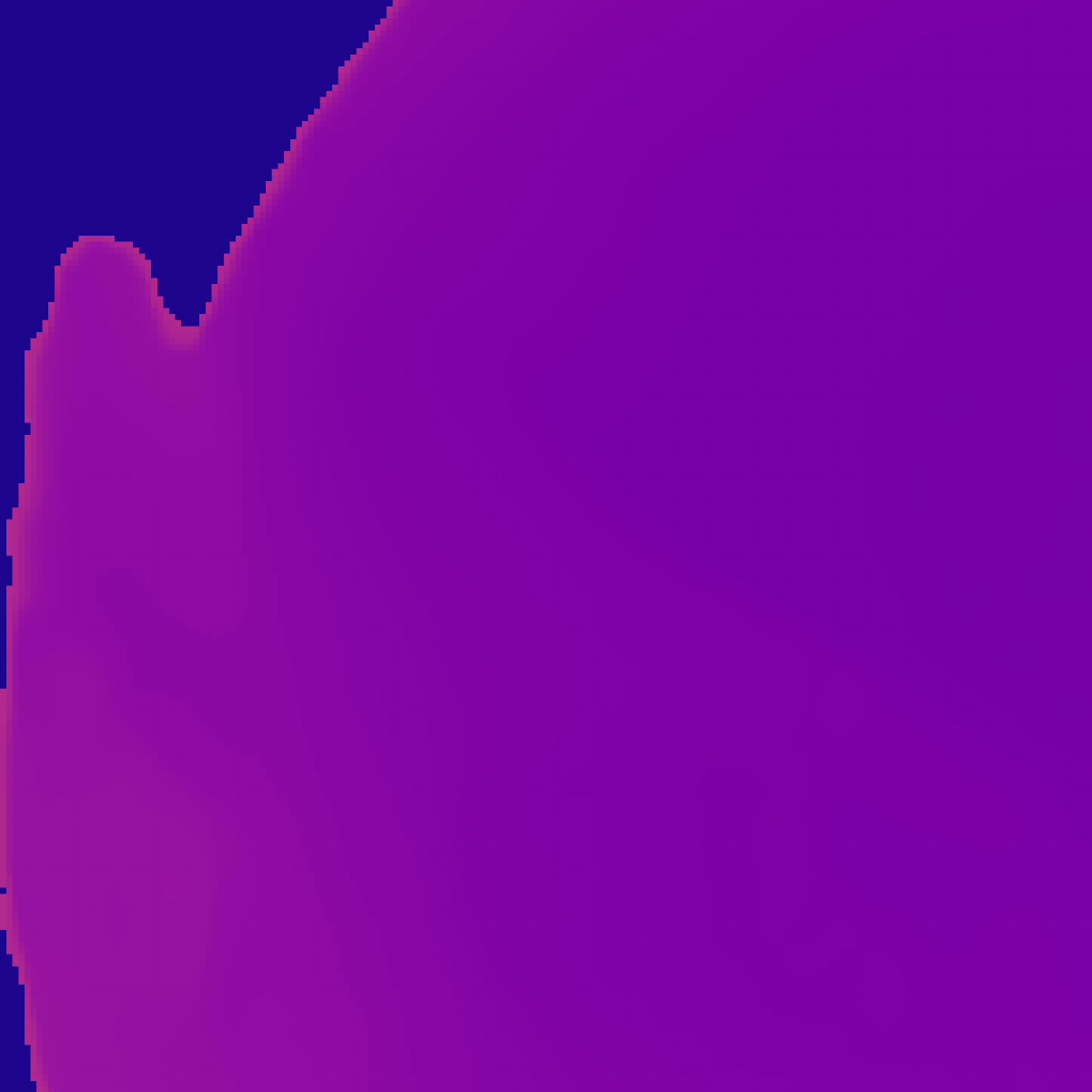}}
	\hspace{-2mm}
	\subfloat[$D_{lr}$]{	
		\centering	
		\label{LR}
		\includegraphics[width=0.8in]{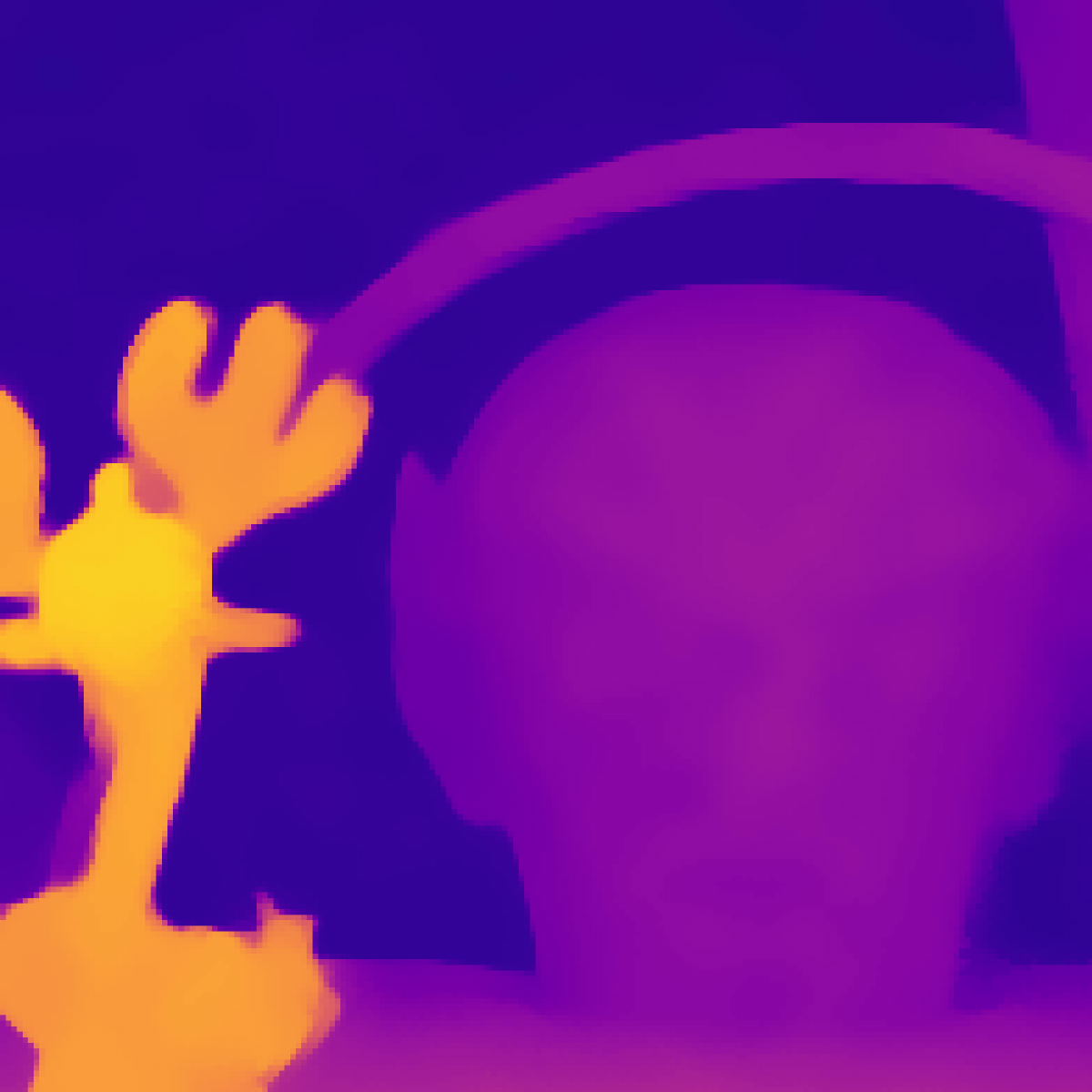}}
		\vspace{-0.3cm}
	\caption{\textbf{Image context processed on Middlebury -- HR vs LR.} (a) RGB image and depth patches $D$ processed when testing on (b) Middlebury\textit{-HR} and (c) Middlebury\textit{-LR}. }	
	\label{hr-lr} 
\end{figure}

\textbf{Cross-dataset Generalization.}
We conclude the comparison with existing methods by conducting cross-dataset experiments with $8\times$ factor. All methods are trained on the NYUv2 dataset and directly evaluated on DIML and Middlebury. Table \ref{cross-data_comparison} collects quantitative results for the 11 selected methods. Again, CNN-based methods attain better performance than traditional approaches, despite the domain gap playing a significant role in performance -- as evident by comparing results with Table \ref{cross-data_comparison}. Nonetheless, \netname{} outperforms any other framework on DIML. 

\begin{table}[t]	    \scriptsize
	\centering
	\renewcommand\tabcolsep{6pt} 
	\caption{\textbf{Abaltion study -- high-frequency information.} Scale $8\times$.}
	\begin{tabular}{@{}ccccccc@{}} 
		\toprule
		\textbf{No.} & \textbf{Gradient} & \tabincell{c}{\textbf{Shallow} \\ \textbf{Feature}} & \textbf{LCF} & \textbf{ResBlock} & \textbf{MSE} & \textbf{MAE}\\
		\midrule
		(\uppercase\expandafter{\romannumeral1}) & \XSolidBrush &  \Checkmark     &  \Checkmark &  & 13.1 & 1.19 \\
		(\uppercase\expandafter{\romannumeral2}) & \Checkmark &    \XSolidBrush   &   &  & 12.4 & 1.14 \\
		(\uppercase\expandafter{\romannumeral3}) & \Checkmark &    \Checkmark     &   & \Checkmark & 12.3 & 1.15 \\
		\gray{(\uppercase\expandafter{\romannumeral4})} & \Checkmark &    \Checkmark     & \Checkmark  &  & \textbf{11.8} & \textbf{1.12} \\
		\bottomrule
	\end{tabular}
	\label{hf_infomation}
\end{table}

When considering the Middlebury dataset, we evaluate using the setting proposed in \cite{de2022learning} -- Middlebury\textit{-HR} in the table. In this case, our results are slightly less accurate compared to a few existing methods. However, given the very high resolution of Middlebury images, we argue that this testing protocol -- i.e., consisting of processing $256\times 256$ crops at a time -- penalizes our network's ability to leverage the global context in the input that results irremediably reduced to a very local area in these images. Therefore, we also evaluate on Middlebury test set defined by~\cite{tang2021joint} -- Middlebury-\textit{LR} in the table. Note that different subsets of images are used in Middlebury\textit{-HR} and Middlebury-\textit{LR} splits. Besides, Middlebury-\textit{LR} images are resized and processed without cropping, i.e., used at full-size after resizing, allowing to fully exploit global context, while this is not feasible with Middlebury-\textit{HR} due to memory constraints. In this case, \netname{} attains the best performance again, confirming our previous analysis, as shown in Tab. \ref{cross-data_comparison}. Such a difference in terms of context is highlighted in Fig. \ref{hr-lr}.

\begin{figure}[t]	
	\centering	
	\captionsetup[subfigure]{font=footnotesize,textfont=footnotesize}
	\subfloat[]{	
		\centering	
		\label{img_diml_7} 
		\includegraphics[width=1.03in]{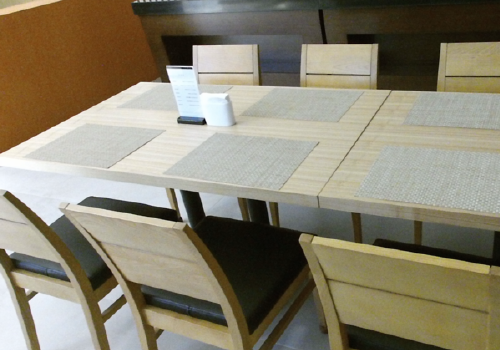}}	
	\hspace{-2.mm}
	\subfloat[]{	
		\centering	
		\label{gt_diml_7}
		\includegraphics[width=1.03in]{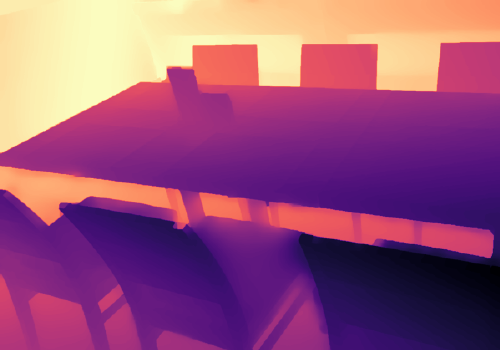}}
	\hspace{-2.mm}
	\subfloat[]{	
		\centering	
		\label{bicubic_diml_7}
		\includegraphics[width=1.03in]{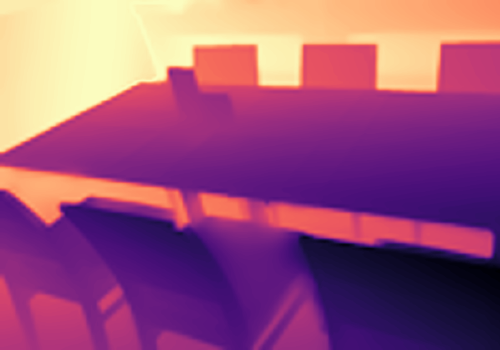}}
	
	\vspace{-0.0cm}
	\subfloat[]{	
		\centering	
		\label{feat0_diml_7}
		\includegraphics[width=1.03in]{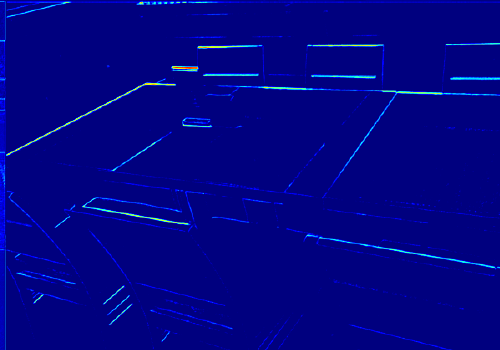}}	
	\hspace{-2.mm}
	\subfloat[]{	
		\centering	
		\label{feat1_diml_7}
		\includegraphics[width=1.03in]{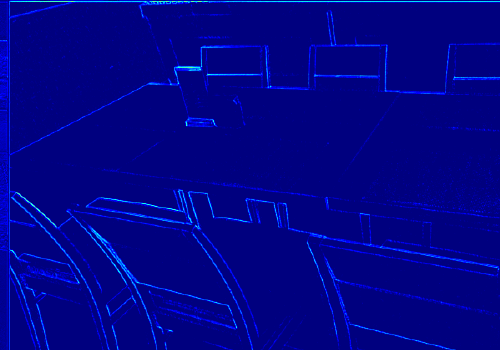}}
	\hspace{-2.mm}
	\subfloat[]{	
		\centering	
		\label{feat2_diml_7}
		\includegraphics[width=1.03in]{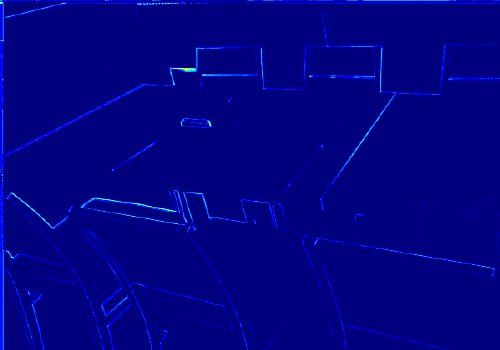}}
	\caption{\textbf{Visual exhibition of high-frequency features generated from HFEB.} (a) RGB image, (b) GT, (c) Bicubic, (d)-(f) high-frequency features.}	
	\label{hfeb_feat} 
\end{figure}

\begin{figure}[t]	
	\centering	
	\captionsetup[subfigure]{font=footnotesize,textfont=footnotesize}
	\subfloat[]{	
		\centering	
		\label{img_diml_1} 
		\includegraphics[width=1.03in]{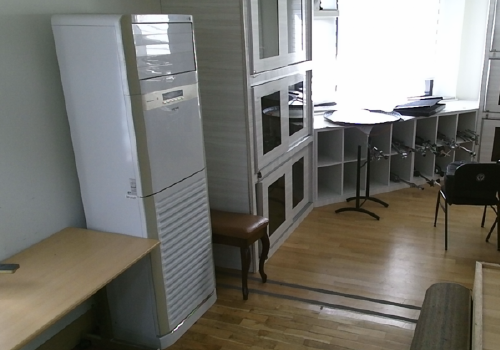}}	
	\hspace{-2mm}
	\subfloat[]{	
		\centering	
		\label{gt_diml_1}
		\includegraphics[width=1.03in]{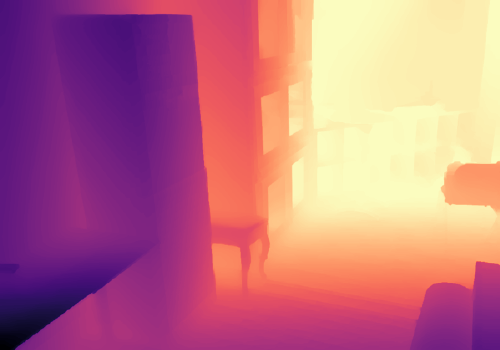}}
	\hspace{-2mm}
	\subfloat[]{	
		\centering	
		\label{bicubic_diml_1}
		\includegraphics[width=1.03in]{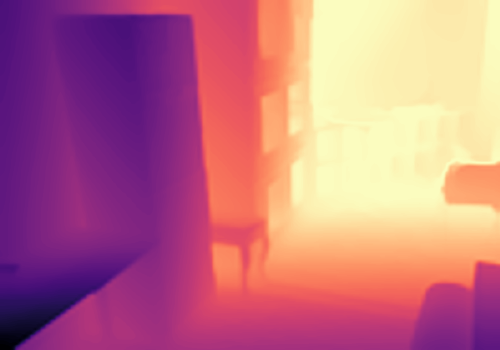}}
	
	\vspace{-0.cm}
	\subfloat[]{	
		\centering	
		\label{feat0_diml_1}
		\includegraphics[width=1.03in]{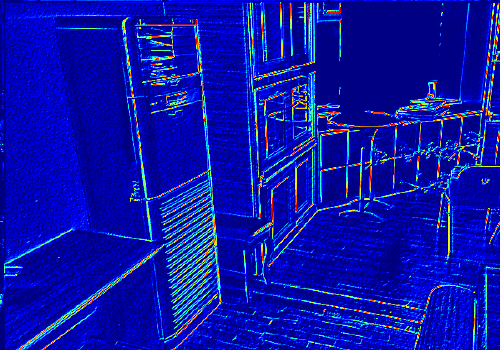}}	
	\hspace{-2mm}
	\subfloat[]{	
		\centering	
		\label{feat1_diml_1}
		\includegraphics[width=1.03in]{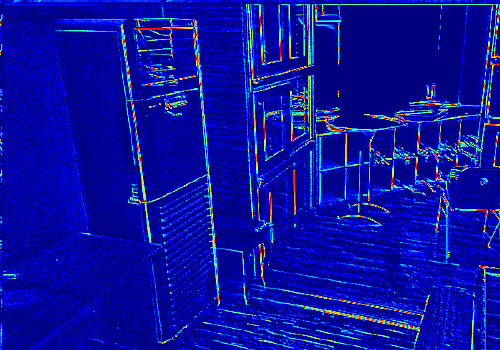}}
	\hspace{-2mm}
	\subfloat[]{	
		\centering	
		\label{feat2_diml_7}
		\includegraphics[width=1.03in]{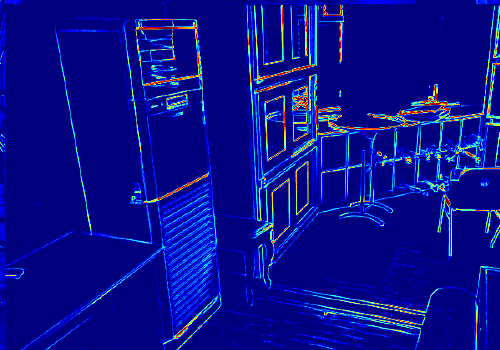}}
	\caption{\textbf{Visual exhibition of shallow high-frequency features generated from LCF.} (a) RGB image, (b) GT, (c) Bicubic, (d)-(f) high-frequency features.}	
	\label{lcf_feat} 
\end{figure}

\begin{table}[t]	\scriptsize
	\centering
	\renewcommand\tabcolsep{15pt} 
        \caption{\textbf{Different configurations for HR information.} Scale $8\times$.}
        \begin{tabular}{@{}clcc@{}} 
		\toprule
		\textbf{No.} & \textbf{HF Information} & \textbf{MSE} & \textbf{MAE}\\
		\midrule
		(\uppercase\expandafter{\romannumeral1}) & {Canny Edge} & 12.0 & 1.13 \\
		(\uppercase\expandafter{\romannumeral2}) & {Gaussian Edge} & 12.1 & 1.16 \\
		(\uppercase\expandafter{\romannumeral3}) & {DCT} & 12.1 & 1.15 \\
		(\uppercase\expandafter{\romannumeral4}) & {Wavelet Transform} & 12.1 & 1.15  \\
		\gray{(\uppercase\expandafter{\romannumeral5})} & {Gradient Map} & \textbf{11.8} & \textbf{1.12} \\
		\bottomrule
	\end{tabular}
	\label{edge_types}
\end{table}

\begin{table}[t]	\scriptsize
	\centering
	\renewcommand\tabcolsep{8pt} 
	\caption{\textbf{Effectiveness of HFEB.} Scale $8\times$.}
	\begin{tabular}{@{}clcccc@{}}
		\toprule
		\textbf{No.} & \textbf{Config.} & \textbf{Params (M)} & \textbf{Flops (G)} & \textbf{MSE} & \textbf{MAE}\\
		\midrule
		(\uppercase\expandafter{\romannumeral1}) & EdgeNet    & 5.78 &  95.6  & 12.0 & \textbf{1.12} \\
		(\uppercase\expandafter{\romannumeral2}) & SCPA  & 0.29 &  13.1  & 12.5 & 1.16 \\
		\gray{(\uppercase\expandafter{\romannumeral3})} & HFEB       & \textbf{0.27} & \textbf{11.6}  & \textbf{11.8} & \textbf{1.12} \\
		\bottomrule
	\end{tabular}
	\label{dsp_ablation}
\end{table}

\subsection{Ablation Study}
We now perform a series of ablation experiments to measure the impact of key components and parameters in \netname. We collects the outcome of these studies, conducted on NYUv2 test set with $8\times$ factor. Without loss of fairness, NLSPN is never used here -- to fully focus on the impact of single components. The configurations marked in gray in Tab.~\ref{hf_infomation}-Tab.~\ref{stage_num} correspond to our final model without NLSPN. 

\textbf{(a) Implicit vs Explicit High-Frequency Features.}
To measure the impact of both implicit and explicit HR features, we compare the performance of the proposed network and its variants when extracting either only one of the two. The quantitative results are collected in Tab.~\ref{hf_infomation}. Without the help of gradient maps (I), the performance of the network significantly degrades. We believe this is caused by the difficulty in effectively extracting fine structures or salient edges required for LR depth maps from implicit HF features alone. Moreover, explicit features highlight regions in the image that need to be focused on, avoiding \netname{} to learn to localize them and easing its task. 
{Fig.~\ref{hfeb_feat} shows three among the high-frequency features $F_{edge}$ from a representative sample. We can notice how each of the three mainly emphasizes object boundaries, confirming the effectiveness of HFEB at extracting gradient information. At the same time, we can notice how the input RGB images expose very low texture, further confirming the effectiveness of HFEB at localizing high-frequency information.}

Nonetheless, explicit HF features alone as guidance (II) are insufficient as well. We argue that the explicit information might neglect some RGB features, whereas implicit HF feature extraction can recover them. Furthermore, to verify the effectiveness of LCF, we replace it with ResBlock~\citep{he2016deep} (III) to extract shallow features from RGB images, highlighting a negative impact on implicit features extraction -- i.e., it results less accurate than (II). Fig.~\ref{lcf_feat} shows some of the features extracted by LCF. We can notice how, in addition to the primary high-frequency information, other information is encoded, such as semantics, which can further provide support for the explicit high-frequency information extracted in parallel by HFEB and improve the guidance for the final, depth super-resolution task.

\textbf{(b) Ablation on Explicit High-Frequency Features.}
Based on the previous analysis, HFEB can significantly improve the network. To determine which high-frequency information is more suitable as guidance for GDSR, we experiment with five kinds of edge maps used as ground truth $E_{gt}$ to train HFEB: (1) the Canny edge map, (2) the Gaussian high-frequency map, (3) the high-frequency map generated by discrete cosine transform, (4) the high-frequency wavelet map and (5) the gradient map, as shown in Tab.~\ref{edge_types}. The Gaussian high-frequency map is obtained using a Gaussian filter, as detailed in~\cite{wang2020depth}. Table~\ref{edge_types} reports the outcome of the evaluation. From it, we can see that the Canny edge and the gradient map allow for better performance. Although DSR-EI with the gradient map attains the best results in terms of MSE and MAE, the different types of high-frequency maps do not significantly affect the final upsampling result.

\textbf{(c) Impact of HFEB.}
To verify the effectiveness of HFEB, we replace it with EdgeNet~\citep{liu2021multi} -- based on the widely-used U-net structure -- and SCPA~\citep{zhao2020efficient}, which inspires our scaling strategy. As shown in Table~\ref{dsp_ablation}, although the parameter size of EdgeNet is 5.6M, its performance is almost the same as our HFEB, while the parameter size of our network is only 0.7M, i.e. only $\frac{1}{8}$ of it. This fact highlights that our network based on a transformer is more efficient at feature extraction. 

Besides, unlike previous works that employ fixed feature scaling rules, we adopt a dynamic scaling strategy to extract high-frequency features from depth maps. Table~\ref{dsp_ablation} also shows that our DSP with the dynamic scale strategy decreases the number of parameters while simultaneously enhancing the performance of GDSR. Compared to the original SCPA~\citep{zhao2020efficient}, DSP can perform dynamic scaling according to the characteristics of the feature map to get a more effective receptive field.

\begin{table}[t]	\scriptsize
	\centering
	\renewcommand\tabcolsep{10pt}
        \caption{\textbf{The impact of scales at which AFFM is applied.}}
	\begin{tabular}{@{}clccc@{}} 
		\toprule
		\textbf{No.} & \textbf{Scales} & \textbf{Params (M)} & \textbf{MSE} & \textbf{MAE}\\
		\midrule
		(\uppercase\expandafter{\romannumeral1}) & H1              & 1.5 & 12.3 & 1.14 \\
		\gray{(\uppercase\expandafter{\romannumeral2})} & H1, H2       & 3.0 & \textbf{11.8} & \textbf{1.12} \\
		(\uppercase\expandafter{\romannumeral3}) & H1, H2, H3              & 4.5 & \textbf{11.8} & \textbf{1.12} \\
		\bottomrule
	\end{tabular}
	\label{affm_scale}
\end{table}

\textbf{(d) Impact of AFFM.}
We now measure the effectiveness of AFFM. Tab.~\ref{affm_setting} shows results obtained by deploying AFFM at different scales, respectively the highest (I), the first two (II) and all of the three scales. We can notice how performing fusion at the highest scale alone results insufficient, whereas using multi-scale features for fusion yields improvements, despite saturating already when using two scales, with the lowest one not providing additional, meaningful details to be taken into account.

Furthermore, we ablate AFFM in its single components. Tab.~\ref{affm_setting} resumes the outcome of this evaluation. 
We first test the performance of \netname{} without AFFM (I), highlighting a large drop in accuracy. By adding dynamic fusion, yet without using attention (II) vastly improves the results already, while replacing the weighted sum in the upper of Fig.~\ref{affm} with concatenation and a ResBlock~\citep{he2016deep} (III) yields worse results compared to our full AFFM (IV). 

\begin{figure}[t]	
	\centering	
	\captionsetup[subfigure]{font=footnotesize,textfont=footnotesize}
	\subfloat[]{	
		\centering	
		\label{img_diml_9} 
		\includegraphics[width=0.8in]{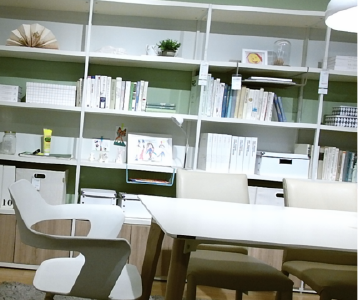}}	
	\hspace{-2mm}
	\subfloat[]{	
		\centering	
		\label{gt_diml_9}
		\includegraphics[width=0.8in]{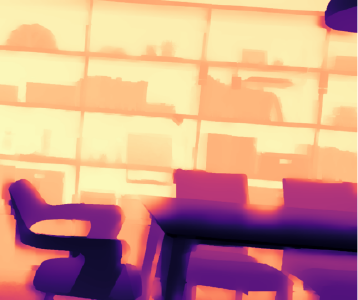}}
	\hspace{-2mm}
	\subfloat[]{	
		\centering	
		\label{bicubic_diml_9}
		\includegraphics[width=0.8in]{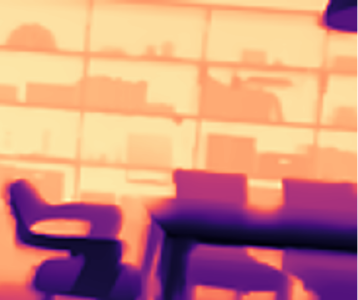}}
		\hspace{-2mm}
	\subfloat[]{	
		\centering	
		\label{feat1_diml_9}
		\includegraphics[width=0.8in]{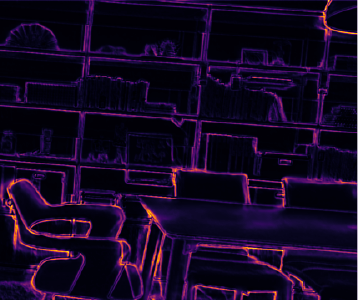}}
		
	\vspace{-0.cm}
	\subfloat[]{	
		\centering	
		\label{feat0_diml_9}
		\includegraphics[width=0.8in]{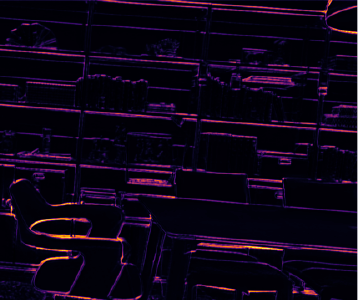}}	
	\hspace{-2mm}
	\subfloat[]{	
		\centering	
		\label{feat1_diml_9}
		\includegraphics[width=0.8in]{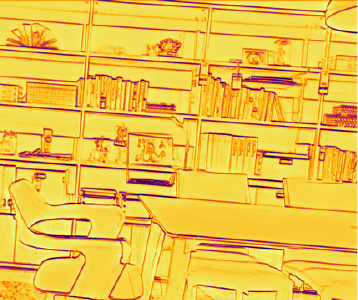}}
	\hspace{-2mm}
	\subfloat[]{	
		\centering	
		\label{feat2_diml_7}
		\includegraphics[width=0.8in]{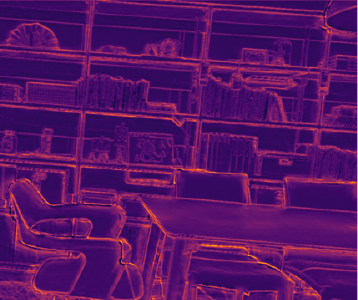}}
	\hspace{-2mm}
	\subfloat[]{	
		\centering	
		\label{feat1_diml_9}
		\includegraphics[width=0.8in]{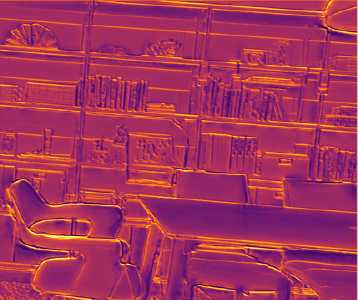}}
	\caption{\textbf{Visual exhibition of attention maps generated from AFFM.} (a) RGB image, (b) GT, (c) Bicubic, (d)-(h) attention maps.}	
	\label{affm_feat} 
\end{figure}

Fig.~\ref{affm_feat} visualizes the attention maps produced by AFFM, highlighting how sharp and accurate they are in correspondence with depth discontinuities, tiny objects, and fine details. Thus, thanks to them AFFM can better focus on reconstructing depth boundaries and details more accurately.

\begin{table}[t]	\scriptsize
	\centering
	\renewcommand\tabcolsep{10pt}
	\caption{\textbf{Ablation study of AFFM.} Scale $8\times$.}
	\begin{tabular}{@{}clccc@{}} 
		\toprule
		\textbf{No.} & \textbf{Config.} & \textbf{Params (M)} & \textbf{MSE} & \textbf{MAE}\\
		\midrule
		(\uppercase\expandafter{\romannumeral1}) & w/o AFFM        & -   & 12.7 & 1.16 \\
		(\uppercase\expandafter{\romannumeral2}) & w/o att         & 1.3 & 12.2 & 1.13 \\
		(\uppercase\expandafter{\romannumeral3}) & Concat.  & 4.5 & 12.2 & 1.13 \\
		\gray{(\uppercase\expandafter{\romannumeral4})} & AFFM & 3.0 & \textbf{11.8} & \textbf{1.12} \\
		\bottomrule
	\end{tabular}
	\label{affm_setting}
\end{table}

\begin{table}[t]	\scriptsize
	\centering
	\renewcommand\tabcolsep{10pt} 
        \caption{\textbf{Comparisons with different stage numbers.} Scale $8\times$.}
	\begin{tabular}{@{}clccc@{}}
            \toprule
		\textbf{No.} & \textbf{Stages} & \textbf{Params (M)} & \textbf{MSE} & \textbf{MAE} \\
		\midrule
		(\uppercase\expandafter{\romannumeral1}) &  $1$   & 14.2 & 13.3 & 1.19 \\
		\gray{(\uppercase\expandafter{\romannumeral2})} & 	$2$   & 25.0 & 11.8 & 1.12 \\
		(\uppercase\expandafter{\romannumeral3}) &  $3$   & 37.5 & \textbf{11.6} & \textbf{1.10} \\
		\bottomrule
	\end{tabular}
	\label{stage_num}
\end{table}

\textbf{(e) Impact of Stages Number.}
To conclude, we evaluate the impact of the multi-stage design.
As shown in Tab.~\ref{stage_num}, a single-stage architecture (I) is vastly outperformed by deploying two stages (II), yet at the expense of doubling the number of parameters. Furthermore, while the three-stage architecture (III) still yields some improvement, the benefit is minor in comparison to the significant increase in parameters. Hence, we choose two stages as the default configuration to balance accuracy and efficiency.

\begin{table*}[t] \scriptsize
	\setlength{\tabcolsep}{5pt}
	\centering
	\label{input_size}
        \caption{\textbf{Results on NYUv2 and DIML dataset -- different input sizes.} We report MSE (cm$^2$) / MAE (cm), the lower the better.}
	\begin{tabular}{@{}ccccccccc@{}}
		\toprule
		  \multirow{2}{*}{\textbf{Method}} & \multirow{2}{*}{\textbf{Size}} & \multicolumn{3}{c}{\textbf{DIML}} & \multirow{2}{*}{\textbf{Size}} & \multicolumn{3}{c}{\textbf{NYUv2}}  \\ 
            \cmidrule(r){3-5} \cmidrule(r){7-9}
                &  & \textbf{4$\times$} & \textbf{8$\times$} & \textbf{16$\times$} &  & \textbf{4$\times$} & \textbf{8$\times$} & \textbf{16$\times$}     \\
            \midrule
		\multirow{2}{*}{\textbf{DSR-EI$^+$}} & 256$\times$256 & 0.65 / \textbf{0.12} & 2.09 / 0.22 & 6.31 / 0.50 & 256$\times$256 & 2.75 / 0.47 & 11.8 / 1.09 & 47.1 / 2.40       \\ 
                                                 & 1344$\times$756 & \textbf{0.58} / \textbf{0.12} & \textbf{1.91} / \textbf{0.20} & \textbf{5.15} / \textbf{0.45} & 640$\times$480 & \textbf{1.93} / \textbf{0.39} & \textbf{8.14} / \textbf{0.89} & \textbf{33.0} / \textbf{2.02}                      \\
 \bottomrule
	\end{tabular}
	\label{tab:fullsize}
\end{table*}

\begin{table*}[t]    \scriptsize
	\renewcommand\tabcolsep{5mm} 
	\centering
         \caption{\textbf{\textbf{Computational requirements at inference}}. Experiments on Nvidia RTX 3090 GPU, with $256\times256$ input and $8\times$ factor.}
	\begin{tabular}{@{}lcccccc@{}}
		\toprule
		 & \textbf{PMBANet} & \textbf{FDSR} & \textbf{JIIF} & \textbf{DCTNet} & \textbf{LGR} & \textbf{Ours} \\ 
		 \midrule
		 \textbf{Runtime (ms)}
		 & 26.9 & 1.03 & 89.8 & 9.03 & 26.4 & 51.5\\
		 \textbf{Memory Peak (GB)}
		 & 3.07 & 2.05 & 2.36 & 0.26 & 0.19 & 18.6 \\ 
		\bottomrule
	\end{tabular}
	\vspace{-0.cm}
	\label{runtime_memory}    
\end{table*}

{\textbf{(f) Results on full-size images.}
In Tab.~\ref{sota_comparison_mid_nyu_diml} and \ref{sota_comparison_rgbdd}, we reported the results achieved by our model when processing $256\times256$ patches, to allow for a fair comparison with LGR~\citep{de2022learning} and DADA~\citep{metzger2022guided}. However, this irremediably reduces the global context processed by \netname{}, hindering its capacity to exploit it enabled by the transformer blocks similar to what was observed in the generalization experiment on Middlebury (Tab. 4). 
In this section, we demonstrate how processing larger images allows \netname{} to further improve its performance. 
Tab. \ref{tab:fullsize} compares the results achieved when switching from $256\times256$ patches to the full resolution images of DIML and NYUv2 -- i.e., $1344\times756$ and $640\times480$, respectively. We can notice consistent improvements, particularly when dealing with larger upsampling factors.}

\subsection{Limitations}
We conclude by listing a few limitations of \netname. As previously pointed out, global context is crucial for it to achieve the best performance. When this is unavailable, some accuracy is lost when generalizing across datasets. Moreover, the significant improvements over existing methods are paid for in terms of time/memory requirements. Tab. \ref{runtime_memory} highlights the higher runtime and, more evidently, peak memory usage. Future work will aim at reducing the overhead, while minimizing the drop in accuracy.

\section{Conclusion}
This paper proposed \netname{}, a depth super-resolution network, which includes a high-frequency extraction branch (HFEB) and a guided depth restoration branch (GDRB). Specifically, implemented as an efficient transformer, HFEB extracts explicit HF features. Then, GDRB deploys a two-stage encoder-decoder network to recover HR depth maps progressively, by adaptively fusing discriminative features while supplementing additional, implicit HF information. Exhaustive experiments demonstrate that \netname{} sets a new state-of-the-art for guided depth super-resolution.
\section*{Acknowledgments}
This work is supported by the National Natural Science Foundation of China (No.61627811), the Natural Science Foundation of Shaanxi Province (No.2021JZ-04), the joint project of key R\&D universities in Shaanxi Province (No.2021GXLH-Z-093, No.2021QFY01-03).

\bibliographystyle{model2-names}
\bibliography{reference}

\begin{thebibliography}{76}
\expandafter\ifx\csname natexlab\endcsname\relax\def\natexlab#1{#1}\fi
\providecommand{\url}[1]{\texttt{#1}}
\providecommand{\href}[2]{#2}
\providecommand{\path}[1]{#1}
\providecommand{\DOIprefix}{doi:}
\providecommand{\ArXivprefix}{arXiv:}
\providecommand{\URLprefix}{URL: }
\providecommand{\Pubmedprefix}{pmid:}
\providecommand{\doi}[1]{\href{http://dx.doi.org/#1}{\path{#1}}}
\providecommand{\Pubmed}[1]{\href{pmid:#1}{\path{#1}}}
\providecommand{\bibinfo}[2]{#2}
\ifx\xfnm\relax \def\xfnm[#1]{\unskip,\space#1}\fi
\bibitem[{Bamji et~al.(2022)Bamji, Godbaz, Oh, Mehta, Payne, Ortiz, Nagaraja,
  Perry and Thompson}]{bamji2022review}
\bibinfo{author}{Bamji, C.}, \bibinfo{author}{Godbaz, J.}, \bibinfo{author}{Oh,
  M.}, \bibinfo{author}{Mehta, S.}, \bibinfo{author}{Payne, A.},
  \bibinfo{author}{Ortiz, S.}, \bibinfo{author}{Nagaraja, S.},
  \bibinfo{author}{Perry, T.}, \bibinfo{author}{Thompson, B.},
  \bibinfo{year}{2022}.
\newblock \bibinfo{title}{A review of indirect time-of-flight technologies}.
\newblock \bibinfo{journal}{IEEE Transactions on Electron Devices} .
\bibitem[{Cai et~al.(2010)Cai, Gallup, Zhang and Zhang}]{cai20103d}
\bibinfo{author}{Cai, Q.}, \bibinfo{author}{Gallup, D.},
  \bibinfo{author}{Zhang, C.}, \bibinfo{author}{Zhang, Z.},
  \bibinfo{year}{2010}.
\newblock \bibinfo{title}{3d deformable face tracking with a commodity depth
  camera}, in: \bibinfo{booktitle}{European conference on computer vision},
  \bibinfo{organization}{Springer}. pp. \bibinfo{pages}{229--242}.
\bibitem[{Carion et~al.(2020)Carion, Massa, Synnaeve, Usunier, Kirillov and
  Zagoruyko}]{carion2020end}
\bibinfo{author}{Carion, N.}, \bibinfo{author}{Massa, F.},
  \bibinfo{author}{Synnaeve, G.}, \bibinfo{author}{Usunier, N.},
  \bibinfo{author}{Kirillov, A.}, \bibinfo{author}{Zagoruyko, S.},
  \bibinfo{year}{2020}.
\newblock \bibinfo{title}{End-to-end object detection with transformers}, in:
  \bibinfo{booktitle}{European conference on computer vision},
  \bibinfo{organization}{Springer}. pp. \bibinfo{pages}{213--229}.
\bibitem[{Chang et~al.(2007)Chang, Lin, Tseng and Tai}]{chang2007reversible}
\bibinfo{author}{Chang, C.C.}, \bibinfo{author}{Lin, C.C.},
  \bibinfo{author}{Tseng, C.S.}, \bibinfo{author}{Tai, W.L.},
  \bibinfo{year}{2007}.
\newblock \bibinfo{title}{Reversible hiding in dct-based compressed images}.
\newblock \bibinfo{journal}{Information Sciences} \bibinfo{volume}{177},
  \bibinfo{pages}{2768--2786}.
\bibitem[{Chen et~al.(2020a)Chen, Lin, Qian, Zeng and Li}]{chen20203d}
\bibinfo{author}{Chen, X.}, \bibinfo{author}{Lin, K.Y.}, \bibinfo{author}{Qian,
  C.}, \bibinfo{author}{Zeng, G.}, \bibinfo{author}{Li, H.},
  \bibinfo{year}{2020}a.
\newblock \bibinfo{title}{3d sketch-aware semantic scene completion via
  semi-supervised structure prior}, in: \bibinfo{booktitle}{Proceedings of the
  IEEE/CVF Conference on Computer Vision and Pattern Recognition}, pp.
  \bibinfo{pages}{4193--4202}.
\bibitem[{Chen et~al.(2020b)Chen, Xing and Zeng}]{chen2020real}
\bibinfo{author}{Chen, X.}, \bibinfo{author}{Xing, Y.}, \bibinfo{author}{Zeng,
  G.}, \bibinfo{year}{2020}b.
\newblock \bibinfo{title}{Real-time semantic scene completion via feature
  aggregation and conditioned prediction}, in: \bibinfo{booktitle}{2020 IEEE
  International Conference on Image Processing (ICIP)},
  \bibinfo{organization}{IEEE}. pp. \bibinfo{pages}{2830--2834}.
\bibitem[{Chen et~al.(2020c)Chen, Dai, Liu, Chen, Yuan and
  Liu}]{chen2020dynamic}
\bibinfo{author}{Chen, Y.}, \bibinfo{author}{Dai, X.}, \bibinfo{author}{Liu,
  M.}, \bibinfo{author}{Chen, D.}, \bibinfo{author}{Yuan, L.},
  \bibinfo{author}{Liu, Z.}, \bibinfo{year}{2020}c.
\newblock \bibinfo{title}{Dynamic convolution: Attention over convolution
  kernels}, in: \bibinfo{booktitle}{Proceedings of the IEEE/CVF Conference on
  Computer Vision and Pattern Recognition}, pp. \bibinfo{pages}{11030--11039}.
\bibitem[{Chen et~al.(2018)Chen, Wang, Peng, Zhang, Yu and
  Sun}]{chen2018cascaded}
\bibinfo{author}{Chen, Y.}, \bibinfo{author}{Wang, Z.}, \bibinfo{author}{Peng,
  Y.}, \bibinfo{author}{Zhang, Z.}, \bibinfo{author}{Yu, G.},
  \bibinfo{author}{Sun, J.}, \bibinfo{year}{2018}.
\newblock \bibinfo{title}{Cascaded pyramid network for multi-person pose
  estimation}, in: \bibinfo{booktitle}{Proceedings of the IEEE conference on
  computer vision and pattern recognition}, pp. \bibinfo{pages}{7103--7112}.
\bibitem[{Chen et~al.(2021)Chen, Cong, Xu and Huang}]{chen2021dpanet}
\bibinfo{author}{Chen, Z.}, \bibinfo{author}{Cong, R.}, \bibinfo{author}{Xu,
  Q.}, \bibinfo{author}{Huang, Q.}, \bibinfo{year}{2021}.
\newblock \bibinfo{title}{Dpanet: Depth potentiality-aware gated attention
  network for rgb-d salient object detection}.
\newblock \bibinfo{journal}{IEEE Transactions on Image Processing}
  \bibinfo{volume}{30}, \bibinfo{pages}{7012--7024}.
\newblock \DOIprefix\doi{10.1109/TIP.2020.3028289}.
\bibitem[{Cho et~al.(2021)Cho, Min, Kim and Sohn}]{cho2021deep}
\bibinfo{author}{Cho, J.}, \bibinfo{author}{Min, D.}, \bibinfo{author}{Kim,
  Y.}, \bibinfo{author}{Sohn, K.}, \bibinfo{year}{2021}.
\newblock \bibinfo{title}{Deep monocular depth estimation leveraging a
  large-scale outdoor stereo dataset}.
\newblock \bibinfo{journal}{Expert Systems with Applications}
  \bibinfo{volume}{178}, \bibinfo{pages}{114877}.
\bibitem[{Deng and Dragotti(2020)}]{deng2020deep}
\bibinfo{author}{Deng, X.}, \bibinfo{author}{Dragotti, P.L.},
  \bibinfo{year}{2020}.
\newblock \bibinfo{title}{Deep convolutional neural network for multi-modal
  image restoration and fusion}.
\newblock \bibinfo{journal}{IEEE transactions on pattern analysis and machine
  intelligence} \bibinfo{volume}{43}, \bibinfo{pages}{3333--3348}.
\bibitem[{Diebel and Thrun(2005)}]{diebel2005application}
\bibinfo{author}{Diebel, J.}, \bibinfo{author}{Thrun, S.},
  \bibinfo{year}{2005}.
\newblock \bibinfo{title}{An application of markov random fields to range
  sensing}.
\newblock \bibinfo{journal}{Advances in neural information processing systems}
  \bibinfo{volume}{18}.
\bibitem[{Dosovitskiy et~al.(2021)Dosovitskiy, Beyer, Kolesnikov, Weissenborn,
  Zhai, Unterthiner, Dehghani, Minderer, Heigold and Gelly}]{2021An}
\bibinfo{author}{Dosovitskiy, A.}, \bibinfo{author}{Beyer, L.},
  \bibinfo{author}{Kolesnikov, A.}, \bibinfo{author}{Weissenborn, D.},
  \bibinfo{author}{Zhai, X.}, \bibinfo{author}{Unterthiner, T.},
  \bibinfo{author}{Dehghani, M.}, \bibinfo{author}{Minderer, M.},
  \bibinfo{author}{Heigold, G.}, \bibinfo{author}{Gelly, S.a.},
  \bibinfo{year}{2021}.
\newblock \bibinfo{title}{An image is worth 16x16 words: Transformers for image
  recognition at scale}, in: \bibinfo{booktitle}{International Conference on
  Learning Representations}.
\bibitem[{Farha and Gall(2019)}]{farha2019ms}
\bibinfo{author}{Farha, Y.A.}, \bibinfo{author}{Gall, J.},
  \bibinfo{year}{2019}.
\newblock \bibinfo{title}{Ms-tcn: Multi-stage temporal convolutional network
  for action segmentation}, in: \bibinfo{booktitle}{Proceedings of the IEEE/CVF
  Conference on Computer Vision and Pattern Recognition}, pp.
  \bibinfo{pages}{3575--3584}.
\bibitem[{Ferstl et~al.(2013)Ferstl, Reinbacher, Ranftl, R{\"u}ther and
  Bischof}]{ferstl2013image}
\bibinfo{author}{Ferstl, D.}, \bibinfo{author}{Reinbacher, C.},
  \bibinfo{author}{Ranftl, R.}, \bibinfo{author}{R{\"u}ther, M.},
  \bibinfo{author}{Bischof, H.}, \bibinfo{year}{2013}.
\newblock \bibinfo{title}{Image guided depth upsampling using anisotropic total
  generalized variation}, in: \bibinfo{booktitle}{Proceedings of the IEEE
  international conference on computer vision}, pp. \bibinfo{pages}{993--1000}.
\bibitem[{Gao et~al.(2019)Gao, Tao, Shen and Jia}]{gao2019dynamic}
\bibinfo{author}{Gao, H.}, \bibinfo{author}{Tao, X.}, \bibinfo{author}{Shen,
  X.}, \bibinfo{author}{Jia, J.}, \bibinfo{year}{2019}.
\newblock \bibinfo{title}{Dynamic scene deblurring with parameter selective
  sharing and nested skip connections}, in: \bibinfo{booktitle}{Proceedings of
  the IEEE/CVF conference on computer vision and pattern recognition}, pp.
  \bibinfo{pages}{3848--3856}.
\bibitem[{Ge et~al.(2019)Ge, Liang, Yuan and Thalmann}]{ge2019real}
\bibinfo{author}{Ge, L.}, \bibinfo{author}{Liang, H.}, \bibinfo{author}{Yuan,
  J.}, \bibinfo{author}{Thalmann, D.}, \bibinfo{year}{2019}.
\newblock \bibinfo{title}{Real-time 3d hand pose estimation with 3d
  convolutional neural networks}.
\newblock \bibinfo{journal}{IEEE Transactions on Pattern Analysis and Machine
  Intelligence} \bibinfo{volume}{41}, \bibinfo{pages}{956--970}.
\newblock \DOIprefix\doi{10.1109/TPAMI.2018.2827052}.
\bibitem[{Ham et~al.(2017)Ham, Cho and Ponce}]{ham2017robust}
\bibinfo{author}{Ham, B.}, \bibinfo{author}{Cho, M.}, \bibinfo{author}{Ponce,
  J.}, \bibinfo{year}{2017}.
\newblock \bibinfo{title}{Robust guided image filtering using nonconvex
  potentials}.
\newblock \bibinfo{journal}{IEEE transactions on pattern analysis and machine
  intelligence} \bibinfo{volume}{40}, \bibinfo{pages}{192--207}.
\bibitem[{Han et~al.(2022)Han, Wang, Chen, Chen, Guo, Liu, Tang, Xiao, Xu, Xu
  et~al.}]{han2022survey}
\bibinfo{author}{Han, K.}, \bibinfo{author}{Wang, Y.}, \bibinfo{author}{Chen,
  H.}, \bibinfo{author}{Chen, X.}, \bibinfo{author}{Guo, J.},
  \bibinfo{author}{Liu, Z.}, \bibinfo{author}{Tang, Y.}, \bibinfo{author}{Xiao,
  A.}, \bibinfo{author}{Xu, C.}, \bibinfo{author}{Xu, Y.}, et~al.,
  \bibinfo{year}{2022}.
\newblock \bibinfo{title}{A survey on vision transformer}.
\newblock \bibinfo{journal}{IEEE transactions on pattern analysis and machine
  intelligence} .
\bibitem[{He et~al.(2010)He, Sun and Tang}]{he2010guided}
\bibinfo{author}{He, K.}, \bibinfo{author}{Sun, J.}, \bibinfo{author}{Tang,
  X.}, \bibinfo{year}{2010}.
\newblock \bibinfo{title}{Guided image filtering}, in:
  \bibinfo{booktitle}{European conference on computer vision},
  \bibinfo{organization}{Springer}. pp. \bibinfo{pages}{1--14}.
\bibitem[{He et~al.(2016)He, Zhang, Ren and Sun}]{he2016deep}
\bibinfo{author}{He, K.}, \bibinfo{author}{Zhang, X.}, \bibinfo{author}{Ren,
  S.}, \bibinfo{author}{Sun, J.}, \bibinfo{year}{2016}.
\newblock \bibinfo{title}{Deep residual learning for image recognition}, in:
  \bibinfo{booktitle}{Proceedings of the IEEE conference on computer vision and
  pattern recognition}, pp. \bibinfo{pages}{770--778}.
\bibitem[{He et~al.(2021)He, Zhu, Li, Bai, Cong, Zhang, Lin, Liu and
  Zhao}]{he2021towards}
\bibinfo{author}{He, L.}, \bibinfo{author}{Zhu, H.}, \bibinfo{author}{Li, F.},
  \bibinfo{author}{Bai, H.}, \bibinfo{author}{Cong, R.},
  \bibinfo{author}{Zhang, C.}, \bibinfo{author}{Lin, C.}, \bibinfo{author}{Liu,
  M.}, \bibinfo{author}{Zhao, Y.}, \bibinfo{year}{2021}.
\newblock \bibinfo{title}{Towards fast and accurate real-world depth
  super-resolution: Benchmark dataset and baseline}, in:
  \bibinfo{booktitle}{Proceedings of the IEEE/CVF Conference on Computer Vision
  and Pattern Recognition}, pp. \bibinfo{pages}{9229--9238}.
\bibitem[{Hirschmuller and Scharstein(2007)}]{hirschmuller2007evaluation}
\bibinfo{author}{Hirschmuller, H.}, \bibinfo{author}{Scharstein, D.},
  \bibinfo{year}{2007}.
\newblock \bibinfo{title}{Evaluation of cost functions for stereo matching},
  in: \bibinfo{booktitle}{2007 IEEE Conference on Computer Vision and Pattern
  Recognition}, \bibinfo{organization}{IEEE}. pp. \bibinfo{pages}{1--8}.
\bibitem[{Huang et~al.(2022)Huang, Huang, You, Wang, Qian and
  Xu}]{huang2022lightvit}
\bibinfo{author}{Huang, T.}, \bibinfo{author}{Huang, L.}, \bibinfo{author}{You,
  S.}, \bibinfo{author}{Wang, F.}, \bibinfo{author}{Qian, C.},
  \bibinfo{author}{Xu, C.}, \bibinfo{year}{2022}.
\newblock \bibinfo{title}{Lightvit: Towards light-weight convolution-free
  vision transformers}.
\newblock \bibinfo{journal}{arXiv preprint arXiv:2207.05557} .
\bibitem[{Hui et~al.(2016)Hui, Loy and Tang}]{hui2016depth}
\bibinfo{author}{Hui, T.W.}, \bibinfo{author}{Loy, C.C.},
  \bibinfo{author}{Tang, X.}, \bibinfo{year}{2016}.
\newblock \bibinfo{title}{Depth map super-resolution by deep multi-scale
  guidance}, in: \bibinfo{booktitle}{European conference on computer vision},
  \bibinfo{organization}{Springer}. pp. \bibinfo{pages}{353--369}.
\bibitem[{Kiechle et~al.(2013)Kiechle, Hawe and
  Kleinsteuber}]{kiechle2013joint}
\bibinfo{author}{Kiechle, M.}, \bibinfo{author}{Hawe, S.},
  \bibinfo{author}{Kleinsteuber, M.}, \bibinfo{year}{2013}.
\newblock \bibinfo{title}{A joint intensity and depth co-sparse analysis model
  for depth map super-resolution}, in: \bibinfo{booktitle}{Proceedings of the
  IEEE international conference on computer vision}, pp.
  \bibinfo{pages}{1545--1552}.
\bibitem[{Kim et~al.(2021)Kim, Ponce and Ham}]{kim2021deformable}
\bibinfo{author}{Kim, B.}, \bibinfo{author}{Ponce, J.}, \bibinfo{author}{Ham,
  B.}, \bibinfo{year}{2021}.
\newblock \bibinfo{title}{Deformable kernel networks for joint image
  filtering}.
\newblock \bibinfo{journal}{International Journal of Computer Vision}
  \bibinfo{volume}{129}, \bibinfo{pages}{579--600}.
\bibitem[{Kim et~al.(2022)Kim, Lee and Cho}]{kim2022mssnet}
\bibinfo{author}{Kim, K.}, \bibinfo{author}{Lee, S.}, \bibinfo{author}{Cho,
  S.}, \bibinfo{year}{2022}.
\newblock \bibinfo{title}{Mssnet: Multi-scale-stage network for single image
  deblurring}.
\newblock \bibinfo{journal}{arXiv preprint arXiv:2202.09652} .
\bibitem[{Kim et~al.(2017)Kim, Min, Ham, Kim and Sohn}]{kim2017deep}
\bibinfo{author}{Kim, S.}, \bibinfo{author}{Min, D.}, \bibinfo{author}{Ham,
  B.}, \bibinfo{author}{Kim, S.}, \bibinfo{author}{Sohn, K.},
  \bibinfo{year}{2017}.
\newblock \bibinfo{title}{Deep stereo confidence prediction for depth
  estimation}, in: \bibinfo{booktitle}{2017 ieee international conference on
  image processing (icip)}, \bibinfo{organization}{IEEE}. pp.
  \bibinfo{pages}{992--996}.
\bibitem[{Kim et~al.(2016)Kim, Ham, Oh and Sohn}]{kim2016structure}
\bibinfo{author}{Kim, Y.}, \bibinfo{author}{Ham, B.}, \bibinfo{author}{Oh, C.},
  \bibinfo{author}{Sohn, K.}, \bibinfo{year}{2016}.
\newblock \bibinfo{title}{Structure selective depth superresolution for rgb-d
  cameras}.
\newblock \bibinfo{journal}{IEEE Transactions on Image Processing}
  \bibinfo{volume}{25}, \bibinfo{pages}{5227--5238}.
\bibitem[{Kim et~al.(2018)Kim, Jung, Min and Sohn}]{kim2018deep}
\bibinfo{author}{Kim, Y.}, \bibinfo{author}{Jung, H.}, \bibinfo{author}{Min,
  D.}, \bibinfo{author}{Sohn, K.}, \bibinfo{year}{2018}.
\newblock \bibinfo{title}{Deep monocular depth estimation via integration of
  global and local predictions}.
\newblock \bibinfo{journal}{IEEE transactions on Image Processing}
  \bibinfo{volume}{27}, \bibinfo{pages}{4131--4144}.
\bibitem[{Kopf et~al.(2007)Kopf, Cohen, Lischinski and
  Uyttendaele}]{kopf2007joint}
\bibinfo{author}{Kopf, J.}, \bibinfo{author}{Cohen, M.F.},
  \bibinfo{author}{Lischinski, D.}, \bibinfo{author}{Uyttendaele, M.},
  \bibinfo{year}{2007}.
\newblock \bibinfo{title}{Joint bilateral upsampling}.
\newblock \bibinfo{journal}{ACM Transactions on Graphics (ToG)}
  \bibinfo{volume}{26}, \bibinfo{pages}{96--es}.
\bibitem[{Kwon et~al.(2015)Kwon, Tai and Lin}]{kwon2015data}
\bibinfo{author}{Kwon, H.}, \bibinfo{author}{Tai, Y.W.}, \bibinfo{author}{Lin,
  S.}, \bibinfo{year}{2015}.
\newblock \bibinfo{title}{Data-driven depth map refinement via multi-scale
  sparse representation}, in: \bibinfo{booktitle}{Proceedings of the IEEE
  conference on computer vision and pattern recognition}, pp.
  \bibinfo{pages}{159--167}.
\bibitem[{Lee et~al.(2022)Lee, Kim, Willette and Hwang}]{lee2022mpvit}
\bibinfo{author}{Lee, Y.}, \bibinfo{author}{Kim, J.},
  \bibinfo{author}{Willette, J.}, \bibinfo{author}{Hwang, S.J.},
  \bibinfo{year}{2022}.
\newblock \bibinfo{title}{Mpvit: Multi-path vision transformer for dense
  prediction}, in: \bibinfo{booktitle}{Proceedings of the IEEE/CVF Conference
  on Computer Vision and Pattern Recognition}, pp. \bibinfo{pages}{7287--7296}.
\bibitem[{Li et~al.(2016a)Li, Huang, Ahuja and Yang}]{li2016deep}
\bibinfo{author}{Li, Y.}, \bibinfo{author}{Huang, J.B.},
  \bibinfo{author}{Ahuja, N.}, \bibinfo{author}{Yang, M.H.},
  \bibinfo{year}{2016}a.
\newblock \bibinfo{title}{Deep joint image filtering}, in:
  \bibinfo{booktitle}{European conference on computer vision},
  \bibinfo{organization}{Springer}. pp. \bibinfo{pages}{154--169}.
\bibitem[{Li et~al.(2019)Li, Huang, Ahuja and Yang}]{li2019joint}
\bibinfo{author}{Li, Y.}, \bibinfo{author}{Huang, J.B.},
  \bibinfo{author}{Ahuja, N.}, \bibinfo{author}{Yang, M.H.},
  \bibinfo{year}{2019}.
\newblock \bibinfo{title}{Joint image filtering with deep convolutional
  networks}.
\newblock \bibinfo{journal}{IEEE transactions on pattern analysis and machine
  intelligence} \bibinfo{volume}{41}, \bibinfo{pages}{1909--1923}.
\bibitem[{Li et~al.(2016b)Li, Min, Do and Lu}]{li2016fast}
\bibinfo{author}{Li, Y.}, \bibinfo{author}{Min, D.}, \bibinfo{author}{Do,
  M.N.}, \bibinfo{author}{Lu, J.}, \bibinfo{year}{2016}b.
\newblock \bibinfo{title}{Fast guided global interpolation for depth and
  motion}, in: \bibinfo{booktitle}{European Conference on Computer Vision},
  \bibinfo{organization}{Springer}. pp. \bibinfo{pages}{717--733}.
\bibitem[{Lin et~al.(2010)Lin, Shie and Guo}]{lin2010improving}
\bibinfo{author}{Lin, S.D.}, \bibinfo{author}{Shie, S.C.},
  \bibinfo{author}{Guo, J.Y.}, \bibinfo{year}{2010}.
\newblock \bibinfo{title}{Improving the robustness of dct-based image
  watermarking against jpeg compression}.
\newblock \bibinfo{journal}{Computer Standards \& Interfaces}
  \bibinfo{volume}{32}, \bibinfo{pages}{54--60}.
\bibitem[{Liu et~al.(2022)Liu, Zhang, Meng, Gao and Wang}]{liu2022pdr}
\bibinfo{author}{Liu, P.}, \bibinfo{author}{Zhang, Z.}, \bibinfo{author}{Meng,
  Z.}, \bibinfo{author}{Gao, N.}, \bibinfo{author}{Wang, C.},
  \bibinfo{year}{2022}.
\newblock \bibinfo{title}{Pdr-net: Progressive depth reconstruction network for
  color guided depth map super-resolution}.
\newblock \bibinfo{journal}{Neurocomputing} \bibinfo{volume}{479},
  \bibinfo{pages}{75--88}.
\bibitem[{Liu et~al.(2021)Liu, Fang, Wang, Li, Sheng and Zhang}]{liu2021multi}
\bibinfo{author}{Liu, Y.}, \bibinfo{author}{Fang, F.}, \bibinfo{author}{Wang,
  T.}, \bibinfo{author}{Li, J.}, \bibinfo{author}{Sheng, Y.},
  \bibinfo{author}{Zhang, G.}, \bibinfo{year}{2021}.
\newblock \bibinfo{title}{Multi-scale grid network for image deblurring with
  high-frequency guidance}.
\newblock \bibinfo{journal}{IEEE Transactions on Multimedia} .
\bibitem[{de~Lutio et~al.(2022)de~Lutio, Becker, D'Aronco, Russo, Wegner and
  Schindler}]{de2022learning}
\bibinfo{author}{de~Lutio, R.}, \bibinfo{author}{Becker, A.},
  \bibinfo{author}{D'Aronco, S.}, \bibinfo{author}{Russo, S.},
  \bibinfo{author}{Wegner, J.D.}, \bibinfo{author}{Schindler, K.},
  \bibinfo{year}{2022}.
\newblock \bibinfo{title}{Learning graph regularisation for guided
  super-resolution}, in: \bibinfo{booktitle}{Proceedings of the IEEE/CVF
  Conference on Computer Vision and Pattern Recognition}, pp.
  \bibinfo{pages}{1979--1988}.
\bibitem[{Lutio et~al.(2019)Lutio, D'aronco, Wegner and
  Schindler}]{lutio2019guided}
\bibinfo{author}{Lutio, R.d.}, \bibinfo{author}{D'aronco, S.},
  \bibinfo{author}{Wegner, J.D.}, \bibinfo{author}{Schindler, K.},
  \bibinfo{year}{2019}.
\newblock \bibinfo{title}{Guided super-resolution as pixel-to-pixel
  transformation}, in: \bibinfo{booktitle}{Proceedings of the IEEE/CVF
  International Conference on Computer Vision}, pp.
  \bibinfo{pages}{8829--8837}.
\bibitem[{Mac~Aodha et~al.(2012)Mac~Aodha, Campbell, Nair and
  Brostow}]{mac2012patch}
\bibinfo{author}{Mac~Aodha, O.}, \bibinfo{author}{Campbell, N.D.},
  \bibinfo{author}{Nair, A.}, \bibinfo{author}{Brostow, G.J.},
  \bibinfo{year}{2012}.
\newblock \bibinfo{title}{Patch based synthesis for single depth image
  super-resolution}, in: \bibinfo{booktitle}{European conference on computer
  vision}, \bibinfo{organization}{Springer}. pp. \bibinfo{pages}{71--84}.
\bibitem[{Metzger et~al.(2022)Metzger, Daudt and Schindler}]{metzger2022guided}
\bibinfo{author}{Metzger, N.}, \bibinfo{author}{Daudt, R.C.},
  \bibinfo{author}{Schindler, K.}, \bibinfo{year}{2022}.
\newblock \bibinfo{title}{Guided depth super-resolution by deep anisotropic
  diffusion}.
\newblock \bibinfo{journal}{arXiv preprint arXiv:2211.11592} .
\bibitem[{Pan et~al.(2019)Pan, Dong, Ren, Lin, Tang and
  Yang}]{pan2019spatially}
\bibinfo{author}{Pan, J.}, \bibinfo{author}{Dong, J.}, \bibinfo{author}{Ren,
  J.S.}, \bibinfo{author}{Lin, L.}, \bibinfo{author}{Tang, J.},
  \bibinfo{author}{Yang, M.H.}, \bibinfo{year}{2019}.
\newblock \bibinfo{title}{Spatially variant linear representation models for
  joint filtering}, in: \bibinfo{booktitle}{Proceedings of the IEEE/CVF
  Conference on Computer Vision and Pattern Recognition}, pp.
  \bibinfo{pages}{1702--1711}.
\bibitem[{Park et~al.(2020)Park, Joo, Hu, Liu and So~Kweon}]{park2020non}
\bibinfo{author}{Park, J.}, \bibinfo{author}{Joo, K.}, \bibinfo{author}{Hu,
  Z.}, \bibinfo{author}{Liu, C.K.}, \bibinfo{author}{So~Kweon, I.},
  \bibinfo{year}{2020}.
\newblock \bibinfo{title}{Non-local spatial propagation network for depth
  completion}, in: \bibinfo{booktitle}{European Conference on Computer Vision},
  \bibinfo{organization}{Springer}. pp. \bibinfo{pages}{120--136}.
\bibitem[{Park et~al.(2011)Park, Kim, Tai, Brown and Kweon}]{park2011high}
\bibinfo{author}{Park, J.}, \bibinfo{author}{Kim, H.}, \bibinfo{author}{Tai,
  Y.W.}, \bibinfo{author}{Brown, M.S.}, \bibinfo{author}{Kweon, I.},
  \bibinfo{year}{2011}.
\newblock \bibinfo{title}{High quality depth map upsampling for 3d-tof
  cameras}, in: \bibinfo{booktitle}{2011 International Conference on Computer
  Vision}, \bibinfo{organization}{IEEE}. pp. \bibinfo{pages}{1623--1630}.
\bibitem[{Paszke et~al.(2019)Paszke, Gross, Massa, Lerer, Bradbury, Chanan,
  Killeen, Lin, Gimelshein, Antiga et~al.}]{paszke2019pytorch}
\bibinfo{author}{Paszke, A.}, \bibinfo{author}{Gross, S.},
  \bibinfo{author}{Massa, F.}, \bibinfo{author}{Lerer, A.},
  \bibinfo{author}{Bradbury, J.}, \bibinfo{author}{Chanan, G.},
  \bibinfo{author}{Killeen, T.}, \bibinfo{author}{Lin, Z.},
  \bibinfo{author}{Gimelshein, N.}, \bibinfo{author}{Antiga, L.}, et~al.,
  \bibinfo{year}{2019}.
\newblock \bibinfo{title}{Pytorch: An imperative style, high-performance deep
  learning library}.
\newblock \bibinfo{journal}{Advances in neural information processing systems}
  \bibinfo{volume}{32}.
\bibitem[{Pu et~al.(2022)Pu, Huang, Liu, Guan and Ling}]{pu2022edter}
\bibinfo{author}{Pu, M.}, \bibinfo{author}{Huang, Y.}, \bibinfo{author}{Liu,
  Y.}, \bibinfo{author}{Guan, Q.}, \bibinfo{author}{Ling, H.},
  \bibinfo{year}{2022}.
\newblock \bibinfo{title}{Edter: Edge detection with transformer}, in:
  \bibinfo{booktitle}{Proceedings of the IEEE/CVF Conference on Computer Vision
  and Pattern Recognition}, pp. \bibinfo{pages}{1402--1412}.
\bibitem[{Qin et~al.(2021)Qin, Zhang, Wu and Li}]{qin2021fcanet}
\bibinfo{author}{Qin, Z.}, \bibinfo{author}{Zhang, P.}, \bibinfo{author}{Wu,
  F.}, \bibinfo{author}{Li, X.}, \bibinfo{year}{2021}.
\newblock \bibinfo{title}{Fcanet: Frequency channel attention networks}, in:
  \bibinfo{booktitle}{Proceedings of the IEEE/CVF international conference on
  computer vision}, pp. \bibinfo{pages}{783--792}.
\bibitem[{Riemens et~al.(2009)Riemens, Gangwal, Barenbrug and
  Berretty}]{riemens2009multistep}
\bibinfo{author}{Riemens, A.}, \bibinfo{author}{Gangwal, O.},
  \bibinfo{author}{Barenbrug, B.}, \bibinfo{author}{Berretty, R.P.},
  \bibinfo{year}{2009}.
\newblock \bibinfo{title}{Multistep joint bilateral depth upsampling}, in:
  \bibinfo{booktitle}{Visual communications and image processing 2009},
  \bibinfo{organization}{SPIE}. pp. \bibinfo{pages}{192--203}.
\bibitem[{Ronneberger et~al.(2015)Ronneberger, Fischer and
  Brox}]{ronneberger2015u}
\bibinfo{author}{Ronneberger, O.}, \bibinfo{author}{Fischer, P.},
  \bibinfo{author}{Brox, T.}, \bibinfo{year}{2015}.
\newblock \bibinfo{title}{U-net: Convolutional networks for biomedical image
  segmentation}, in: \bibinfo{booktitle}{International Conference on Medical
  image computing and computer-assisted intervention},
  \bibinfo{organization}{Springer}. pp. \bibinfo{pages}{234--241}.
\bibitem[{Sarlin et~al.(2019)Sarlin, Cadena, Siegwart and
  Dymczyk}]{sarlin2019coarse}
\bibinfo{author}{Sarlin, P.E.}, \bibinfo{author}{Cadena, C.},
  \bibinfo{author}{Siegwart, R.}, \bibinfo{author}{Dymczyk, M.},
  \bibinfo{year}{2019}.
\newblock \bibinfo{title}{From coarse to fine: Robust hierarchical localization
  at large scale}, in: \bibinfo{booktitle}{Proceedings of the IEEE/CVF
  Conference on Computer Vision and Pattern Recognition}, pp.
  \bibinfo{pages}{12716--12725}.
\bibitem[{Scharstein et~al.(2014)Scharstein, Hirschm{\"u}ller, Kitajima,
  Krathwohl, Ne{\v{s}}i{\'c}, Wang and Westling}]{scharstein2014high}
\bibinfo{author}{Scharstein, D.}, \bibinfo{author}{Hirschm{\"u}ller, H.},
  \bibinfo{author}{Kitajima, Y.}, \bibinfo{author}{Krathwohl, G.},
  \bibinfo{author}{Ne{\v{s}}i{\'c}, N.}, \bibinfo{author}{Wang, X.},
  \bibinfo{author}{Westling, P.}, \bibinfo{year}{2014}.
\newblock \bibinfo{title}{High-resolution stereo datasets with
  subpixel-accurate ground truth}, in: \bibinfo{booktitle}{German conference on
  pattern recognition}, \bibinfo{organization}{Springer}. pp.
  \bibinfo{pages}{31--42}.
\bibitem[{Scharstein and Pal(2007)}]{scharstein2007learning}
\bibinfo{author}{Scharstein, D.}, \bibinfo{author}{Pal, C.},
  \bibinfo{year}{2007}.
\newblock \bibinfo{title}{Learning conditional random fields for stereo}, in:
  \bibinfo{booktitle}{2007 IEEE Conference on Computer Vision and Pattern
  Recognition}, \bibinfo{organization}{IEEE}. pp. \bibinfo{pages}{1--8}.
\bibitem[{Scharstein and Szeliski(2003)}]{scharstein2003high}
\bibinfo{author}{Scharstein, D.}, \bibinfo{author}{Szeliski, R.},
  \bibinfo{year}{2003}.
\newblock \bibinfo{title}{High-accuracy stereo depth maps using structured
  light}, in: \bibinfo{booktitle}{2003 IEEE Computer Society Conference on
  Computer Vision and Pattern Recognition, 2003. Proceedings.},
  \bibinfo{organization}{IEEE}. pp. \bibinfo{pages}{I--I}.
\bibitem[{Silberman et~al.(2012)Silberman, Hoiem, Kohli and
  Fergus}]{silberman2012indoor}
\bibinfo{author}{Silberman, N.}, \bibinfo{author}{Hoiem, D.},
  \bibinfo{author}{Kohli, P.}, \bibinfo{author}{Fergus, R.},
  \bibinfo{year}{2012}.
\newblock \bibinfo{title}{Indoor segmentation and support inference from rgbd
  images}, in: \bibinfo{booktitle}{European conference on computer vision},
  \bibinfo{organization}{Springer}. pp. \bibinfo{pages}{746--760}.
\bibitem[{Su et~al.(2019)Su, Jampani, Sun, Gallo, Learned-Miller and
  Kautz}]{su2019pixel}
\bibinfo{author}{Su, H.}, \bibinfo{author}{Jampani, V.}, \bibinfo{author}{Sun,
  D.}, \bibinfo{author}{Gallo, O.}, \bibinfo{author}{Learned-Miller, E.},
  \bibinfo{author}{Kautz, J.}, \bibinfo{year}{2019}.
\newblock \bibinfo{title}{Pixel-adaptive convolutional neural networks}, in:
  \bibinfo{booktitle}{Proceedings of the IEEE/CVF Conference on Computer Vision
  and Pattern Recognition}, pp. \bibinfo{pages}{11166--11175}.
\bibitem[{Tang et~al.(2021a)Tang, Chen and Zeng}]{tang2021joint}
\bibinfo{author}{Tang, J.}, \bibinfo{author}{Chen, X.}, \bibinfo{author}{Zeng,
  G.}, \bibinfo{year}{2021}a.
\newblock \bibinfo{title}{Joint implicit image function for guided depth
  super-resolution}, in: \bibinfo{booktitle}{Proceedings of the 29th ACM
  International Conference on Multimedia}, pp. \bibinfo{pages}{4390--4399}.
\bibitem[{Tang et~al.(2021b)Tang, Cong, Sheng, He, Zhang, Zhao and
  Kwong}]{tang2021bridgenet}
\bibinfo{author}{Tang, Q.}, \bibinfo{author}{Cong, R.}, \bibinfo{author}{Sheng,
  R.}, \bibinfo{author}{He, L.}, \bibinfo{author}{Zhang, D.},
  \bibinfo{author}{Zhao, Y.}, \bibinfo{author}{Kwong, S.},
  \bibinfo{year}{2021}b.
\newblock \bibinfo{title}{Bridgenet: A joint learning network of depth map
  super-resolution and monocular depth estimation}, in:
  \bibinfo{booktitle}{Proceedings of the 29th ACM International Conference on
  Multimedia}, pp. \bibinfo{pages}{2148--2157}.
\bibitem[{Touvron et~al.(2021)Touvron, Cord, Douze, Massa, Sablayrolles and
  J{\'e}gou}]{touvron2021training}
\bibinfo{author}{Touvron, H.}, \bibinfo{author}{Cord, M.},
  \bibinfo{author}{Douze, M.}, \bibinfo{author}{Massa, F.},
  \bibinfo{author}{Sablayrolles, A.}, \bibinfo{author}{J{\'e}gou, H.},
  \bibinfo{year}{2021}.
\newblock \bibinfo{title}{Training data-efficient image transformers \&
  distillation through attention}, in: \bibinfo{booktitle}{International
  Conference on Machine Learning}, \bibinfo{organization}{PMLR}. pp.
  \bibinfo{pages}{10347--10357}.
\bibitem[{Vaswani et~al.(2017)Vaswani, Shazeer, Parmar, Uszkoreit, Jones,
  Gomez, Kaiser and Polosukhin}]{vaswani2017attention}
\bibinfo{author}{Vaswani, A.}, \bibinfo{author}{Shazeer, N.},
  \bibinfo{author}{Parmar, N.}, \bibinfo{author}{Uszkoreit, J.},
  \bibinfo{author}{Jones, L.}, \bibinfo{author}{Gomez, A.N.},
  \bibinfo{author}{Kaiser, {\L}.}, \bibinfo{author}{Polosukhin, I.},
  \bibinfo{year}{2017}.
\newblock \bibinfo{title}{Attention is all you need}.
\newblock \bibinfo{journal}{Advances in neural information processing systems}
  \bibinfo{volume}{30}.
\bibitem[{Wang et~al.(2019)Wang, Kembhavi, Farhadi, Yuille and
  Rastegari}]{wang2019elastic}
\bibinfo{author}{Wang, H.}, \bibinfo{author}{Kembhavi, A.},
  \bibinfo{author}{Farhadi, A.}, \bibinfo{author}{Yuille, A.L.},
  \bibinfo{author}{Rastegari, M.}, \bibinfo{year}{2019}.
\newblock \bibinfo{title}{Elastic: Improving cnns with dynamic scaling
  policies}, in: \bibinfo{booktitle}{Proceedings of the IEEE/CVF Conference on
  Computer Vision and Pattern Recognition}, pp. \bibinfo{pages}{2258--2267}.
\bibitem[{Wang et~al.(2021)Wang, Xie, Li, Fan, Song, Liang, Lu, Luo and
  Shao}]{wang2021pyramid}
\bibinfo{author}{Wang, W.}, \bibinfo{author}{Xie, E.}, \bibinfo{author}{Li,
  X.}, \bibinfo{author}{Fan, D.P.}, \bibinfo{author}{Song, K.},
  \bibinfo{author}{Liang, D.}, \bibinfo{author}{Lu, T.}, \bibinfo{author}{Luo,
  P.}, \bibinfo{author}{Shao, L.}, \bibinfo{year}{2021}.
\newblock \bibinfo{title}{Pyramid vision transformer: A versatile backbone for
  dense prediction without convolutions}, in: \bibinfo{booktitle}{Proceedings
  of the IEEE/CVF International Conference on Computer Vision}, pp.
  \bibinfo{pages}{568--578}.
\bibitem[{Wang et~al.(2014)Wang, Ortega, Tian and Vetro}]{wang2014graph}
\bibinfo{author}{Wang, Y.}, \bibinfo{author}{Ortega, A.},
  \bibinfo{author}{Tian, D.}, \bibinfo{author}{Vetro, A.},
  \bibinfo{year}{2014}.
\newblock \bibinfo{title}{A graph-based joint bilateral approach for depth
  enhancement}, in: \bibinfo{booktitle}{2014 IEEE International Conference on
  Acoustics, Speech and Signal Processing (ICASSP)},
  \bibinfo{organization}{IEEE}. pp. \bibinfo{pages}{885--889}.
\bibitem[{Wang et~al.(2020)Wang, Ye, Sun, Yang, Xu and Li}]{wang2020depth}
\bibinfo{author}{Wang, Z.}, \bibinfo{author}{Ye, X.}, \bibinfo{author}{Sun,
  B.}, \bibinfo{author}{Yang, J.}, \bibinfo{author}{Xu, R.},
  \bibinfo{author}{Li, H.}, \bibinfo{year}{2020}.
\newblock \bibinfo{title}{Depth upsampling based on deep edge-aware learning}.
\newblock \bibinfo{journal}{Pattern Recognition} \bibinfo{volume}{103},
  \bibinfo{pages}{107274}.
\bibitem[{Yang et~al.(2007)Yang, Yang, Davis and Nist{\'e}r}]{yang2007spatial}
\bibinfo{author}{Yang, Q.}, \bibinfo{author}{Yang, R.}, \bibinfo{author}{Davis,
  J.}, \bibinfo{author}{Nist{\'e}r, D.}, \bibinfo{year}{2007}.
\newblock \bibinfo{title}{Spatial-depth super resolution for range images}, in:
  \bibinfo{booktitle}{2007 IEEE Conference on Computer Vision and Pattern
  Recognition}, \bibinfo{organization}{IEEE}. pp. \bibinfo{pages}{1--8}.
\bibitem[{Ye et~al.(2020)Ye, Sun, Wang, Yang, Xu, Li and Li}]{ye2020pmbanet}
\bibinfo{author}{Ye, X.}, \bibinfo{author}{Sun, B.}, \bibinfo{author}{Wang,
  Z.}, \bibinfo{author}{Yang, J.}, \bibinfo{author}{Xu, R.},
  \bibinfo{author}{Li, H.}, \bibinfo{author}{Li, B.}, \bibinfo{year}{2020}.
\newblock \bibinfo{title}{Pmbanet: Progressive multi-branch aggregation network
  for scene depth super-resolution}.
\newblock \bibinfo{journal}{IEEE Transactions on Image Processing}
  \bibinfo{volume}{29}, \bibinfo{pages}{7427--7442}.
\bibitem[{Yuan et~al.(2023)Yuan, Jiang, Li, Qian, Li and
  Yang}]{yuan2023recurrent}
\bibinfo{author}{Yuan, J.}, \bibinfo{author}{Jiang, H.}, \bibinfo{author}{Li,
  X.}, \bibinfo{author}{Qian, J.}, \bibinfo{author}{Li, J.},
  \bibinfo{author}{Yang, J.}, \bibinfo{year}{2023}.
\newblock \bibinfo{title}{Recurrent structure attention guidance for depth
  super-resolution}.
\newblock \bibinfo{journal}{arXiv preprint arXiv:2301.13419} .
\bibitem[{Zamir et~al.(2022)Zamir, Arora, Khan, Hayat, Khan and
  Yang}]{zamir2022restormer}
\bibinfo{author}{Zamir, S.W.}, \bibinfo{author}{Arora, A.},
  \bibinfo{author}{Khan, S.}, \bibinfo{author}{Hayat, M.},
  \bibinfo{author}{Khan, F.S.}, \bibinfo{author}{Yang, M.H.},
  \bibinfo{year}{2022}.
\newblock \bibinfo{title}{Restormer: Efficient transformer for high-resolution
  image restoration}, in: \bibinfo{booktitle}{Proceedings of the IEEE/CVF
  Conference on Computer Vision and Pattern Recognition}, pp.
  \bibinfo{pages}{5728--5739}.
\bibitem[{Zamir et~al.(2021)Zamir, Arora, Khan, Hayat, Khan, Yang and
  Shao}]{zamir2021multi}
\bibinfo{author}{Zamir, S.W.}, \bibinfo{author}{Arora, A.},
  \bibinfo{author}{Khan, S.}, \bibinfo{author}{Hayat, M.},
  \bibinfo{author}{Khan, F.S.}, \bibinfo{author}{Yang, M.H.},
  \bibinfo{author}{Shao, L.}, \bibinfo{year}{2021}.
\newblock \bibinfo{title}{Multi-stage progressive image restoration}, in:
  \bibinfo{booktitle}{Proceedings of the IEEE/CVF conference on computer vision
  and pattern recognition}, pp. \bibinfo{pages}{14821--14831}.
\bibitem[{Zhang et~al.(2018)Zhang, Li, Li, Wang, Zhong and Fu}]{zhang2018image}
\bibinfo{author}{Zhang, Y.}, \bibinfo{author}{Li, K.}, \bibinfo{author}{Li,
  K.}, \bibinfo{author}{Wang, L.}, \bibinfo{author}{Zhong, B.},
  \bibinfo{author}{Fu, Y.}, \bibinfo{year}{2018}.
\newblock \bibinfo{title}{Image super-resolution using very deep residual
  channel attention networks}, in: \bibinfo{booktitle}{Proceedings of the
  European conference on computer vision (ECCV)}, pp.
  \bibinfo{pages}{286--301}.
\bibitem[{Zhao et~al.(2020)Zhao, Kong, He, Qiao and Dong}]{zhao2020efficient}
\bibinfo{author}{Zhao, H.}, \bibinfo{author}{Kong, X.}, \bibinfo{author}{He,
  J.}, \bibinfo{author}{Qiao, Y.}, \bibinfo{author}{Dong, C.},
  \bibinfo{year}{2020}.
\newblock \bibinfo{title}{Efficient image super-resolution using pixel
  attention}, in: \bibinfo{booktitle}{European Conference on Computer Vision
  Workshops}, \bibinfo{organization}{Springer}. pp. \bibinfo{pages}{56--72}.
\bibitem[{Zhao et~al.(2022)Zhao, Zhang, Xu, Lin and Pfister}]{zhao2022discrete}
\bibinfo{author}{Zhao, Z.}, \bibinfo{author}{Zhang, J.}, \bibinfo{author}{Xu,
  S.}, \bibinfo{author}{Lin, Z.}, \bibinfo{author}{Pfister, H.},
  \bibinfo{year}{2022}.
\newblock \bibinfo{title}{Discrete cosine transform network for guided depth
  map super-resolution}, in: \bibinfo{booktitle}{Proceedings of the IEEE/CVF
  Conference on Computer Vision and Pattern Recognition}, pp.
  \bibinfo{pages}{5697--5707}.
\bibitem[{Zou et~al.(2022)Zou, Xiao, Yu, Li and Lee}]{zou2022delving}
\bibinfo{author}{Zou, X.}, \bibinfo{author}{Xiao, F.}, \bibinfo{author}{Yu,
  Z.}, \bibinfo{author}{Li, Y.}, \bibinfo{author}{Lee, Y.J.},
  \bibinfo{year}{2022}.
\newblock \bibinfo{title}{Delving deeper into anti-aliasing in convnets}.
\newblock \bibinfo{journal}{International Journal of Computer Vision} ,
  \bibinfo{pages}{1--15}.
\bibitem[{Zuo et~al.(2020)Zuo, Fang, An, Shang and Yang}]{zuo2020frequency}
\bibinfo{author}{Zuo, Y.}, \bibinfo{author}{Fang, Y.}, \bibinfo{author}{An,
  P.}, \bibinfo{author}{Shang, X.}, \bibinfo{author}{Yang, J.},
  \bibinfo{year}{2020}.
\newblock \bibinfo{title}{Frequency-dependent depth map enhancement via
  iterative depth-guided affine transformation and intensity-guided
  refinement}.
\newblock \bibinfo{journal}{IEEE Transactions on Multimedia}
  \bibinfo{volume}{23}, \bibinfo{pages}{772--783}.

\end{thebibliography}

\end{document}